# Quality Control Using Convolutional Neural Networks Applied to Samples of Very Small Size

*Rallou A. Chatzimichail, PhD [a], Aristides T. Hatjimihail, MD, PhD [b]*

[a] rc@hcsl.com; [b] ath@hcsl.com;

[a,b] *Hellenic Complex Systems Laboratory,Kostis Palamas 21, 66131 Drama, Greece.*

# 1. Abstract

Although there is extensive literature on the application of artificial neural networks (NNs) in quality control (QC), to monitor the conformity of a process to quality specifications, at least five QC measurements are required, increasing the related cost. To explore the application of neural networks to samples of QC measurements of very small size, four one-dimensional (1-D) convolutional neural networks (CNNs) were designed, trained, and tested with datasets of $n$-tuples of simulated standardized normally distributed QC measurements, for $1 \leq n \leq 4$.

The designed neural networks were compared to statistical QC functions with equal probabilities for false rejection, applied to samples of the same size. When the $n$-tuples included at least two QC measurements distributed as $\mathcal{N}(\mu, \sigma^2)$, where $0.2 < |\mu| \leq 6.0$, and $1.0 < \sigma \leq 7.0$, the designed neural networks outperformed the respective statistical QC functions. Therefore, 1-D CNNs applied to samples of 2-4 quality control measurements can be used to increase the probability of detection of the nonconformity of a process to the quality specifications, with lower cost.

# 2. Introduction
## 1.1. QC

Alternative quality control procedures can be applied to a process to test the null hypothesis, that the process conforms to the quality specifications and consequently is in control, against the alternative, that the process is out of control. When a true null hypothesis is rejected, a statistical type I error is committed. We have then a false rejection of a run of the process. The probability of a type I error is called probability for false rejection. When a false null hypothesis is accepted, a statistical type II error is committed. We fail then to detect a significant change of the probability density function of a quality characteristic of the process. The probability for rejection of a false null hypothesis equals the probability of detection of the nonconformity of the process to the quality specifications. A QC procedure can be formulated as a Boolean expression of QC functions, applied to a sample of QC measurements. If it is true, then the null hypothesis is considered as false, the process as out of control, and the run is rejected (A. T. Hatjimihail 1992).

A statistical QC procedure is evaluated as a Boolean proposition of one or more statistical QC functions. Each QC function is a decision rule, evaluated by calculating a statistic of the measured quality characteristic of a sample of QC measurements. Then, if the statistic is out of the interval between the decision limits, the decision rule is considered as true. If it is true, then the null hypothesis is considered as false, the process as out of control, and the run is rejected. Control charts are plots of the statistics, in time order.

## 1.2. NNs

NNs are adaptive computational algorithms, for statistical data modeling and classification of arbitrary precision, inspired by the brain structure and information processing. They are equivalent to nonlinear mathematical functions $\mathbf{W} = \mathrm{F}(\mathbf{Q}),$ where $\mathbf{W}$ and $\mathbf{Q}$ are vectors, matrices, or tensors of finite dimensions.

NNs consist of interconnected input, processing, and output nodes. Information flows from input to output, through weighed directed interconnections (see Figure 1). During forward propagation, each processing node receives as input the sum of the weighed outputs of other nodes plus a bias. Its output is calculated by an activation function and is propagated to other nodes. During back propagation, the weights and biases of each node are adapted to minimize an output dependent loss function. In training, forward and back propagation are applied repeatedly, using appropriate datasets.

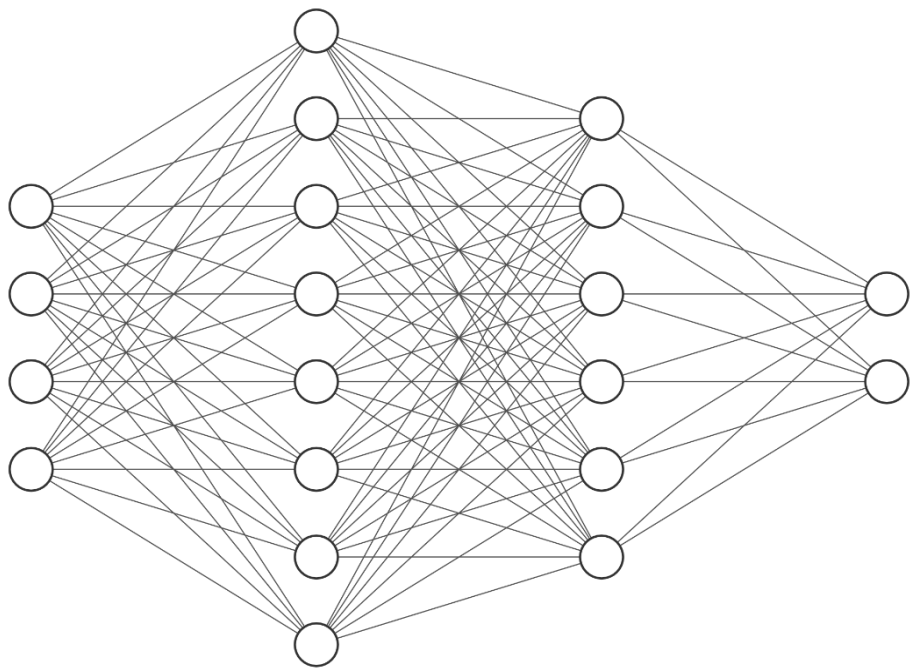

**Figure 1**. An NNs with an input layer of four nodes, two hidden layers of eight and six nodes, and an output layer of two nodes.

Since 1943 (McCulloch and Pitts 1943), there have been designed many NNs architectures, growing in complexity (Leijnen and van Veen 2020), evolving into multilayer deep neural networks (DNNs).

Common types of NNs are:

(1) Feedforward NNs, directed acyclic networks with fully connected input, hidden and output layers.
(2) Convolutional NNs (CNNs), directed acyclic networks with connected input, hidden and output layers.
(3) Recurrent NNs, directed networks, with connected input, hidden and output layers, including cyclic subnetworks.

NNs are bioinspired computational algorithms, as they are the genetic algorithms (GA). Since 1989, NNs have been proposed as alternative to statistical quality control procedures (Zorriassatine and Tannock 1998) while GA have been used for the design of QC since 1993 (A. T. Hatjimihail 1993). The first application of NNs to QC was published by Pugh (Pugh 1989), who applied NNs to samples of 5 QC measurements, with performance comparable to that of simple statistical quality control rules. Hwarng and Hubele (Hwarng and Hubele 1993) applied NNs to samples of 8 QC measurements for control charts pattern recognition among 8 unnatural control chart patterns. Other early applications of NNs to QC were published by Smith (Smith 1994), Stützle (Stützle 1995), Cheng (Cheng 1995) and Chang and Aw (Chang and Aw 1996). The application of NNs to statistical process control was reviewed comprehensively by Psarakis (Psarakis 2011). Hachicha and Ghorbel (Hachicha and Ghorbel 2012) reviewed extensively their application to control charts pattern recognition.

The input of the NNs applied to statistical process control was either $n$-tuples of QC measurements or derived statistics. To the best of our knowledge there have not been any publication in quality control about NNs applied to $n$-tuples of QC measurements for $n < 5$.

QC costs are substantial (Howanitz, Tetrault, and Steindel 1997). As the cost of QC sampling increases with the size of the sample, especially when it involves expensive QC materials or it is destructive, we explored the application of NNs to samples of QC measurements of very small size.

For that purpose, four 1-D CNNs were designed and applied to $n$-tuples of QC measurements, for $1 \leq n \leq 4$, as described in detail in the *Methods* section. CNNs are compact DNNs with low computational complexity, easier to implement in relatively low-cost systems, as they include convolutional and pooling layers. Through small kernels, each neuron of a convolutional layer receives input from only a restricted area of the previous layer, called the neuron's *receptive field*. Moreover, pooling layers combine the outputs of neuron clusters at one layer into a single neuron in the next layer, reducing further the dimensions of data. Since CNNs merge both feature extraction and classification tasks into a single structure, they have been used extensively for feature extraction in image processing (Khan et al. 2020). One-dimensional (1-D) CNNs have been used in signal processing and classification, as in anomaly detection in quality control (Kiranyaz et al. 2021).

# 3. Methods
## 3.1. Simulated QC Sample Tuples

It is assumed that $u$ are standardized normally distributed measurements of control samples when the measurement process is in control and $v$ when it is out of control. Then:

$$U \sim \mathcal{N}(0,1)$$

$$V \sim \mathcal{N}(\mu, \sigma^2)$$

where:

$$|\mu| > 0 \vee \sigma > 1$$

Series of QC measurements $n$-tuples (QCMT) were simulated, with up to $k$ measurements out of control, in all possible combinations, where:

$$1 \leq n \leq 4$$

$$1 \leq k \leq n$$

The combinations of the elements of the QCMT are presented in Table 1.

**Table 1**

*The possible control sample measurements n-tuples (QCMT)*

|   |   |   | k |   |   |   |
|---|---|---|---|---|---|---|
|   |   |   | 1 | 2 | 3 | 4 |
| n | 1 | $[u_1]$ | $[v_1]$ |   |   |   |
|   | 2 | $[u_1, u_2]$ | $[u_1, v_1]$ $[v_1, u_1]$ | $[v_1, v_2]$ |   |   |
|   | 3 | $[u_1, u_2, u_3]$ | $[u_1, u_2, v_1]$ $[u_1, v_1, u_2]$ $[v_1, u_1, u_2]$ | $[u_1, v_1, v_2]$ $[v_1, u_1, v_2]$ $[v_1, v_2, u_1]$ | $[v_1, v_2, v_3]$ |   |
|   | 4 | $[u_1, u_2, u_3, u_4]$ | $[u_1, u_2, u_3, v_1]$ $[u_1, u_2, v_1, u_3]$ $[u_1, v_1, u_2, u_3]$ $[v_1, u_1, u_2, u_3]$ | $[u_1, u_2, v_1, v_2]$ $[u_1, v_1, u_2, v_2]$ $[u_1, v_1, v_2, u_2]$ $[v_1, u_1, u_2, v_2]$ $[v_1, u_1, v_2, u_2]$ $[v_1, v_2, u_1, u_2]$ | $[u_1, v_1, v_2, v_3]$ $[v_1, u_1, v_2, v_3]$ $[v_1, v_2, u_1, v_3]$ $[v_1, v_2, v_3, u_1]$ | $[v_1, v_2, v_3, v_4]$ |

*For $1 \leq n \leq 4$, assuming $U \sim \mathcal{N}(0,1)$ and $V \sim \mathcal{N}(\mu, \sigma^2)$*

## 3.2. Simulated datasets

Training and testing datasets were created with simulated QC measurements, normally distributed and standardized.

### 3.2.1. Training Datasets

For $a \in \{1,2,3,4,6,8,12,16\}$, and $1 \leq n \leq 4$, each training dataset $\mathbf{T}_a(n)$ consists of:

1. $a\, 2^{n-1} 10^5$ $n$-tuples of random simulated QC measurements distributed as $\mathcal{N}(0,1)$,
2. $2^{n-1} 10^5$ $n$-tuples of random simulated QC measurements, $k$ distributed as $\mathcal{N}(0, \sigma^2)$ and $n-k$ as $\mathcal{N}(0,1)$, for $1 \leq k \leq n$ and $1 < \sigma \leq 11$.
3. $2^{n-1}\, 10^5$ $n$-tuples of random simulated QC measurements, $k$ distributed as $\mathcal{N}(\mu, 1)$ and $n-k$ as $\mathcal{N}(0,1)$, for $1 \leq k \leq n$ and $0 < |\mu| \leq 10$.

### 3.2.2. Testing Datasets

The following testing datasets were created:

1. $\mathbf{D}(n, 0, 1)$, for $1 \leq n \leq 4$, with $10^8$ $n$-tuples of random simulated QC measurements distributed as $\mathcal{N}(0,1)$.
2. $\mathbf{D}(n, k, 0, \sigma)$ for $\sigma = 1.1, 1.2, \ldots, 7.0$, and $1 \leq n \leq 4$ and $1 \leq k \leq n$, with $\binom{n}{k} 10^5$ $n$-tuples of random simulated QC measurements, $k$ distributed as $\mathcal{N}(0, \sigma^2)$ and $n-k$ as $\mathcal{N}(0,1)$.
3. $\mathbf{D}(n, k, \mu, 1)$ for $|\mu| = 0.1, 0.2, \ldots, 6.0$, and $1 \leq n \leq 4$ and $1 \leq k \leq n$, with $\binom{n}{k} 10^5$ $n$-tuples of random simulated QC measurements, $k$ distributed as $\mathcal{N}(\mu, 1)$ and $n-k$ as $\mathcal{N}(0,1)$

Tables 2 and 3 present the elements of the simulated training and testing datasets, with the respective numbers of the QCMT.

**Table 2:** The simulated training datasets, for $\alpha \in \{4,6,8,12,16\}$

|  |  | $n$ | $U$ | $V$ |
|---|---|---|---|---|
| **$T_a(1)$** | $\{[u_1], False\}_{i=1}^n$ | $a\,10^5$ | $\mathcal{N}(0,1)$ | |
| | $\{[v_1], True\}_{i=1}^n$ | $10^5$ | | $\mathcal{N}(0,\sigma^2), 1 < \sigma \leq 11$ |
| | | $10^5$ | | $\mathcal{N}(\mu,1), 0 < |\mu| \leq 10$ |
| **$T_a(2)$** | $\{[u_1,u_2], False\}_{i=1}^n$ | $3a\,10^5$ | $\mathcal{N}(0,1)$ | |
| | $\{[u_1,v_1], True\}_{i=1}^n$ | $10^5$ | $\mathcal{N}(0,1)$ | $\mathcal{N}(0,\sigma^2), 1 < \sigma \leq 11$ |
| | | $10^5$ | $\mathcal{N}(0,1)$ | $\mathcal{N}(\mu,1), 0 < |\mu| \leq 10$ |
| | $\{[v_1,u_1], True\}_{i=1}^n$ | $10^5$ | $\mathcal{N}(0,1)$ | $\mathcal{N}(0,\sigma^2), 1 < \sigma \leq 11$ |
| | | $10^5$ | $\mathcal{N}(0,1)$ | $\mathcal{N}(\mu,1), 0 < |\mu| \leq 10$ |
| | $\{[v_1,v_2], True\}_{i=1}^n$ | $10^5$ | $\mathcal{N}(0,1)$ | $\mathcal{N}(0,\sigma^2), 1 < \sigma \leq 11$ |
| | | $10^5$ | $\mathcal{N}(0,1)$ | $\mathcal{N}(\mu,1), 0 < |\mu| \leq 10$ |
| **$T_a(3)$** | $\{[u_1,u_2,u_3], False\}_{i=1}^n$ | $7a\,10^5$ | $\mathcal{N}(0,1)$ | |
| | $\{[u_1,u_2,v_1], True\}_{i=1}^n$ | $10^5$ | $\mathcal{N}(0,1)$ | $\mathcal{N}(0,\sigma^2), 1 < \sigma \leq 11$ |
| | | $10^5$ | $\mathcal{N}(0,1)$ | $\mathcal{N}(\mu,1), 0 < |\mu| \leq 10$ |
| | $\{[u_1,v_1,u_2], True\}_{i=1}^n$ | $10^5$ | $\mathcal{N}(0,1)$ | $\mathcal{N}(0,\sigma^2), 1 < \sigma \leq 11$ |
| | | $10^5$ | $\mathcal{N}(0,1)$ | $\mathcal{N}(\mu,1), 0 < |\mu| \leq 10$ |
| | $\{[v_1,u_1,u_2], True\}_{i=1}^n$ | $10^5$ | $\mathcal{N}(0,1)$ | $\mathcal{N}(0,\sigma^2), 1 < \sigma \leq 11$ |
| | | $10^5$ | $\mathcal{N}(0,1)$ | $\mathcal{N}(\mu,1), 0 < |\mu| \leq 10$ |
| | $\{[u_1,v_1,v_2], True\}_{i=1}^n$ | $10^5$ | $\mathcal{N}(0,1)$ | $\mathcal{N}(0,\sigma^2), 1 < \sigma \leq 11$ |
| | | $10^5$ | $\mathcal{N}(0,1)$ | $\mathcal{N}(\mu,1), 0 < |\mu| \leq 10$ |
| | $\{[v_1,u_1,v_2], True\}_{i=1}^n$ | $10^5$ | $\mathcal{N}(0,1)$ | $\mathcal{N}(0,\sigma^2), 1 < \sigma \leq 11$ |
| | | $10^5$ | $\mathcal{N}(0,1)$ | $\mathcal{N}(\mu,1), 0 < |\mu| \leq 10$ |
| | $\{[v_1,v_2,u_1], True\}_{i=1}^n$ | $10^5$ | $\mathcal{N}(0,1)$ | $N(0,\sigma^2), 1 < \sigma \leq 11$ |
| | | $10^5$ | $\mathcal{N}(0,1)$ | $\mathcal{N}(\mu,1), 0 < |\mu| \leq 10$ |
| | $\{[v_1,v_2,v_3], True\}_{i=1}^n$ | $10^5$ | | $\mathcal{N}(0,\sigma^2), 1 < \sigma \leq 11$ |
| | | $10^5$ | | $\mathcal{N}(\mu,1), 0 < |\mu| \leq 10$ |
| **$T_a(4)$** | $\{[u_1,u_2,u_3,u_4], False\}_{i=1}^n$ | $1.5a\,10^6$ | $\mathcal{N}(0,1)$ | |
| | $\{[u_1,u_2,u_3,v_1], True\}_{i=1}^n$ | $10^5$ | $\mathcal{N}(0,1)$ | $\mathcal{N}(0,\sigma^2), 1 < \sigma \leq 11$ |
| | | $10^5$ | $\mathcal{N}(0,1)$ | $\mathcal{N}(\mu,1), 0 < |\mu| \leq 10$ |
| | $\{[u_1,u_2,v_1,u_3], True\}_{i=1}^n$ | $10^5$ | $\mathcal{N}(0,1)$ | $\mathcal{N}(0,\sigma^2), 1 < \sigma \leq 11$ |
| | | $10^5$ | $\mathcal{N}(0,1)$ | $\mathcal{N}(\mu,1), 0 < |\mu| \leq 10$ |
| | $\{[u_1,v_1,u_2,u_3], True\}_{i=1}^n$ | $10^5$ | $\mathcal{N}(0,1)$ | $\mathcal{N}(0,\sigma^2), 1 < \sigma \leq 11$ |
| | | $10^5$ | $\mathcal{N}(0,1)$ | $\mathcal{N}(\mu,1), 0 < |\mu| \leq 10$ |
| | $\{[v_1,u_1,u_2,u_3], True\}_{i=1}^n$ | $10^5$ | $\mathcal{N}(0,1)$ | $\mathcal{N}(0,\sigma^2), 1 < \sigma \leq 11$ |
| | | $10^5$ | $\mathcal{N}(0,1)$ | $\mathcal{N}(\mu,1), 0 < |\mu| \leq 10$ |
| | $\{[u_1,u_2,v_1,v_2], True\}_{i=1}^n$ | $10^5$ | $\mathcal{N}(0,1)$ | $\mathcal{N}(0,\sigma^2), 1 < \sigma \leq 11$ |
| | | $10^5$ | $\mathcal{N}(0,1)$ | $\mathcal{N}(\mu,1), 0 < |\mu| \leq 10$ |
| | $\{[u_1,v_1,u_2,v_2], True\}_{i=1}^n$ | $10^5$ | $\mathcal{N}(0,1)$ | $\mathcal{N}(0,\sigma^2), 1 < \sigma \leq 11$ |
| | | $10^5$ | $\mathcal{N}(0,1)$ | $\mathcal{N}(\mu,1), 0 < |\mu| \leq 10$ |
| | $\{[u_1,v_1,v_2,u_2], True\}_{i=1}^n$ | $10^5$ | $\mathcal{N}(0,1)$ | $\mathcal{N}(0,\sigma^2), 1 < \sigma \leq 11$ |
| | | $10^5$ | $\mathcal{N}(0,1)$ | $\mathcal{N}(\mu,1), 0 < |\mu| \leq 10$ |
| | $\{[v_1,u_1,u_2,v_2], True\}_{i=1}^n$ | $10^5$ | $\mathcal{N}(0,1)$ | $\mathcal{N}(0,\sigma^2), 1 < \sigma \leq 11$ |
| | | $10^5$ | $\mathcal{N}(0,1)$ | $\mathcal{N}(\mu,1), 0 < |\mu| \leq 10$ |
| | $\{[v_1,u_1,v_2,u_2], True\}_{i=1}^n$ | $10^5$ | $\mathcal{N}(0,1)$ | $\mathcal{N}(0,\sigma^2), 1 < \sigma \leq 11$ |
| | | $10^5$ | $\mathcal{N}(0,1)$ | $\mathcal{N}(\mu,1), 0 < |\mu| \leq 10$ |
| | $\{[v_1,v_2,u_1,u_2], True\}_{i=1}^n$ | $10^5$ | $\mathcal{N}(0,1)$ | $\mathcal{N}(0,\sigma^2), 1 < \sigma \leq 11$ |
| | | $10^5$ | $\mathcal{N}(0,1)$ | $\mathcal{N}(\mu,1), 0 < |\mu| \leq 10$ |
| | $\{[u_1,v_1,v_2,v_3], True\}_{i=1}^n$ | $10^5$ | $\mathcal{N}(0,1)$ | $\mathcal{N}(0,\sigma^2), 1 < \sigma \leq 11$ |
| | | $10^5$ | $\mathcal{N}(0,1)$ | $\mathcal{N}(\mu,1), 0 < |\mu| \leq 10$ |
| | $\{[v_1,u_1,v_2,v_3], True\}_{i=1}^n$ | $10^5$ | $\mathcal{N}(0,1)$ | $\mathcal{N}(0,\sigma^2), 1 < \sigma \leq 11$ |
| | | $10^5$ | $\mathcal{N}(0,1)$ | $\mathcal{N}(\mu,1), 0 < |\mu| \leq 10$ |
| | $\{[v_1,v_2,u_1,v_3], True\}_{i=1}^n$ | $10^5$ | $\mathcal{N}(0,1)$ | $\mathcal{N}(0,\sigma^2), 1 < \sigma \leq 11$ |
| | | $10^5$ | $\mathcal{N}(0,1)$ | $\mathcal{N}(\mu,1), 0 < |\mu| \leq 10$ |
| | $\{[v_1,v_2,v_3,u_1], True\}_{i=1}^n$ | $10^5$ | $\mathcal{N}(0,1)$ | $\mathcal{N}(0,\sigma^2), 1 < \sigma \leq 11$ |
| | | $10^5$ | $\mathcal{N}(0,1)$ | $\mathcal{N}(\mu,1), 0 < |\mu| \leq 10$ |

| | | n | U | V |
|---|---|---|---|---|
| | $\{[v_1, v_2, v_3, v_4], True\}_{i=1}^n$ | $10^5$ | | $\mathcal{N}(0, \sigma^2), 1 < \sigma \leq 11$ |
| | | $10^5$ | | $\mathcal{N}(\mu, 1), 0 < |\mu| \leq 10$ |

**Table 3:** The simulated testing datasets

| | | n | U | V |
|---|---|---|---|---|
| **D(1,0,1)** | $\{[u_1], False\}_{i=1}^n$ | $8\ 10^5$ | $\mathcal{N}(0,1)$ | |
| **D(1,1,$\mu,\sigma$)** | $\{[v_1], True\}_{i=1}^n$ | $10^5$ | | $\mathcal{N}(\mu, \sigma^2), |\mu| > 0 \lor \sigma > 1$ |
| D(2,0,1) | $\{[u_1, u_2], False\}_{i=1}^n$ | $8\ 10^5$ | $\mathcal{N}(0,1)$ | |
| **D(2,1,$\mu,\sigma$)** | $\{[u_1, v_1], True\}_{i=1}^n$ | $10^5$ | $\mathcal{N}(0,1)$ | $\mathcal{N}(\mu, \sigma^2), |\mu| > 0 \lor \sigma > 1$ |
| | $\{[v_1, u_1], True\}_{i=1}^n$ | $10^5$ | | |
| **D(2,2,$\mu,\sigma$)** | $\{[v_1, v_2], True\}_{i=1}^n$ | $10^5$ | | $\mathcal{N}(\mu, \sigma^2), |\mu| > 0 \lor \sigma > 1$ |
| **D(3,0,1)** | $\{[u_1, u_2, u_3], False\}_{i=1}^n$ | $8\ 10^5$ | $\mathcal{N}(0,1)$ | |
| **D(3,1,$\mu,\sigma$)** | $\{[u_1, u_2, v_1], True\}_{i=1}^n$ | $10^5$ | $\mathcal{N}(0,1)$ | $\mathcal{N}(\mu, \sigma^2), |\mu| > 0 \lor \sigma > 1$ |
| | $\{[u_1, v_1, u_2], True\}_{i=1}^n$ | $10^5$ | | |
| | $\{[u_1, u_2, v_1], True\}_{i=1}^n$ | $10^5$ | | |
| **D(3,2,$\mu,\sigma$)** | $\{[u_1, v_1, v_2], True\}_{i=1}^n$ | $10^5$ | | |
| | $\{[v_1, u_1, v_2], True\}_{i=1}^n$ | $10^5$ | | |
| | $\{[v_1, v_2, u_1], True\}_{i=1}^n$ | $10^5$ | | |
| **D(3,3,$\mu,\sigma$)** | $\{[v_1, v_2, v_3], True\}_{i=1}^n$ | $10^5$ | | $\mathcal{N}(\mu, \sigma^2), |\mu| > 0 \lor \sigma > 1$ |
| **D(4,0,1)** | $\{[u_1, u_2, u_3, u_4], False\}_{i=1}^n$ | $8\ 10^5$ | $\mathcal{N}(0,1)$ | |
| **D(4,1,$\mu,\sigma$)** | $\{[u_1, u_2, u_3, v_1], True\}_{i=1}^n$ | $10^5$ | | |
| | $\{[u_1, u_2, v_1, u_3], True\}_{i=1}^n$ | $10^5$ | | |
| | $\{[u_1, v_1, u_2, u_3], True\}_{i=1}^n$ | $10^5$ | | |
| | $\{[v_1, u_1, u_2, u_3], True\}_{i=1}^n$ | $10^5$ | | |
| **D(4,2,$\mu,\sigma$)** | $\{[u_1, u_2, v_1, v_2], True\}_{i=1}^n$ | $10^5$ | $\mathcal{N}(0,1)$ | $\mathcal{N}(\mu, \sigma^2), |\mu| > 0 \lor \sigma > 1$ |
| | $\{[u_1, v_1, u_2, v_2], True\}_{i=1}^n$ | $10^5$ | | |
| | $\{[u_1, v_1, v_2, u_2], True\}_{i=1}^n$ | $10^5$ | | |
| | $\{[v_1, u_1, u_2, v_2], True\}_{i=1}^n$ | $10^5$ | | |
| | $\{[v_1, u_1, v_2, u_2], True\}_{i=1}^n$ | $10^5$ | | |
| | $\{[v_1, v_2, u_1, u_2], True\}_{i=1}^n$ | $10^5$ | | |
| **D(4,3,$\mu,\sigma$)** | $\{[u_1, v_1, v_2, v_3], True\}_{i=1}^n$ | $10^5$ | | |
| | $\{[v_1, u_1, v_2, v_3], True\}_{i=1}^n$ | $10^5$ | | |
| | $\{[v_1, v_2, u_1, v_3], True\}_{i=1}^n$ | $10^5$ | | |
| | $\{[v_1, v_2, v_3, u_1], True\}_{i=1}^n$ | $10^5$ | | |
| **D(4,4,$\mu,\sigma$)** | $\{[v_1, v_2, v_3, v_4], True\}_{i=1}^n$ | $10^5$ | | $\mathcal{N}(\mu, \sigma^2), |\mu| > 0 \lor \sigma > 1$ |

### 3.3. QC functions

As QC function we define a Boolean valued function applied to a n-tuple **x** of QC measurements of a run of a process. If it is *true,* the process is considered out of control, and the run is rejected.

#### 3.3.1. Neural Network QC functions

As $N_\alpha(\mathbf{x})$ is denoted a neural network QC function. For this project, there were designed four 1-D CNNs with 69 layers, 486 – 11634 nodes and 1x1 kernels. Figure 2 presents their template. In Table 4 there is a summary presentation of their layers and nodes.

**Table 4:** CNNs Layers and nodes

| CNNs | $N_a(x_1)$ | | $N_a(x_1, x_2)$ | | $N_a(x_1, x_2, x_3)$ | | $N_a(x_1, x_2, x_3, x_4)$ | |
|---|---|---|---|---|---|---|---|---|
| | Layers | Nodes | Layers | Nodes | Layers | Nodes | Layers | Nodes |
| Input | 1 | 1 | 1 | 2 | 1 | 3 | 1 | 4 |
| Hidden | 67 | 483 | 67 | 2102 | 67 | 5577 | 67 | 11628 |
| Output | 1 | 2 | 1 | 2 | 1 | 2 | 1 | 2 |

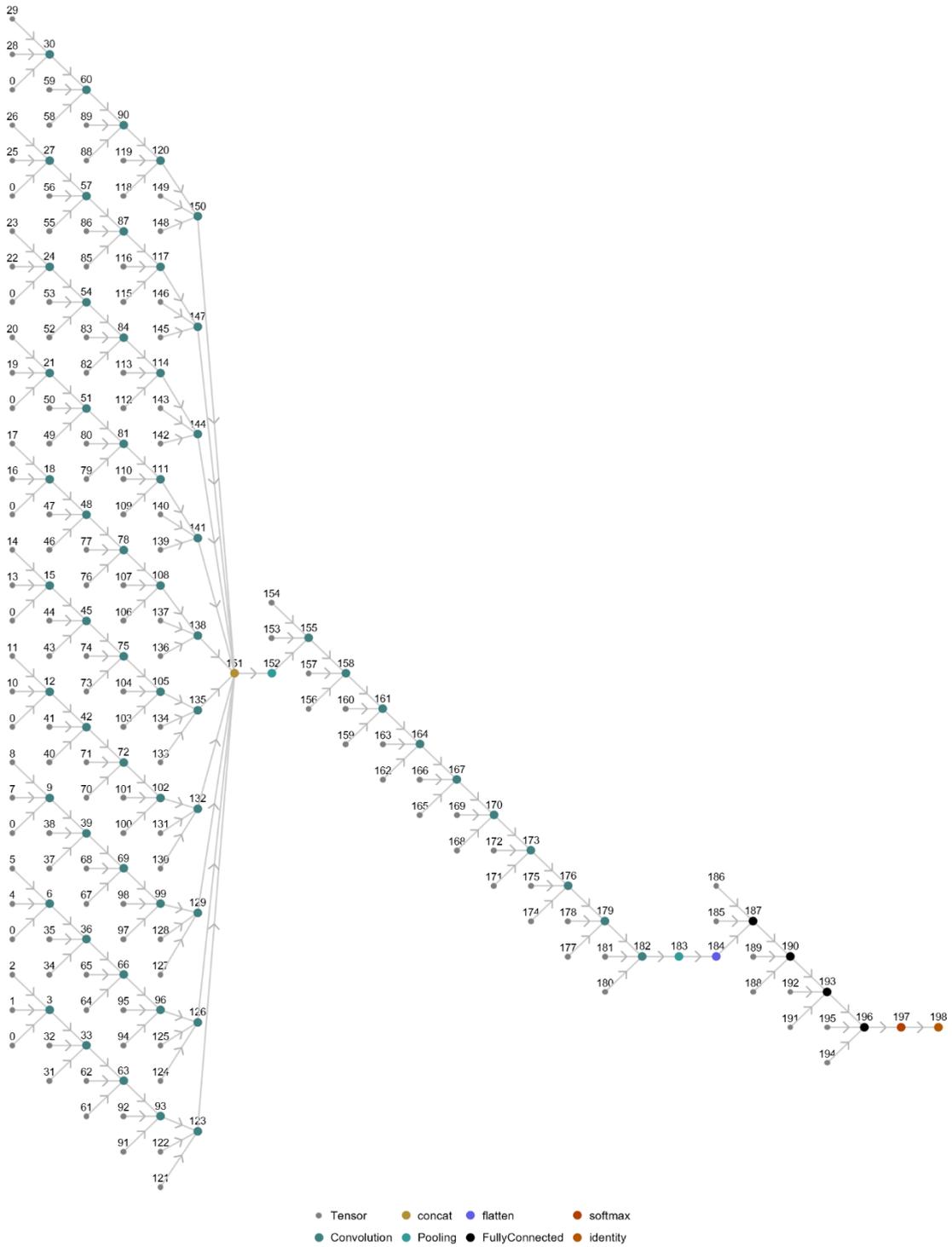

**Figure 2** The template of the CNN $N_a(\boldsymbol{x})$

For $a \in \{1,2,3,4,6,8,12,16\}$, $n = \#\boldsymbol{x}$ (that is $n$ equal to the cardinality of $\boldsymbol{x}$), and $1 \leq n \leq 4$, each 1-D CNN $N_\alpha(\boldsymbol{x})$ consists of:

(1) The input layer, with $n$ nodes.
(2) Ten parallel arrays of five convolutional layers each. Every convolutional layer has $n$ output channels, and kernels of size 1x1.
(3) A catenating layer, which catenates the ten parallel arrays.
(4) A pooling layer, performing 1-D pooling with kernels of size 2x1.
(5) Ten convolutional net layers having $n$ output channels, and kernels of size 1x1.
(6) A pooling net layer, performing 1-D pooling with kernels of size 2x1.
(7) Four linear net layers with output vectors of decreasing size.
(8) A *softmax* net layer, normalizing the exponential of the output vector of the fourth linear layer.
(9) The output layer, which is a decoder with two nodes, classifying the input as *true* (out of control) or *false* (out of control).

For $a \in \{1,2,3,4,6,8,12,16\}$, each CNN $N_\alpha(\boldsymbol{x})$ was trained using the respective training set $\mathbf{T}_\alpha(n)$, and then tested on the datasets $\mathbf{D}(n,0,1)$ and $\mathbf{D}(n,k,\mu,\sigma)$, for each combination of $k$ and $n$, for $1 \leq n \leq 4$ and $1 \leq k \leq n$. The misclassification rate of the application of $N_\alpha(\boldsymbol{x})$ to the dataset $\mathbf{D}(n,0,1)$ equals its probability for false rejection, denoted as $P_{N,a}(n,\mu,\sigma)$. The correct classification rate of the application of $N_\alpha(\boldsymbol{x})$ to the dataset $\mathbf{D}(n,k,\mu,\sigma)$, for $|\mu| > 0$ and $\sigma > 1$, equals its probability for rejection, denoted as $P_{N,a}(n,k,\mu,\sigma)$ (see *Notation and Formalism*).

The thirty two trained CNNs are available at https://www.hcsl.com/Supplements/hcsltr21s1.zip, in Apache MXNet, ONNX, and Wolfram Language frameworks formats, as described in *Supplement 1*.

### 3.3.2. Statistical QC Functions

Let $m$ the expected mean and $s$ the standard deviation of the QC measurements $x_i$ of a measurement process when the process is in control. As $S_a(\boldsymbol{x}; l, m, s)$ is denoted a statistical QC function, applied to a n tuple $\boldsymbol{x}$ of QC measurements $x_i$, which is true if $|x_i - m| > l\,s$, for any of them. Furthermore, let:

$$S_\alpha(\boldsymbol{x};\ l) = S_a(\boldsymbol{x}; l, 0,1)$$

For $a \in \{1,2,3,4,6,8,12,16\}$, the probability for rejection of the application of $S_a(\boldsymbol{x}; l)$ to the dataset $D(n,0,1)$ is denoted as $P_{S,a}(n,0,1)$. The decision limit $l$ of $S_a(\boldsymbol{x}; l)$ *is* defined (see *Formalism and Notation*) so that:

$$P_{S,a}(n,0,1) \ = \ P_{N,a}(n,0,1)$$

For $a \in \{1,2,3,4,6,8,12,16\}$, $1 \leq n \leq 4$ and $1 \leq k \leq n$, the probability for rejection of the application of $S_a(\boldsymbol{x}; l)$ to the dataset $D(n,k,\mu,\sigma)$ is denoted as $P_{S,a}(n,k,\mu,\sigma)$.

The calculated values of $l$ are presented in Table 5.

**Table 5:** The probabilities for false rejection $P_{N,a}(n,0,1)$ and the calculated values of the decision limits $l$

| CNN | $n$ | $\alpha$ | $P_{N,a}(n,0,1)$ | $l$ |
|---|---|---|---|---|
| $N_a(x_1)$ | 1 | 1 | 0.127240 | 1.525077 |
| | | 2 | 0.034726 | 2.111536 |
| | | 3 | 0.027421 | 2.205468 |
| | | 4 | 0.023750 | 2.261149 |
| | | 6 | 0.012258 | 2.504643 |
| | | 8 | 0.008197 | 2.643835 |
| | | 12 | 0.005178 | 2.795778 |
| | | 16 | 0.004186 | 2.863775 |
| $N_a(x_1, x_2)$ | 2 | 1 | 0.114545 | 1.888090 |
| | | 2 | 0.046078 | 2.268308 |
| | | 3 | 0.031890 | 2.407227 |
| | | 4 | 0.008733 | 2.849716 |
| | | 6 | 0.016204 | 2.646416 |
| | | 8 | 0.009049 | 2.838333 |
| | | 12 | 0.003311 | 3.145682 |
| | | 16 | 0.002574 | 3.218706 |
| $N_a(x_1, x_2, x_3)$ | 3 | 1 | 0.120814 | 2.033406 |
| | | 2 | 0.041528 | 2.456272 |
| | | 3 | 0.024233 | 2.646056 |
| | | 4 | 0.007836 | 3.009274 |
| | | 6 | 0.009234 | 2.958895 |
| | | 8 | 0.005146 | 3.135029 |
| | | 12 | 0.005211 | 3.131339 |
| | | 16 | 0.002915 | 3.298332 |
| $N_a(x_1, x_2, x_3, x_4)$ | 4 | 1 | 0.101483 | 2.220312 |
| | | 2 | 0.040159 | 2.56916 |
| | | 3 | 0.029011 | 2.681327 |
| | | 4 | 0.004909 | 3.231921 |
| | | 6 | 0.010381 | 3.010814 |
| | | 8 | 0.005599 | 3.194108 |
| | | 12 | 0.003848 | 3.301041 |

### 3.4. Computer System and Software

The datasets were created and the CNNs were designed, trained, and tested using Wolfram Mathematica Ver. 13.0, Wolfram Research, Inc., Champaign, IL, USA, on a computer system with an Intel Core i9-11900K CPU, 128 GB RAM, a NVIDIA GeForce RTX 3080 Ti GPU, under Microsoft Windows 11 Professional operating system.

## 4. Results

The plots of the probabilities for rejection $P_{N,a}(n,k,\mu,\sigma)$ and $P_{S,a}(n,k,\mu,\sigma)$ of the application of the QC functions to the respective datasets, their differences $\Delta P_a(n,k,\mu,\sigma)$, and their relative differences $\Delta P_{R,a}(n,k,\mu,\sigma)$ are presented in *Appendix*. The probabilities for rejection $P_{N,a}(n,k,\mu,\sigma)$ and $P_{S,a}(n,k,\mu,\sigma)$, their differences $\Delta P_a(n,k,\mu,\sigma)$ with the respective statistical significance ($p$-values), and their relative differences $\Delta P_{R,a}(n,k,\mu,\sigma)$ are presented in detail in *Supplement 2*.

As examples, Figure 2 presents the probabilities for rejection $P_{N,a}(2,2,\mu,\sigma)$ of the application of the CNN functions to the respective datasets, while Figure 3 presents the probabilities for rejection $P_{N,6}(3,3,\mu,\sigma)$ and $P_{S,6}(3,3,\mu,\sigma)$ of the application of the QC functions to the respective datasets.

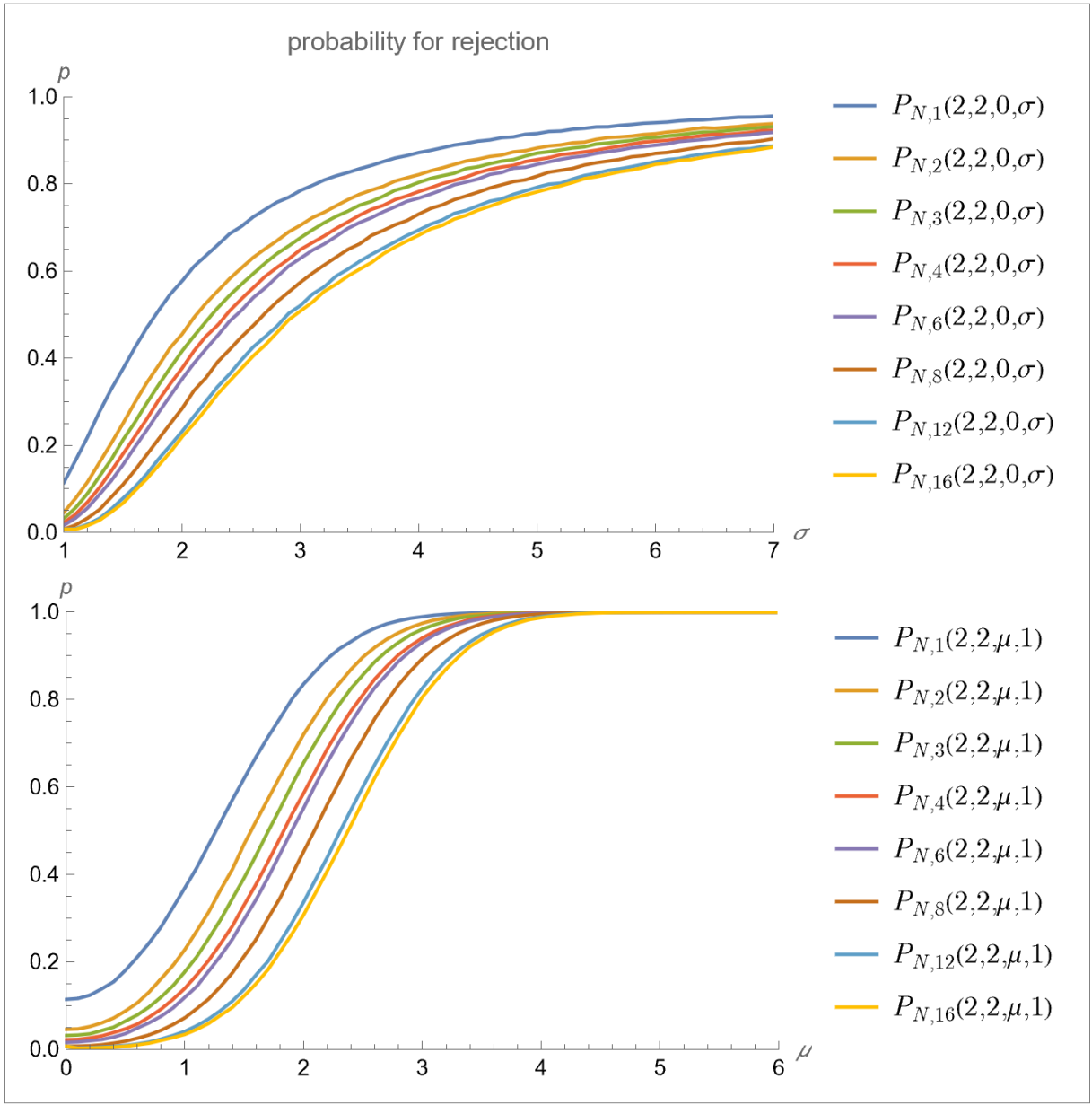

**Figure 2** The probabilities for rejection of the CNN $N_a(x_1, x_2)$, for $a \in \{1,2,3,4,6,8,12,16\}$, when applied to two QC measurements, distributed as $\mathcal{N}(0, \sigma^2)$ (upper plot) or as $\mathcal{N}(\mu, 1)$ (lower plot)

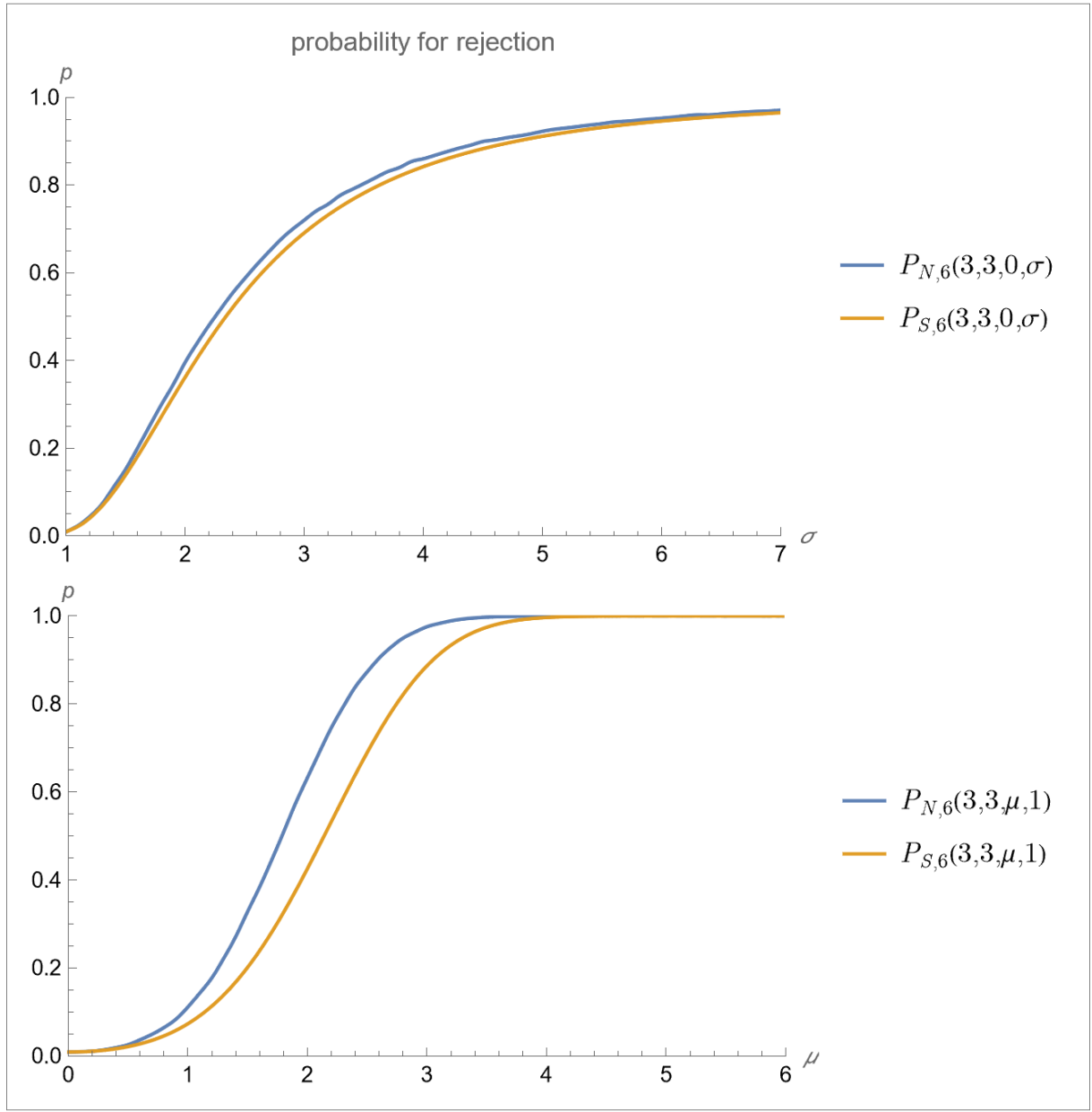

**Figure 3**. The probabilities for rejection of the CNN $N_6(x_1, x_2, x_3)$ when applied to three QC measurements, distributed as $\mathcal{N}(0, \sigma^2)$ *(upper plot)* or as $\mathcal{N}(\mu, 1)$ *(lower plot)*

Likewise, Figures 4 and 5 present the differences $\Delta P_8(n, k, \mu, \sigma)$ and the relative differences $\Delta P_{R,8}(n, k, \mu, \sigma)$ of the probabilities for rejection $P_{N,8}(n, k, \mu, \sigma)$ and $P_{S,8}(n, k, \mu, \sigma)$ of the application of the QC functions to the respective datasets.

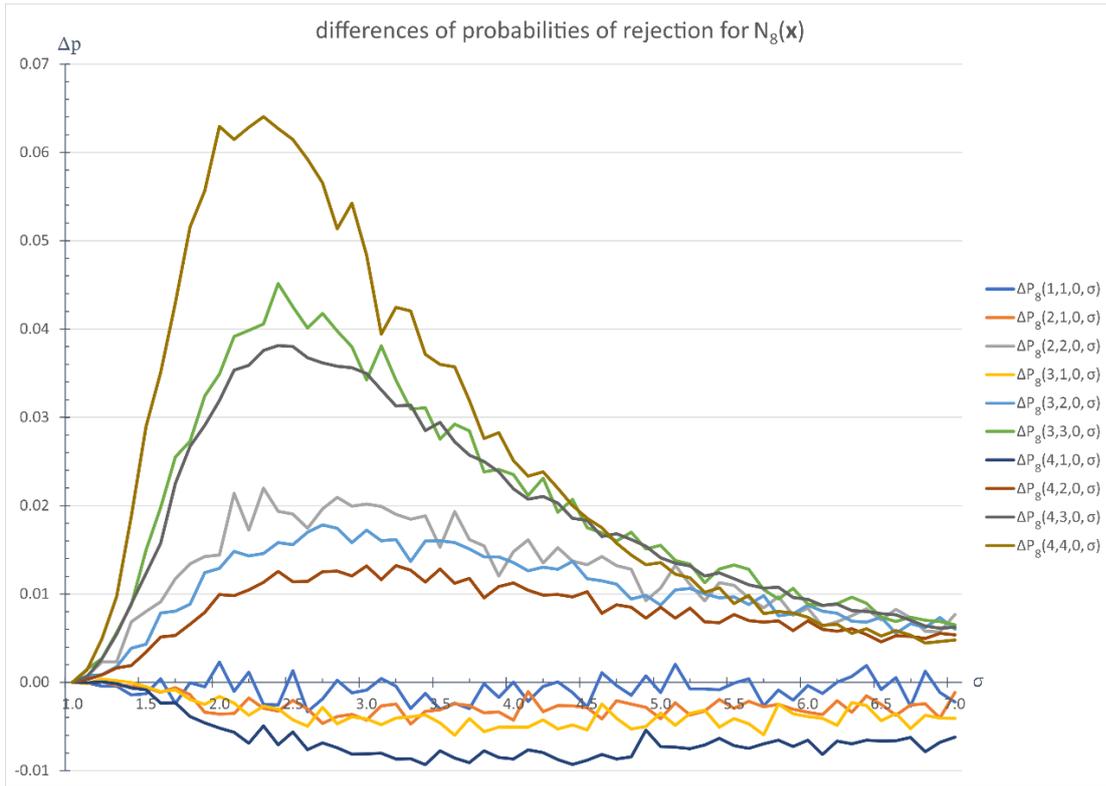

**Figure 4** The differences between the probabilities for rejection of $N_8(x)$ and the respective statistical QC functions vs $\sigma$

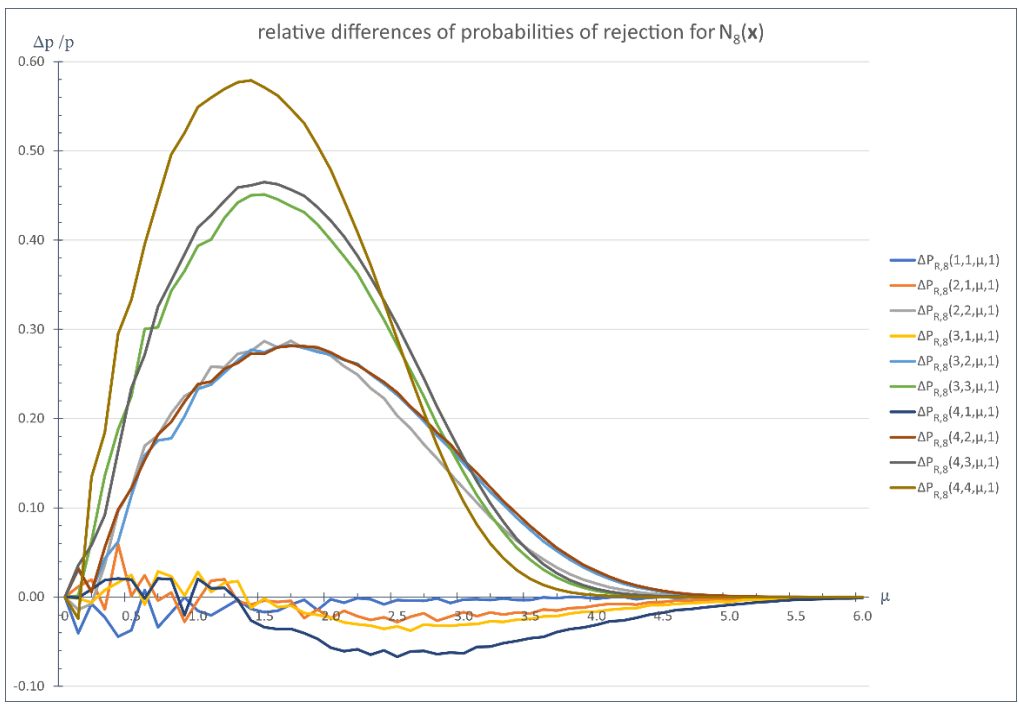

**Figure 5** The relative differences between the probabilities for rejection of $N_8(x)$ and the respective statistical QC functions vs $\mu$

The results show that for $a \in \{1,2,3,4,6,8,12,16\}$, and $1 < n \leq 4$, $1 < k \leq n$, $0.2 < |\mu| \leq 6.0$, and $1.0 \leq \sigma \leq 7.0$ (see *Supplement 2* and Figures 1-80 and 147-178 *of Appendix*):

$$\Delta P_a(n, k, \mu, \sigma) > 0$$

For $a \in \{1,2,3,4,6,8,12,16\}$, $1 < n \leq 4$, $1 < k \leq n$, $0.2 < |\mu| \leq 6.0$, and $1.2 < \sigma \leq 7.0$, the differences are statistically significant ($p < 0.01$), except for some differences where $\Delta P_a(n, k, \mu, \sigma) < 0.002$ (see *Supplement 2*).

For $a \in \{1,2,3,4,6,8,12,16\}$, and $1 < n \leq 4$, most of the differences $\Delta P_a(n, 1, \mu, \sigma)$ are negative, with $-0.03 < \Delta P_a(n, 1, \mu, \sigma) < 0$ (see *Supplement 2*).

For $a \in \{1,2,3,4,6,8,12,16\}$, and $1 < n \leq 4$, and $1 < k \leq n$ and $|\mu| < 0.2$ there are some negative differences $\Delta P_a(n, k, \mu, \sigma)$, however $|\Delta P_a(n, k, \mu, \sigma)| < 0.0008$ (see Table 6).

**Table 6**

The negative $\Delta P_a(n, k, \mu, \sigma)$ and the respective $p$-values for $k > 1$

|  | value | $p$-value |
|---|---|---|
| $\Delta P_1(2,2,0.1,1)$ | −0.000329 | 0.185258 |
| $\Delta P_1(2,2,0.2,1)$ | −0.000086 | 0.414529 |
| $\Delta P_1(3,2,0.1,1)$ | −0.000798 | 0.016306 |
| $\Delta P_1(3,3,0.1,1)$ | −0.000029 | 0.474765 |
| $\Delta P_1(4,3,0.1,1)$ | −0.000252 | 0.235606 |
| $\Delta P_1(4,4,0.1,1)$ | −0.000124 | 0.365589 |
| $\Delta P_2(2,2,0.1,1)$ | −0.000499 | 0.018740 |
| $\Delta P_3(2,2,0.2,1)$ | −0.000474 | 0.011969 |
| $\Delta P_3(3,2,0.2,1)$ | −0.000171 | 0.165979 |
| $\Delta P_4(2,2,0.1,1)$ | −0.000138 | 0.214109 |
| $\Delta P_4(3,2,0.1,1)$ | −0.000247 | 0.035223 |
| $\Delta P_6(2,2,0.1,1)$ | −0.000176 | 0.053050 |
| $\Delta P_6(4,2,0.1,1)$ | −0.000076 | 0.258990 |
| $\Delta P_8(2,2,0.1,1)$ | −0.000106 | 0.148377 |
| $\Delta P_8(2,2,0.1,1)$ | −0.000066 | 0.278841 |
| $\Delta P_8(4,4,0.2,1)$ | −0.000140 | 0.056779 |
| $\Delta P_{12}(2,2,0,1.1)$ | −0.000213 | 0.020407 |
| $\Delta P_{12}(2,2,0.1,1)$ | −0.000130 | 0.027128 |
| $\Delta P_{12}(2,2,0.2,1)$ | −0.000189 | 0.004022 |
| $\Delta P_{12}(4,2,0.1,1)$ | −0.000001 | 0.501188 |
| $\Delta P_{16}(4,2,0,1.1)$ | −0.000007 | 0.477516 |
| $\Delta P_{16}(3,2,0.1,1)$ | −0.000176 | 0.053050 |
| $\Delta P_{16}(4,2,0.2,1)$ | −0.000076 | 0.258990 |

Furthermore (see *Supplement 2*):

(1) For $a \in \{1,2,3,4,6,8,12,16\}$, $1 < n \leq 4$, $1 \leq k \leq n$, $0 \leq |\mu_h| < |\mu_j| \leq 6.0$ and $1.0 \leq \sigma \leq 7.0$ (see Figures 1-146 of *Appendix)*:

$$P_{N,\alpha}(n, k, \mu_h, \sigma) < P_{N,\alpha}(n, k, \mu_j, \sigma)$$

(2) For $a \in \{1,2,3,4,6,8,12,16\}, 1 < n \leq 4, 1 \leq k \leq n, 0 \leq |\mu| \leq 6.0, 1.0 \leq \sigma_h < \sigma_j \leq 7.0$ (see Figures 1-146 of *Appendix*):

$$P_{N,a}(n, k, \mu, \sigma_h) < P_{N,a}(n, k, \mu, \sigma_j)$$

(3) For $a \in \{1,2,3,4,6,8,12,16\}, 1 < n \leq 4, 1 \leq k_h < k_j \leq n, 0 \leq |\mu| \leq 6.0$, and $1.0 \leq \sigma \leq 7.0$ (see Figures 81-104 of *Appendix*):

$$P_{N,a}(n, k_h, \mu, \sigma) < P_{N,a}(n, k_j, \mu, \sigma)$$

(4) For $a_h \in \{1,2,3,4,6,8,12\}$, $a_j \in \{2,3,4,6,8,12,16\}$, $a_h < a_j$, $1 \leq n \leq 4$, $1 \leq k \leq n$, $0 < |\mu| \leq 6.0$, and $1.0 < \sigma \leq 7.0$ (see Figures 105-114 of *Appendix*):

$$P_{N, a_h}(n, k, \mu, \sigma) > P_{N, a_j}(n, k, \mu, \sigma)$$

(5) For $a \in \{1,2,3,4,6,8,12,16\}, 1 \leq n_h < n_j \leq 4, 0 < |\mu| \leq 6.0$, and $1.0 < \sigma \leq 7.0$ (see Figures 115-122 of *Appendix*):

$$P_{N,a}(n_h, n_h, \mu, \sigma) < P_{N,a}(n_j, n_j, \mu, \sigma)$$

(6) For $a \in \{1,2,3,4,6,8,12,16\}, 1 \leq n_h < n_j \leq 4, 1 \leq k \leq n_h, 0 < |\mu| \leq 6.0$, and $1.0 < \sigma \leq 7.0$ (see Figures 123-146 of *Appendix*):

$$P_{N,a}(n_j, k, \mu, \sigma) < P_{N,a}(n_h, k, \mu, \sigma)$$

# 5. Illustrative example

The performance of the designed CNNs was compared to the performance of the respective statistical functions with equal probabilities for false rejection, for the detection of the critical systematic error of a clinical chemistry assay.

Briefly, the coefficient of variation and the bias of the uricase and peroxidase method uric acid assay of the AU5800 (Beckman, Coulter, Brea, Ca) analyser, were estimated as equal to 1.44% and 1.46% respectively (Xia et al. 2018). The optimal total allowable analytical error for the uric acid, based on the biological variation is equal to 5.98% (Fraser et al. 1997). Therefore, the standardized critical random and systematic errors of the assay, as defined by Linnet (Linnet 1989), are equal to 3.33 and 2.67 respectively

Tables 7 and 8 show the probabilities for critical error detection of the QC functions $N_a(x)$ and the respective statistical QC functions $S_a(x; l)$, their differences and relative differences, as well as their probabilities for false rejection, for $2 \leq n \leq 4$, $2 \leq k \leq n$, and $a \in \{1,2,3,4,6,8,12,16\}$. All the differences are positive and statistically significant ($p < 10^{-6}$).

The selection of the NN QC function to be applied depends upon the risk of critical error, and the cost of the control measurements and the false rejections. . Although the costs can be calculated, the risk can only be approximated. Hence, an applicable QC function, with a relatively high probability for critical random error detection ($> 0.75$), a high probability for critical systematic error detection ($> 0.9$), and a low probability for false rejection ($< 0.01$), could be the $N_6(x_1, x_2, x_3)$ (see Figure 4), as:

**Table 7:** The probabilities for critical systematic error detection of the QC functions $N_a(\boldsymbol{x})$ and the respective statistical QC functions $S_a(\boldsymbol{x}; l)$, their differences and relative differences, as well as their probabilities for false rejection, for 2≤$n$≤4, 2≤$k$≤$n$, $a \in \{1,2,3,4,6,8,12,16\}$, and $\mu_c$ = 2.67

| $n$ | $k$ | $a$ | $P_{N,a}(n,0,1)$ | $l$ | $P_{N,a}(n,k,\mu_c,1)$ | $P_{S,a}(n,k,\mu_c,1)$ | $\Delta P_a(n,k,\mu_c,1)$ | $\Delta P_{R,a}(n,k,\mu_c,1)$ |
|---|---|---|---|---|---|---|---|---|
| 2 | 2 | 1 | 0.114545 | 1.888090 | 0.970429 | 0.952854 | 0.017575 | 0.018444 |
| 2 | 2 | 2 | 0.046078 | 2.268308 | 0.932883 | 0.881695 | 0.051188 | 0.058056 |
| 2 | 2 | 3 | 0.031890 | 2.407227 | 0.904121 | 0.842897 | 0.061224 | 0.072635 |
| 2 | 2 | 4 | 0.022235 | 2.537022 | 0.867865 | 0.800097 | 0.067768 | 0.084700 |
| 2 | 2 | 6 | 0.016204 | 2.646416 | 0.849032 | 0.759319 | 0.089713 | 0.118149 |
| 2 | 2 | 8 | 0.007229 | 2.909426 | 0.785067 | 0.646436 | 0.138631 | 0.214454 |
| 2 | 2 | 12 | 0.003311 | 3.145682 | 0.686446 | 0.533716 | 0.152730 | 0.286163 |
| 2 | 2 | 16 | 0.002574 | 3.218706 | 0.656111 | 0.498175 | 0.157936 | 0.317030 |
| 3 | 2 | 1 | 0.120814 | 2.033406 | 0.957183 | 0.934143 | 0.023040 | 0.024664 |
| 3 | 2 | 2 | 0.041528 | 2.456273 | 0.886290 | 0.829882 | 0.056408 | 0.067971 |
| 3 | 2 | 3 | 0.024233 | 2.646056 | 0.845290 | 0.761419 | 0.083871 | 0.110151 |
| 3 | 2 | 4 | 0.014565 | 2.814931 | 0.794489 | 0.690580 | 0.103910 | 0.150467 |
| 3 | 2 | 6 | 0.009234 | 2.958895 | 0.733737 | 0.624573 | 0.109164 | 0.174782 |
| 3 | 2 | 8 | 0.005146 | 3.135030 | 0.675917 | 0.539690 | 0.136226 | 0.252416 |
| 3 | 2 | 12 | 0.005211 | 3.131339 | 0.666497 | 0.541492 | 0.125005 | 0.230854 |
| 3 | 2 | 16 | 0.002915 | 3.298332 | 0.600938 | 0.460144 | 0.140794 | 0.305979 |
| 3 | 3 | 1 | 0.120814 | 2.033406 | 0.994133 | 0.981975 | 0.012158 | 0.012381 |
| 3 | 3 | 2 | 0.041528 | 2.456273 | 0.977577 | 0.928330 | 0.049247 | 0.053049 |
| 3 | 3 | 3 | 0.024233 | 2.646056 | 0.964508 | 0.882028 | 0.082480 | 0.093512 |
| 3 | 3 | 4 | 0.014565 | 2.814931 | 0.950179 | 0.826616 | 0.123563 | 0.149481 |
| 3 | 3 | 6 | 0.009234 | 2.958895 | 0.922430 | 0.768898 | 0.153532 | 0.199677 |
| 3 | 3 | 8 | 0.005146 | 3.135030 | 0.896336 | 0.686891 | 0.209445 | 0.304917 |
| 3 | 3 | 12 | 0.005211 | 3.131339 | 0.887244 | 0.688717 | 0.198527 | 0.288256 |
| 3 | 3 | 16 | 0.002915 | 3.298332 | 0.850374 | 0.602762 | 0.247612 | 0.410796 |
| 4 | 2 | 1 | 0.101483 | 2.220312 | 0.933884 | 0.898972 | 0.034912 | 0.038836 |
| 4 | 2 | 2 | 0.040159 | 2.569160 | 0.854466 | 0.792838 | 0.061629 | 0.077732 |
| 4 | 2 | 3 | 0.029011 | 2.681327 | 0.828322 | 0.749180 | 0.079142 | 0.105638 |
| 4 | 2 | 4 | 0.014786 | 2.901210 | 0.756794 | 0.652813 | 0.103981 | 0.159281 |
| 4 | 2 | 6 | 0.010381 | 3.010815 | 0.713196 | 0.600919 | 0.112276 | 0.186841 |
| 4 | 2 | 8 | 0.005599 | 3.194108 | 0.642783 | 0.511516 | 0.131267 | 0.256624 |
| 4 | 2 | 12 | 0.003848 | 3.301042 | 0.599206 | 0.459357 | 0.139849 | 0.304446 |
| 4 | 2 | 16 | 0.004080 | 3.284519 | 0.588973 | 0.467366 | 0.121607 | 0.260196 |
| 4 | 3 | 1 | 0.101483 | 2.220312 | 0.990978 | 0.966123 | 0.024854 | 0.025726 |
| 4 | 3 | 2 | 0.040159 | 2.569160 | 0.968567 | 0.903758 | 0.064810 | 0.071711 |
| 4 | 3 | 3 | 0.029011 | 2.681327 | 0.963287 | 0.872522 | 0.090765 | 0.104026 |
| 4 | 3 | 4 | 0.014786 | 2.901210 | 0.936527 | 0.793899 | 0.142628 | 0.179655 |
| 4 | 3 | 6 | 0.010381 | 3.010815 | 0.910047 | 0.746571 | 0.163476 | 0.218969 |
| 4 | 3 | 8 | 0.005599 | 3.194108 | 0.881661 | 0.657630 | 0.224031 | 0.340664 |
| 4 | 3 | 12 | 0.003848 | 3.301042 | 0.852930 | 0.601707 | 0.251223 | 0.417518 |
| 4 | 3 | 16 | 0.004080 | 3.284519 | 0.843337 | 0.610479 | 0.232858 | 0.381435 |
| 4 | 4 | 1 | 0.101483 | 2.220312 | 0.999255 | 0.988641 | 0.010614 | 0.010736 |
| 4 | 4 | 2 | 0.040159 | 2.569160 | 0.995497 | 0.955288 | 0.040209 | 0.042091 |
| 4 | 4 | 3 | 0.029011 | 2.681327 | 0.994943 | 0.935210 | 0.059733 | 0.063871 |
| 4 | 4 | 4 | 0.014786 | 2.901210 | 0.988630 | 0.877652 | 0.110978 | 0.126449 |
| 4 | 4 | 6 | 0.010381 | 3.010815 | 0.979138 | 0.839064 | 0.140074 | 0.166940 |
| 4 | 4 | 8 | 0.005599 | 3.194108 | 0.973312 | 0.760039 | 0.213273 | 0.280607 |
| 4 | 4 | 12 | 0.003848 | 3.301042 | 0.963679 | 0.706576 | 0.257103 | 0.363871 |
| 4 | 4 | 16 | 0.004080 | 3.284519 | 0.957798 | 0.715139 | 0.017575 | 0.339318 |

**Table 8:** The probabilities for critical random error detection of the QC functions $N_a(\mathbf{x})$ and the respective statistical QC functions $S_a(\mathbf{x}; l)$, their differences and relative differences, as well as their probabilities for false rejection, for 2≤n≤4, 2≤k≤n, a ∈ {1,2,3,4,6,8,12,16}, and $\sigma_c$ = 3.33

| n | k | a | $P_{N,a}(n,0,1)$ | $l$ | $P_{N,a}(n,k,0,\sigma_c)$ | $P_{S,a}(n,k,0,\sigma_c)$ | $\Delta P_a(n,k,0,\sigma_c)$ | $\Delta P_{R,a}(n,k,0,\sigma_c)$ |
|---|---|---|---|---|---|---|---|---|
| 2 | 2 | 1 | 0.114545 | 1.888090 | 0.820710 | 0.815717 | 0.004993 | 0.006121 |
| 2 | 2 | 2 | 0.046078 | 2.268308 | 0.753781 | 0.745744 | 0.008038 | 0.010778 |
| 2 | 2 | 3 | 0.031890 | 2.407227 | 0.728892 | 0.718832 | 0.010061 | 0.013996 |
| 2 | 2 | 4 | 0.022235 | 2.537022 | 0.702068 | 0.693237 | 0.008831 | 0.012739 |
| 2 | 2 | 6 | 0.016204 | 2.646416 | 0.685062 | 0.671415 | 0.013647 | 0.020325 |
| 2 | 2 | 8 | 0.007229 | 2.909426 | 0.637138 | 0.618423 | 0.018715 | 0.030262 |
| 2 | 2 | 12 | 0.003311 | 3.145682 | 0.593274 | 0.570763 | 0.022511 | 0.039441 |
| 2 | 2 | 16 | 0.002574 | 3.218706 | 0.575099 | 0.556117 | 0.018982 | 0.034133 |
| 3 | 2 | 1 | 0.120814 | 2.033406 | 0.804366 | 0.798559 | 0.005807 | 0.007272 |
| 3 | 2 | 2 | 0.041528 | 2.456273 | 0.722860 | 0.713287 | 0.009573 | 0.013420 |
| 3 | 2 | 3 | 0.024233 | 2.646056 | 0.683709 | 0.674162 | 0.009546 | 0.014160 |
| 3 | 2 | 4 | 0.014565 | 2.814931 | 0.651198 | 0.639280 | 0.011919 | 0.018644 |
| 3 | 2 | 6 | 0.009234 | 2.958895 | 0.622393 | 0.609635 | 0.012758 | 0.020927 |
| 3 | 2 | 8 | 0.005146 | 3.135030 | 0.587840 | 0.573638 | 0.014202 | 0.024758 |
| 3 | 2 | 12 | 0.005211 | 3.131339 | 0.588805 | 0.574388 | 0.014417 | 0.025100 |
| 3 | 2 | 16 | 0.002915 | 3.298332 | 0.557886 | 0.540675 | 0.017211 | 0.031832 |
| 3 | 3 | 1 | 0.120814 | 2.033406 | 0.909988 | 0.903577 | 0.006410 | 0.007095 |
| 3 | 3 | 2 | 0.041528 | 2.456273 | 0.857034 | 0.843187 | 0.013847 | 0.016422 |
| 3 | 3 | 3 | 0.024233 | 2.646056 | 0.829792 | 0.811709 | 0.018083 | 0.022278 |
| 3 | 3 | 4 | 0.014565 | 2.814931 | 0.805116 | 0.781756 | 0.023360 | 0.029881 |
| 3 | 3 | 6 | 0.009234 | 2.958895 | 0.780677 | 0.754969 | 0.025708 | 0.034052 |
| 3 | 3 | 8 | 0.005146 | 3.135030 | 0.751832 | 0.720882 | 0.030951 | 0.042934 |
| 3 | 3 | 12 | 0.005211 | 3.131339 | 0.750725 | 0.721609 | 0.029117 | 0.040349 |
| 3 | 3 | 16 | 0.002915 | 3.298332 | 0.724862 | 0.688244 | 0.036618 | 0.053205 |
| 4 | 2 | 1 | 0.101483 | 2.220312 | 0.772341 | 0.767670 | 0.004671 | 0.006085 |
| 4 | 2 | 2 | 0.040159 | 2.569160 | 0.701798 | 0.693199 | 0.008599 | 0.012404 |
| 4 | 2 | 3 | 0.029011 | 2.681327 | 0.677069 | 0.669318 | 0.007751 | 0.011581 |
| 4 | 2 | 4 | 0.014786 | 2.901210 | 0.632607 | 0.622903 | 0.009705 | 0.015580 |
| 4 | 2 | 6 | 0.010381 | 3.010815 | 0.610869 | 0.600031 | 0.010838 | 0.018062 |
| 4 | 2 | 8 | 0.005599 | 3.194108 | 0.574481 | 0.562274 | 0.012207 | 0.021710 |
| 4 | 2 | 12 | 0.003848 | 3.301042 | 0.555940 | 0.540575 | 0.015365 | 0.028424 |
| 4 | 2 | 16 | 0.004080 | 3.284519 | 0.552367 | 0.543910 | 0.008457 | 0.015548 |
| 4 | 3 | 1 | 0.101483 | 2.220312 | 0.890943 | 0.881861 | 0.009083 | 0.010299 |
| 4 | 3 | 2 | 0.040159 | 2.569160 | 0.842058 | 0.826546 | 0.015513 | 0.018768 |
| 4 | 3 | 3 | 0.029011 | 2.681327 | 0.825227 | 0.807022 | 0.018205 | 0.022558 |
| 4 | 3 | 4 | 0.014786 | 2.901210 | 0.790938 | 0.766700 | 0.024238 | 0.031614 |
| 4 | 3 | 6 | 0.010381 | 3.010815 | 0.772276 | 0.745724 | 0.026553 | 0.035607 |
| 4 | 3 | 8 | 0.005599 | 3.194108 | 0.740107 | 0.709583 | 0.030524 | 0.043017 |
| 4 | 3 | 12 | 0.003848 | 3.301042 | 0.724164 | 0.687997 | 0.036167 | 0.052569 |
| 4 | 3 | 16 | 0.004080 | 3.284519 | 0.719965 | 0.691352 | 0.028612 | 0.041386 |
| 4 | 4 | 1 | 0.101483 | 2.220312 | 0.949559 | 0.939926 | 0.009632 | 0.010248 |
| 4 | 4 | 2 | 0.040159 | 2.569160 | 0.917611 | 0.901935 | 0.015676 | 0.017381 |
| 4 | 4 | 3 | 0.029011 | 2.681327 | 0.907597 | 0.887382 | 0.020215 | 0.022780 |
| 4 | 4 | 4 | 0.014786 | 2.901210 | 0.882921 | 0.855663 | 0.027258 | 0.031856 |
| 4 | 4 | 6 | 0.010381 | 3.010815 | 0.870035 | 0.838346 | 0.031689 | 0.037799 |
| 4 | 4 | 8 | 0.005599 | 3.194108 | 0.847989 | 0.807317 | 0.040672 | 0.050379 |
| 4 | 4 | 12 | 0.003848 | 3.301042 | 0.834971 | 0.788113 | 0.046858 | 0.059456 |
| 4 | 4 | 16 | 0.004080 | 3.284519 | 0.830135 | 0.791130 | 0.039005 | 0.049303 |

1. $P_{N,6}(3,0,1) = P_{S,6}(3,0,1) = 0.009234$

2. $P_{N,6}(3,3,0,3.3) = 0.780677$

3. $P_{S,6}(3,3,0,3.3) = 0.754969$

4. $\Delta P_6(3,3,0,3.3) = 0.025708$

5. $\Delta P_{R,6}(3,3,0,3.3) = 0.034052$

6. $P_{N,6}(3,3,2.67,1) = 0.922430$

7. $P_{S,6}(3,3,2.67,1) = 0.768898$

8. $\Delta P_6(3,3,2.67,1) = 0.153532$

9. $\Delta P_{R,6}(3,3,2.67,1) = 0.199677$

Therefore, the NN QC function $N_6(x_1, x_2, x_3)$ outperforms significantly the statistical $S(x_1, x_2, x_3; 2.958895)$, although they have equal probabilities for false rejection.

# 5. Discussion

The complexity of statistical QC has increased significantly since the 1920s, when it was introduced into industry by Shewhart (Shewhart 1930), especially after the development of low cost digital computers on integrated circuits with substantial processing power. Since 1989, the wide availability of more processing power led to the development of NNs and later of DNNs, including CNNs, and their application to statistical QC. Deep learning transforms low level representations to a higher order abstract one, extracting features from raw data in multiple stages. CNNs are relatively simple and compact DNNs, with exemplary learning ability and performance (Khan et al. 2020). Particularly, 1-D CNNs simple enough to be implemented in low cost computer systems have been successfully used to process 1-D signals (Kiranyaz et al. 2021; Kiranyaz, Ince, and Gabbouj 2016), including control charts (Xu et al. 2019).

Although the relationship among a QC procedure, the reliability of a system, the risk of nonconformity to quality specifications and the QC related cost is complex (Aristides T. Hatjimihail 2009), it is obvious that cost increases with the number of the QC measurements required for the safe detection of nonconformity. The application of statistical QC functions to very small samples of QC measurements has been extensively studied (Duncan 1986) (Montgomery 2019). However, NNs QC functions have only been applied to $n$-tuples of QC measurements with $n > 5$. Consequently, it is meaningful to explore their application to $n$-tuples of QC measurements for $1 \leq n \leq 4$.

For this project, there were designed 4 1-D CNNs with 1x1 kernels, applicable to 1 to 4 QC measurements respectively. As the directly cost related probability for false rejection is a key index of the performance of a QC function, each of the 4 1-D CNNs was trained with eight training datasets, with different ratios of $n$-tuples of random simulated QC measurements distributed as $\mathcal{N}(0,1)$ (see Tables 1-4). As a result, thirty two trained 1-D CNNs were obtained, which were applied to testing datasets containing all the possible combinations of measurements. The range of their probabilities for false rejection was from 0.002574 to 0.127400 (see Table 5). Another approach for obtaining varying probabilities for false rejection could be weighing the output of the CNN, with differing penalization of the probability for false rejection.

The comparison of the trained 1-D CNNs with the respective statistical QC functions with decision limits shown on Table 5, shows that the designed CNNs outperform the respective statistical QC functions, for $2 \leq n \leq 4$ and $2 \leq k \leq n$, that is when the samples they are applied to include at least two QC measurements out of control, distributed either as $\mathcal{N}(\mu, 1)$, for $0.2 < |\mu| \leq 6.0$, or as $\mathcal{N}(0, \sigma^2)$, for $1.0 < \sigma \leq 7.0$. However, statistical QC functions outperform the 1-D CNNs, for $1 \leq n \leq 4$ and $k = 1$, that is when the samples they are applied to include only one control measurement distributed as either as $\mathcal{N}(\mu, 1)$, for $0 < |\mu| \leq 6.0$, or as $\mathcal{N}(0, \sigma^2)$, for $1.0 < \sigma \leq 7.0$, but then $|\Delta P_a(n, 1, \mu, \sigma)| < 0.028$ (see Figures 1-85 and 147-178 of *Appendix* and *Supplement 2*).

Besides, the probability for rejection of the 1-D CNNs increases with the sample size, when all the QC measurements of the sample are out of control (see Figures 115-122 of *Appendix* and *Supplement 2*), while decreases with the sample size when the numbers of the out of control measurements of each sample are equal (see Figures 123-146 of *Appendix* and *Supplement 2*).

Limitations of this project, that could be improved by further research, are the following:

(7) The assumption of normality of either the measurements or their applicable transforms (Gillard 2012; Atkinson, Riani, and Corbellini 2021; Sakia 1992; Box and Cox 1964), however, this is usually valid. Furthermore, the four designed CNNs can be trained and tested using datasets of any parametric or nonparametric distribution.

(8) The arbitrary selection of the different ratios of $n$-tuples of random simulated QC measurements distributed as $\mathcal{N}(0,1)$ of the simulated datasets, for obtaining varying probabilities for false rejection. Another approach could be weighing the output of the CNNs with differing penalization of the probability for false rejection.

(9) The arbitrary design of the four CNNs. As the processing power of the computer systems is increasing exponentially, efficient methods of optimal design of NNs will be developed, including evolutionary algorithms (Ding et al. 2013).

The thirty two trained CNNs can be used as they are or they can be retrained with different datasets. Their models, including their weights and biases, are available in various formats, to be implemented in any neural networks' framework (see *Supplement 1*).

# 6. Conclusion

One-dimensional CNNs applied to samples of 2-4 QC measurements can improve the detection of nonconformity of a process to quality specifications, with lower cost.

# 8. Formalisms and Notation

## 8.1. Abbreviations

QC: quality control

NN: artificial neural network

DNN: deep neural network

CNN: convolutional neural network

1-D: one dimensional

GA: genetic algorithms

QCMT: quality control measurements tuple(s)

## 8.2. Operators

*# : cardinality*

$\sim$ : distributed as

## 8.3. Datasets

$\mathbf{Z_\theta} = \{z_i\}_{i=1}^n = \{x_i, y_i\}_{i=1}^n = \{x_i, f(x_i; \theta)\}_{i=1}^n$: *dataset of n-tuples $x_i$ of QC measurements with* $y_i \in \{True, False\}$

$\mathbf{T}_a(n)$: *training dataset of n-tuples $x_i$ of QC measurements.*

$\mathbf{D}(n, 0, 1)$: *testing dataset of n-tuples $x_i$ of QC measurements, distributed as $\mathcal{N}(0, 1)$.*

$\mathbf{D}(n, k, \mu, \sigma)$: *testing dataset of n-tuples $x_i$ of QC measurements, where k of them are distributed as $\mathcal{N}(\mu, \sigma^2)$ and n - k as $\mathcal{N}(0, 1)$.*

## 8.4. Functions

$\binom{n}{k}$: binomial coefficient

$\mathcal{N}(\mu, \sigma^2)$: normal distribution with mean $\mu$ and variance $\sigma^2$

$N_a(\boldsymbol{x})$: *neural network QC function applied to a tuple $\boldsymbol{x}$ of QC measurements;* $N_a(\boldsymbol{x}) \in \{True, False\}$

$S_a(\boldsymbol{x}; \boldsymbol{\theta}) = S_a(\boldsymbol{x}; l, m, s)$: *statistical QC function, applied to a tuple $\boldsymbol{x}$ of QC measurements;*
$S_a(\boldsymbol{x}; l, m, s) = \begin{cases} True \text{ if } \exists\, x_i \in \boldsymbol{x}, |x_i - m| > l\, s \\ False \text{ otherwise} \end{cases}$

$S_a(x; \text{l}) = S_a(x; l, 0, 1)$

$L(f_q; \theta) = \frac{1}{n}\sum_{i=1}^{n} I\left(f_q(x_i; \theta) \neq f(x_i; \theta)\right)$: misclassification rate of QC function $f_q$, where $I(e) = \begin{cases} 1 \text{ if } e \text{ True,} \\ 0 \text{ if } e \text{ False} \end{cases}$

$f(x; \theta) = \begin{cases} \text{True if } \exists\, x_i \in x, X_i \sim \mathcal{N}(\mu, \sigma^2) \wedge (\mu > 0 \vee \sigma > 1), \\ \text{False if } \forall\, x_i \in x, X_i \sim \mathcal{N}(0,1) \end{cases}$ and $f_q(x; \theta) = \begin{cases} N_\alpha(x) \\ \vee \\ S_\alpha(x; \theta) \end{cases}$

$P_{N,a}(n, 0, 1)$: probability for false rejection of the $N_a(x)$ *applied* to the dataset $D(n, 0, 1)$ for $n = \#x$, and $1 \leq n \leq 4$.

$P_{N,a}(n, k, \mu, \sigma)$: probability for rejection of the $N_a(x)$ *applied* to the dataset $D(n, k, \mu, \sigma)$ for $n = \#x$, $1 \leq n \leq 4$, and $1 \leq k \leq n$.

$P_{S,a}(n, 0, 1)$: probability for rejection of the $S_a(x; l)$ applied to the dataset $D(n, 0, 1)$ for $n = \#x$, $1 \leq n \leq 4$, and decision limit $l$.

$P_{S,a}(n, k, \mu, \sigma)$: probability for rejection of the $S_a(x; l)$ applied to the dataset $D(n, k, \mu, \sigma)$ for $n = \#x$, $1 \leq n \leq 4$, $1 \leq k \leq n$, and decision limit $l$.

The decision limit *l* is defined as following.

Let $P_{C,a}(x; k, l, \mu, \sigma)$ be the probability for rejection of a *n*-tuple of QC measurements, for $n = \#x$, $1 \leq n \leq 4$, $1 \leq k \leq n$, where k of them are distributed as $\mathcal{N}(\mu, \sigma^2)$ and (n - k) as $\mathcal{N}(0,1)$. Then:

$$P_{C,a}(x; k, l, \mu, \sigma) = 1 - \frac{1}{2^k} \text{erf}\left(\frac{1}{\sqrt{2}}\right)^{n-k} \left(\text{erf}\left(\frac{\mu+l}{\sigma\sqrt{2}}\right) - \text{erf}\left(\frac{\mu-l}{\sigma\sqrt{2}}\right)\right)^k \text{ for } 0 > k \leq n$$

To define *l*, we solve the following equation:

$$P_{C,a}(x; n, l, 0, 1) = P_{N,a}(n, 0, 1) \Rightarrow l = \sqrt{2}\,\text{erf}^{-1}\left(\left(1 - P_N(n, 0, 1)\right)^{\frac{1}{n}}\right)$$

Therefore:

$$P_{S,a}(n, k, \mu, \sigma) = P_{C,a}\left(x; k, \sqrt{2}\,\text{erf}^{-1}\left(\left(1 - P_{N,a}(n, 0, 1)\right)^{\frac{1}{n}}\right), \mu, \sigma\right)$$

In addition:

$$\Delta P_a(n, \mu, \sigma) = P_{N,a}(n, \mu, \sigma) - P_{S,a}(n, \mu, \sigma)$$

$$\Delta P_a(n, k, \mu, \sigma) = P_{N,a}(n, k, \mu, \sigma) - P_{S,a}(n, k, \mu, \sigma)$$

$$\Delta P_{R,a}(n, k, \mu, \sigma) = \frac{\Delta P_a(n, k, \mu, \sigma)}{P_{S,a}(n, k, \mu, \sigma)}$$

# 9. Supplementary Materials

The following resources are available online:

1. *Supplement 1,* with the 1D-CNN models in various formats, at:
   https://www.hcsl.com/Supplements/hcsltr21s1.zip
2. *Supplement 2,* with spreadsheets presenting the probabilities for rejection $P_{N,a}(n,k,\mu,\sigma)$ and $P_{S,a}(n,k,\mu,\sigma)$, their differences $\Delta P_a(n,k,\mu,\sigma)$ and the respective statistical significance ($p$-values), and their relative differences $\Delta P_{R,a}(n,k,\mu,\sigma)$, at:
   https://www.hcsl.com/Supplements/hcsltr21s2.zip
3. In addition, *Intelligent Quality*, a computer program in Wolfram Language, for calculating and plotting the probabilities for rejection $P_{N,a}(n,k,\mu,\sigma)$ and $P_{S,a}(n,k,\mu,\sigma)$, their differences $\Delta P_a(n,k,\mu,\sigma)$, and their relative differences $\Delta P_{R,a}(n,k,\mu,\sigma)$, is available at:
   https://www.hcsl.com/Tools/IntelligentQuality/.

# 10. Appendix
## Plots of the Probabilities for Rejection of QC Functions

### Table of Figures

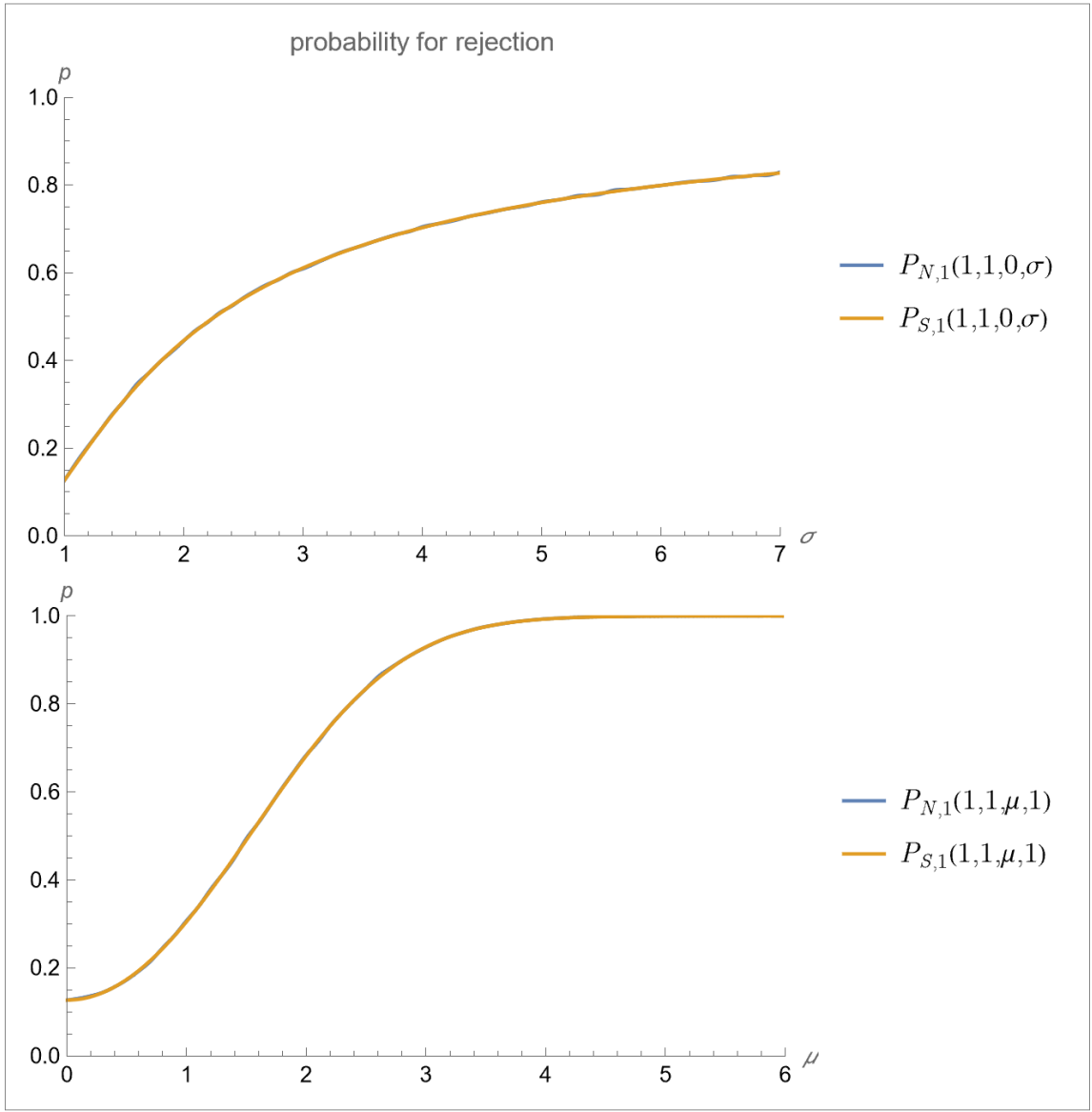

Figure 1: $P_{N,1}(1,1,\mu,\sigma)$ vs $P_{S,1}(1,1,\mu,\sigma)$

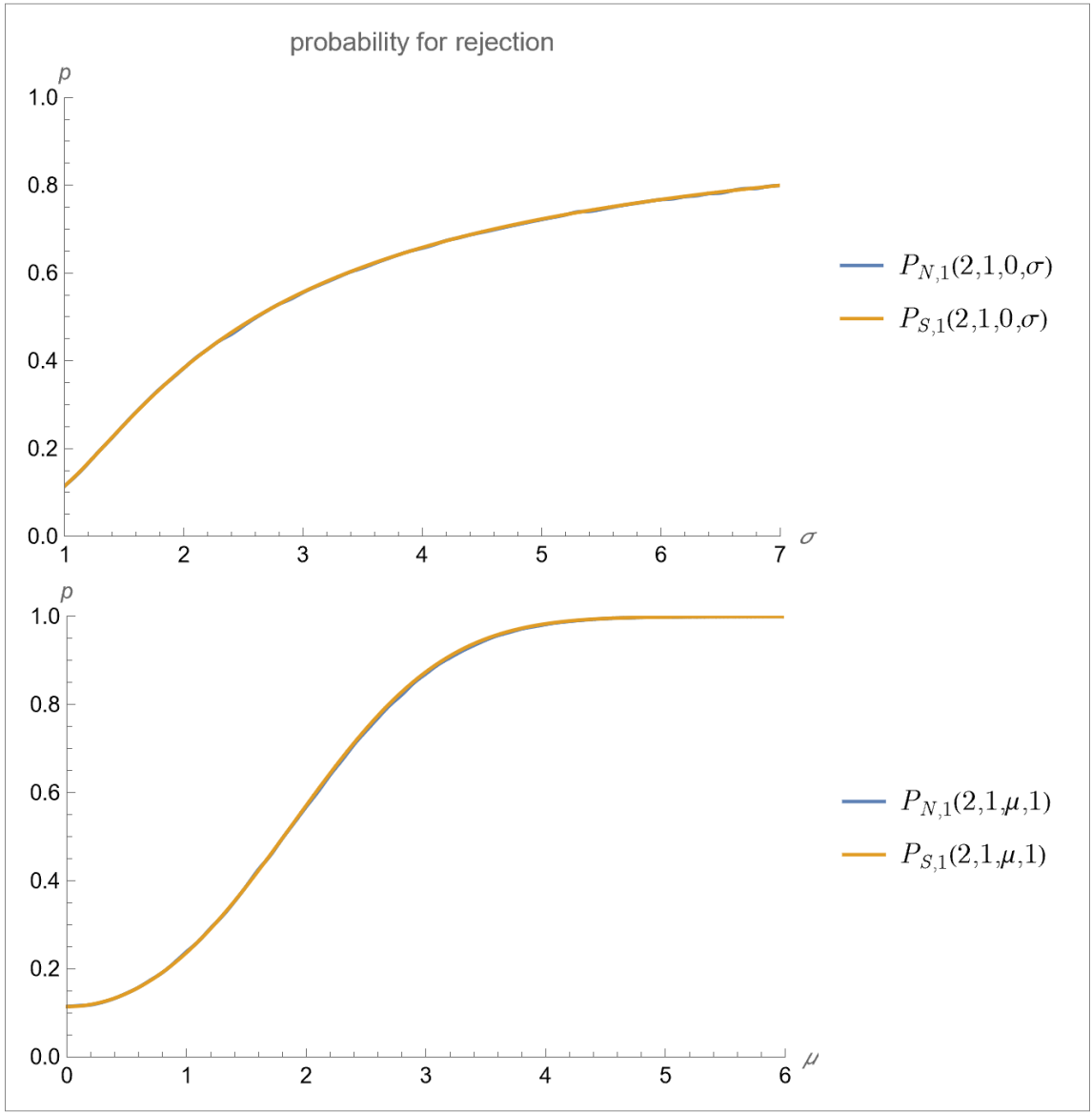

*Figure 2: $P_{N,1}(2,1,\mu,\sigma)$ vs $P_{S,1}(2,1,\mu,\sigma)$*

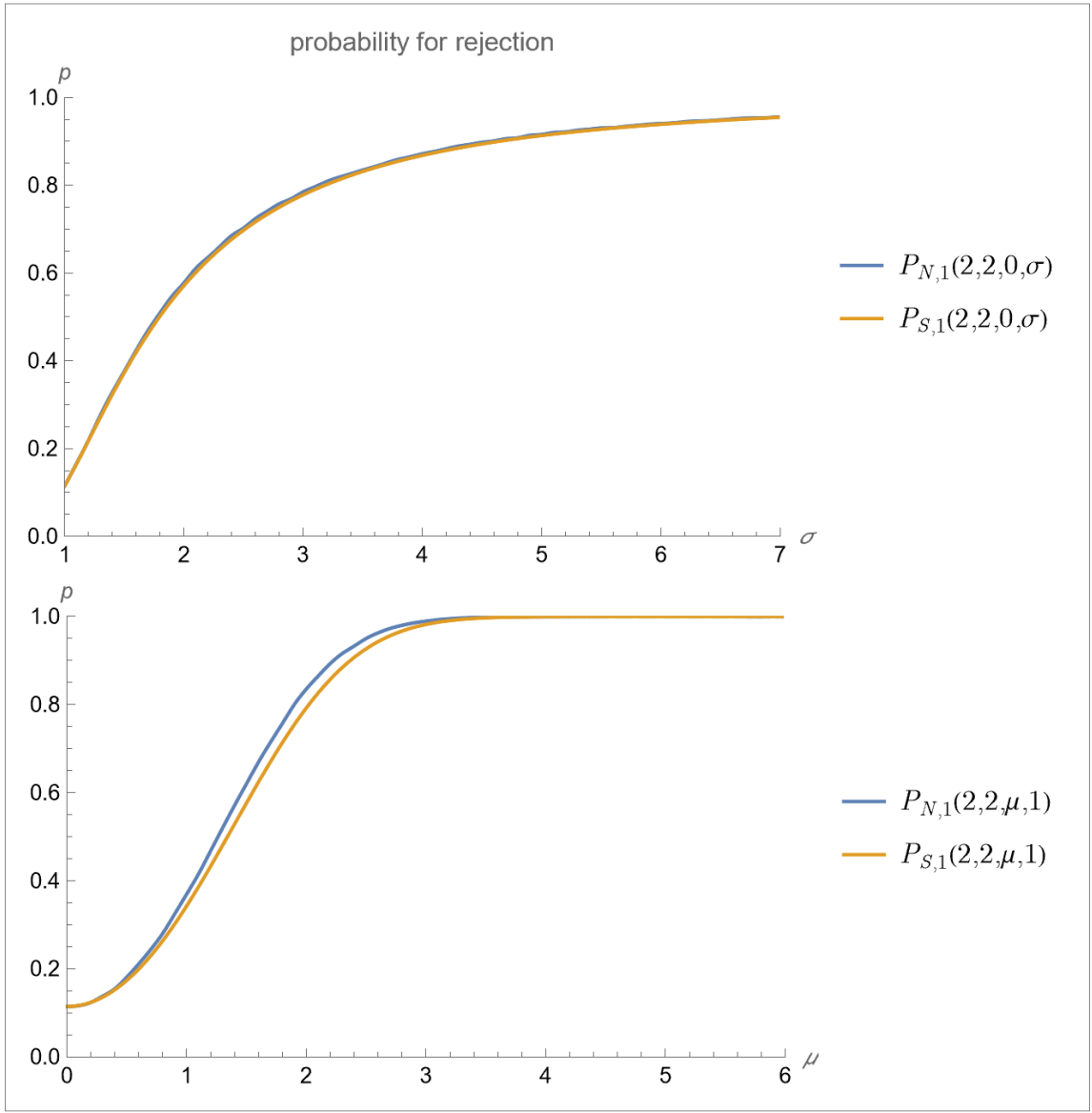

*Figure 3:* $P_{N,1}(2,2,\mu,\sigma)$ *vs* $P_{S,1}(2,2,\mu,\sigma)$

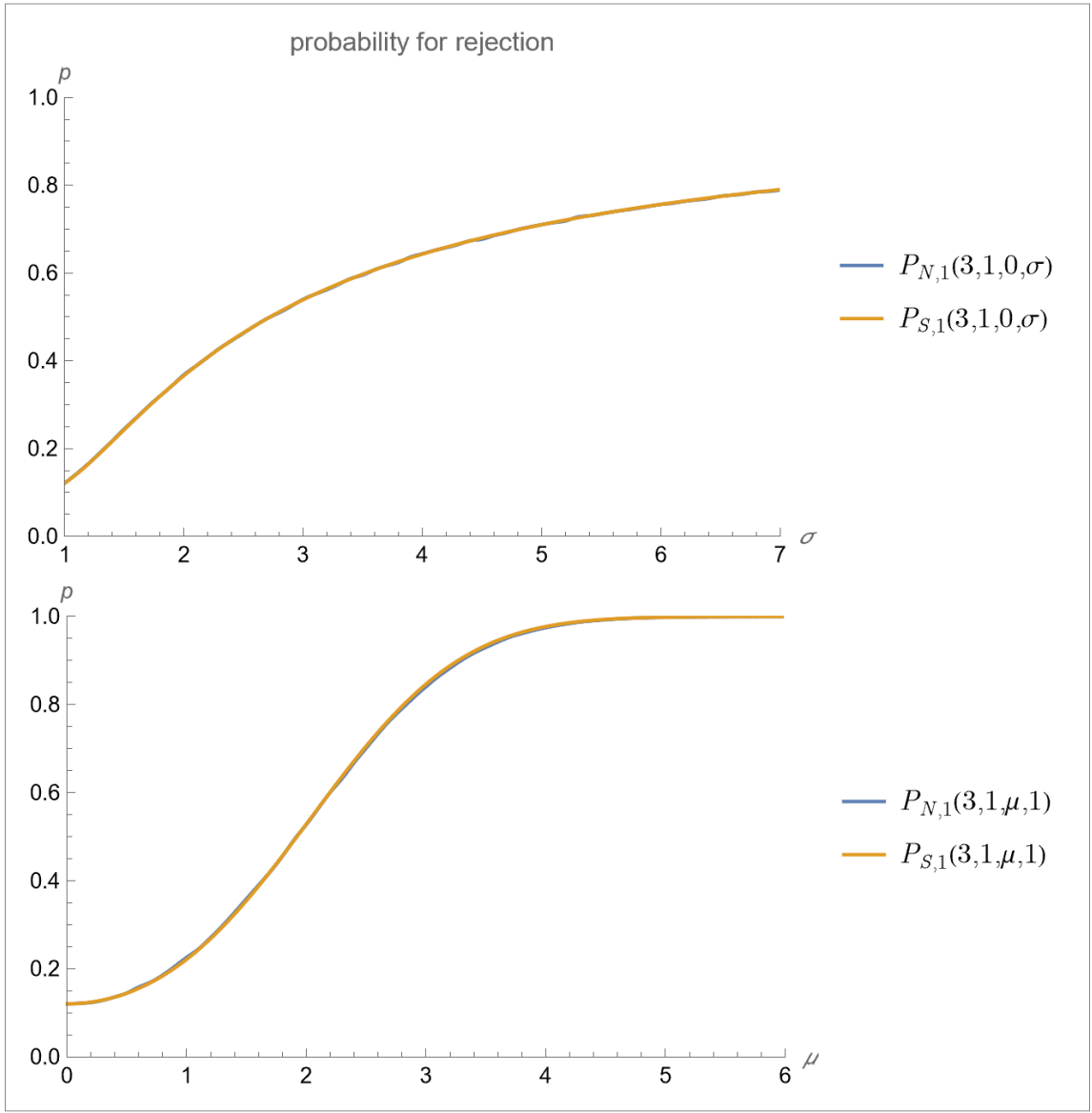

*Figure 4: $P_{N,1}(3,1,\mu,\sigma)$ vs $P_{S,1}(3,1,\mu,\sigma)$*

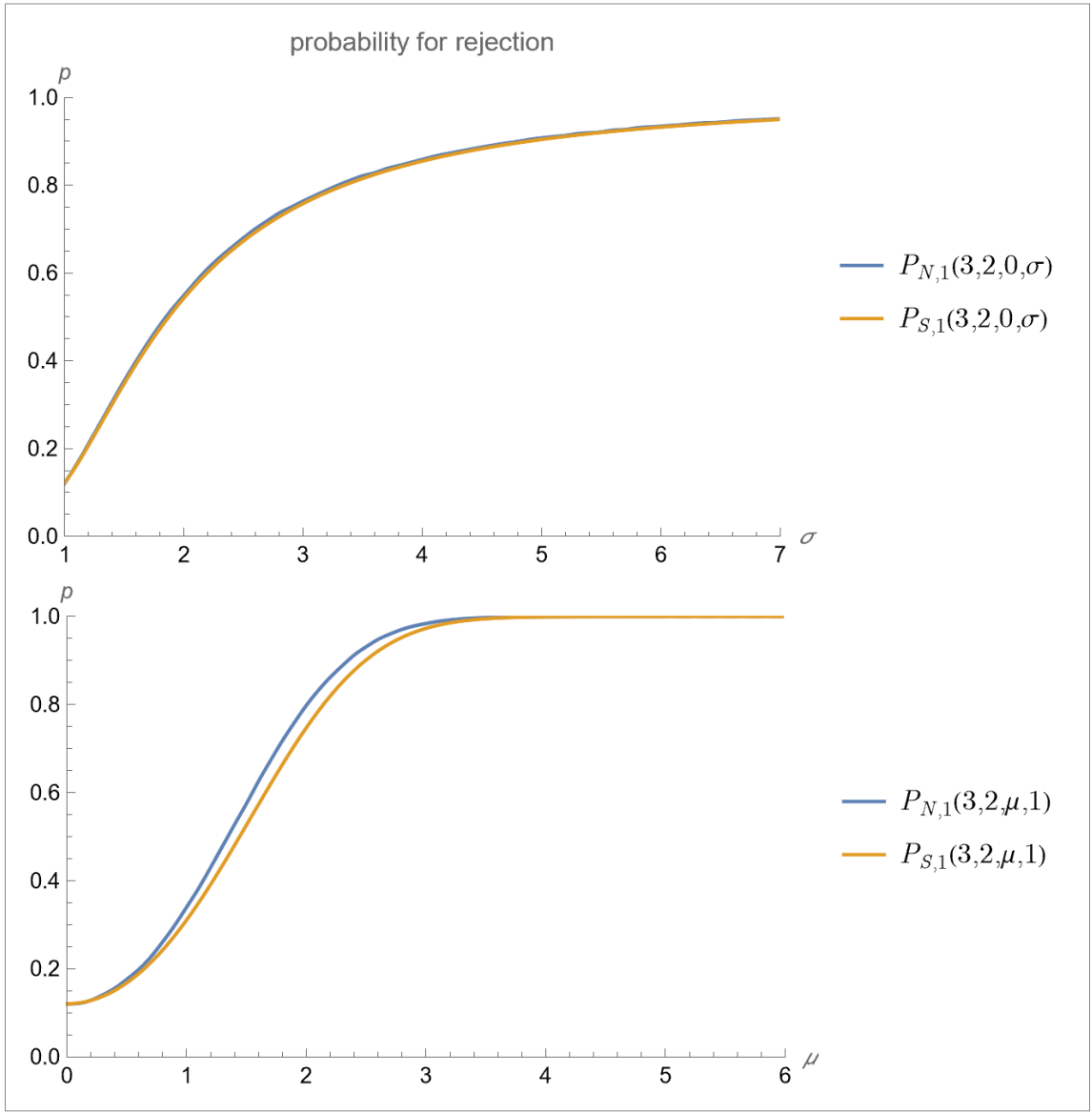

*Figure 5: $P_{N,1}(3,2,\mu,\sigma)$ vs $P_{S,1}(3,2,\mu,\sigma)$*

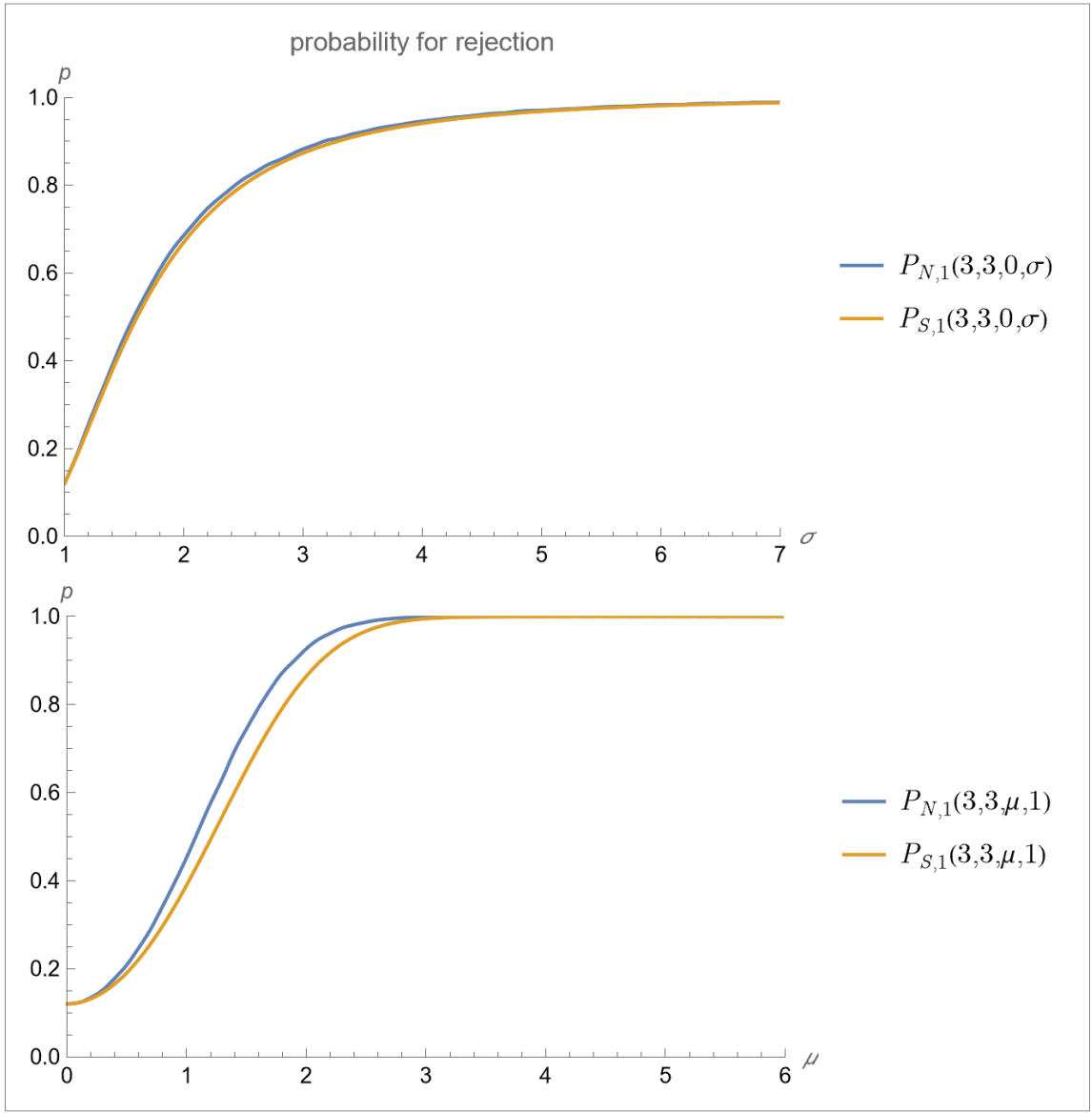

*Figure 6: $P_{N,1}(3,3,\mu,\sigma)$ vs $P_{S,1}(3,3,\mu,\sigma)$*

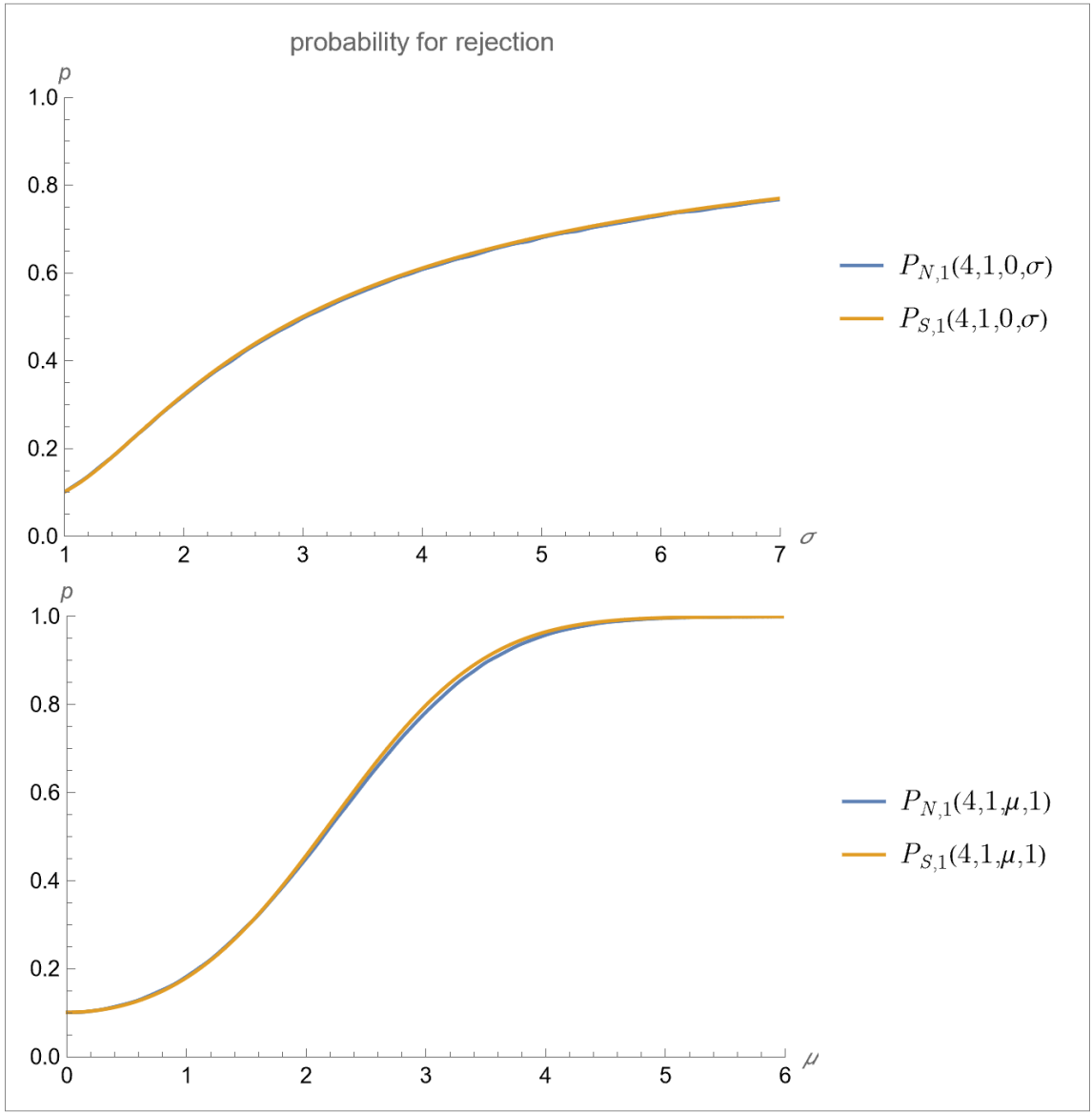

*Figure 7: $P_{N,1}(4,1,\mu,\sigma)$ vs $P_{S,1}(4,1,\mu,\sigma)$*

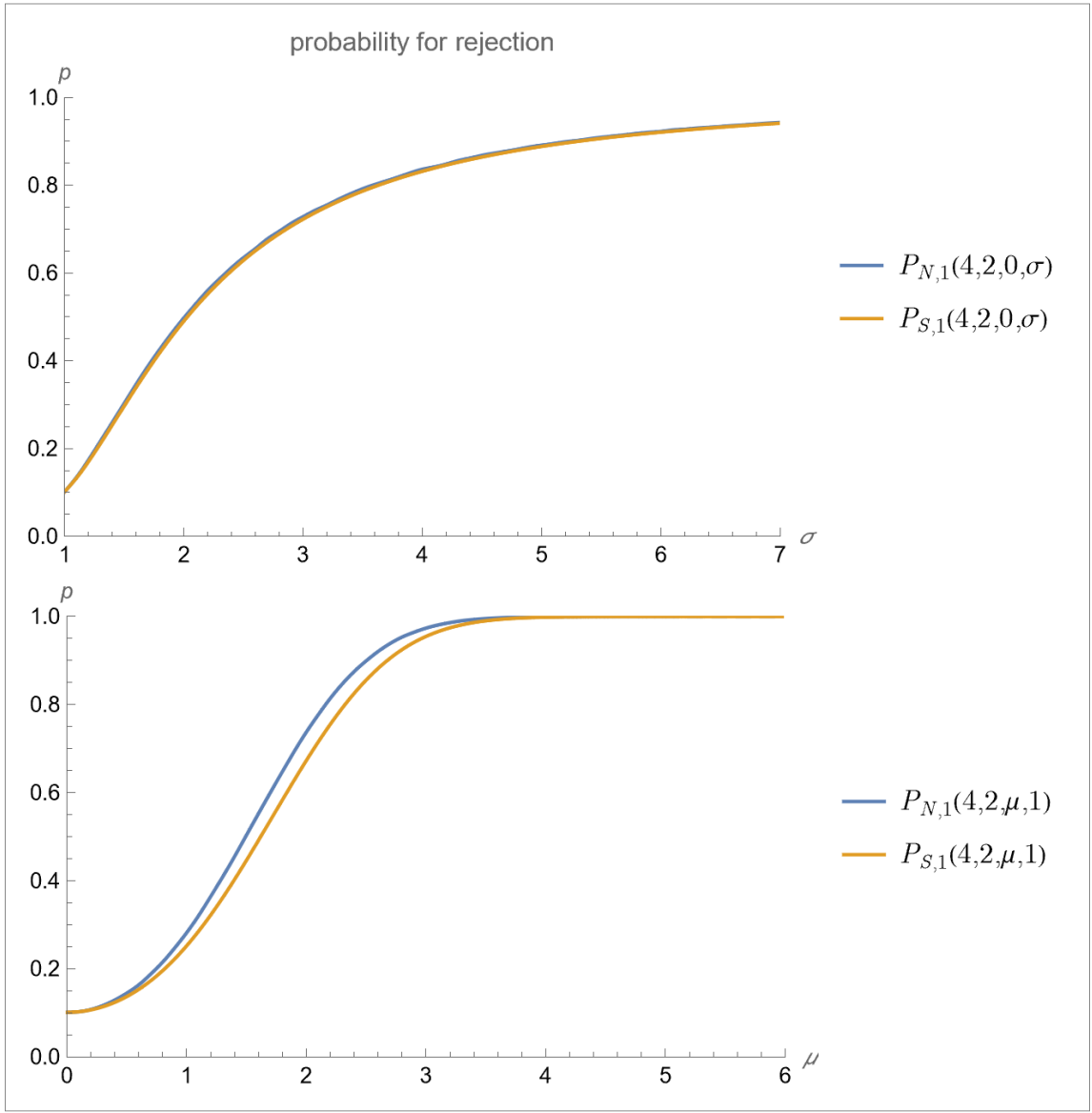

*Figure 8: $P_{N,1}(4,2,\mu,\sigma)$ vs $P_{S,1}(4,2,\mu,\sigma)$*

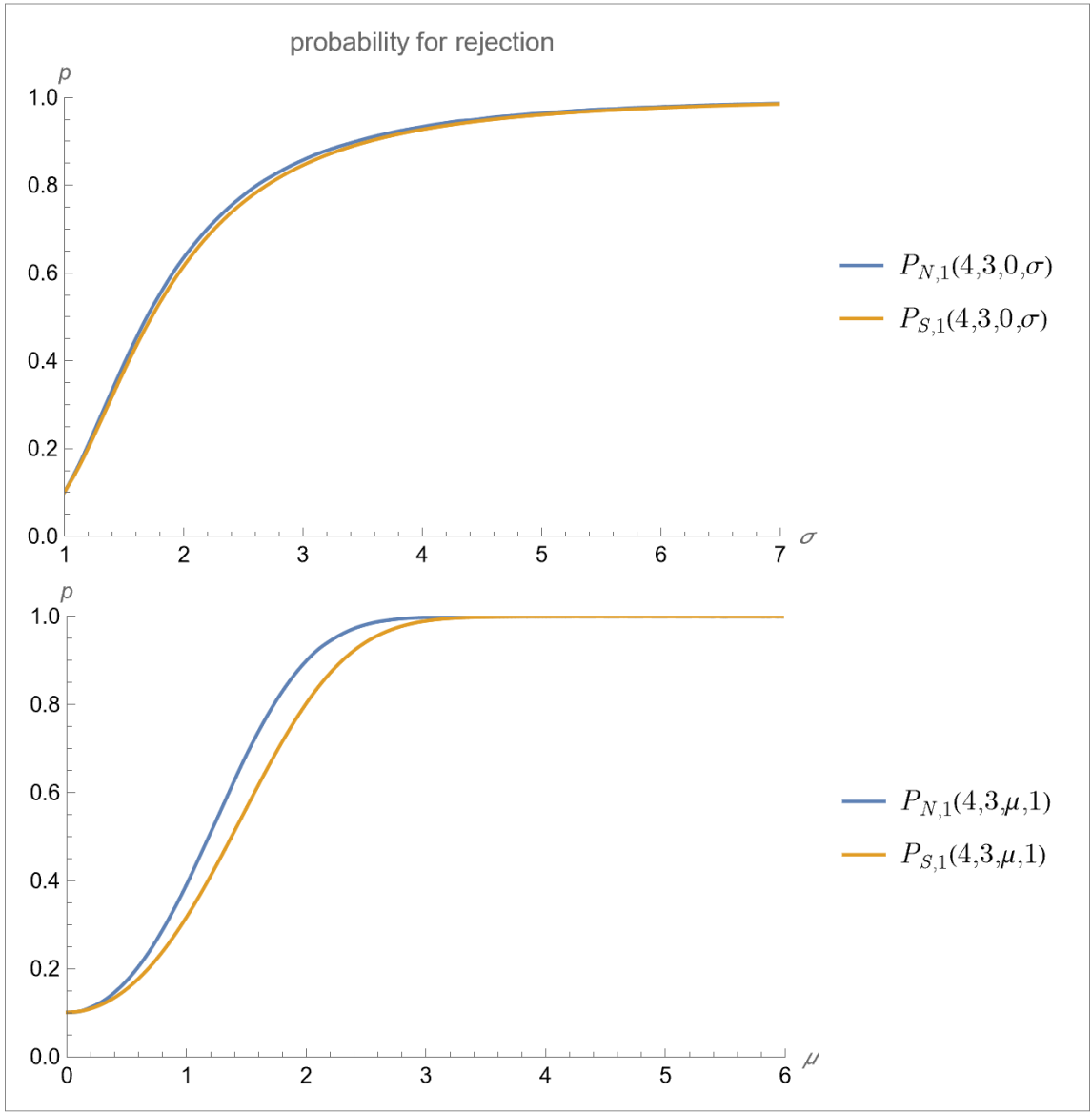

*Figure 9: $P_{N,1}(4,3,\mu,\sigma)$ vs $P_{S,1}(4,3,\mu,\sigma)$*

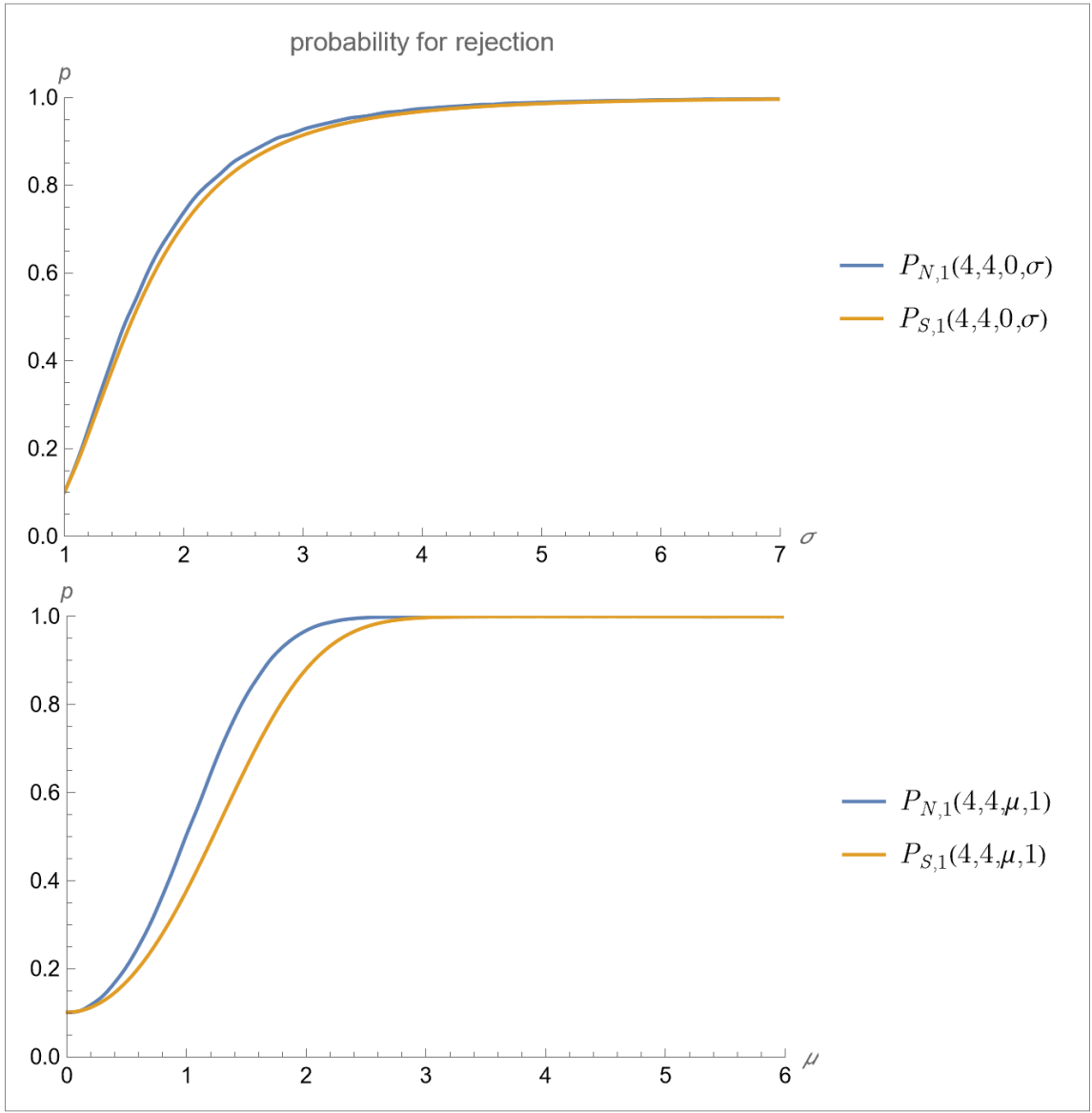

*Figure 10: $P_{N,1}(4,4,\mu,\sigma)$ vs $P_{S,1}(4,4,\mu,\sigma)$*

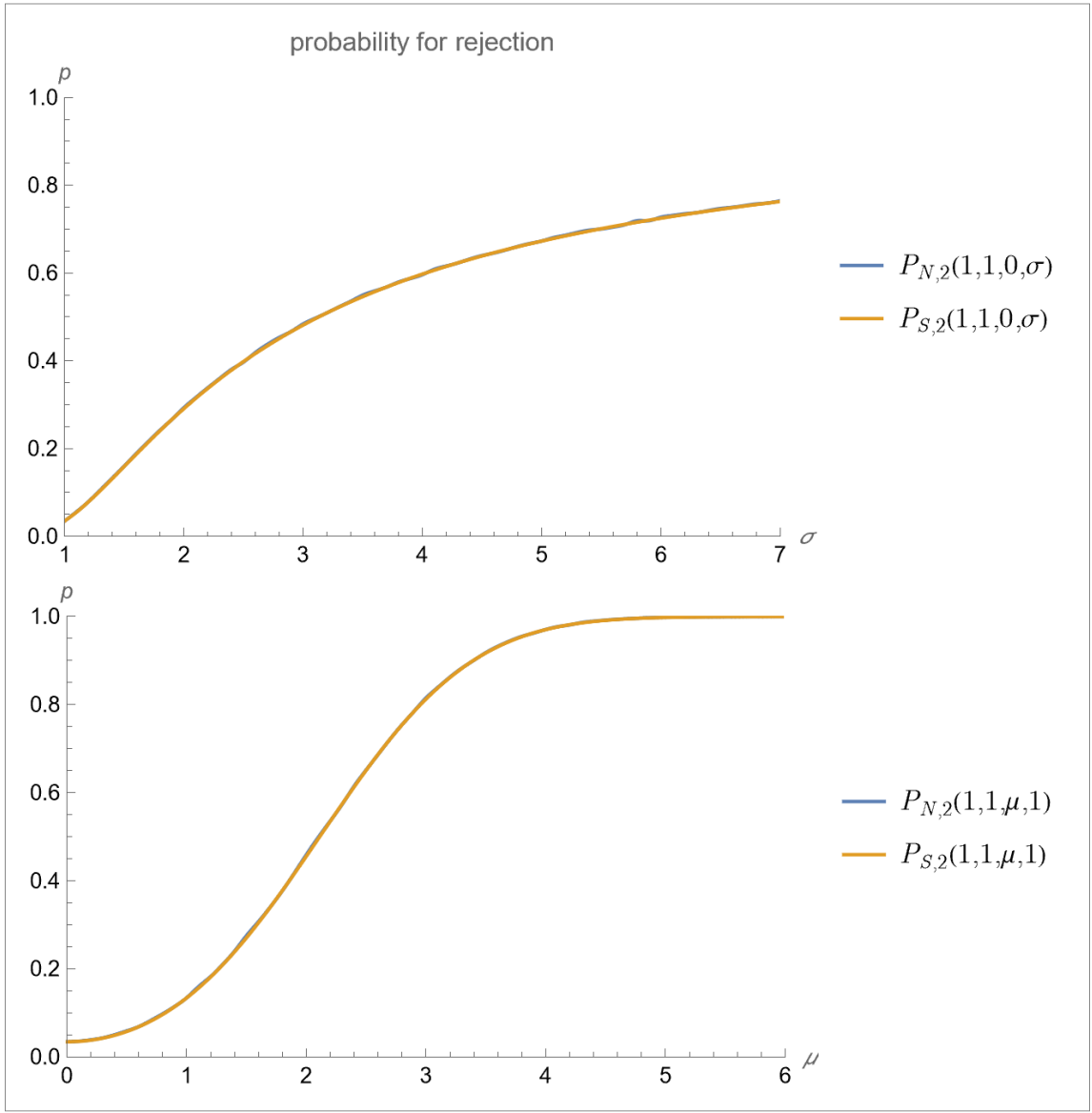

*Figure 11: $P_{N,2}(1,1,\mu,\sigma)$ vs $P_{S,2}(1,1,\mu,\sigma)$*

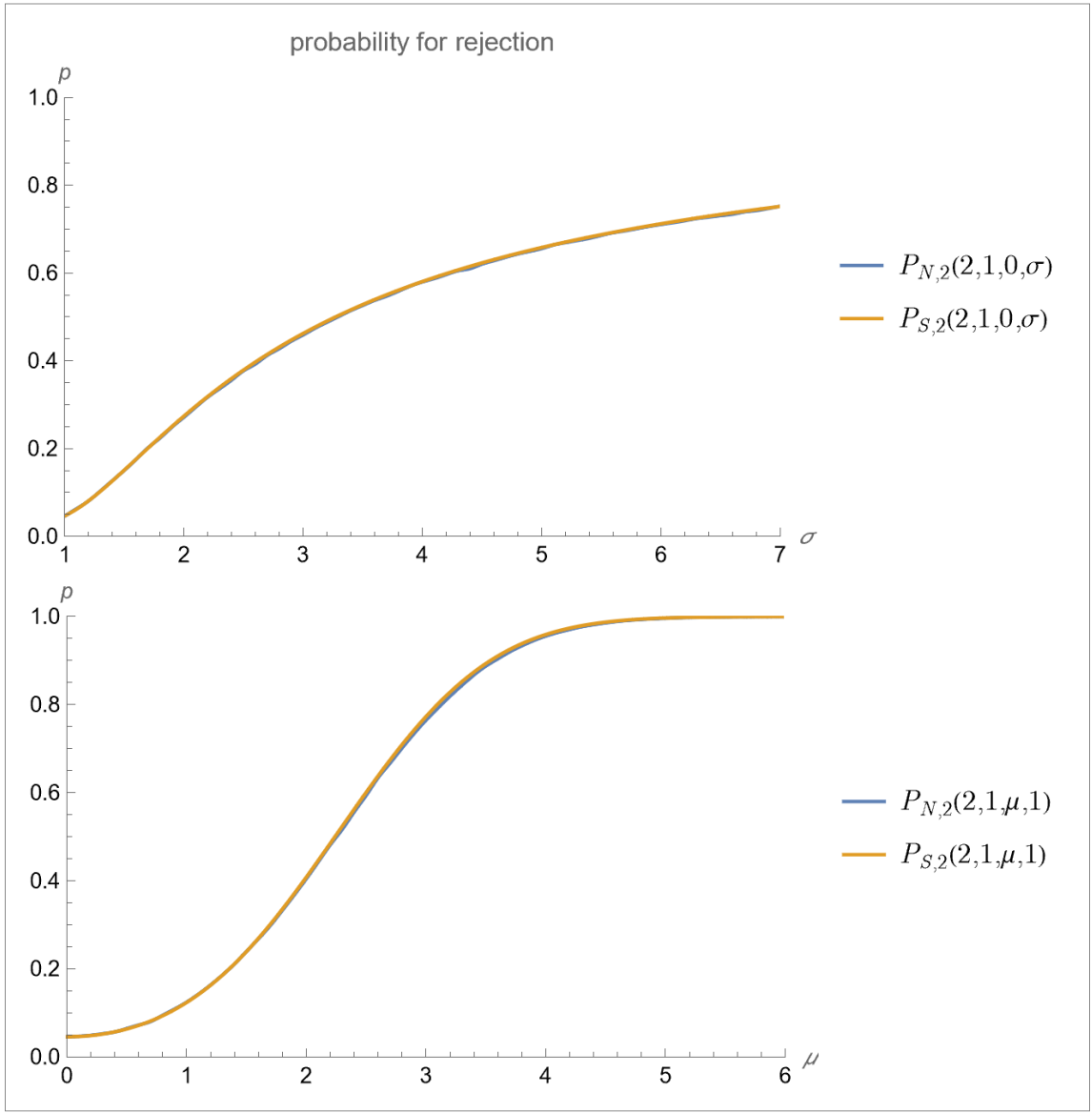

*Figure 12: $P_{N,2}(2,1,\mu,\sigma)$ vs $P_{S,2}(2,1,\mu,\sigma)$*

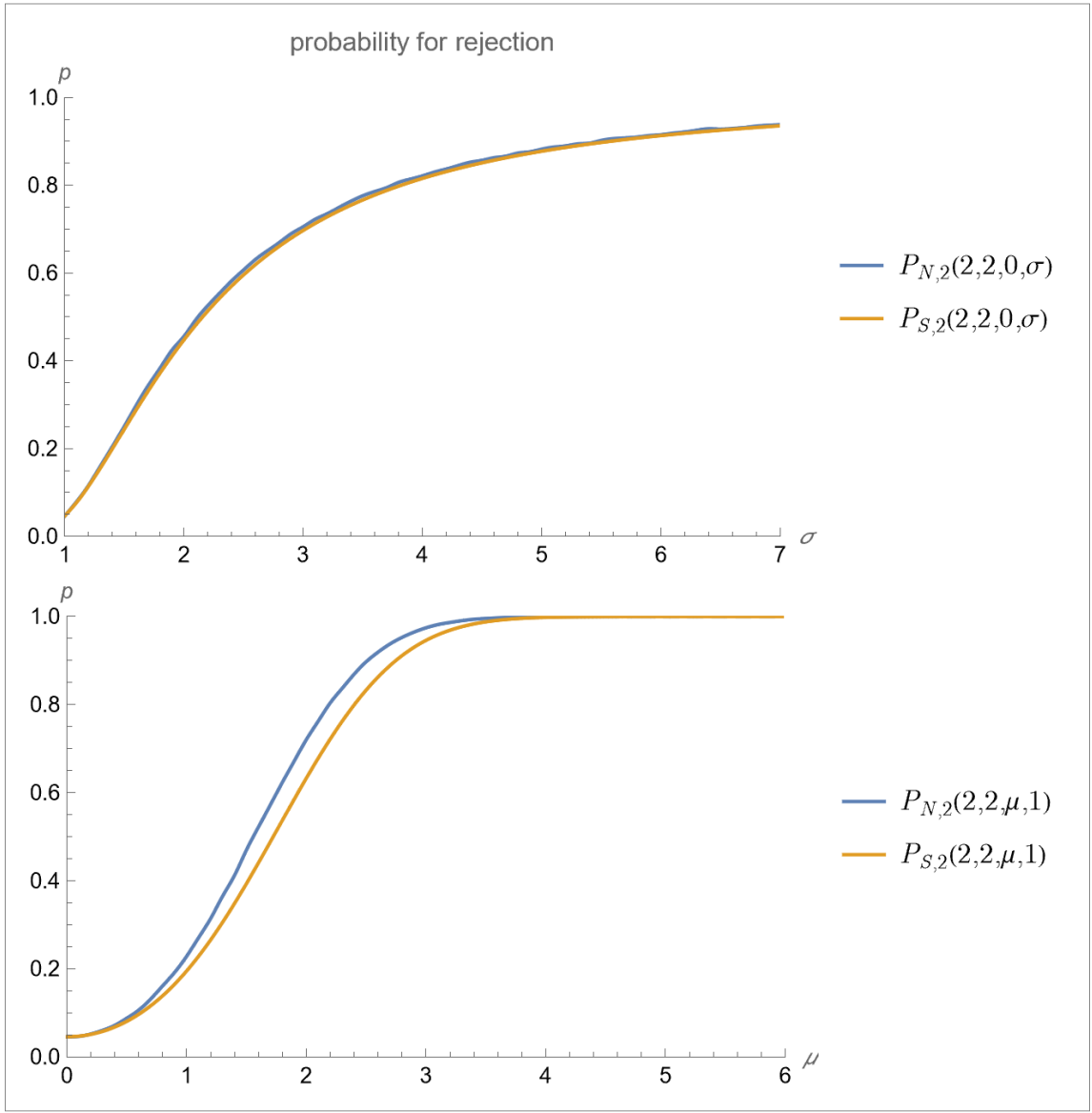

*Figure 13: $P_{N,2}(2,2,\mu,\sigma)$ vs $P_{S,2}(2,2,\mu,\sigma)$*

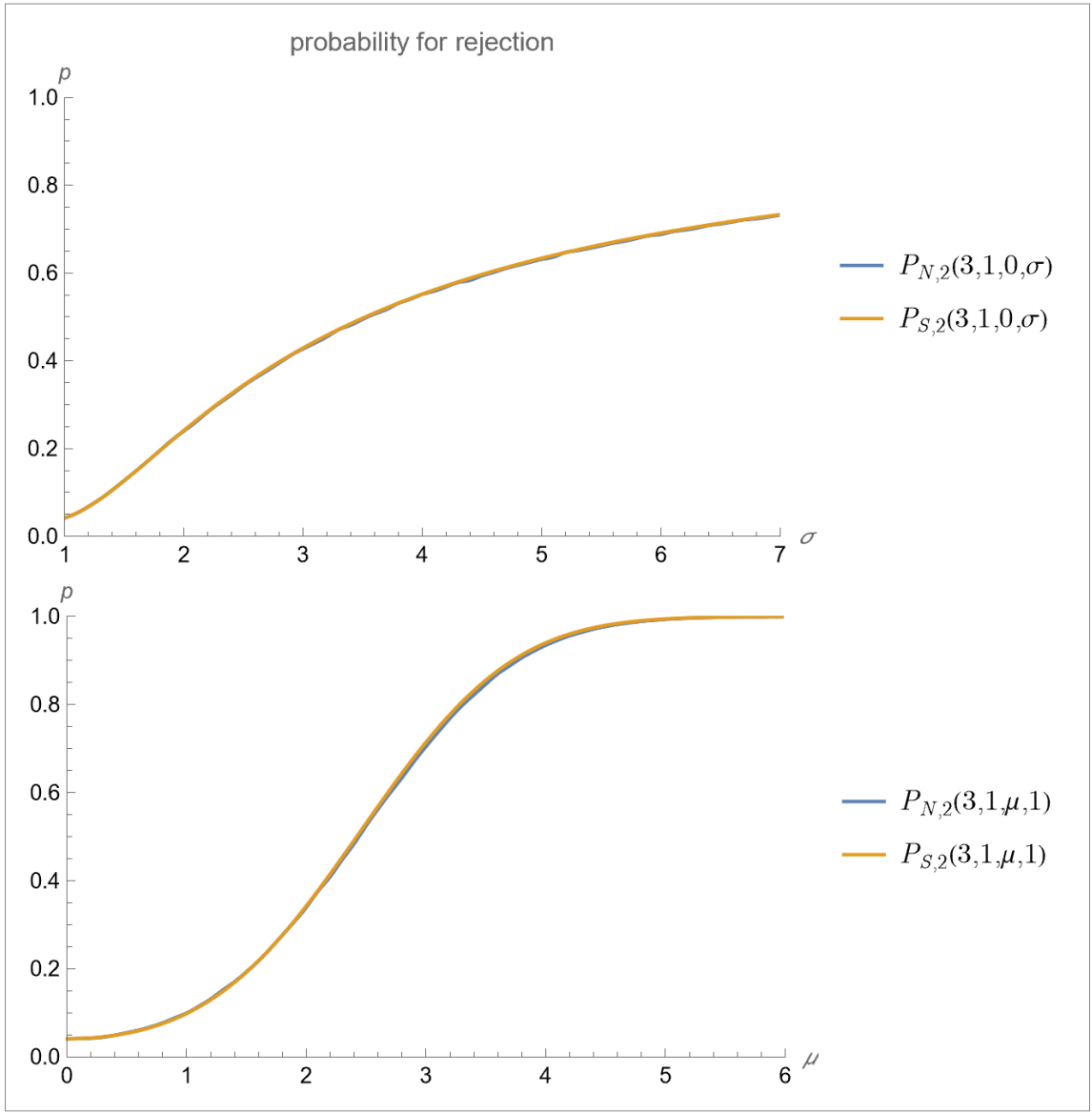

*Figure 14: $P_{N,2}(3,1,\mu,\sigma)$ vs $P_{S,2}(3,1,\mu,\sigma)$*

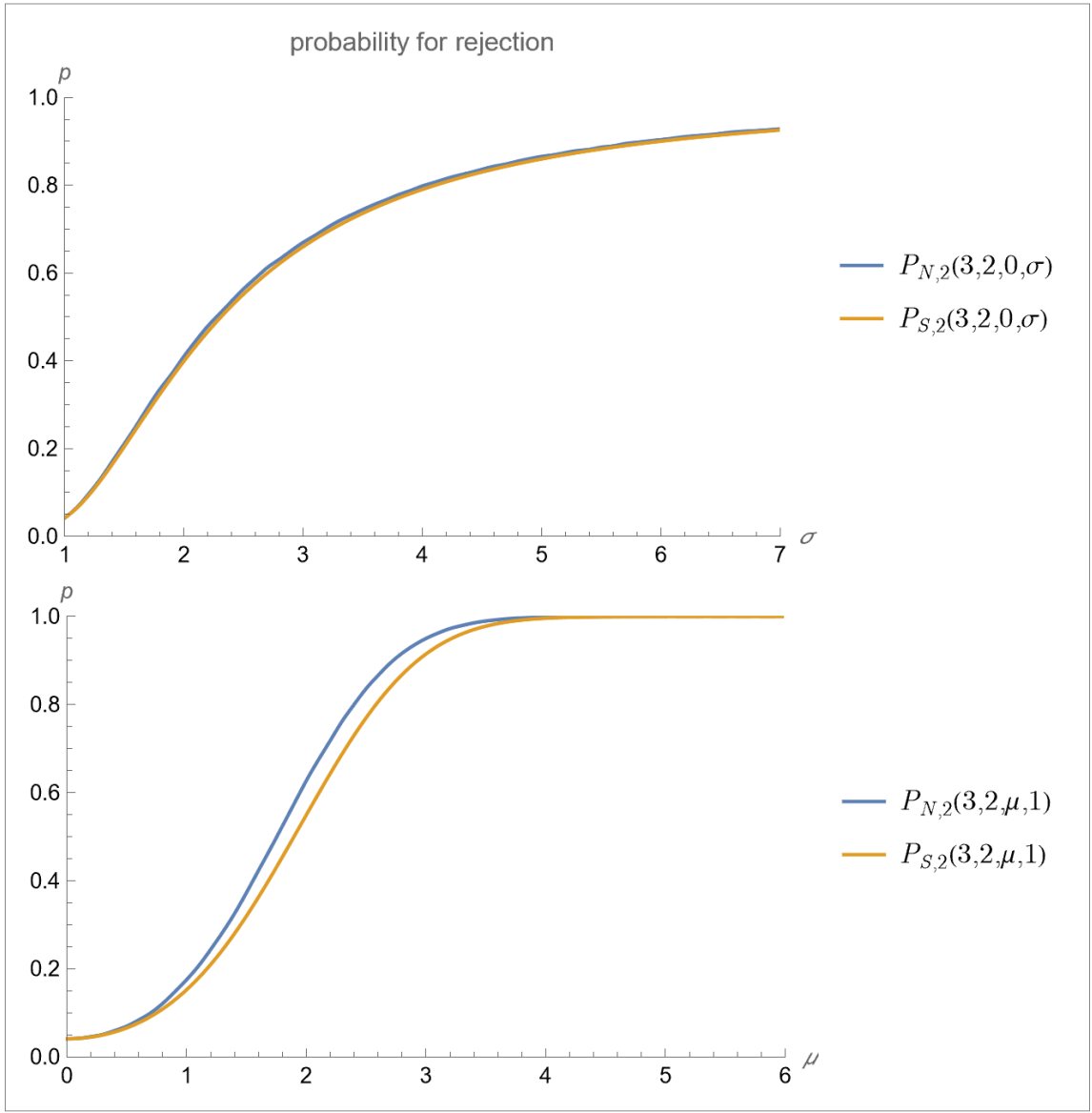

*Figure 15: $P_{N,2}(3,2,\mu,\sigma)$ vs $P_{S,2}(3,2,\mu,\sigma)$*

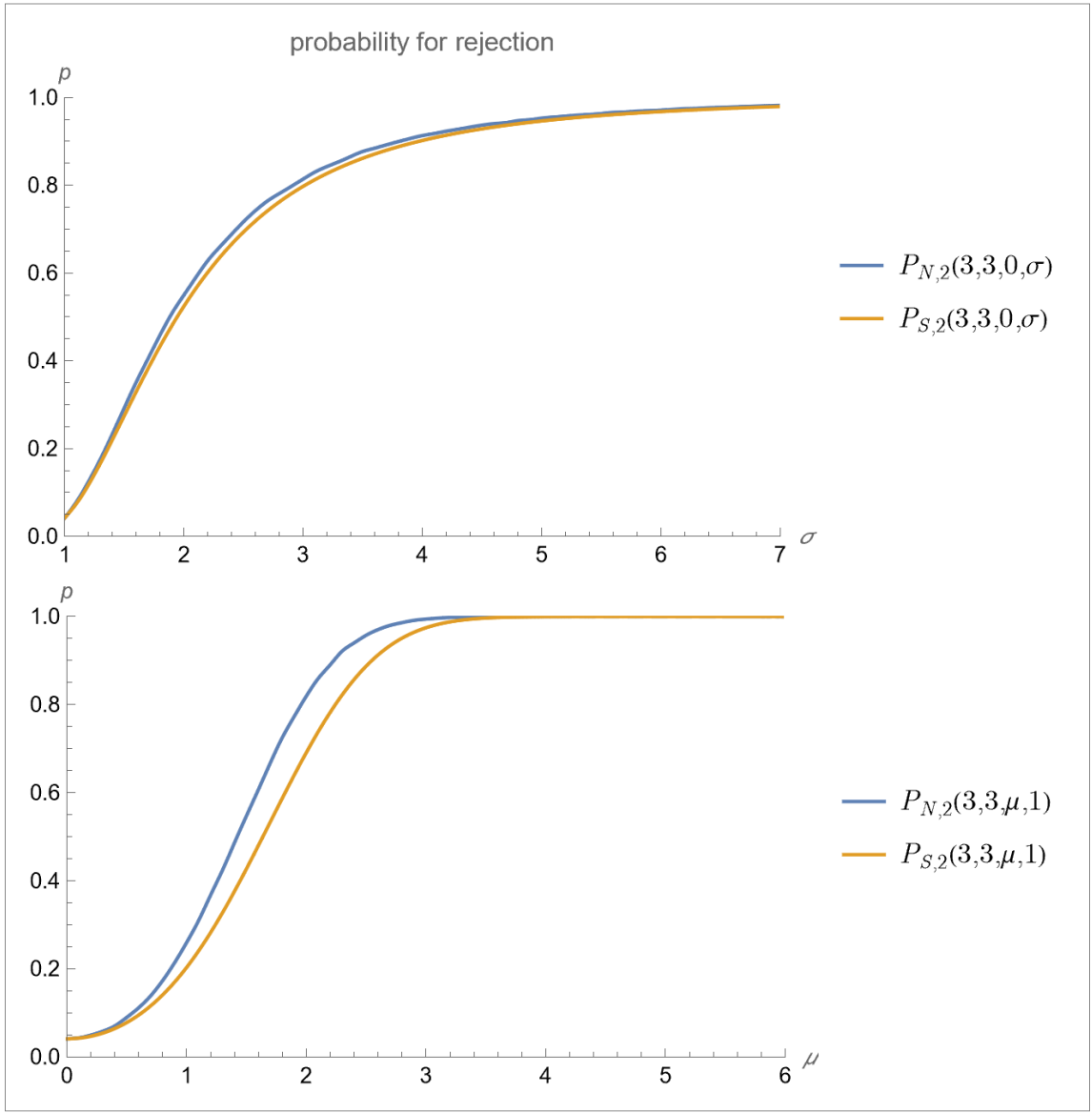

*Figure 16: $P_{N,2}(3,3,\mu,\sigma)$ vs $P_{S,2}(3,3,\mu,\sigma)$*

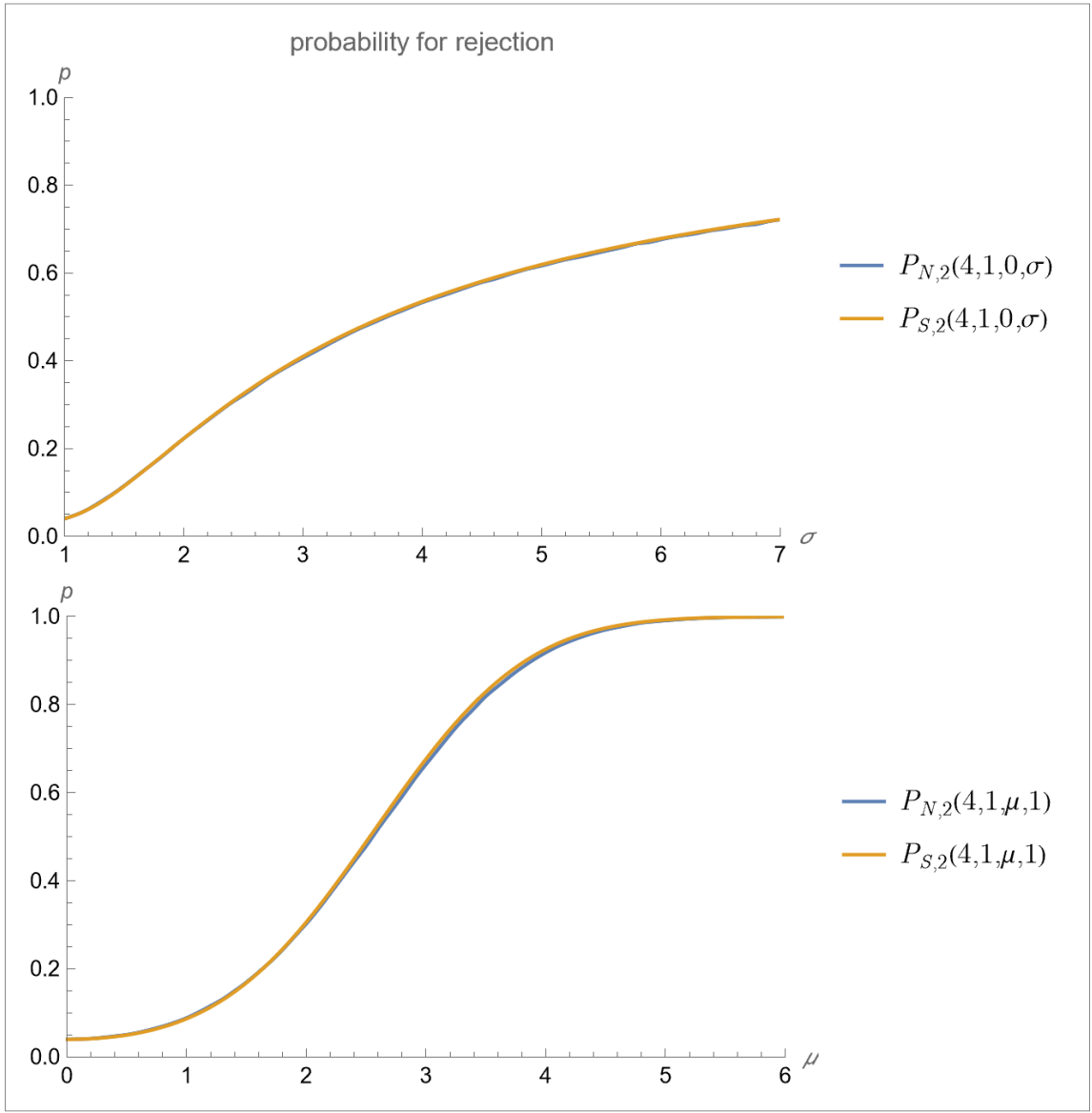

*Figure 17: $P_{N,2}(4,1,\mu,\sigma)$ vs $P_{S,2}(4,1,\mu,\sigma)$*

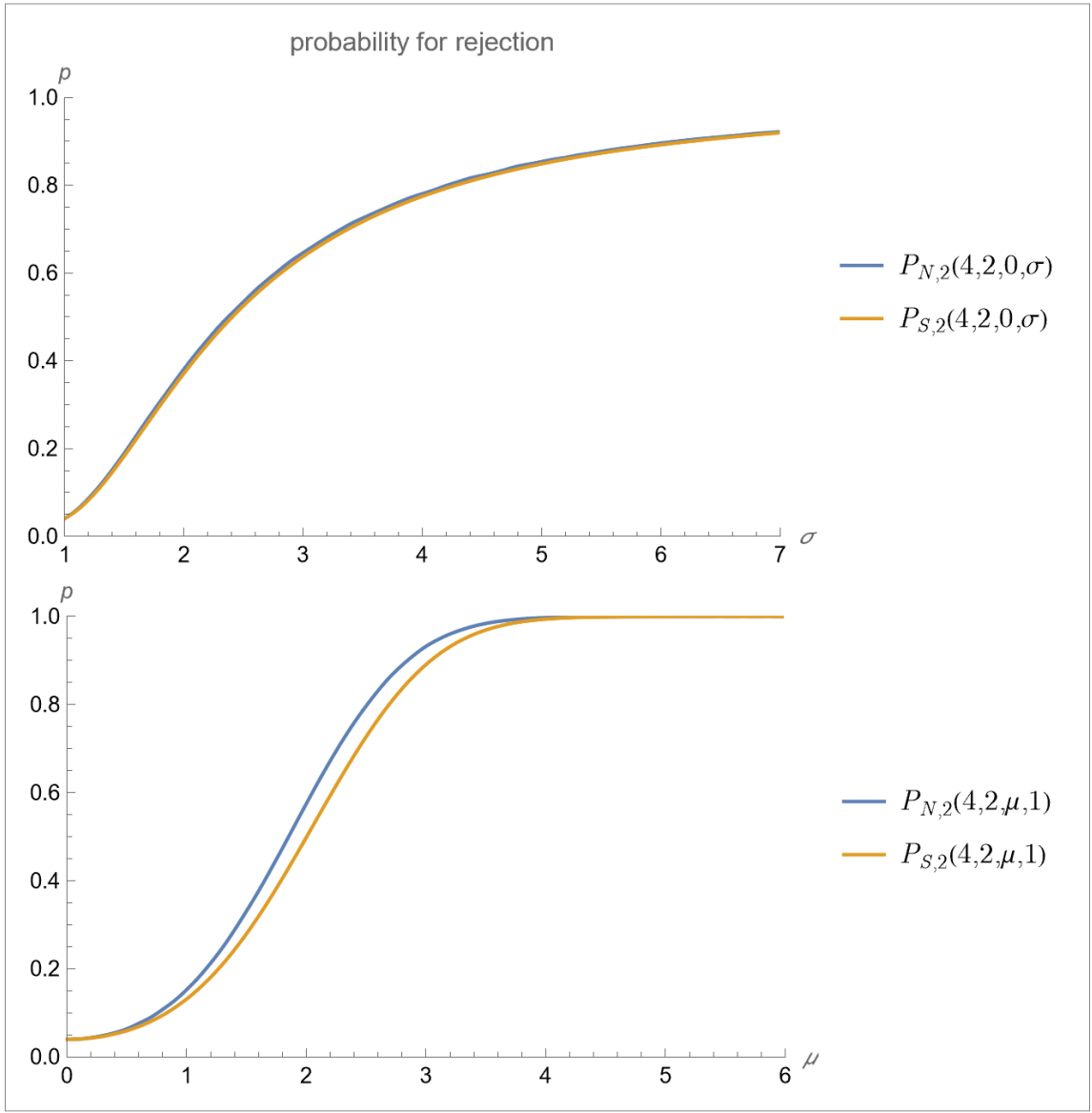

*Figure 18: $P_{N,2}(4,2,\mu,\sigma)$ vs $P_{S,2}(4,2,\mu,\sigma)$*

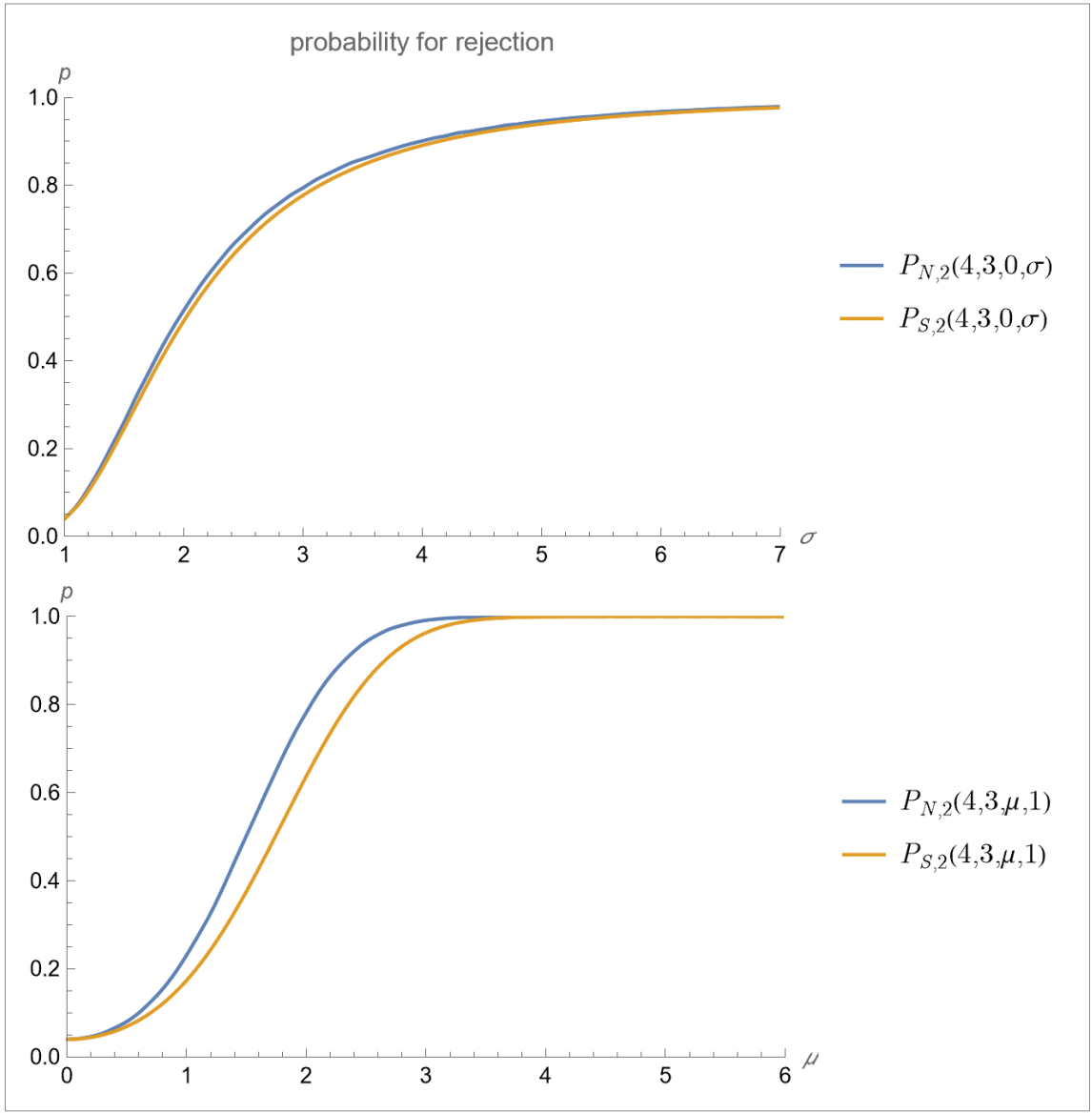

*Figure 19: $P_{N,2}(4,3,\mu,\sigma)$ vs $P_{S,2}(4,3,\mu,\sigma)$*

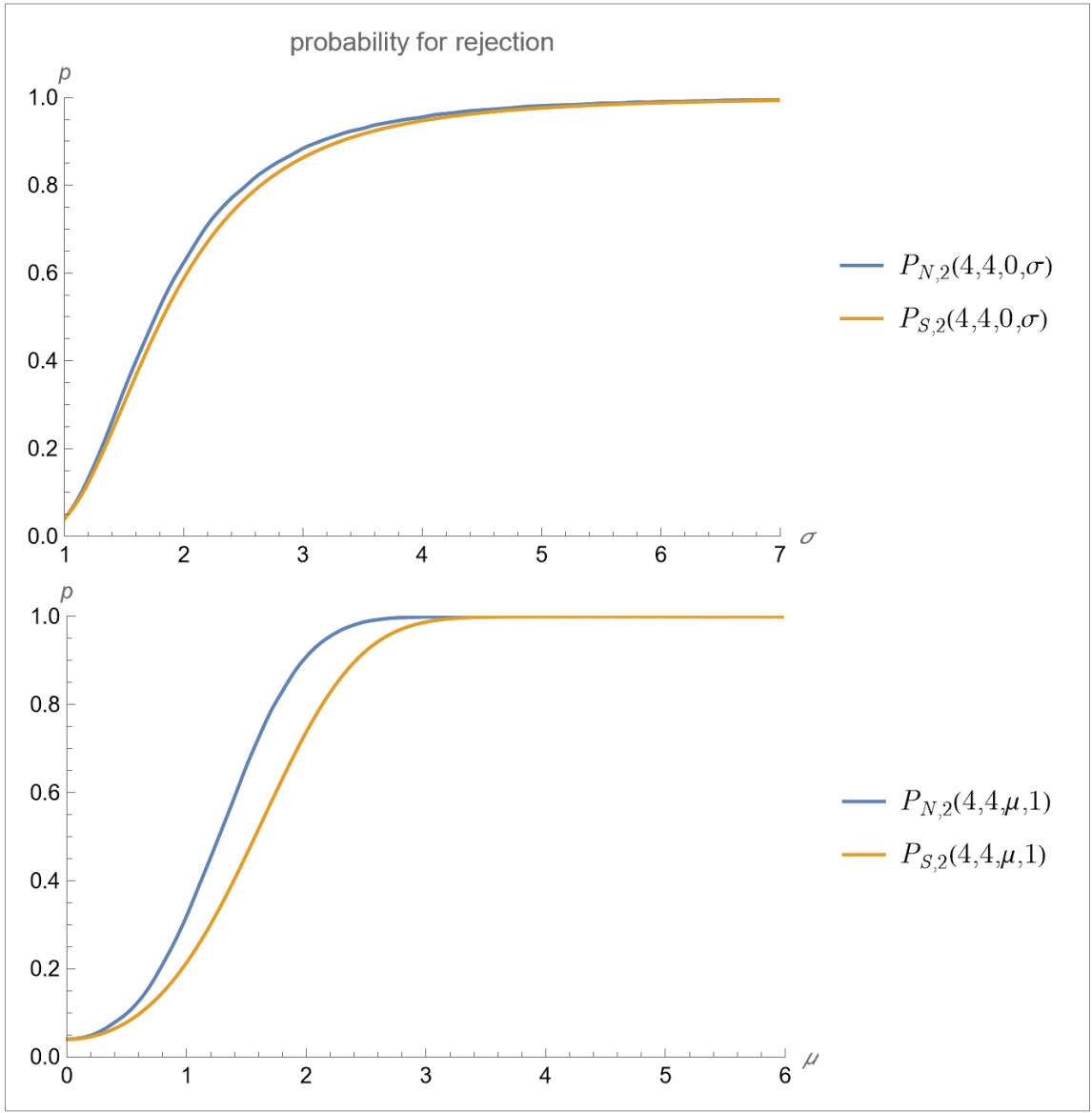

*Figure 20: $P_{N,2}(4,4,\mu,\sigma)$ vs $P_{S,2}(4,4,\mu,\sigma)$*

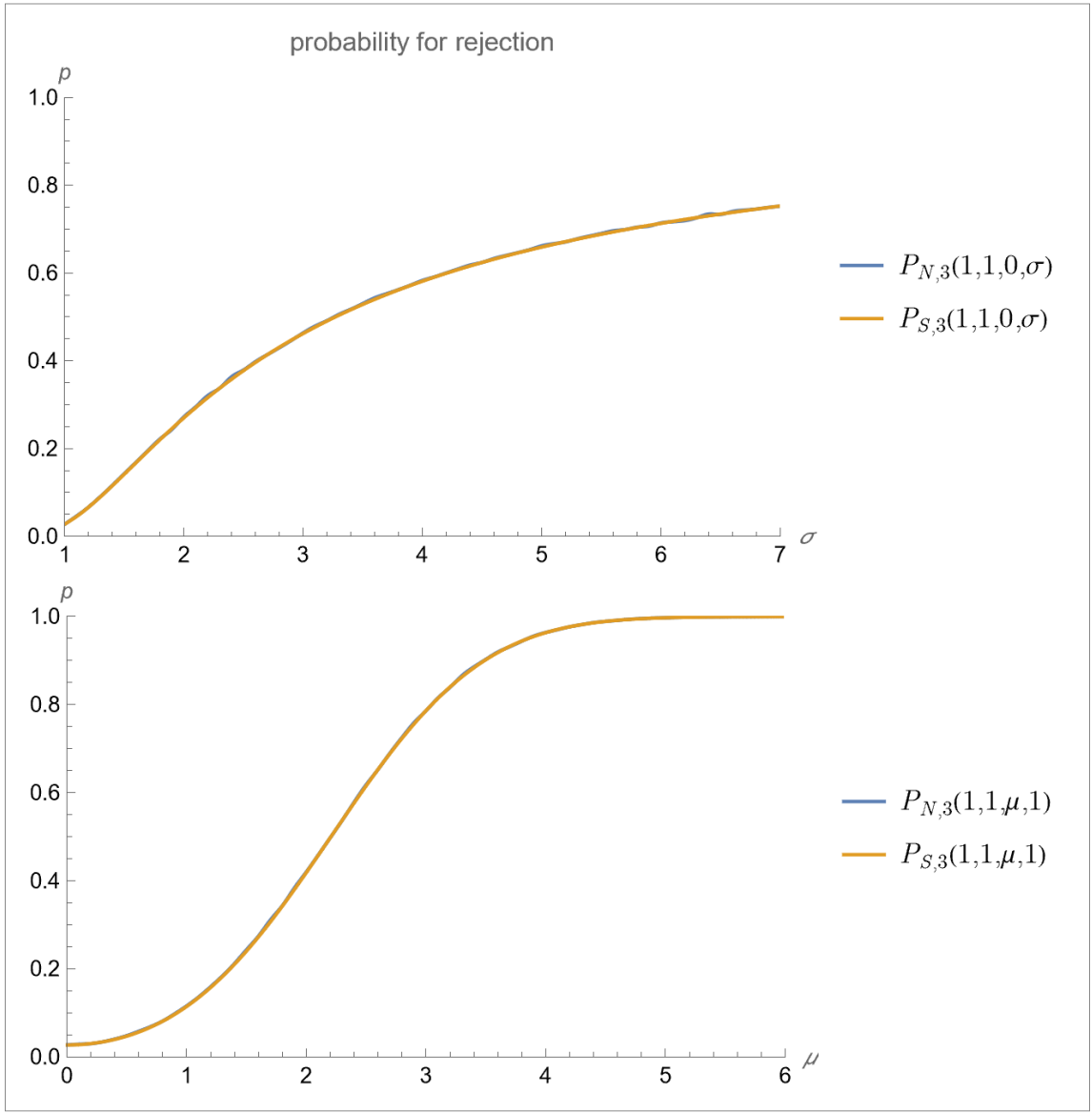

*Figure 21: $P_{N,3}(1,1,\mu,\sigma)$ vs $P_{S,3}(1,1,\mu,\sigma)$*

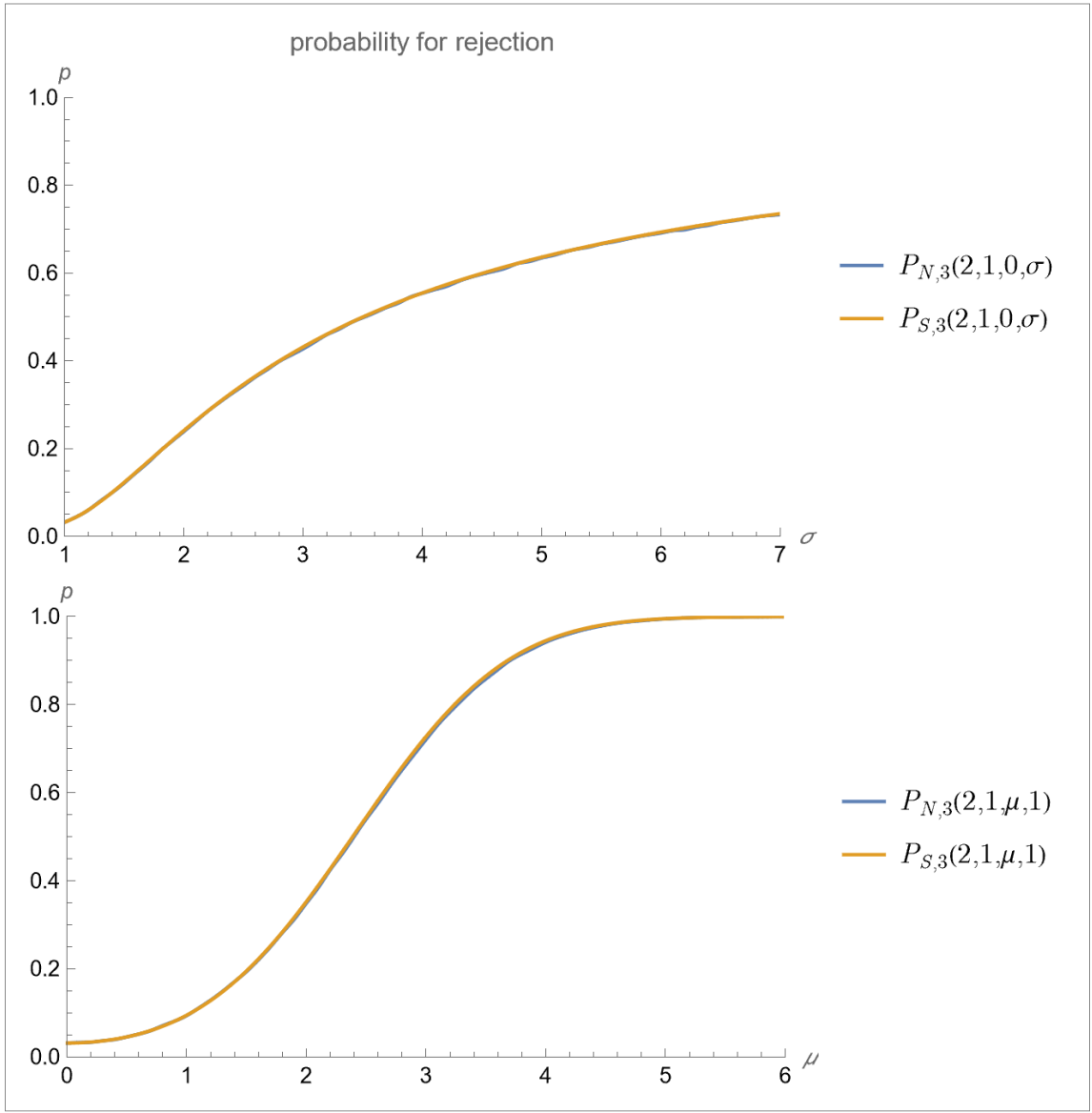

*Figure 22: $P_{N,3}(2,1,\mu,\sigma)$ vs $P_{S,3}(2,1,\mu,\sigma)$*

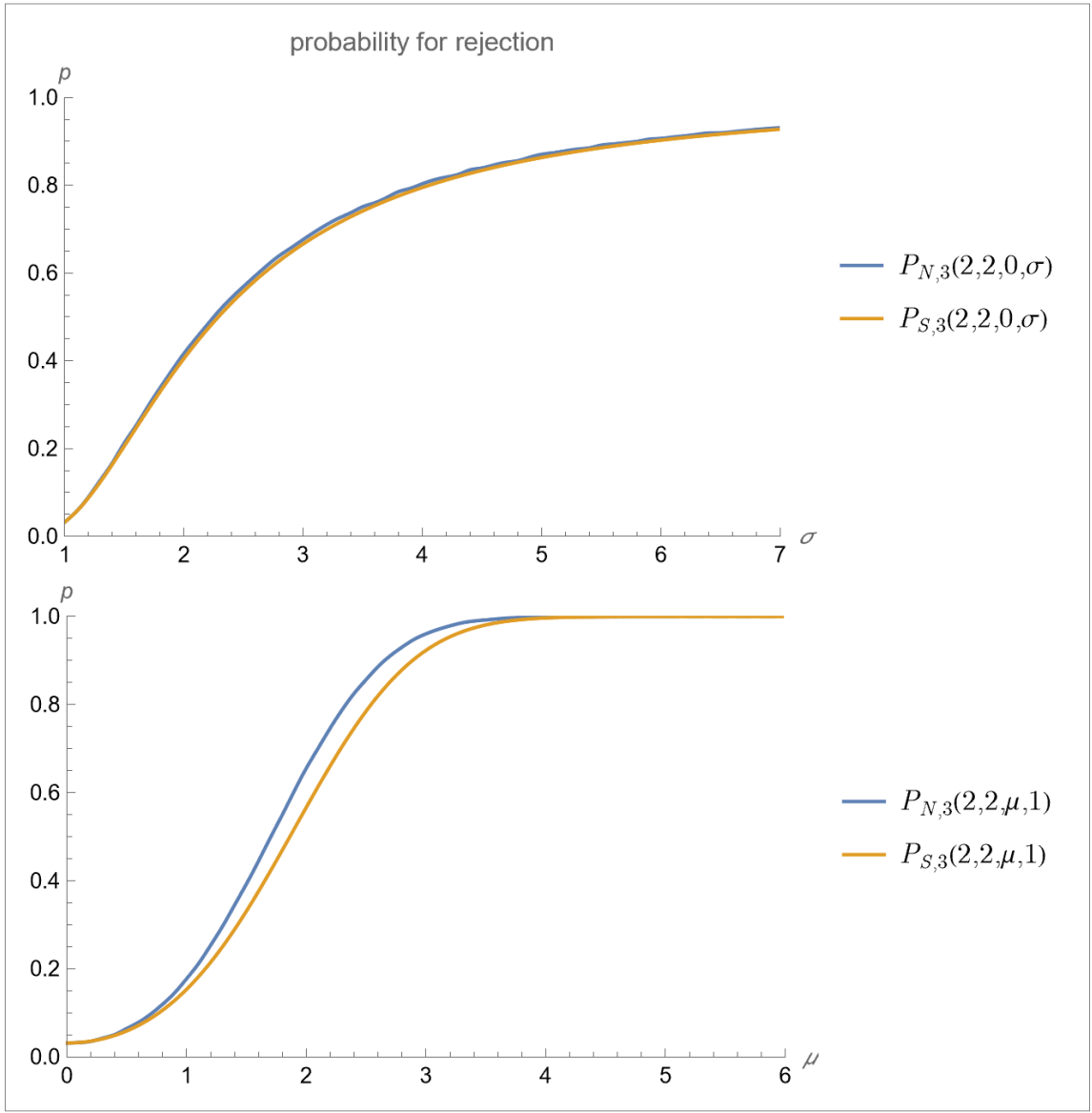

*Figure 23: $P_{N,3}(2,2,\mu,\sigma)$ vs $P_{S,3}(2,2,\mu,\sigma)$*

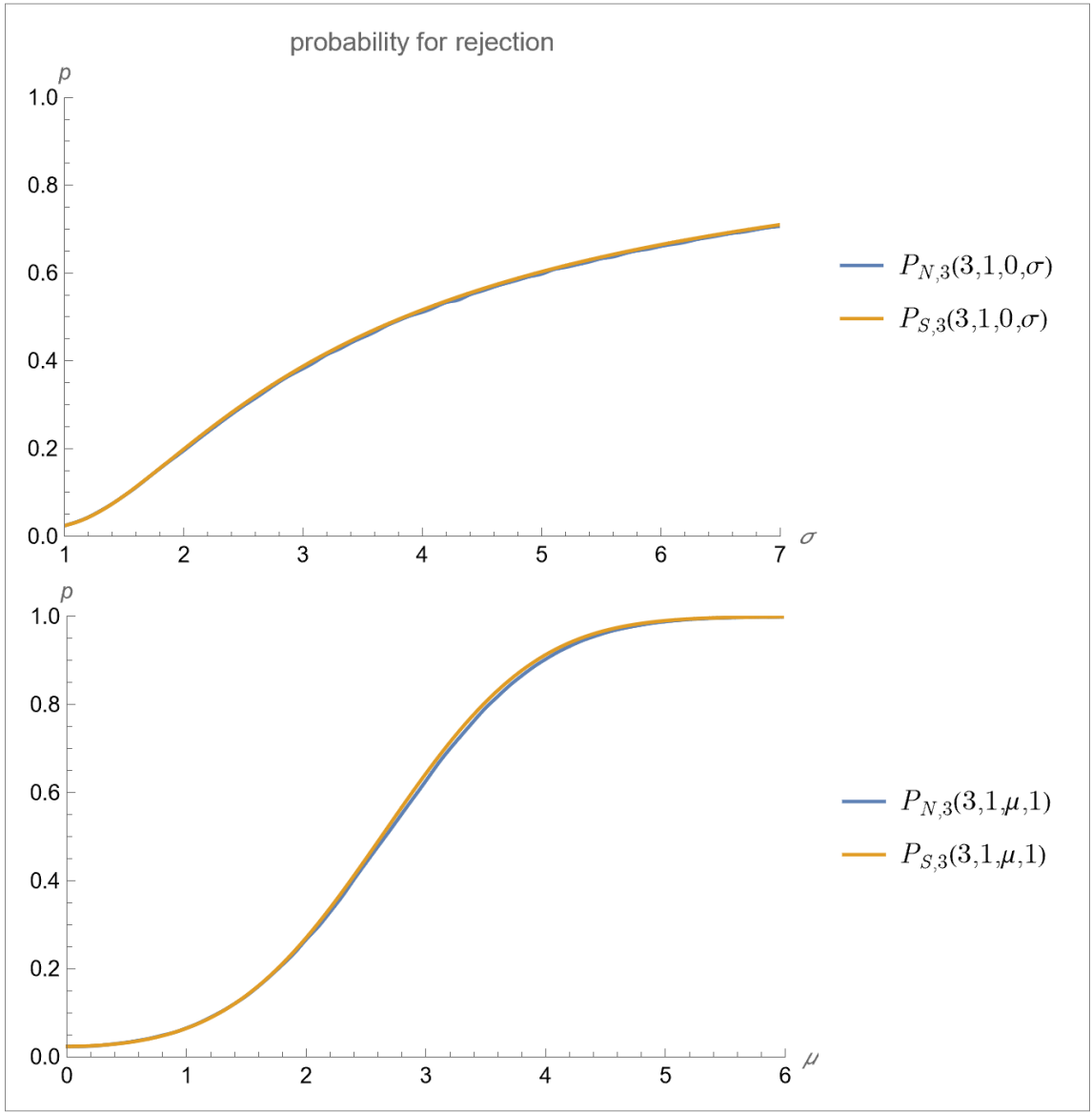

*Figure 24: $P_{N,3}(3,1,\mu,\sigma)$ vs $P_{S,3}(3,1,\mu,\sigma)$*

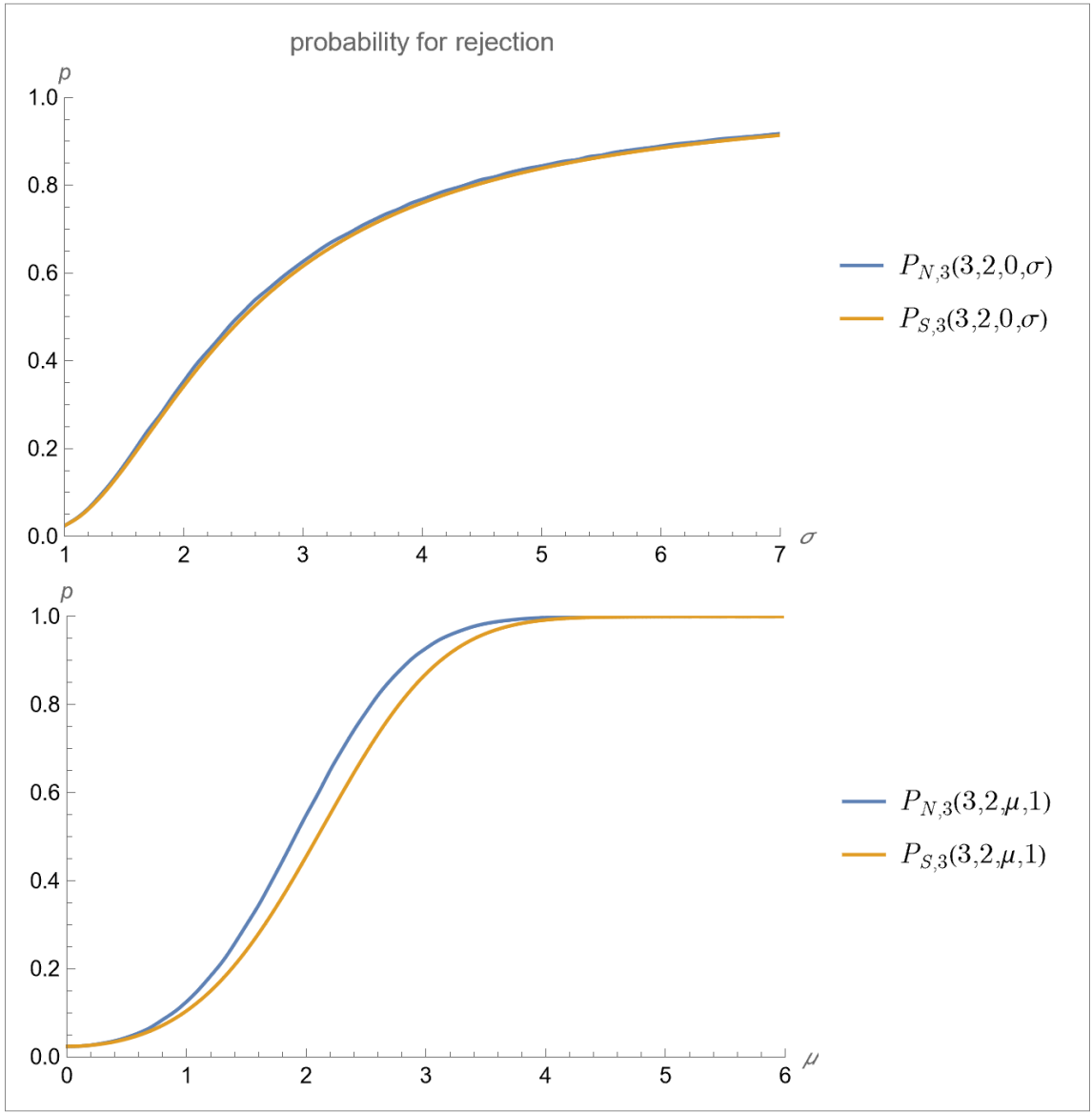

*Figure 25: $P_{N,3}(3,2,\mu,\sigma)$ vs $P_{S,3}(3,2,\mu,\sigma)$*

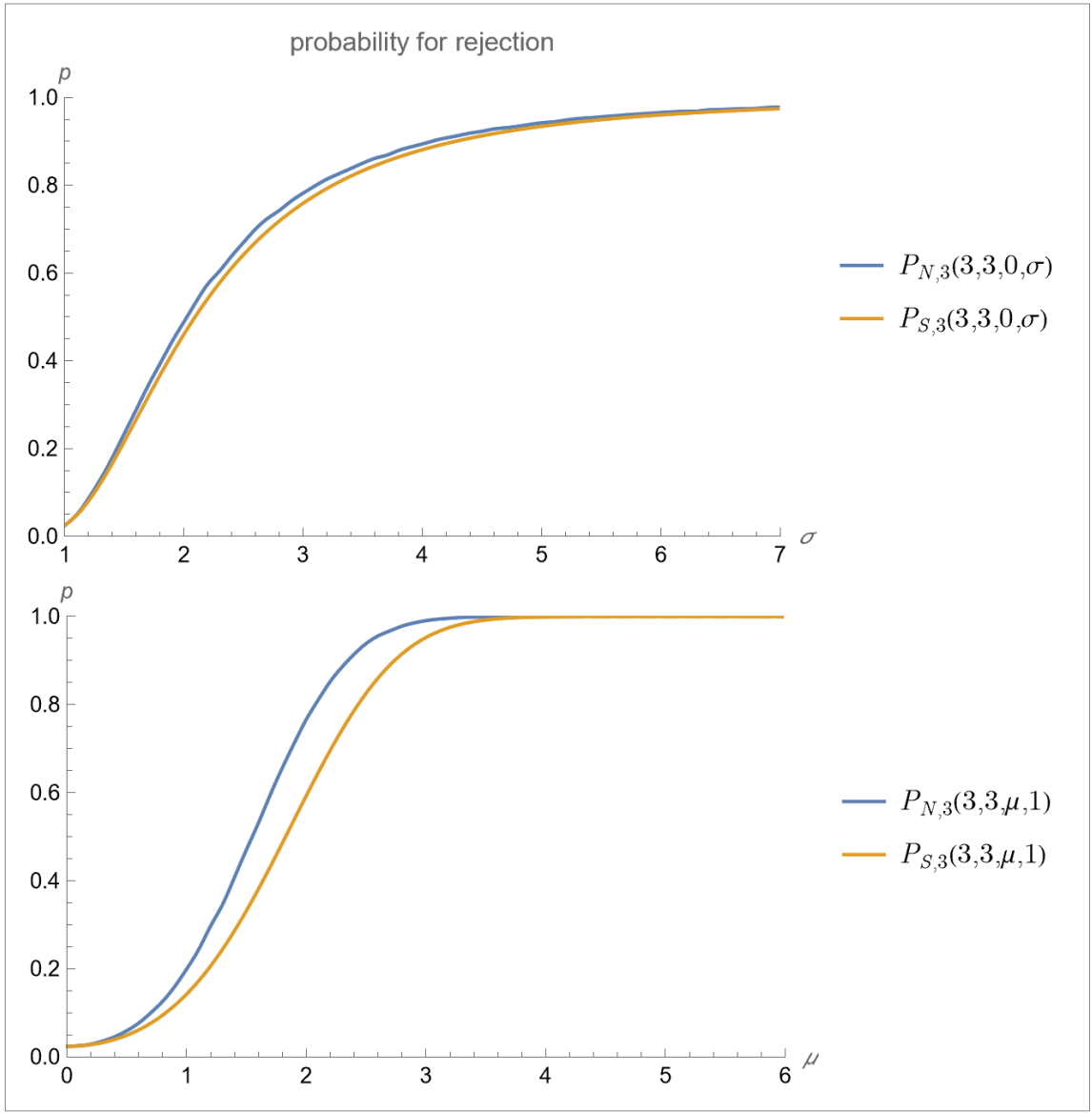

*Figure 26: $P_{N,3}(3,3,\mu,\sigma)$ vs $P_{S,3}(3,3,\mu,\sigma)$*

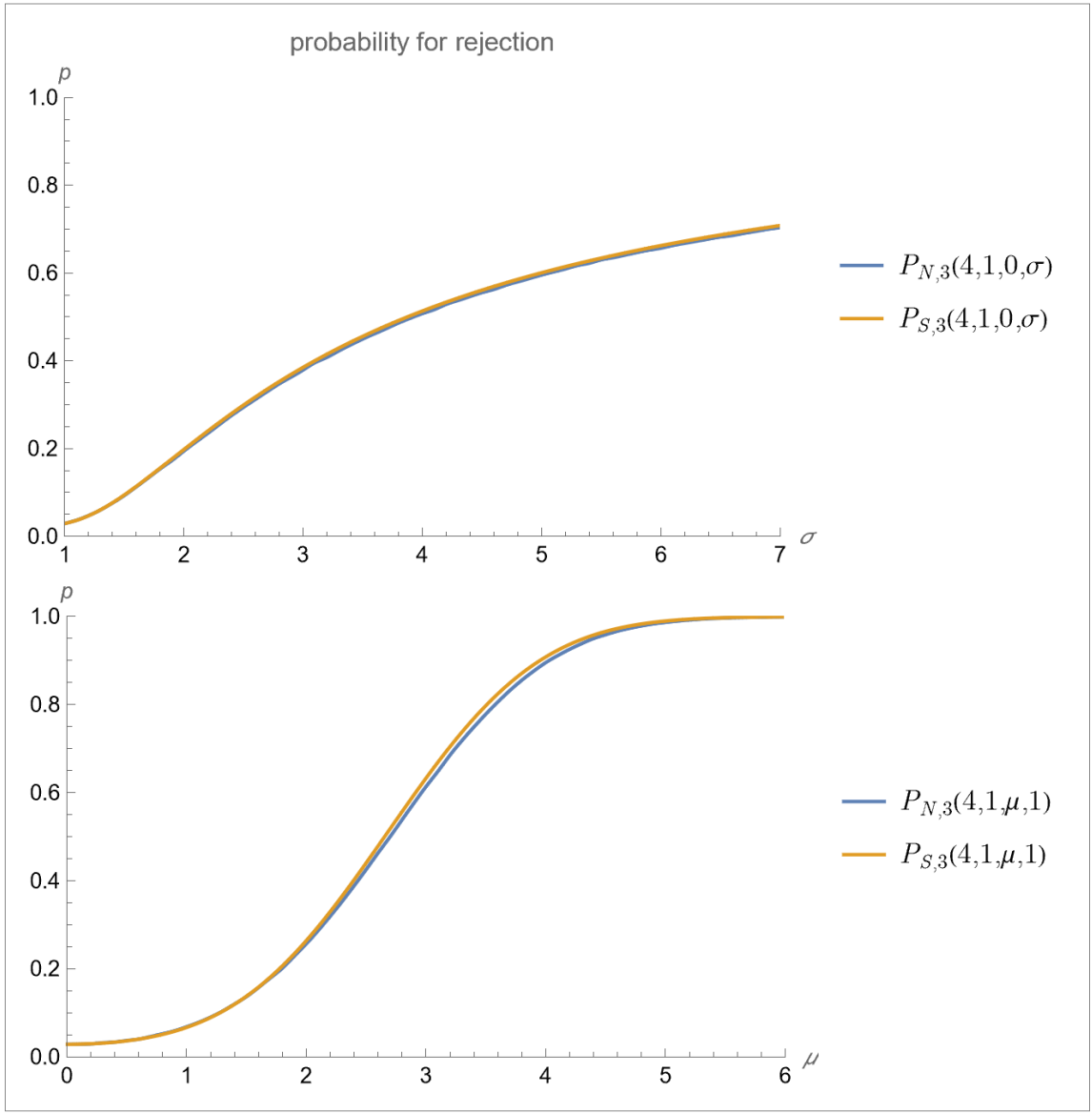

*Figure 27: $P_{N,3}(4,1,\mu,\sigma)$ vs $P_{S,3}(4,1,\mu,\sigma)$*

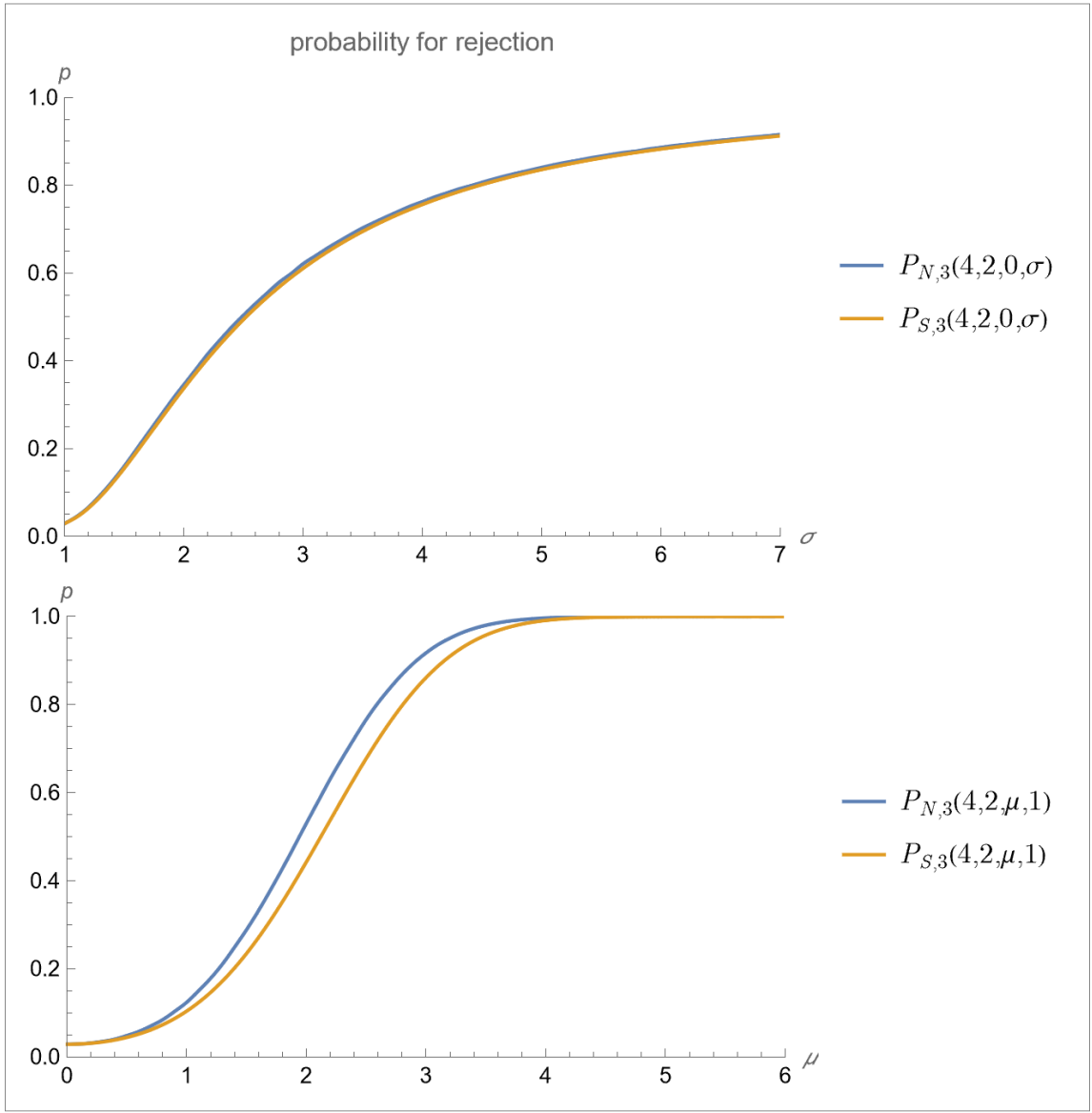

*Figure 28: $P_{N,3}(4,2,\mu,\sigma)$ vs $P_{S,3}(4,2,\mu,\sigma)$*

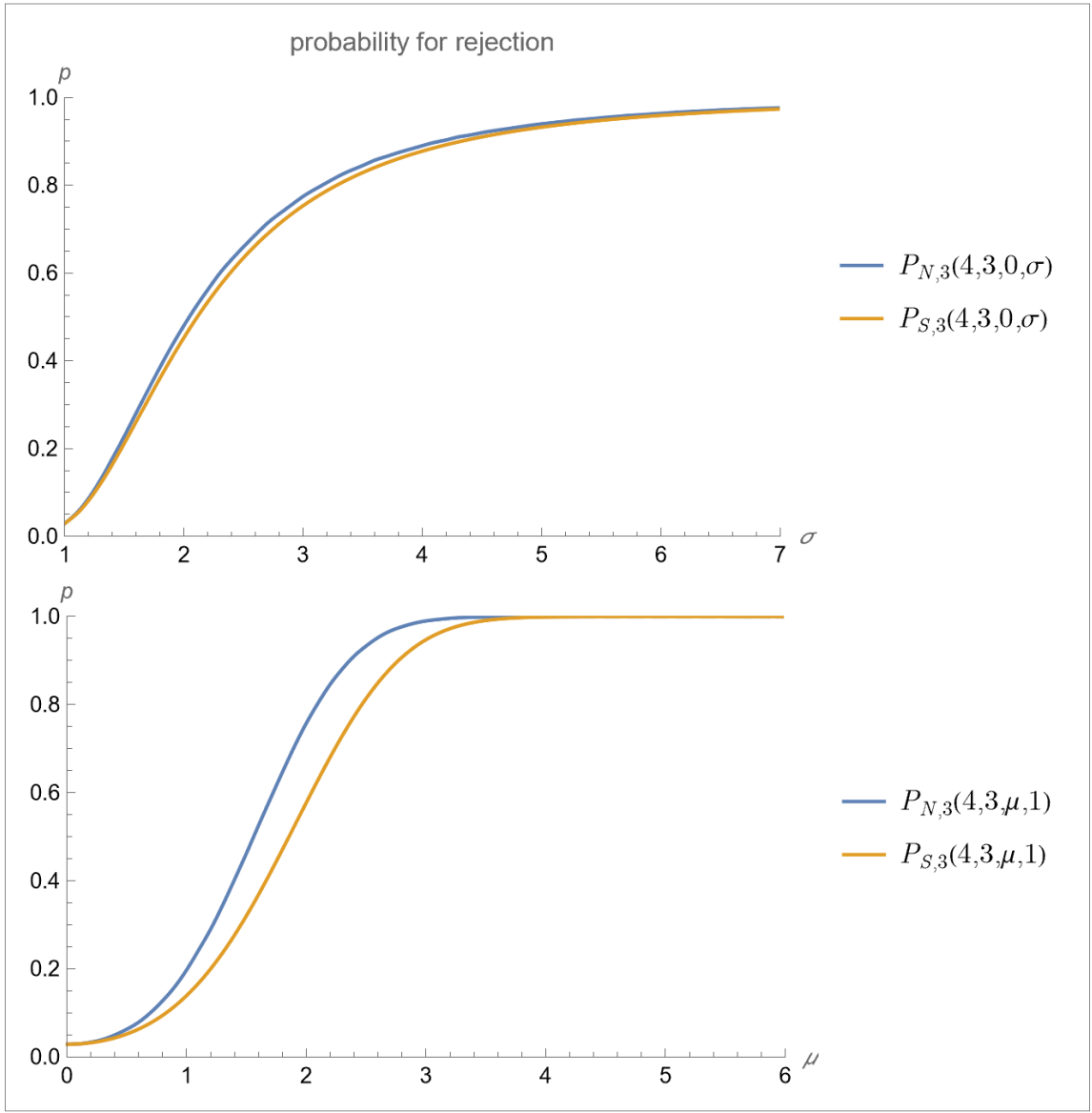

*Figure 29: $P_{N,3}(4,3,\mu,\sigma)$ vs $P_{S,3}(4,3,\mu,\sigma)$*

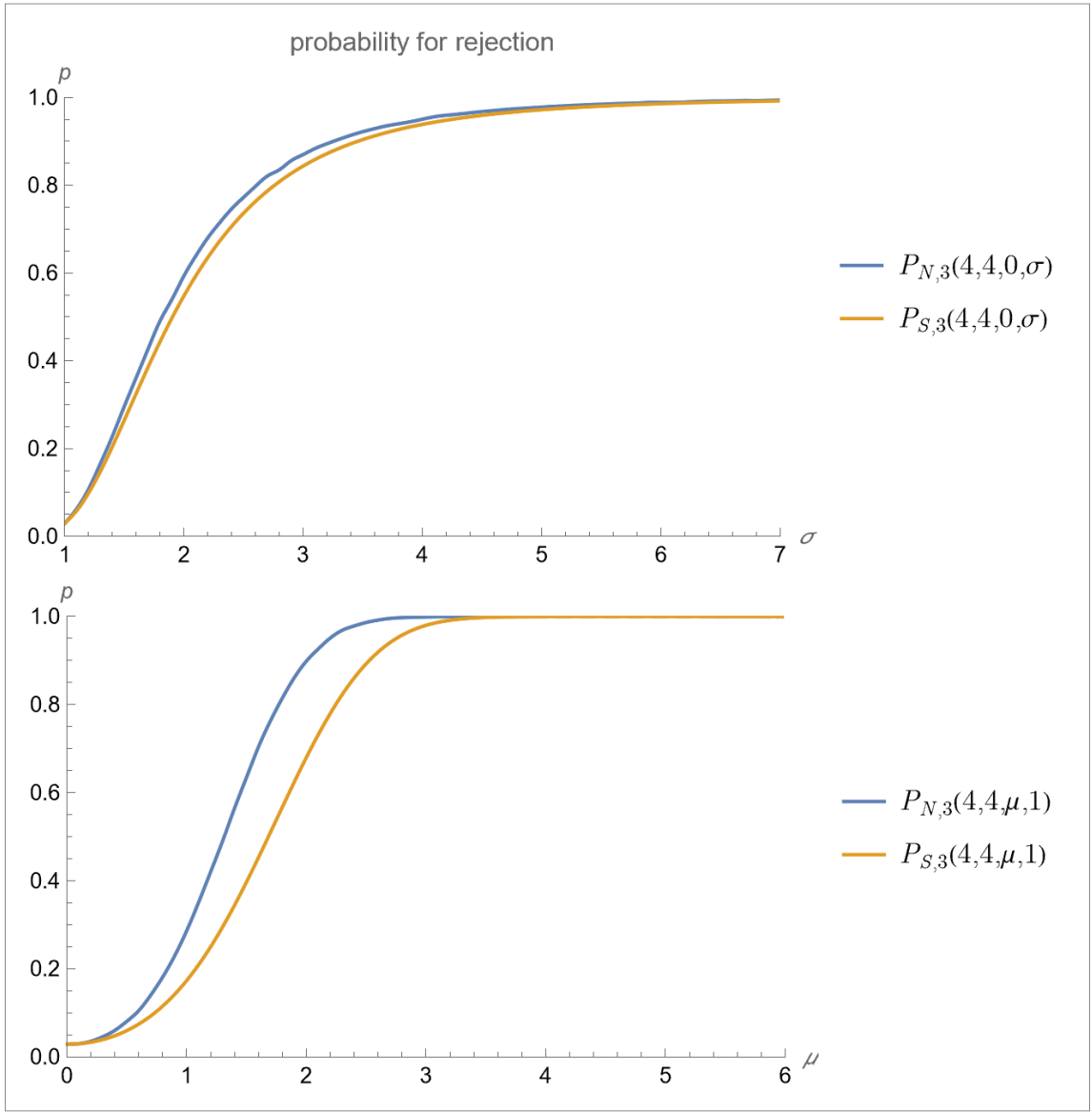

*Figure 30: $P_{N,3}(4,4,\mu,\sigma)$ vs $P_{S,3}(4,4,\mu,\sigma)$*

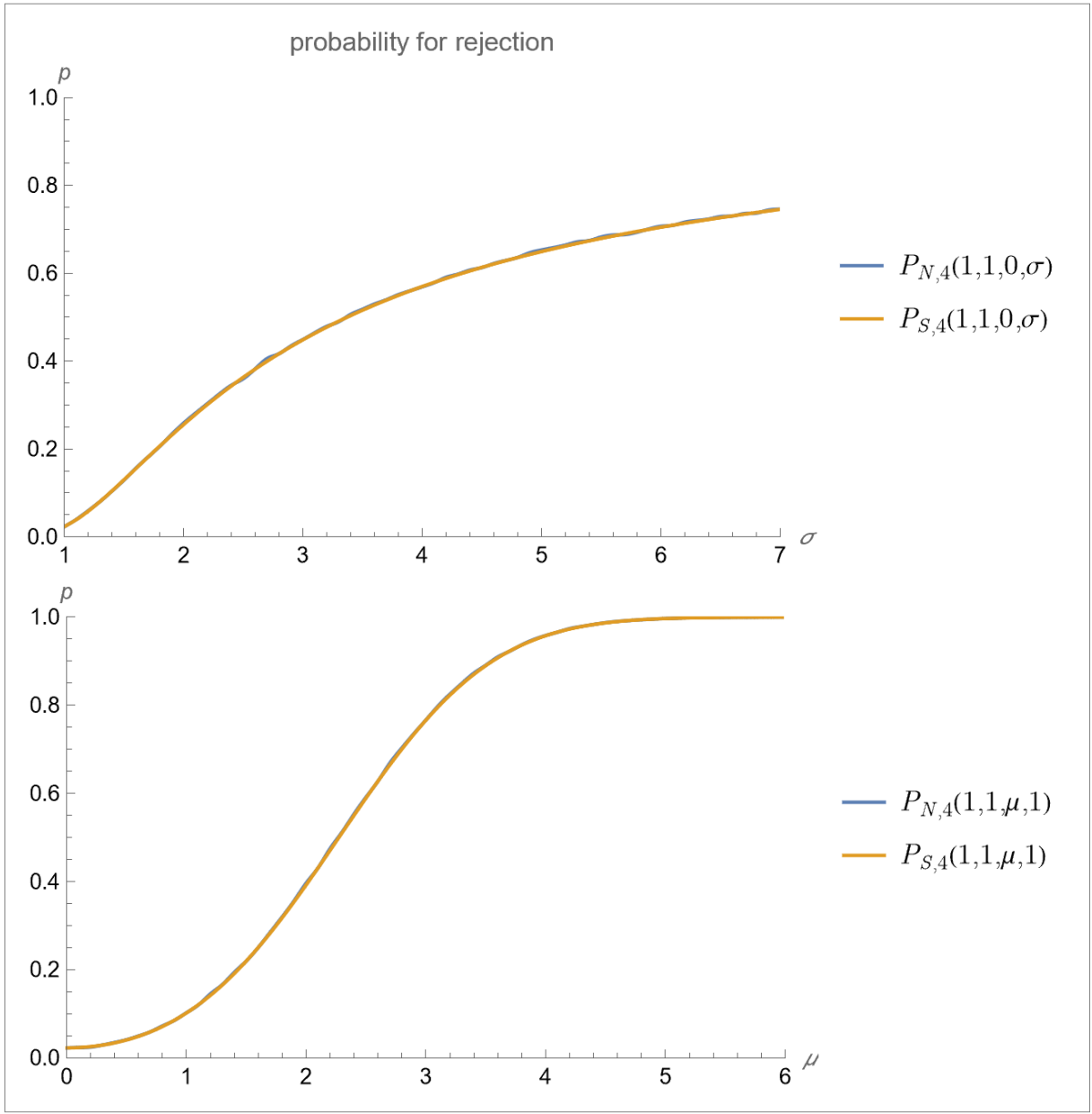

*Figure 31: $P_{N,4}(1,1,\mu,\sigma)$ vs $P_{S,4}(1,1,\mu,\sigma)$*

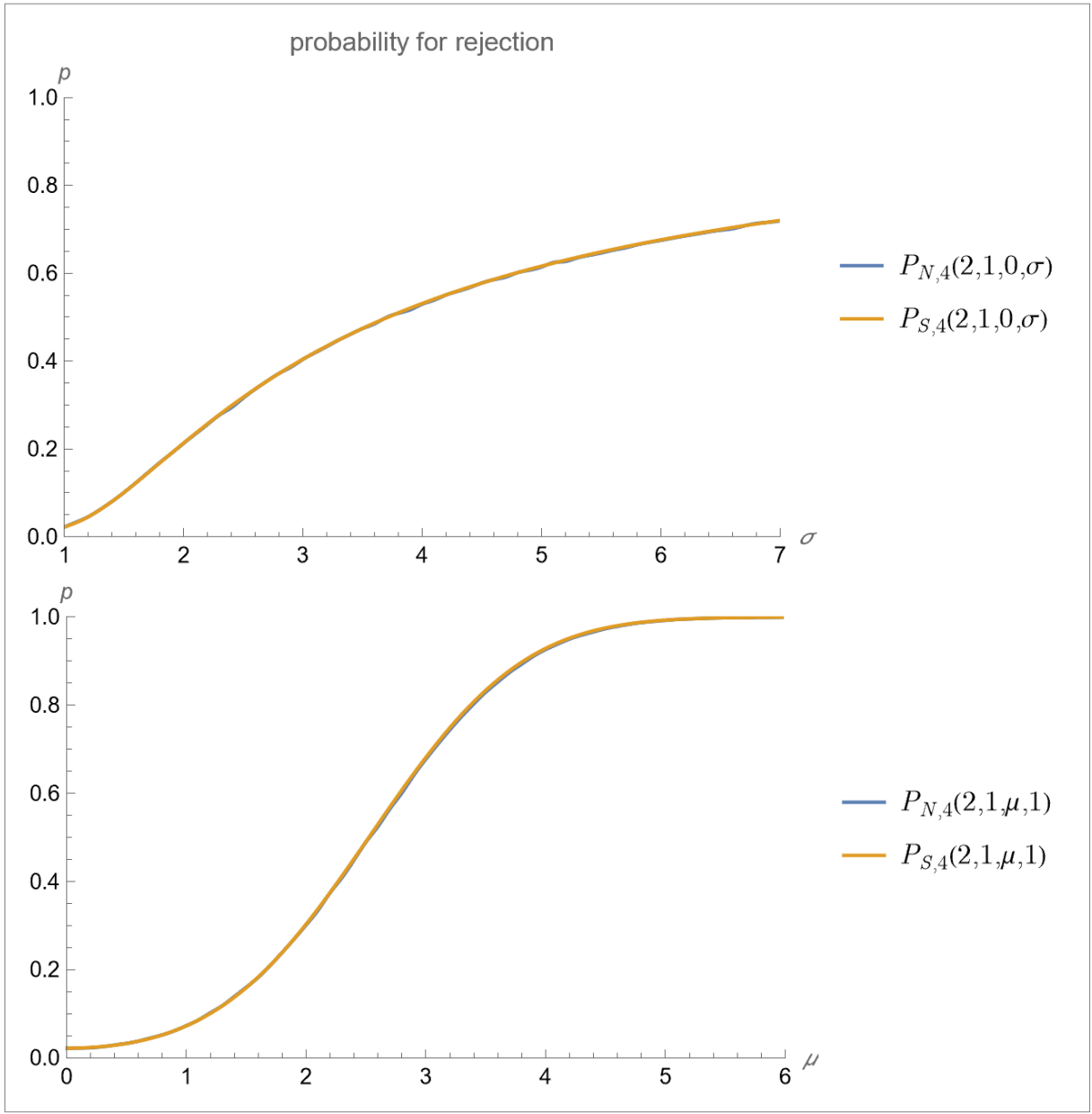

*Figure 32: $P_{N,4}(2,1,\mu,\sigma)$ vs $P_{S,4}(2,1,\mu,\sigma)$*

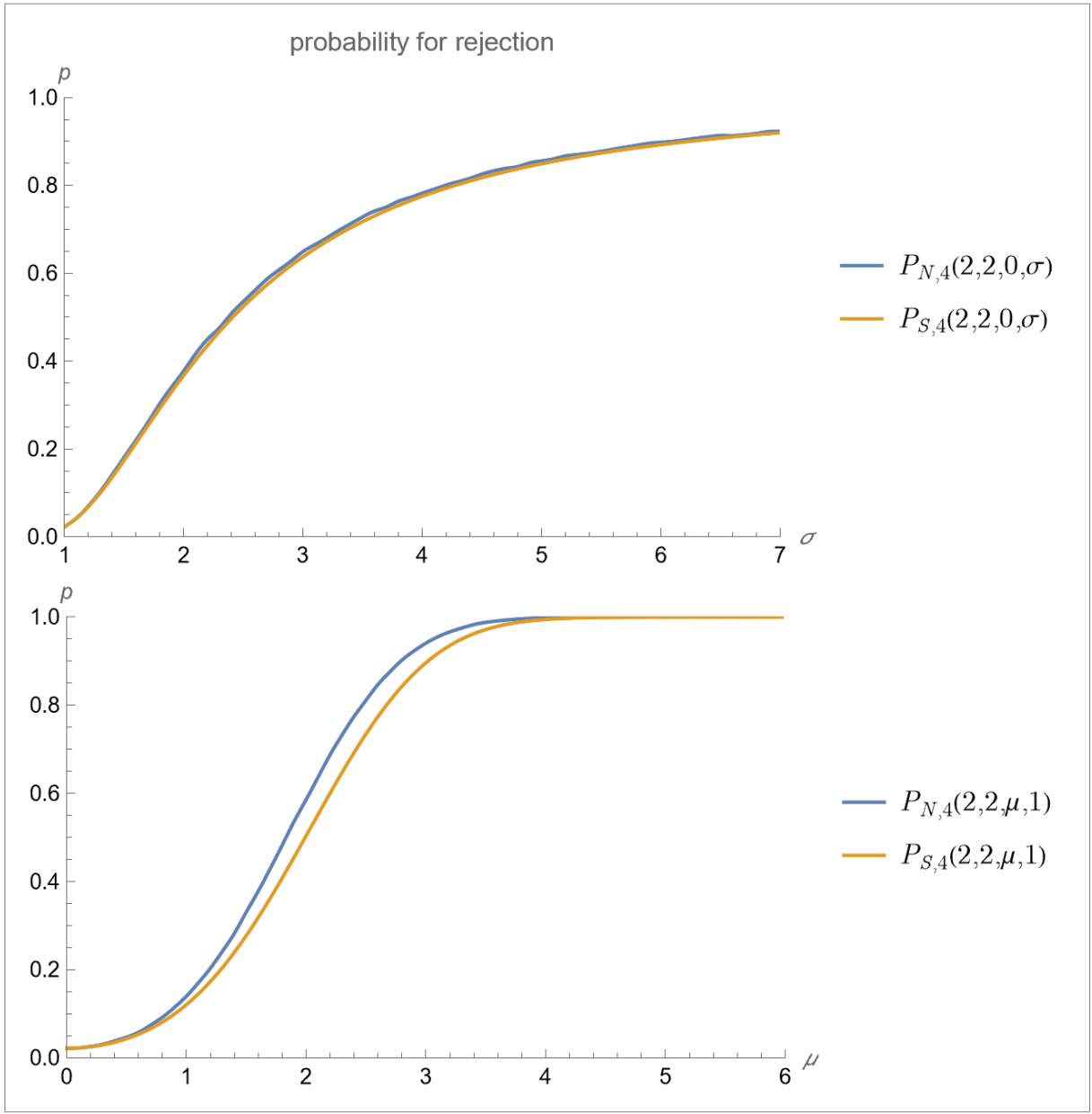

*Figure 33: $P_{N,4}(2,2,\mu,\sigma)$ vs $P_{S,4}(2,2,\mu,\sigma)$*

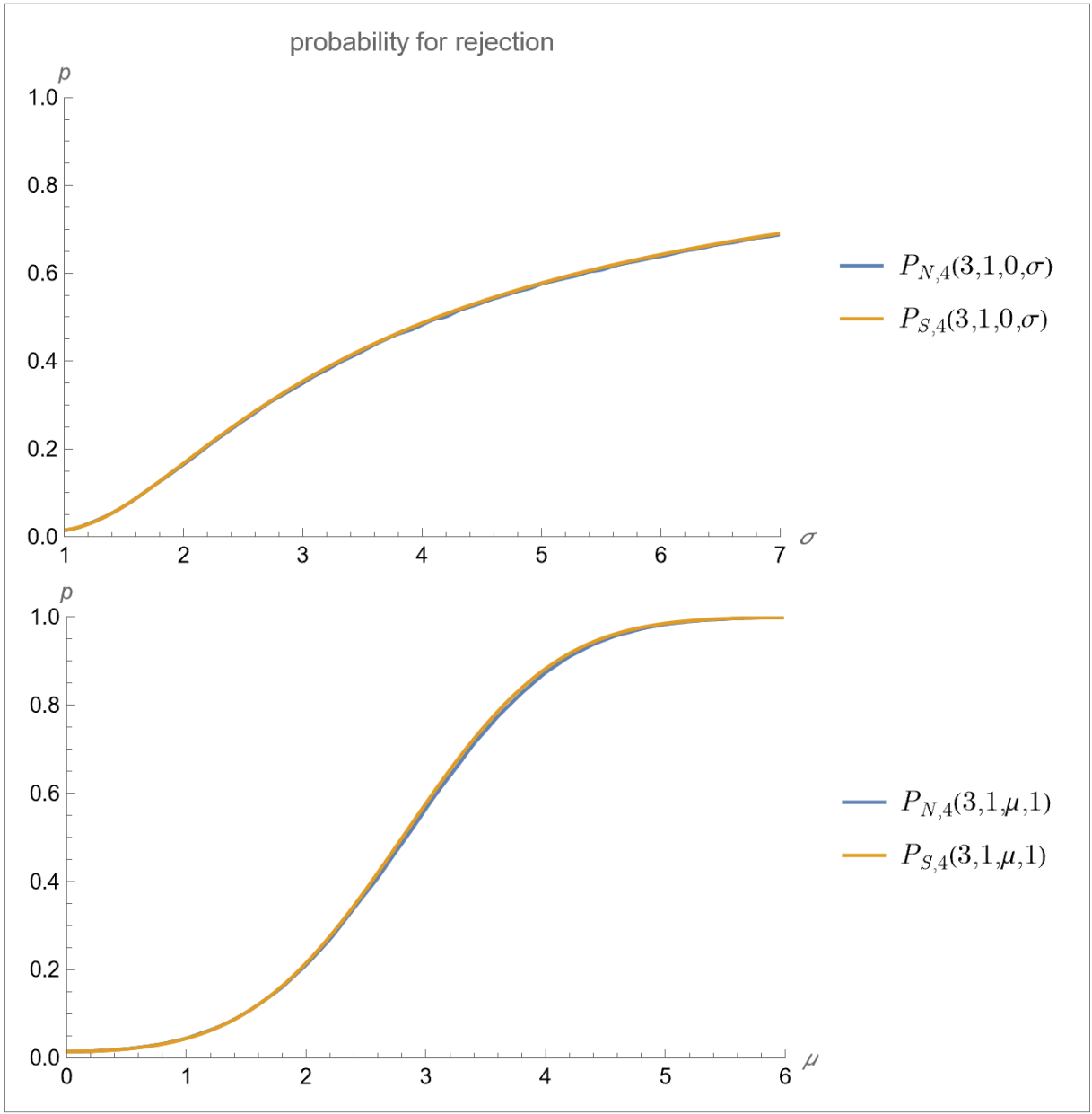

*Figure 34: $P_{N,4}(3,1,\mu,\sigma)$ vs $P_{S,4}(3,1,\mu,\sigma)$*

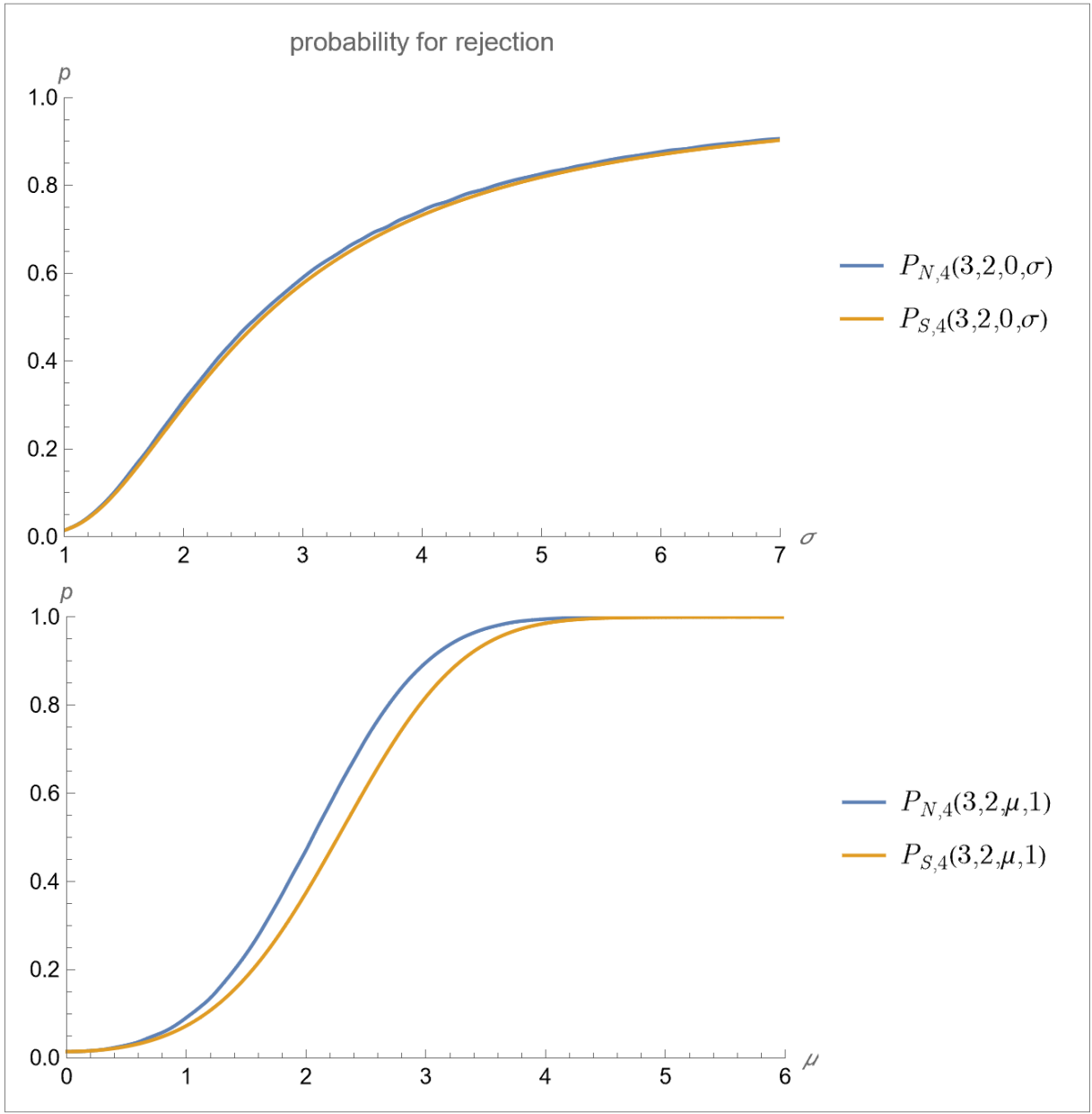

*Figure 35: $P_{N,4}(3,2,\mu,\sigma)$ vs $P_{S,4}(3,2,\mu,\sigma)$*

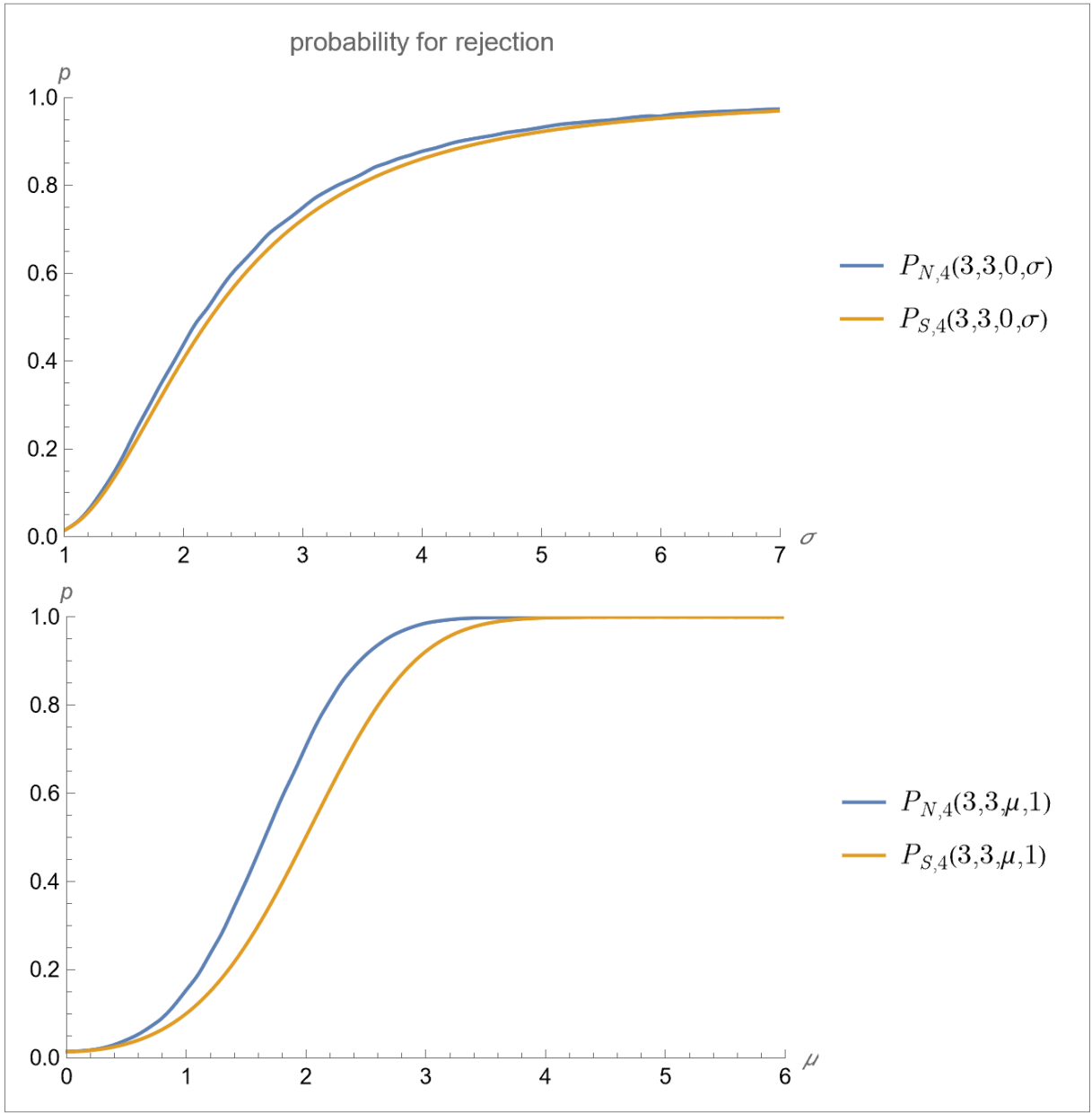

*Figure 36: $P_{N,4}(3,3,\mu,\sigma)$ vs $P_{S,4}(3,3,\mu,\sigma)$*

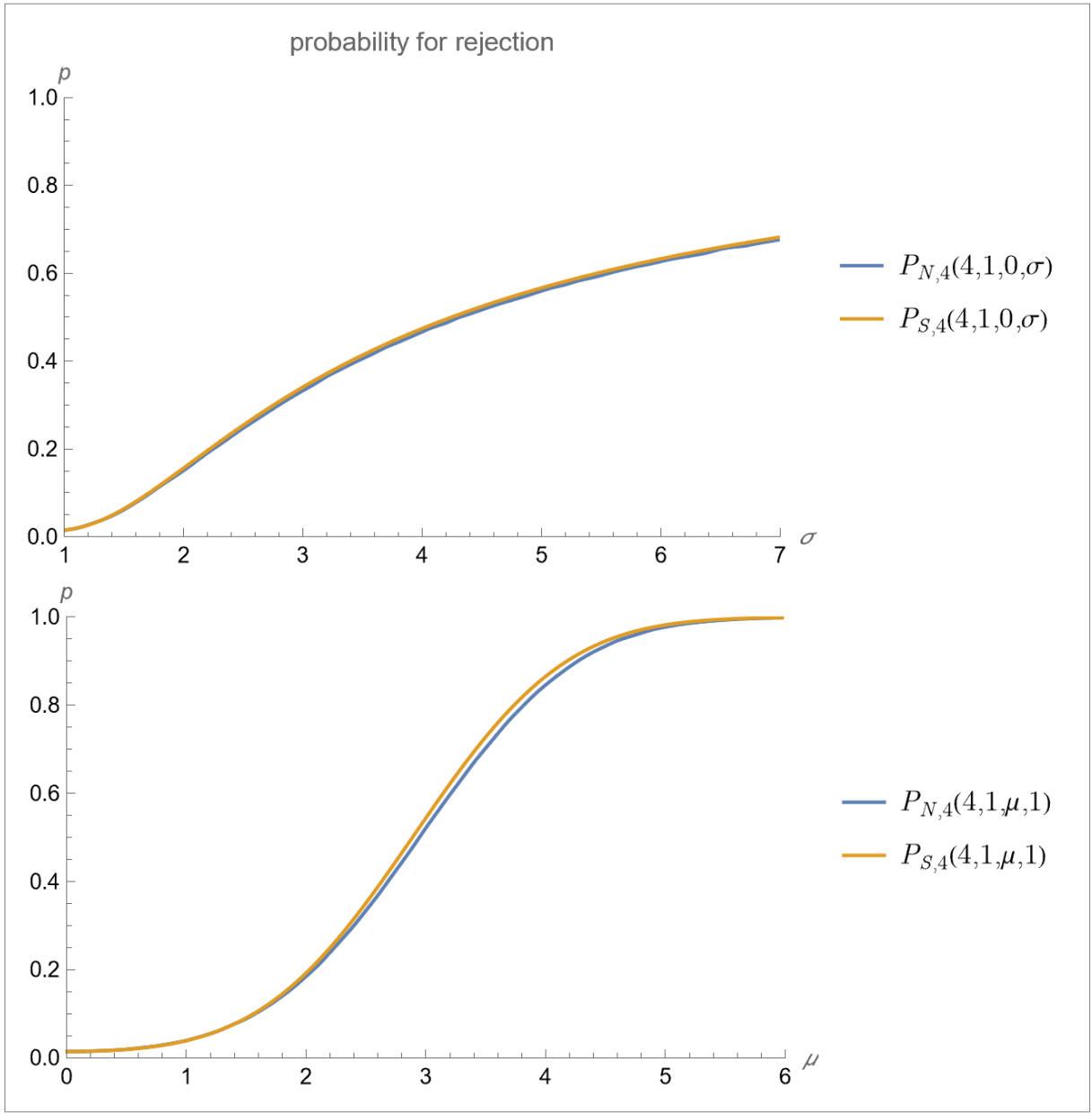

*Figure 37: $P_{N,4}(4,1,\mu,\sigma)$ vs $P_{S,4}(4,1,\mu,\sigma)$*

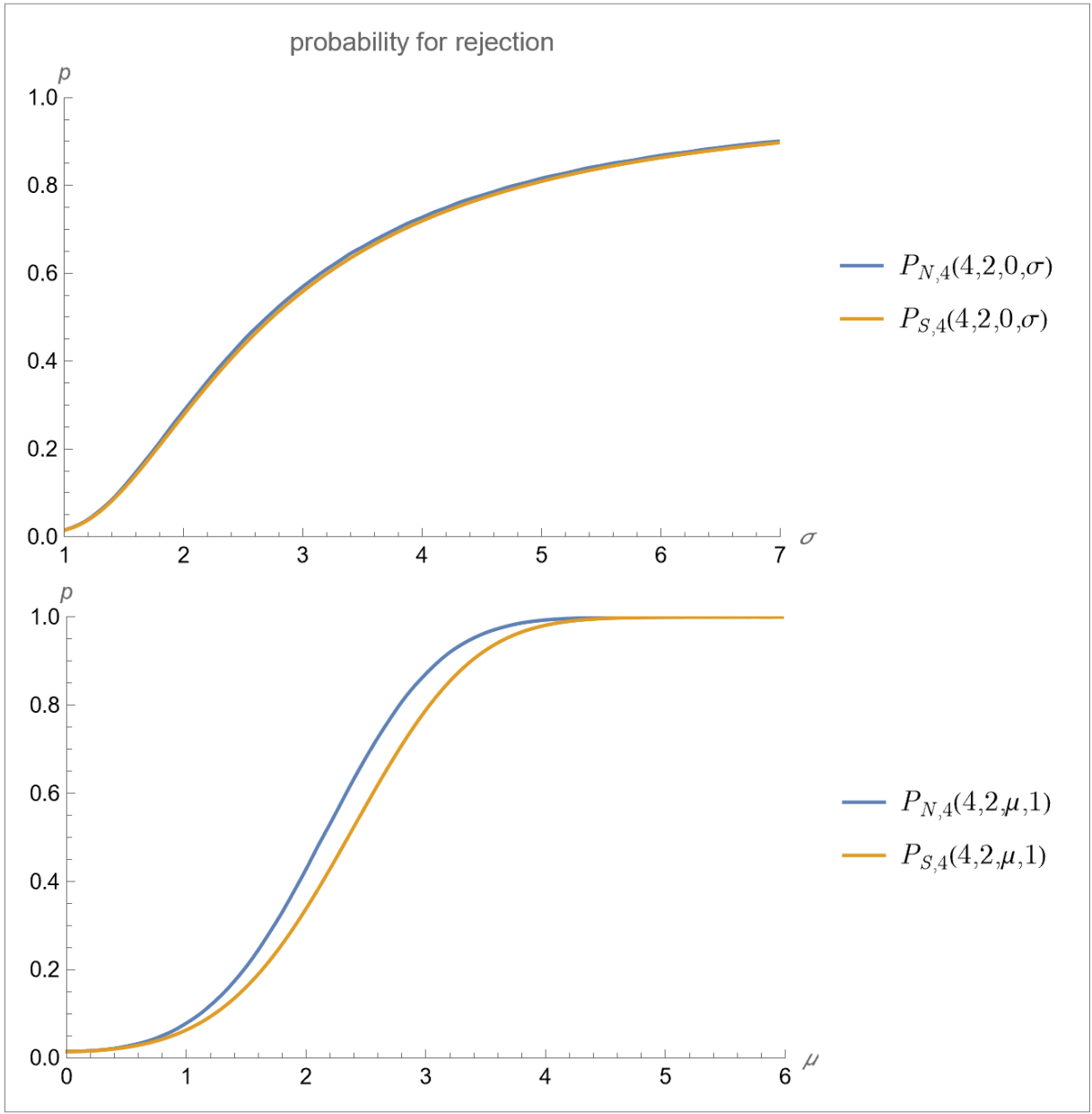

*Figure 38:* $P_{N,4}(4,2,\mu,\sigma)$ vs $P_{S,4}(4,2,\mu,\sigma)$

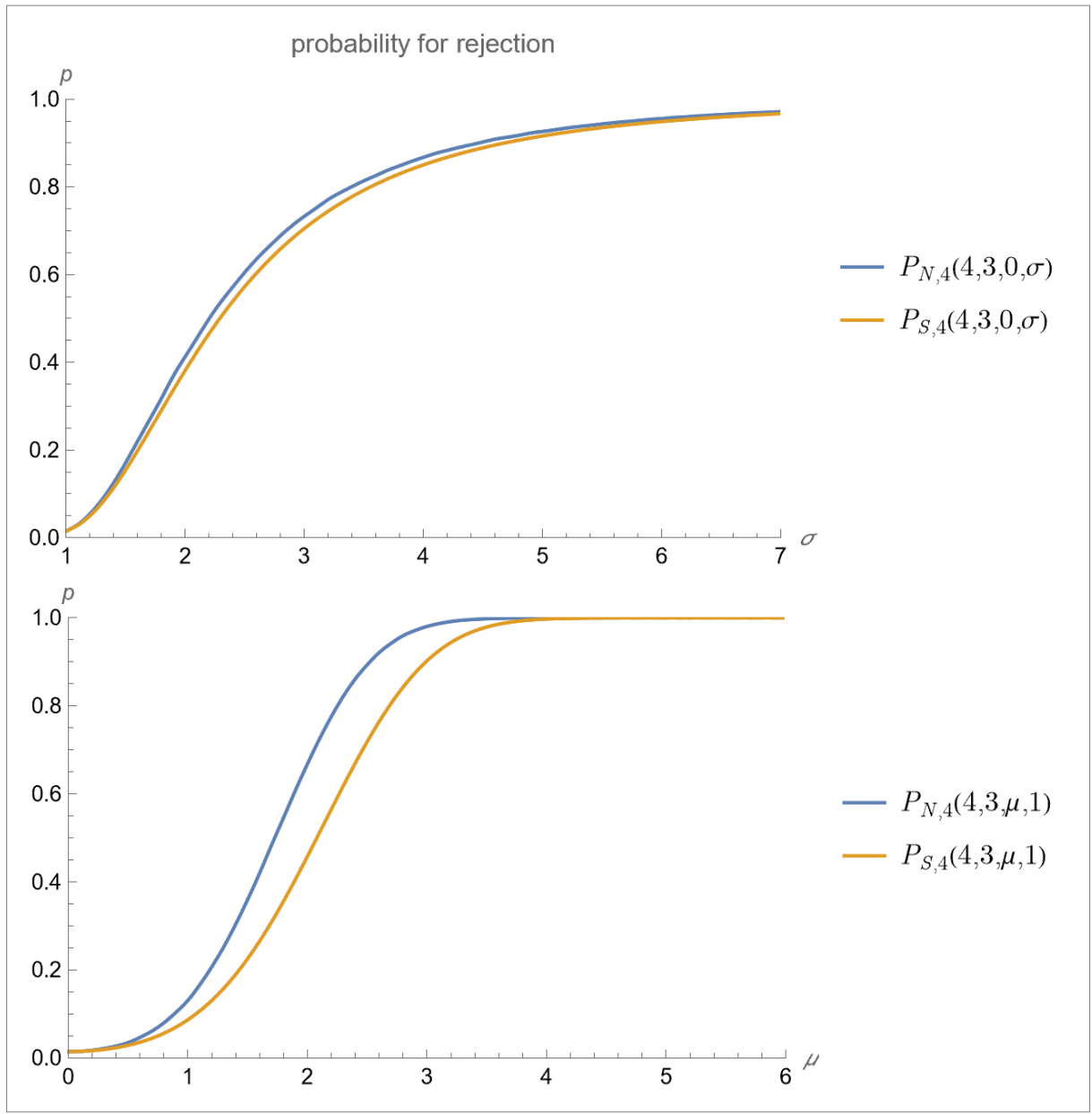

*Figure 39: $P_{N,4}(4,3,\mu,\sigma)$ vs $P_{S,4}(4,3,\mu,\sigma)$*

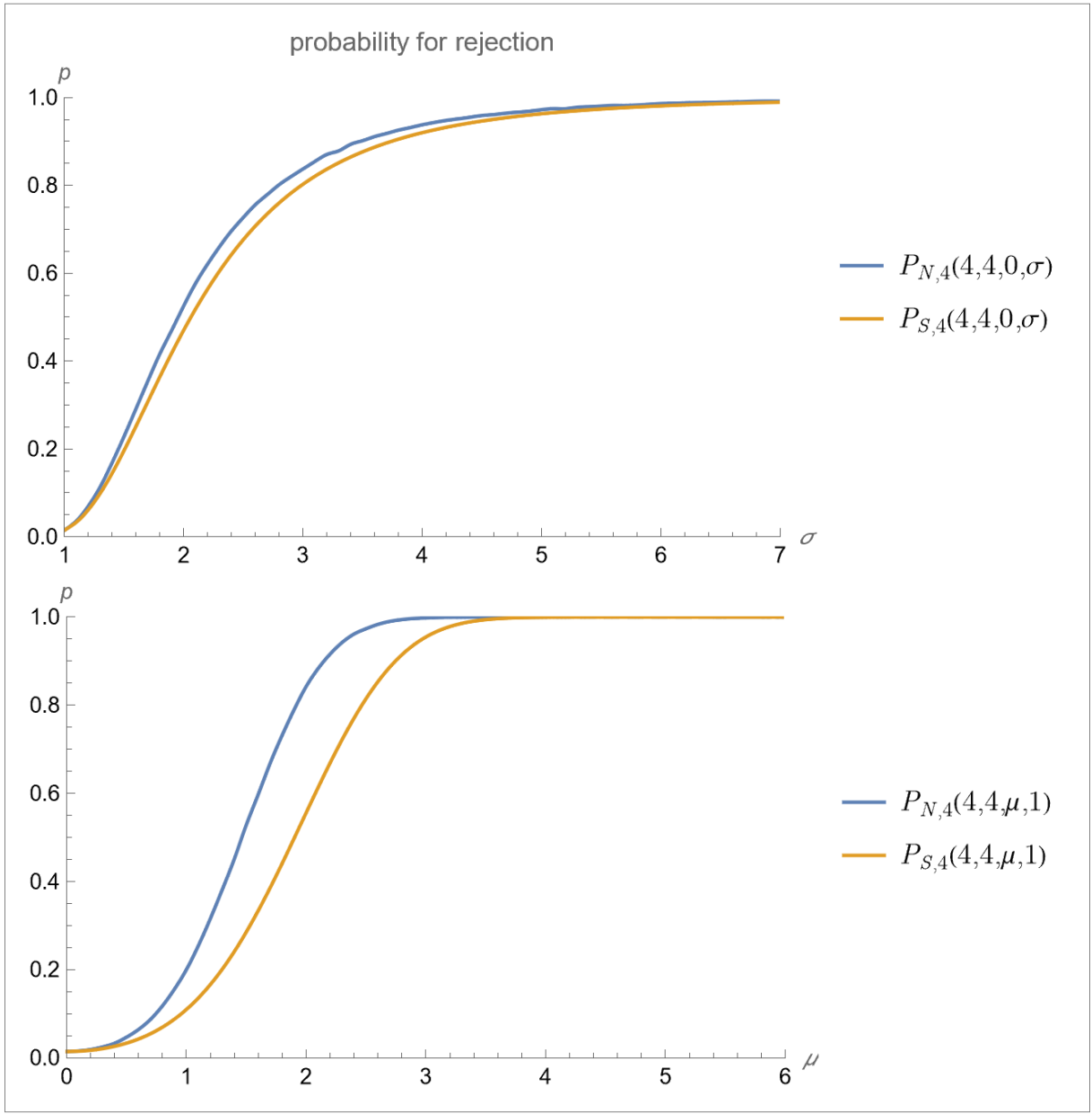

*Figure 40: $P_{N,4}(4,4,\mu,\sigma)$ vs $P_{S,4}(4,4,\mu,\sigma)$*

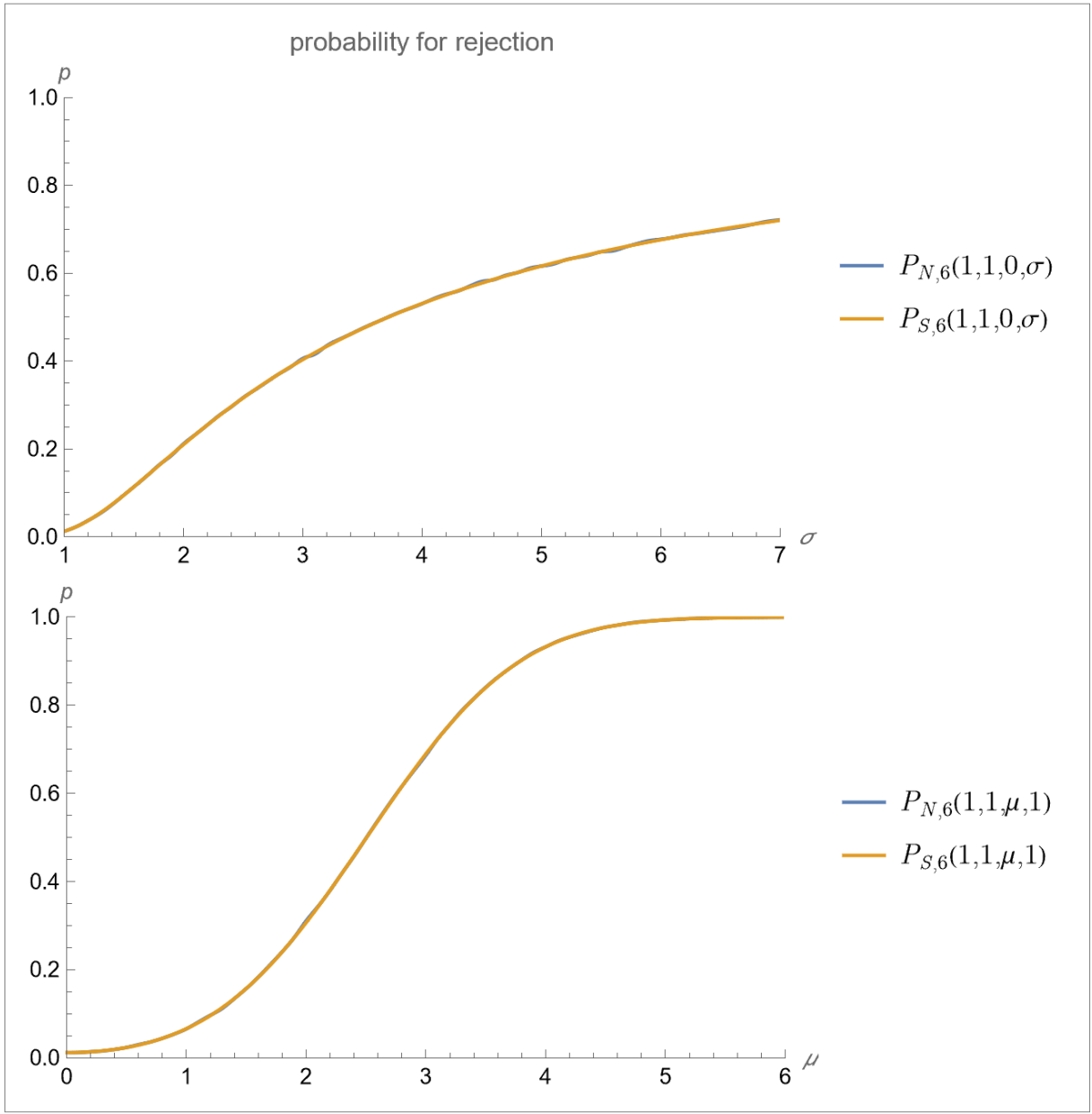

*Figure 41: $P_{N,6}(1,1,\mu,\sigma)$ vs $P_{S,6}(1,1,\mu,\sigma)$*

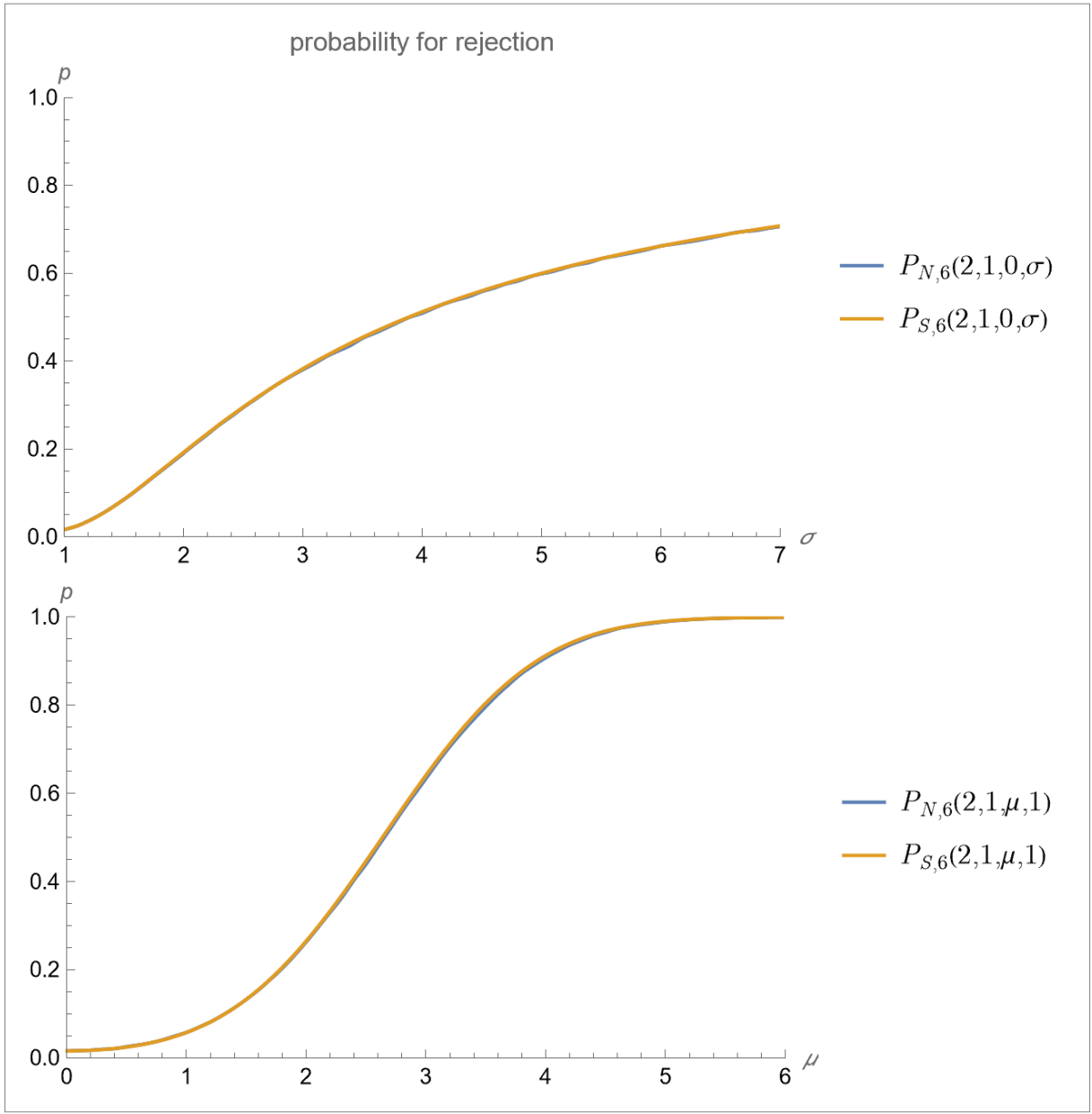

*Figure 42: $P_{N,6}(2,1,\mu,\sigma)$ vs $P_{S,6}(2,1,\mu,\sigma)$*

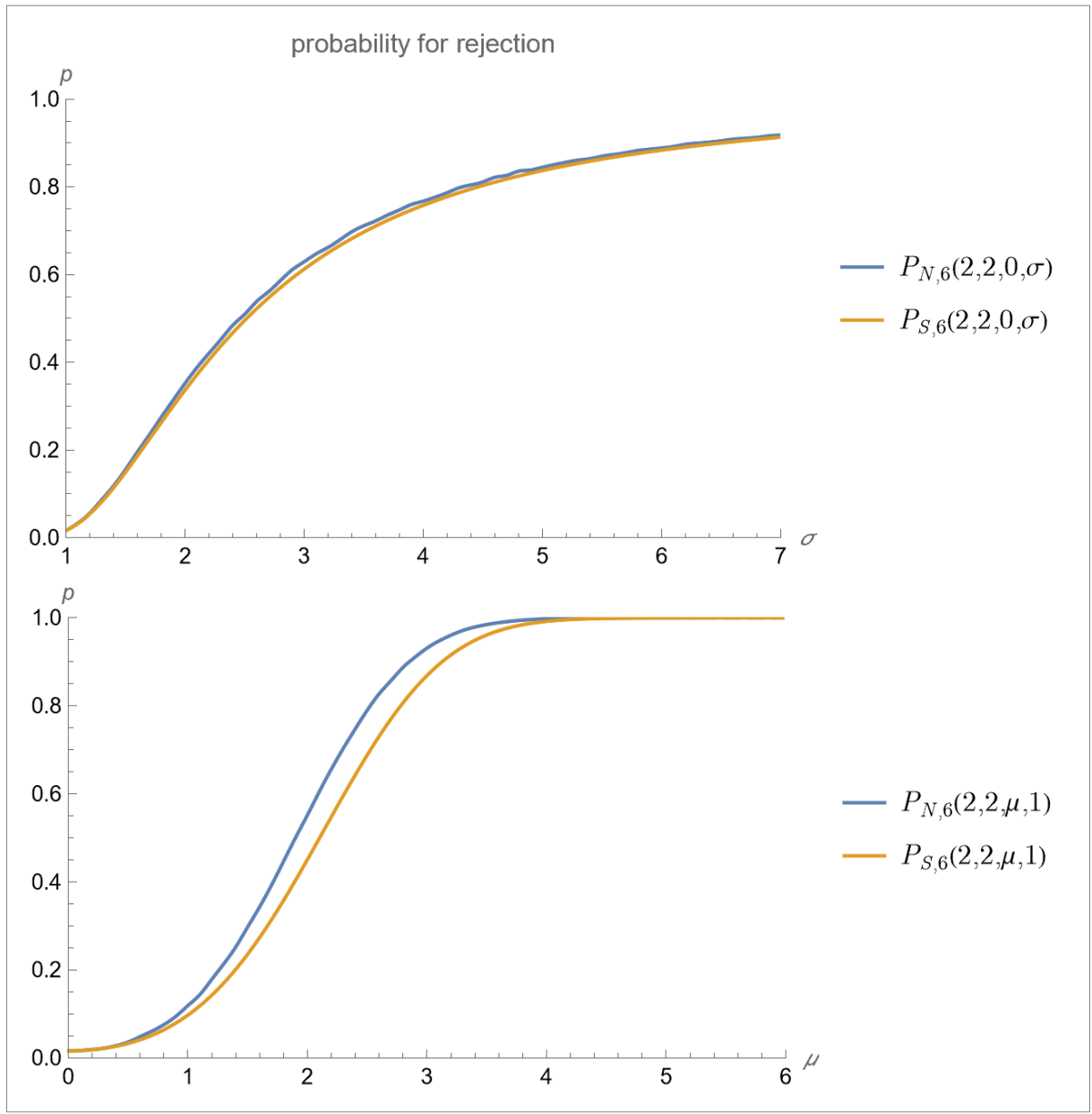

*Figure 43: $P_{N,6}(2,2,\mu,\sigma)$ vs $P_{S,6}(2,2,\mu,\sigma)$*

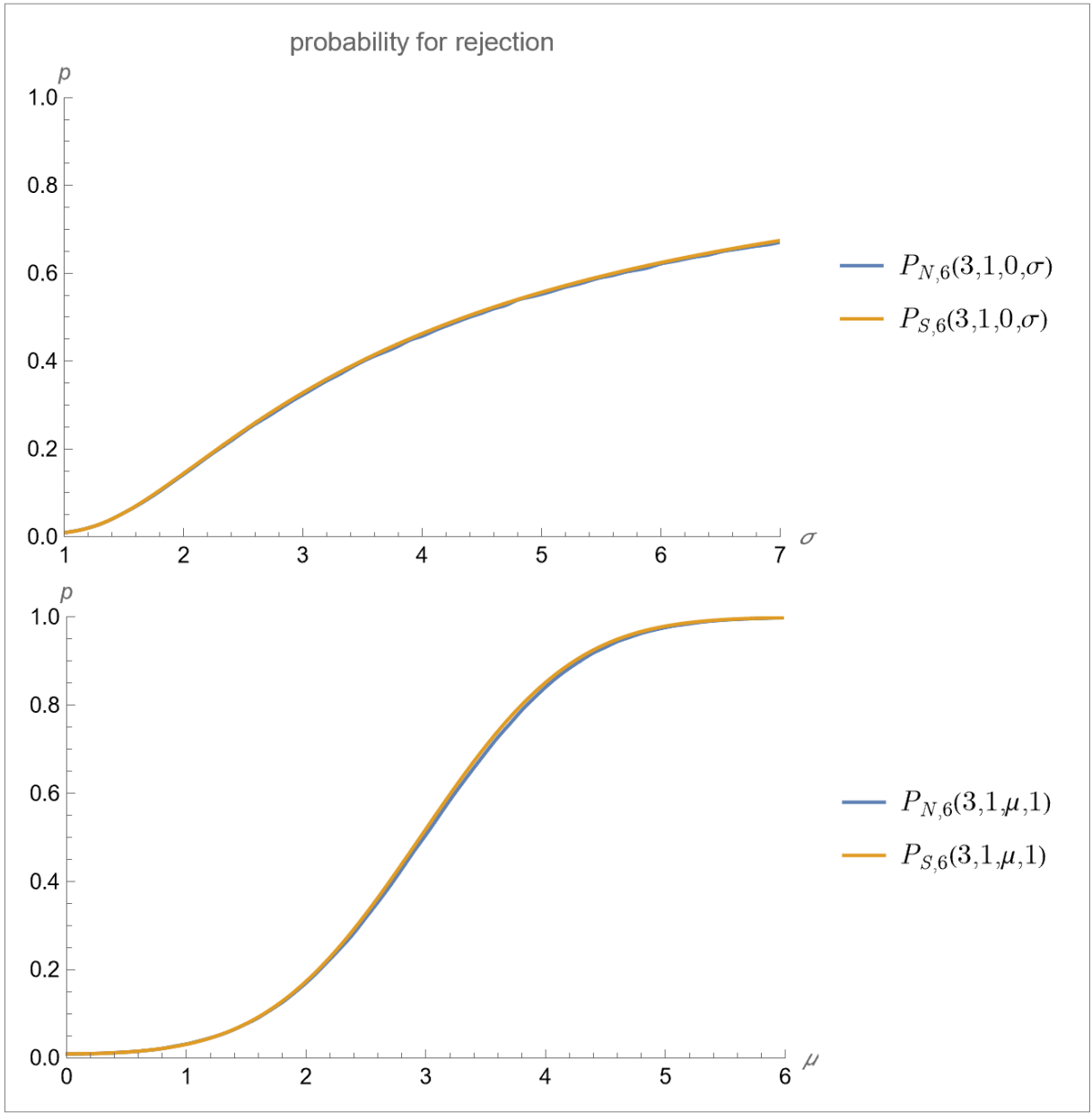

*Figure 44: $P_{N,6}(3,1,\mu,\sigma)$ vs $P_{S,6}(3,1,\mu,\sigma)$*

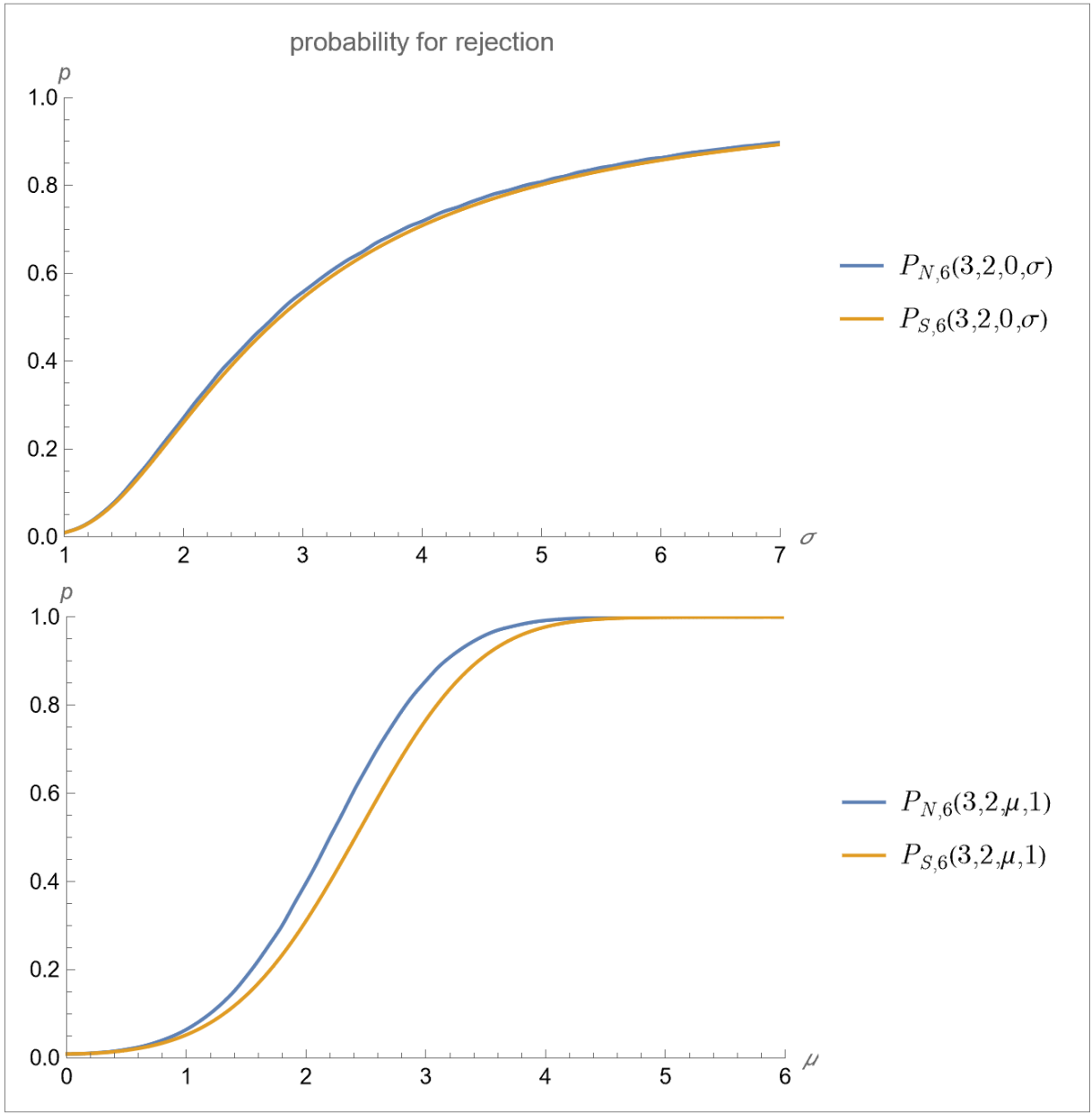

*Figure 45: $P_{N,6}(3,2,\mu,\sigma)$ vs $P_{S,6}(3,2,\mu,\sigma)$*

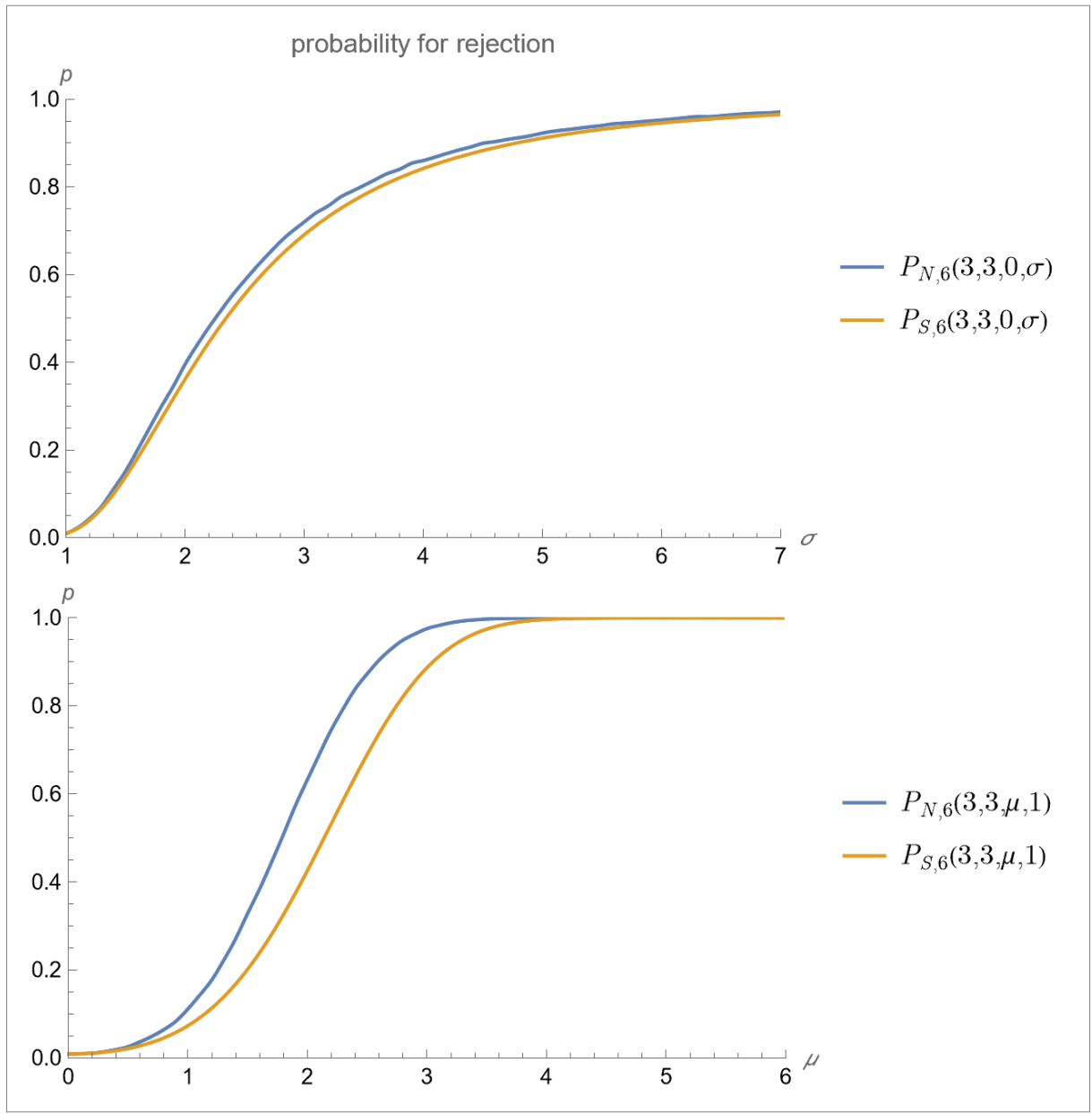

*Figure 46: $P_{N,6}(3,3,\mu,\sigma)$ vs $P_{S,6}(3,3,\mu,\sigma)$*

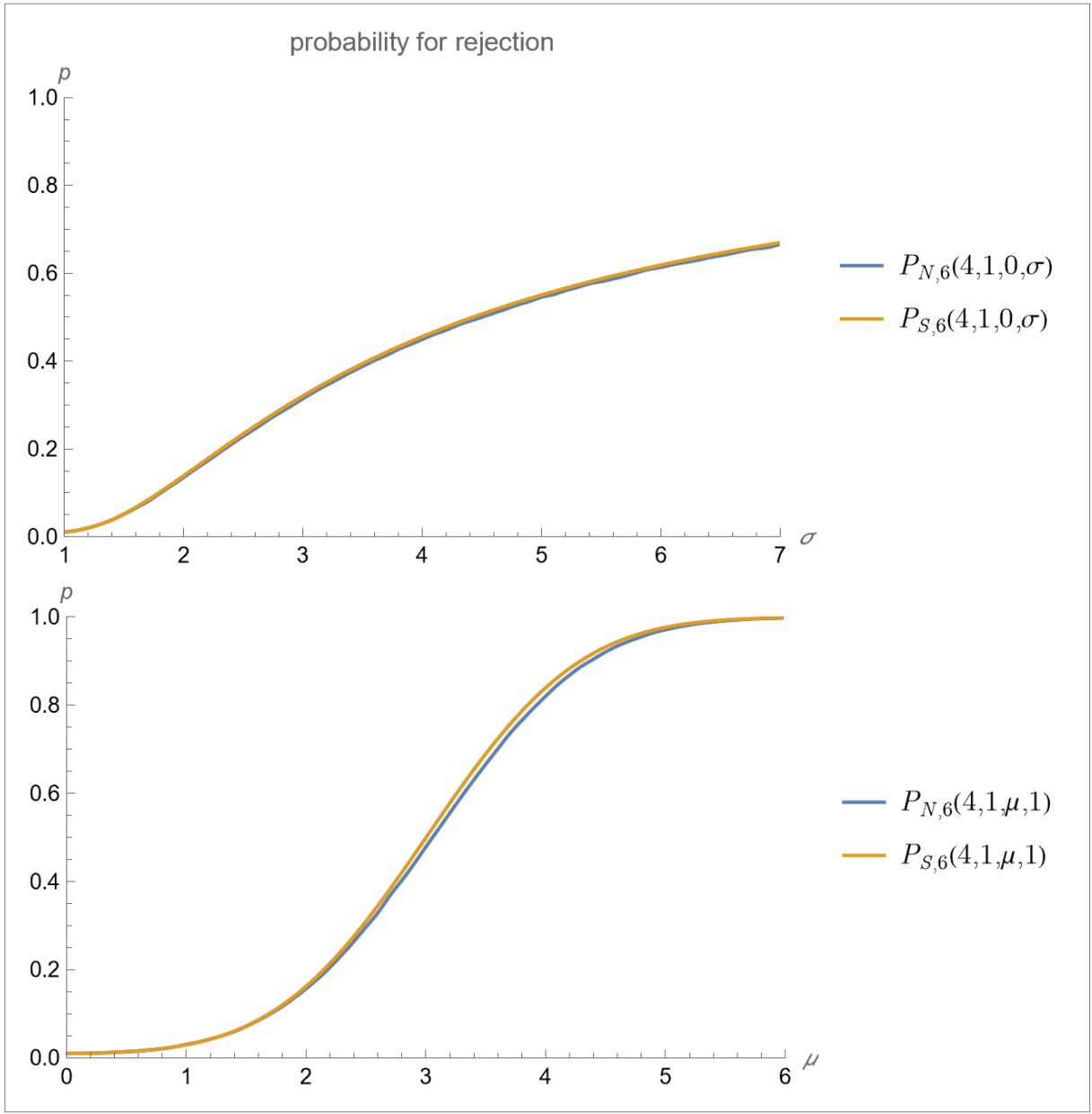

*Figure 47: $P_{N,6}(4,1,\mu,\sigma)$ vs $P_{S,6}(4,1,\mu,\sigma)$*

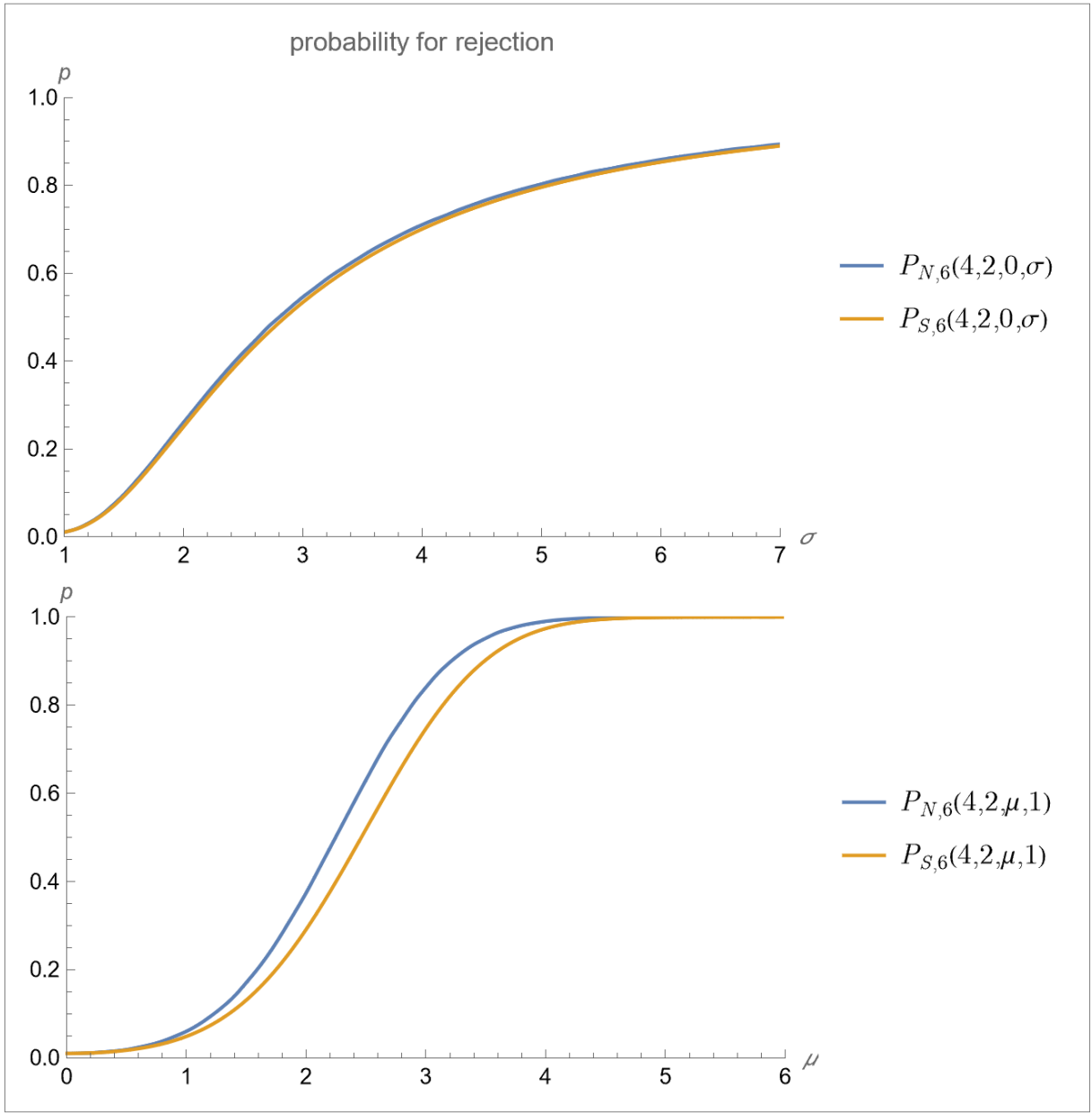

*Figure 48: $P_{N,6}(4,2,\mu,\sigma)$ vs $P_{S,6}(4,2,\mu,\sigma)$*

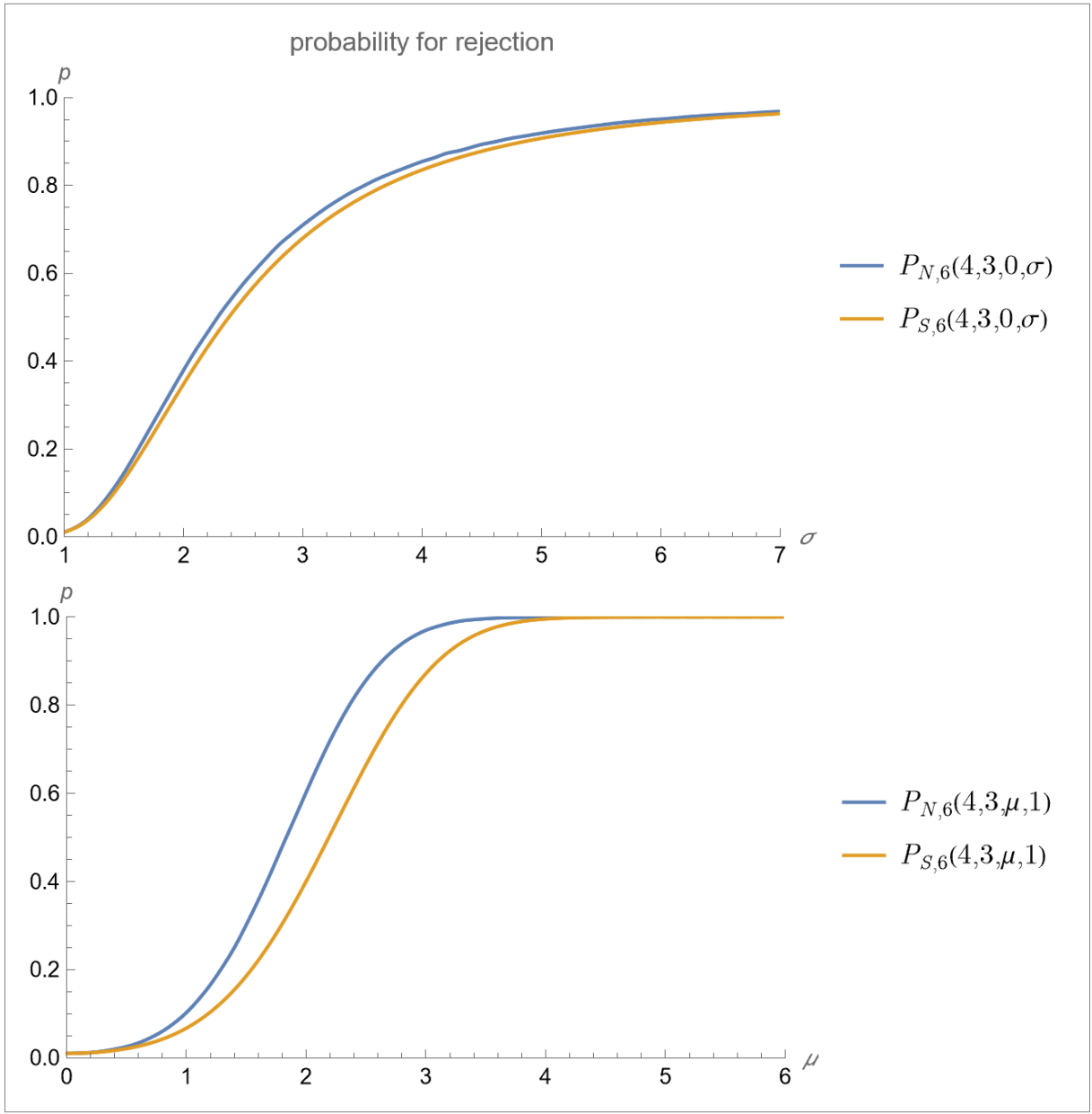

*Figure 49: $P_{N,6}(4,3,\mu,\sigma)$ vs $P_{S,6}(4,3,\mu,\sigma)$*

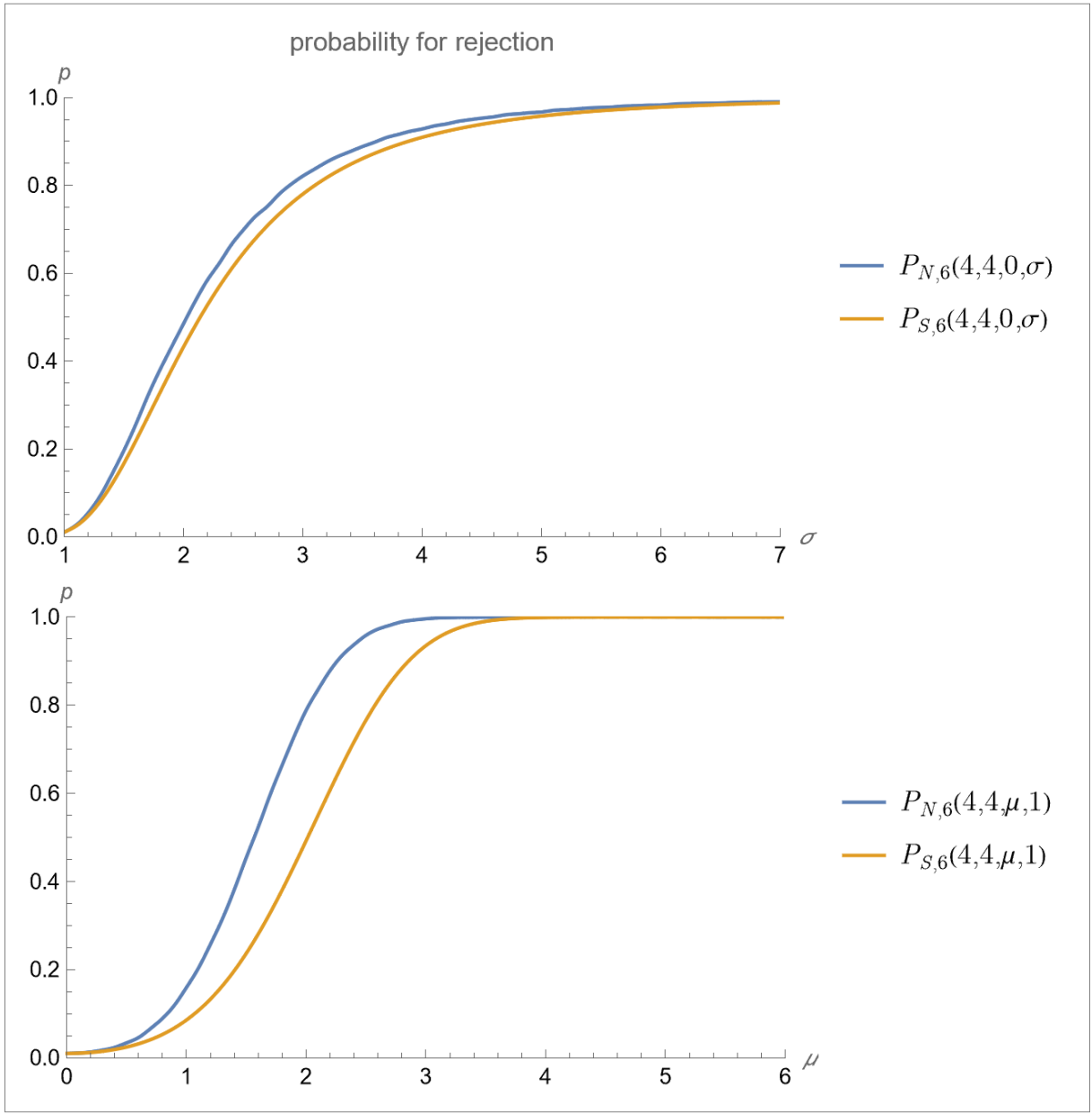

*Figure 50: $P_{N,6}(4,4,\mu,\sigma)$ vs $P_{S,6}(4,4,\mu,\sigma)$*

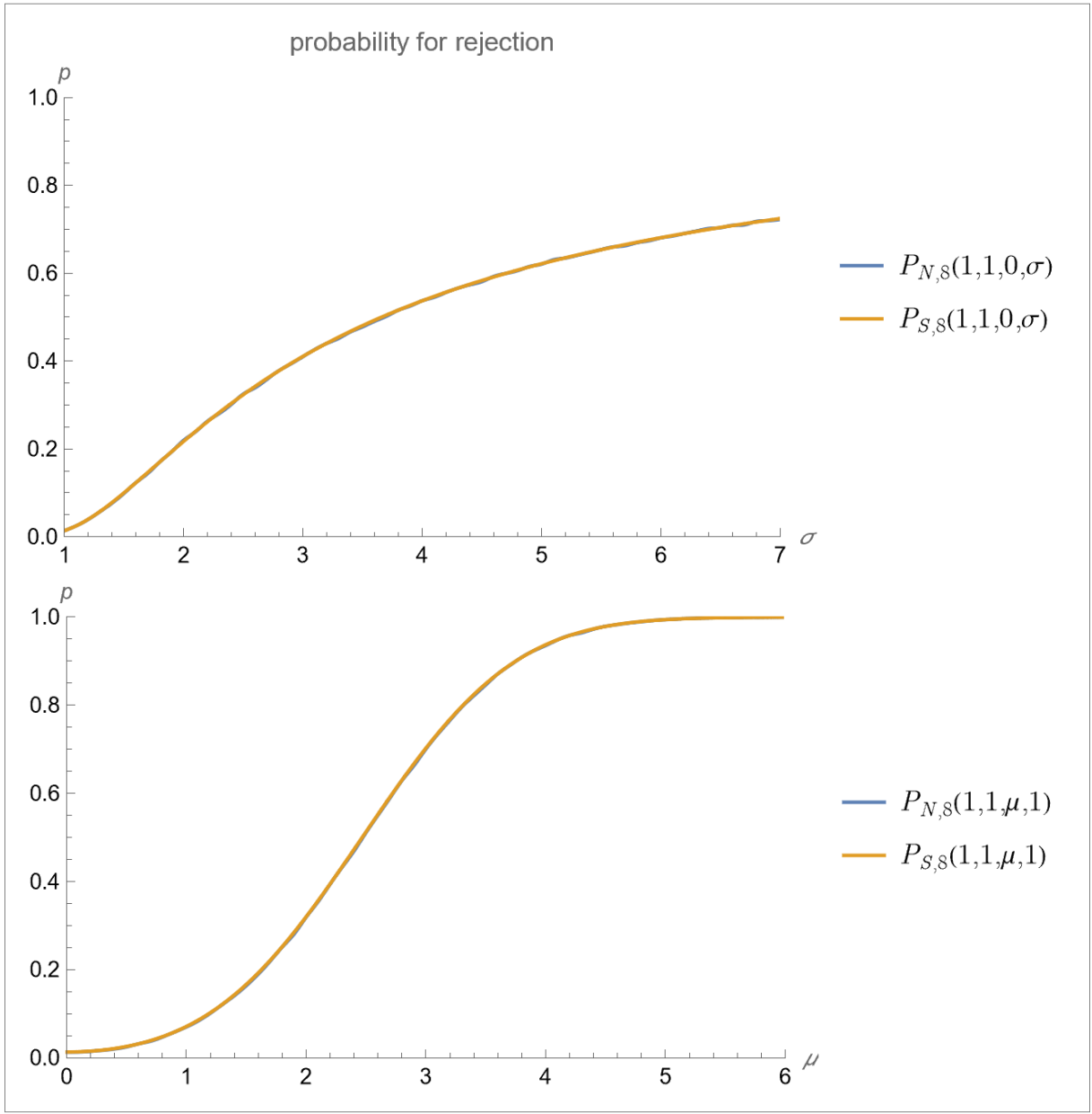

*Figure 51: $P_{N,8}(1,1,\mu,\sigma)$ vs $P_{S,8}(1,1,\mu,\sigma)$*

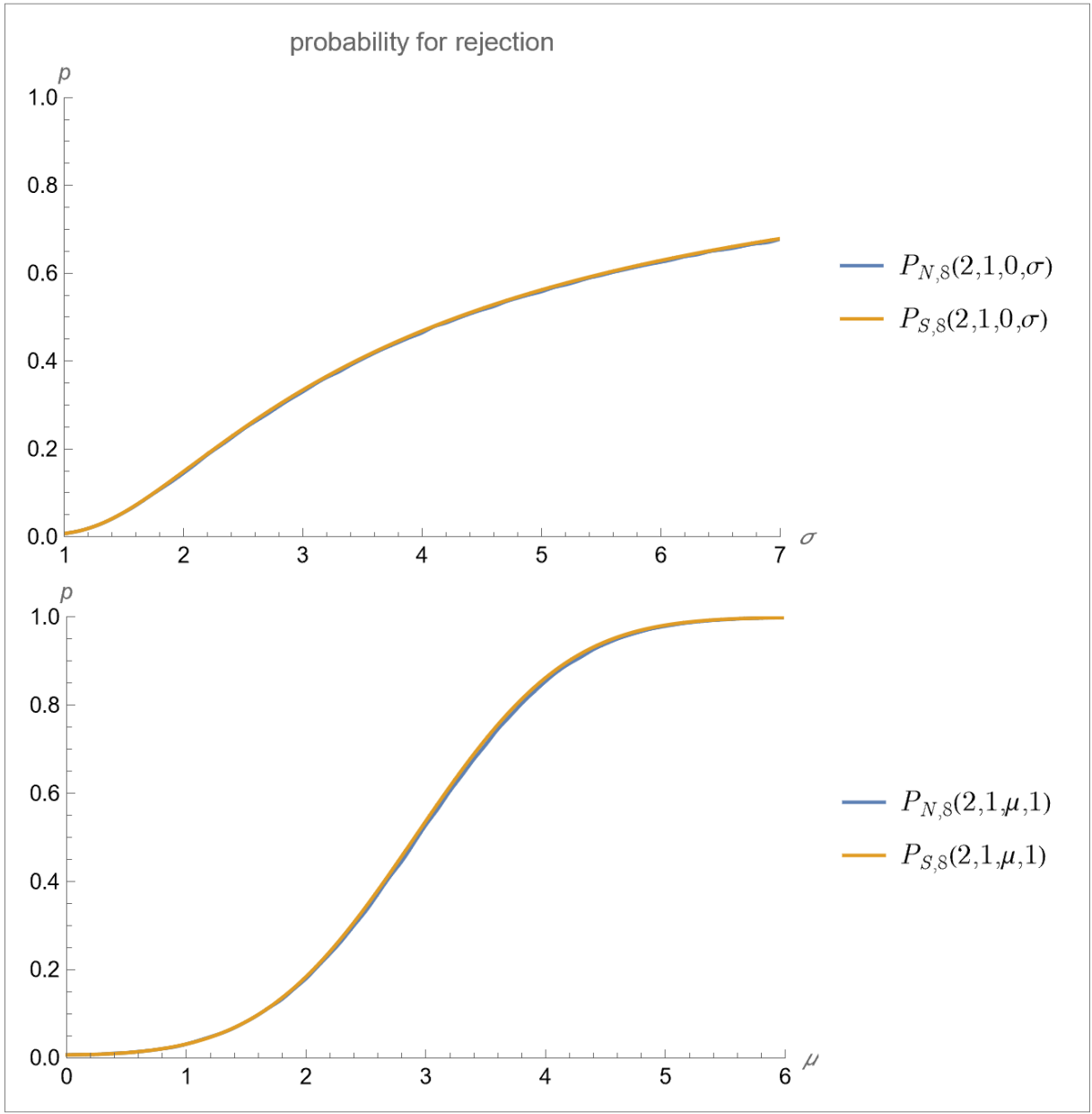

*Figure 52: $P_{N,8}(2,1,\mu,\sigma)$ vs $P_{S,8}(2,1,\mu,\sigma)$*

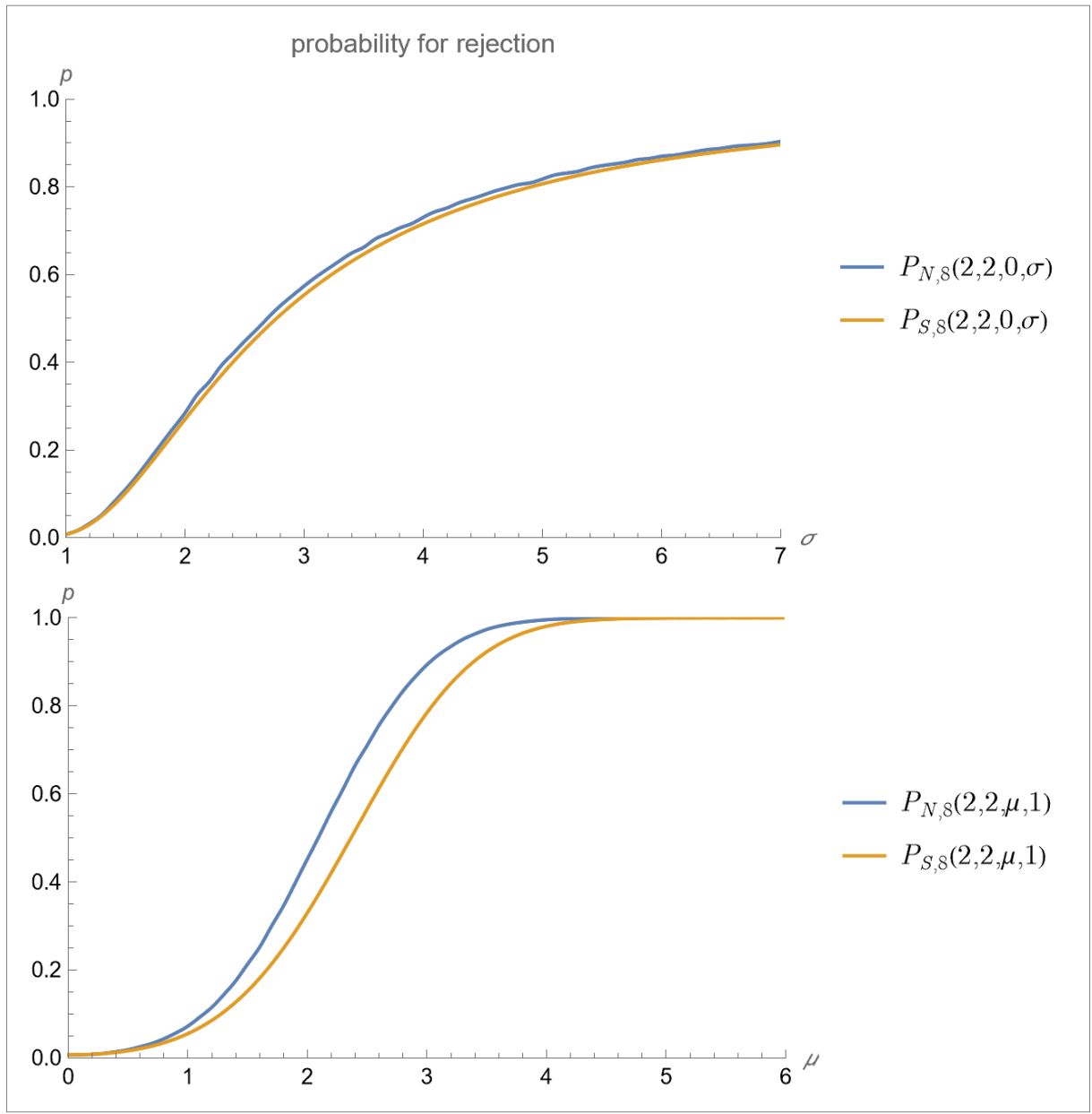

*Figure 53: $P_{N,8}(2,2,\mu,\sigma)$ vs $P_{S,8}(2,2,\mu,\sigma)$*

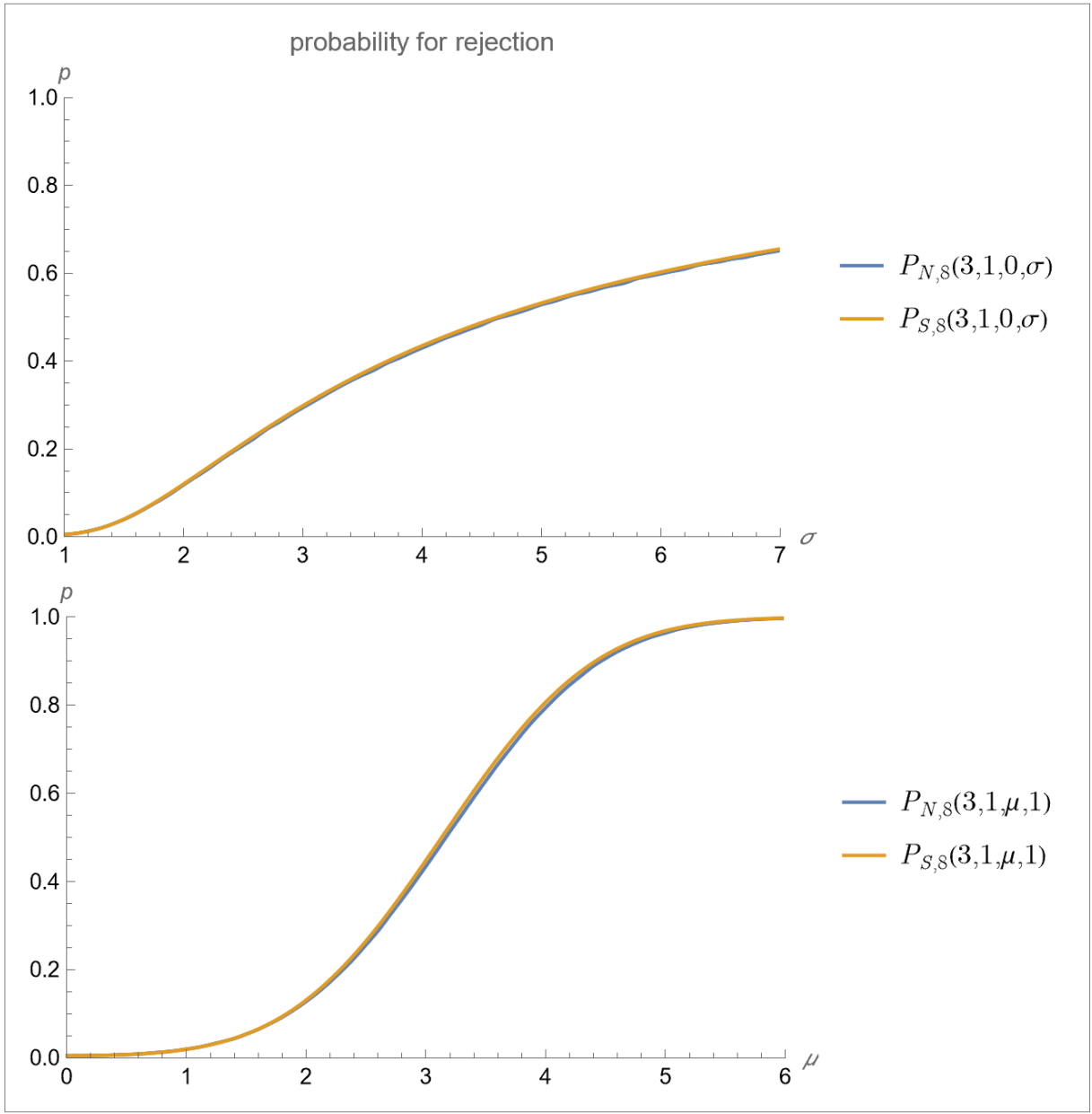

*Figure 54: $P_{N,8}(3,1,\mu,\sigma)$ vs $P_{S,8}(3,1,\mu,\sigma)$*

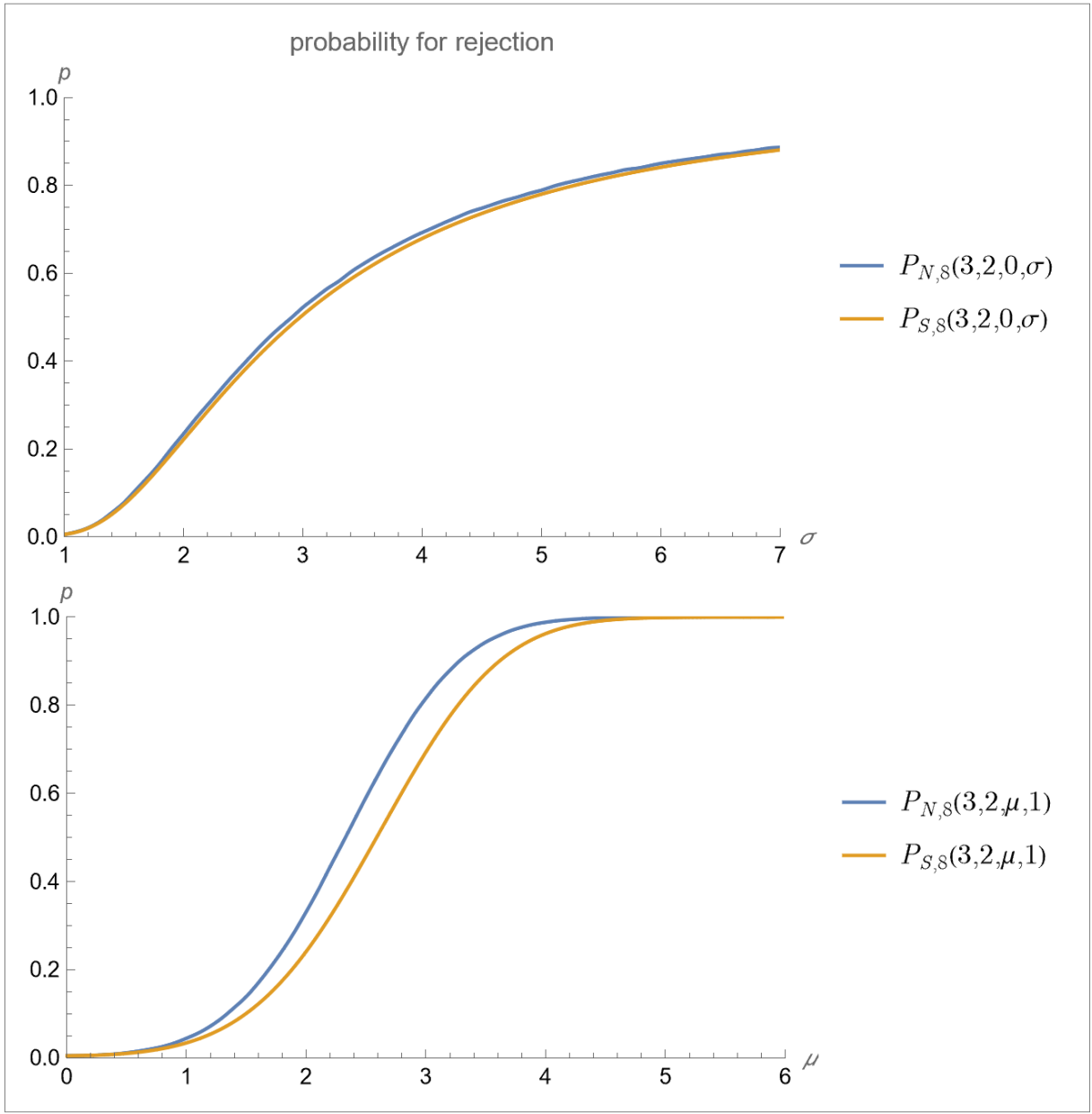

*Figure 55: $P_{N,8}(3,2,\mu,\sigma)$ vs $P_{S,8}(3,2,\mu,\sigma)$*

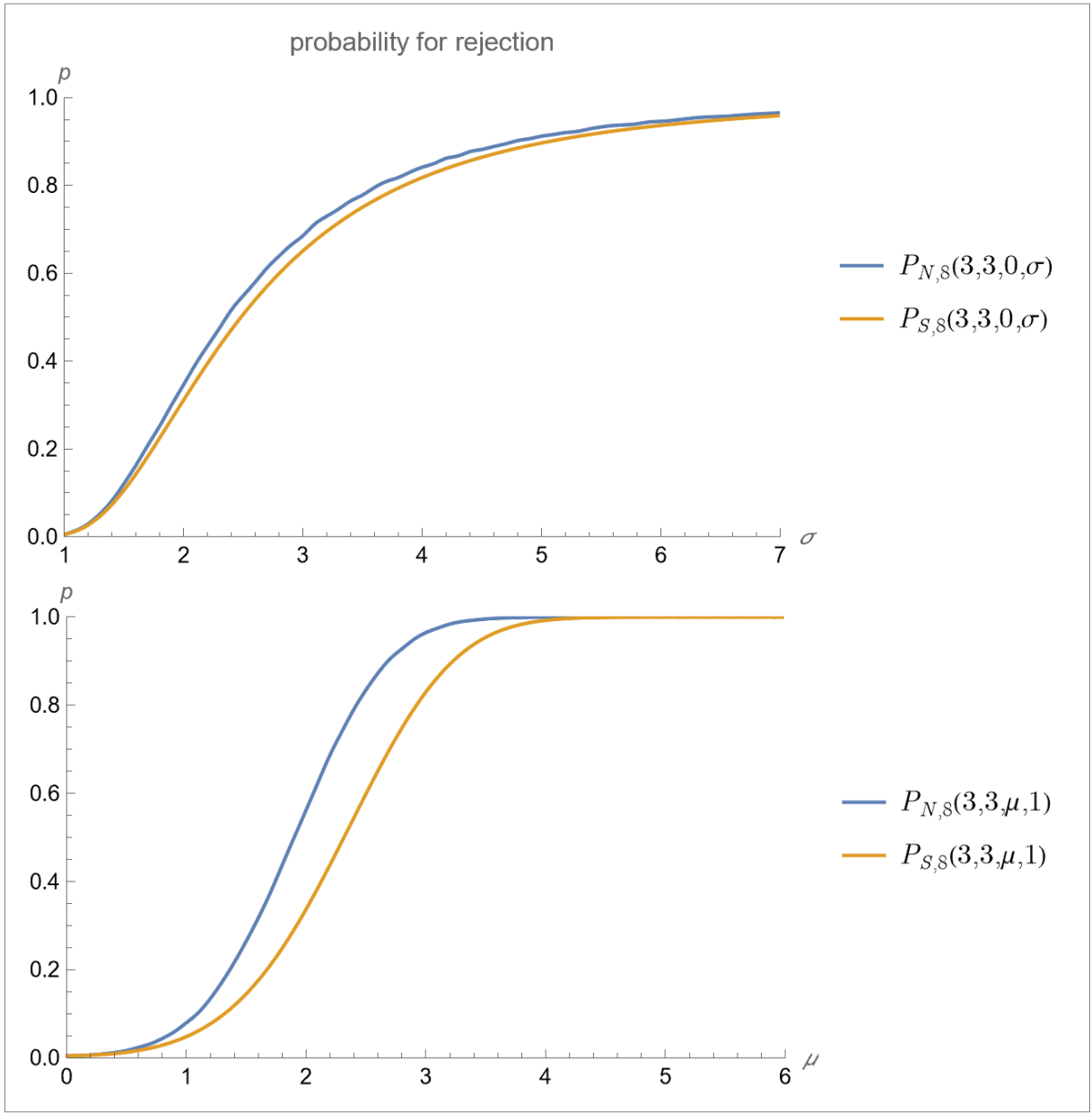

*Figure 56:* $P_{N,8}(3,3,\mu,\sigma)$ vs $P_{S,8}(3,3,\mu,\sigma)$

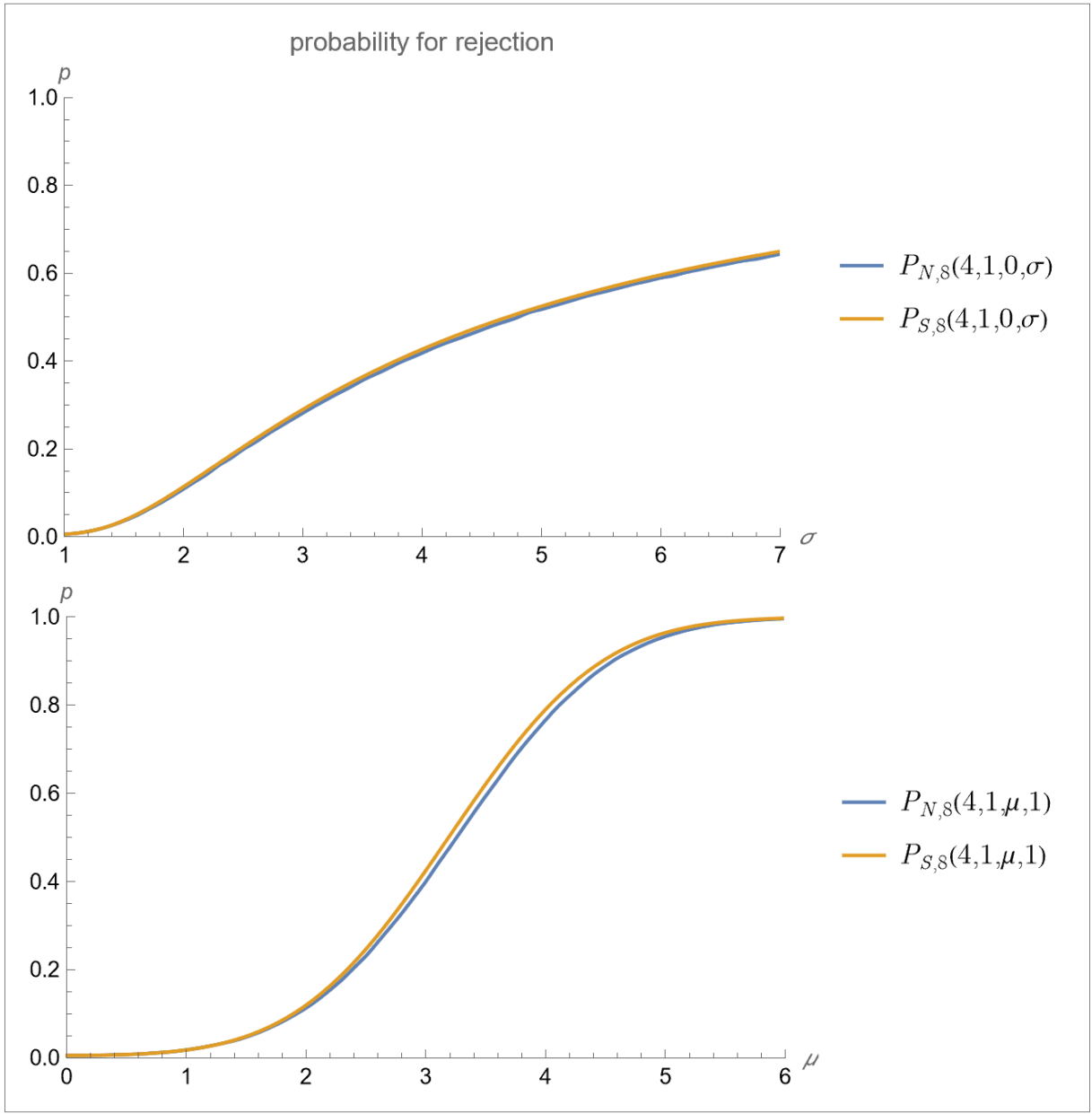

*Figure 57: $P_{N,8}(4,1,\mu,\sigma)$ vs $P_{S,8}(4,1,\mu,\sigma)$*

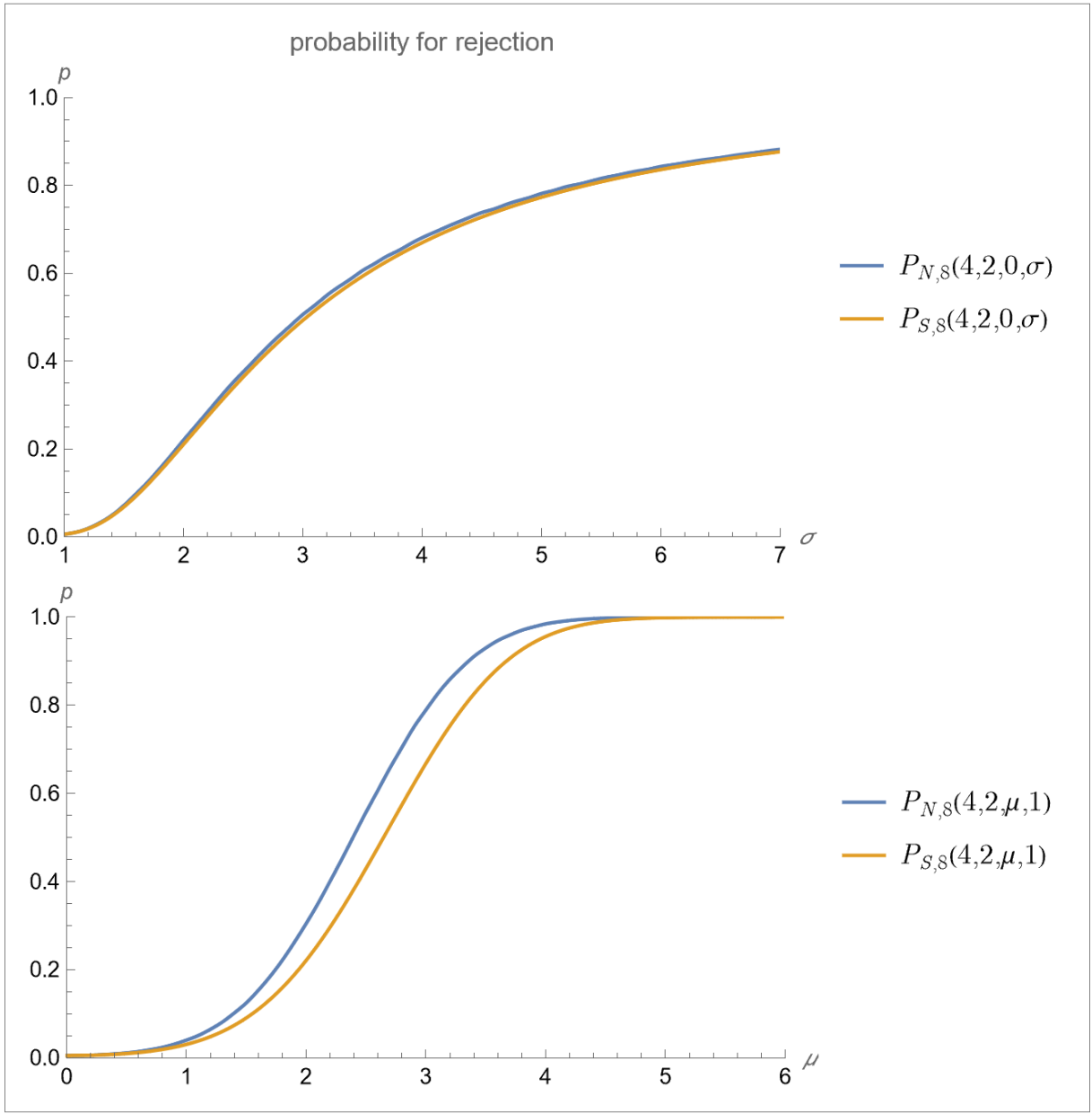

*Figure 58: $P_{N,8}(4,2,\mu,\sigma)$ vs $P_{S,8}(4,2,\mu,\sigma)$*

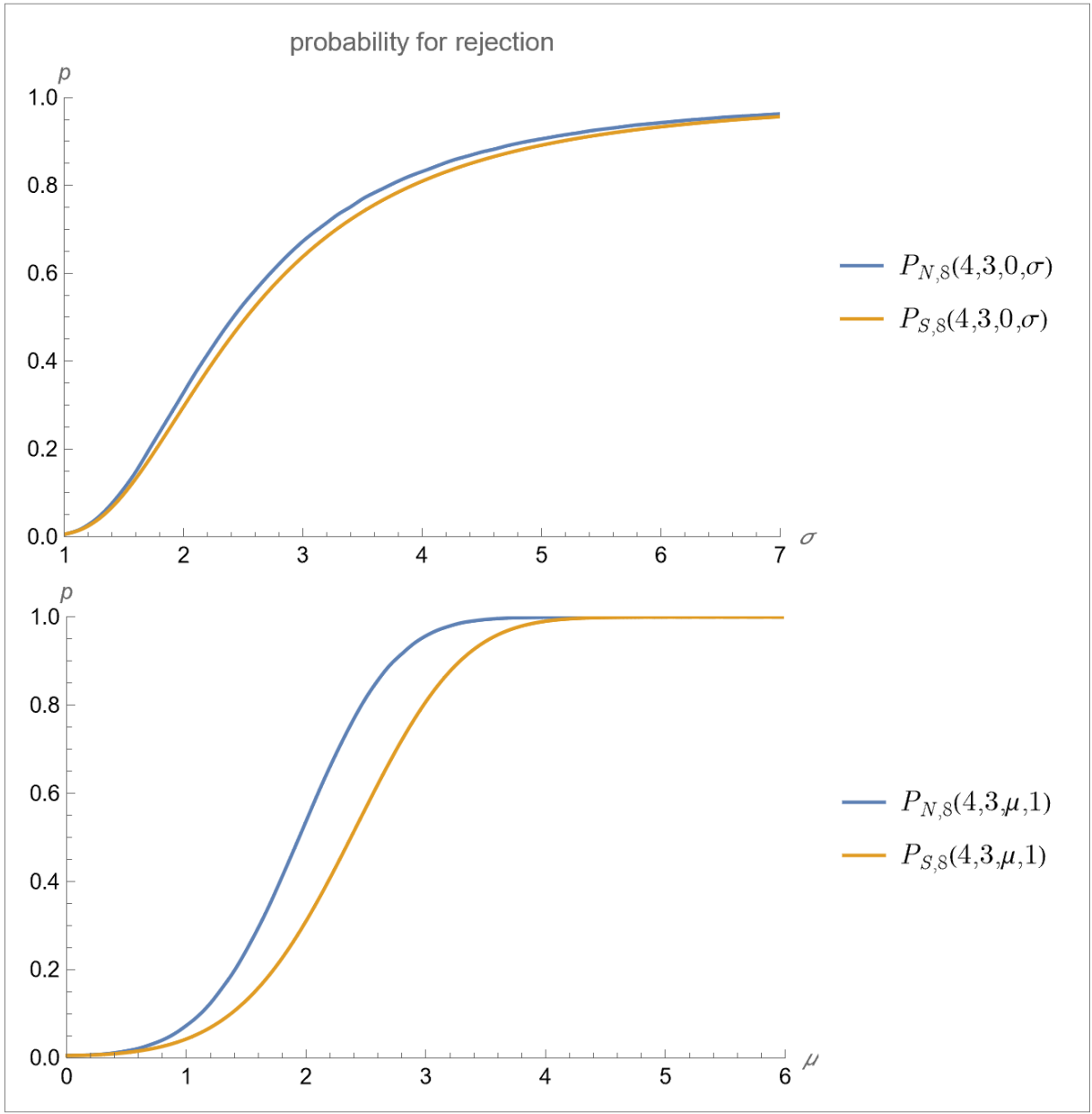

*Figure 59: $P_{N,8}(4,3,\mu,\sigma)$ vs $P_{S,8}(4,3,\mu,\sigma)$*

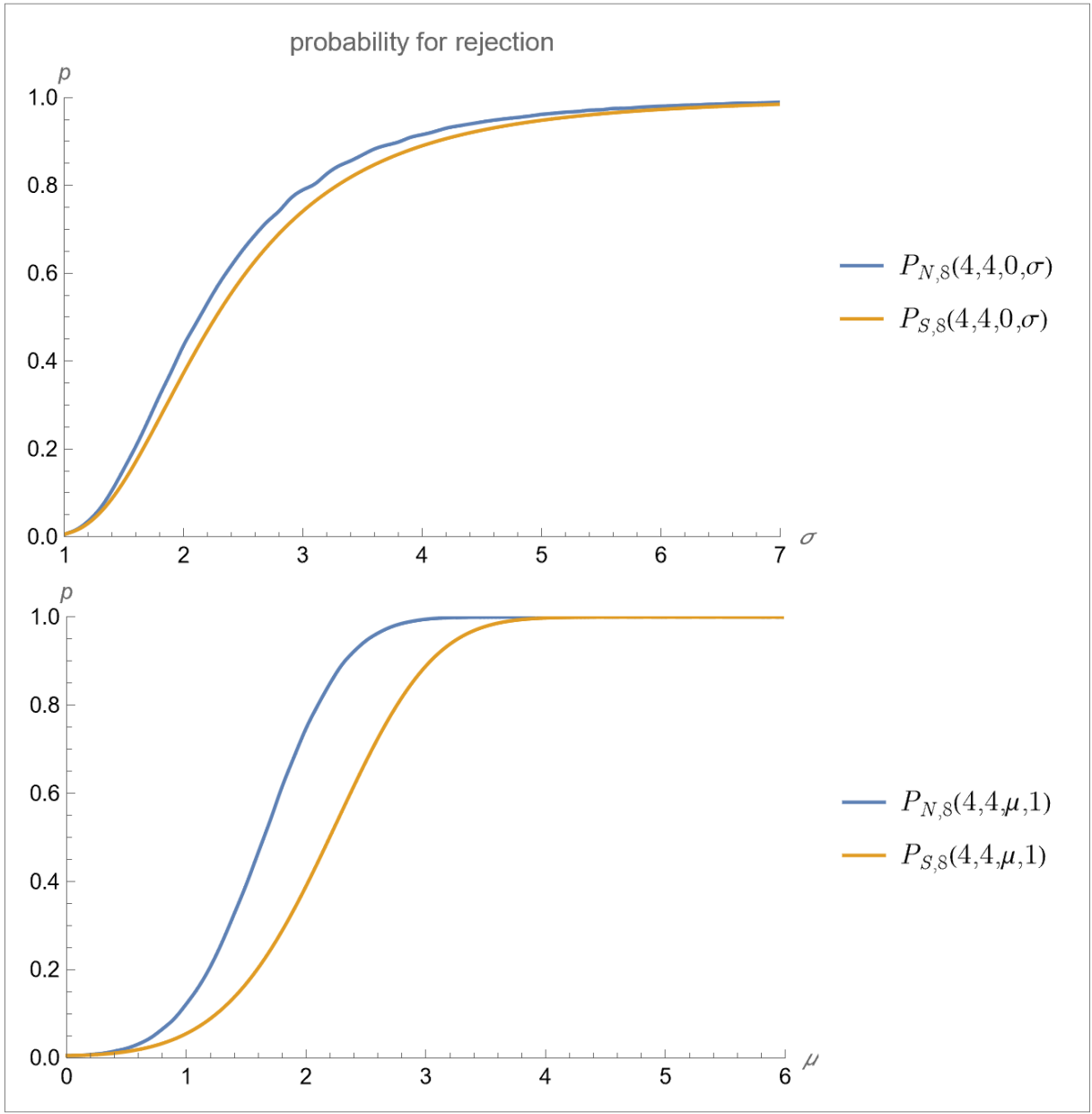

*Figure 60: $P_{N,8}(4,4,\mu,\sigma)$ vs $P_{S,8}(4,4,\mu,\sigma)$*

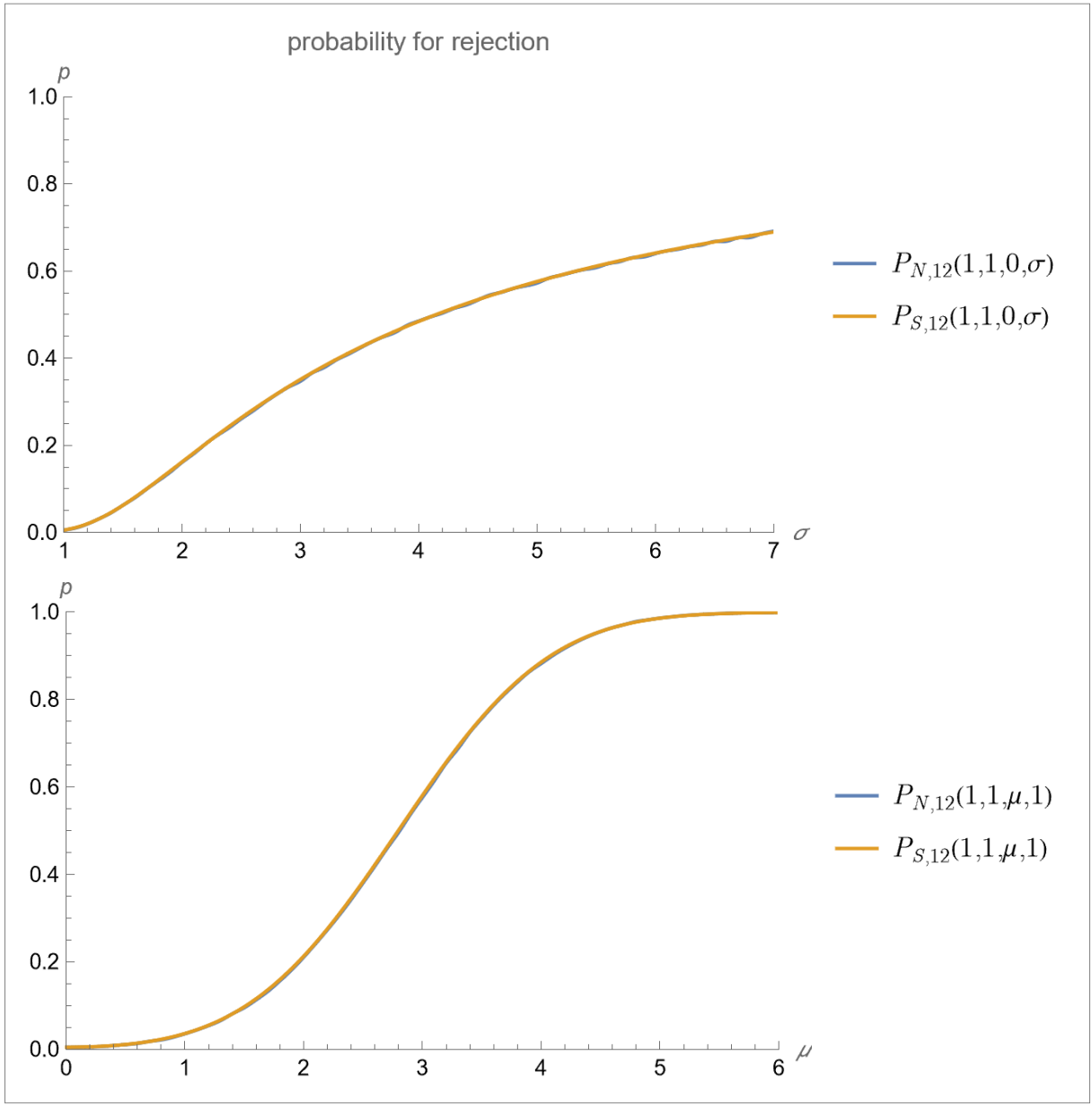

*Figure 61: $P_{N,12}(1,1,\mu,\sigma)$ vs $P_{S,12}(1,1,\mu,\sigma)$*

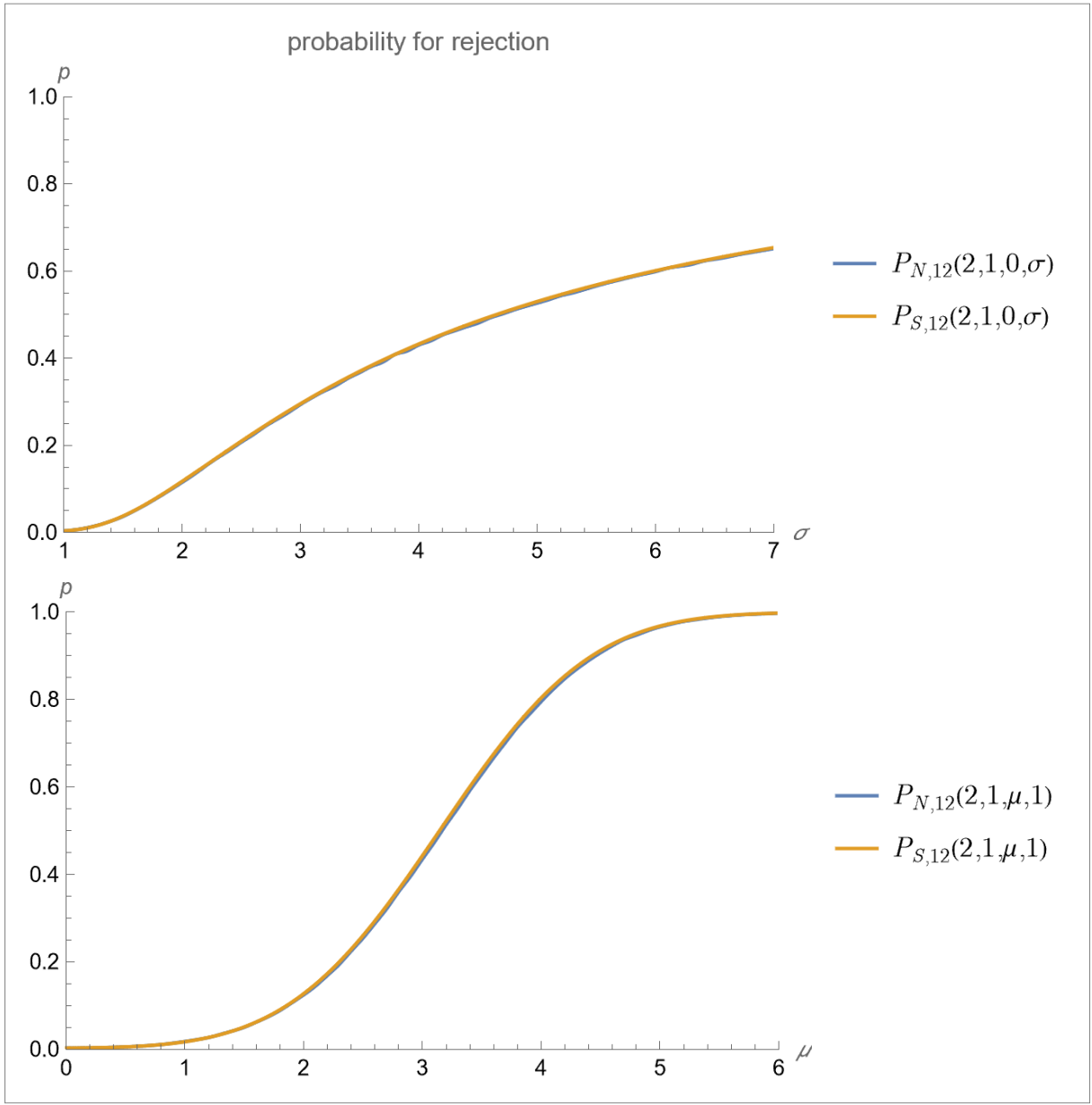

*Figure 62: $P_{N,12}(2,1,\mu,\sigma)$ vs $P_{S,12}(2,1,\mu,\sigma)$*

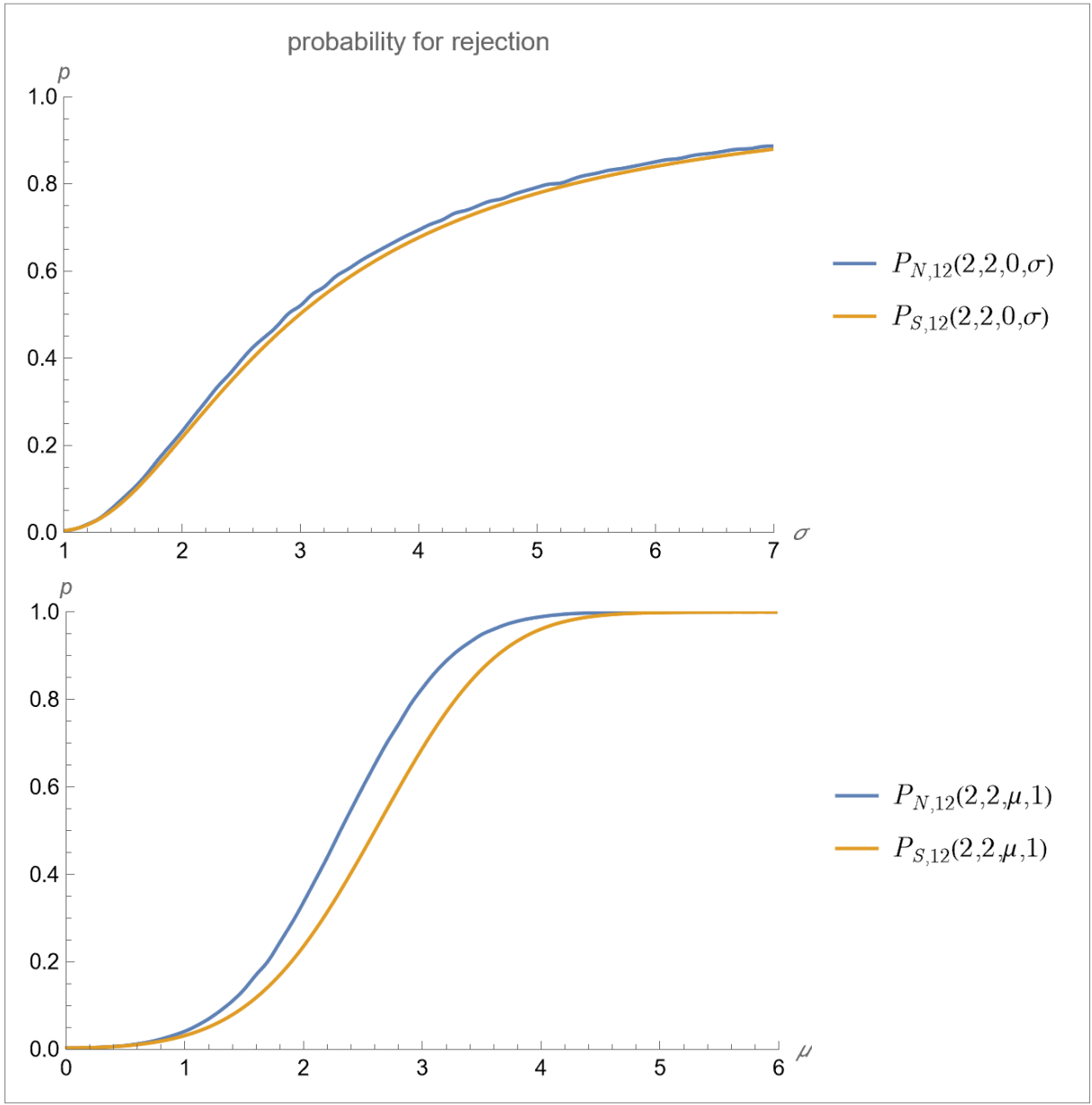

*Figure 63: $P_{N,12}(2,2,\mu,\sigma)$ vs $P_{S,12}(2,2,\mu,\sigma)$*

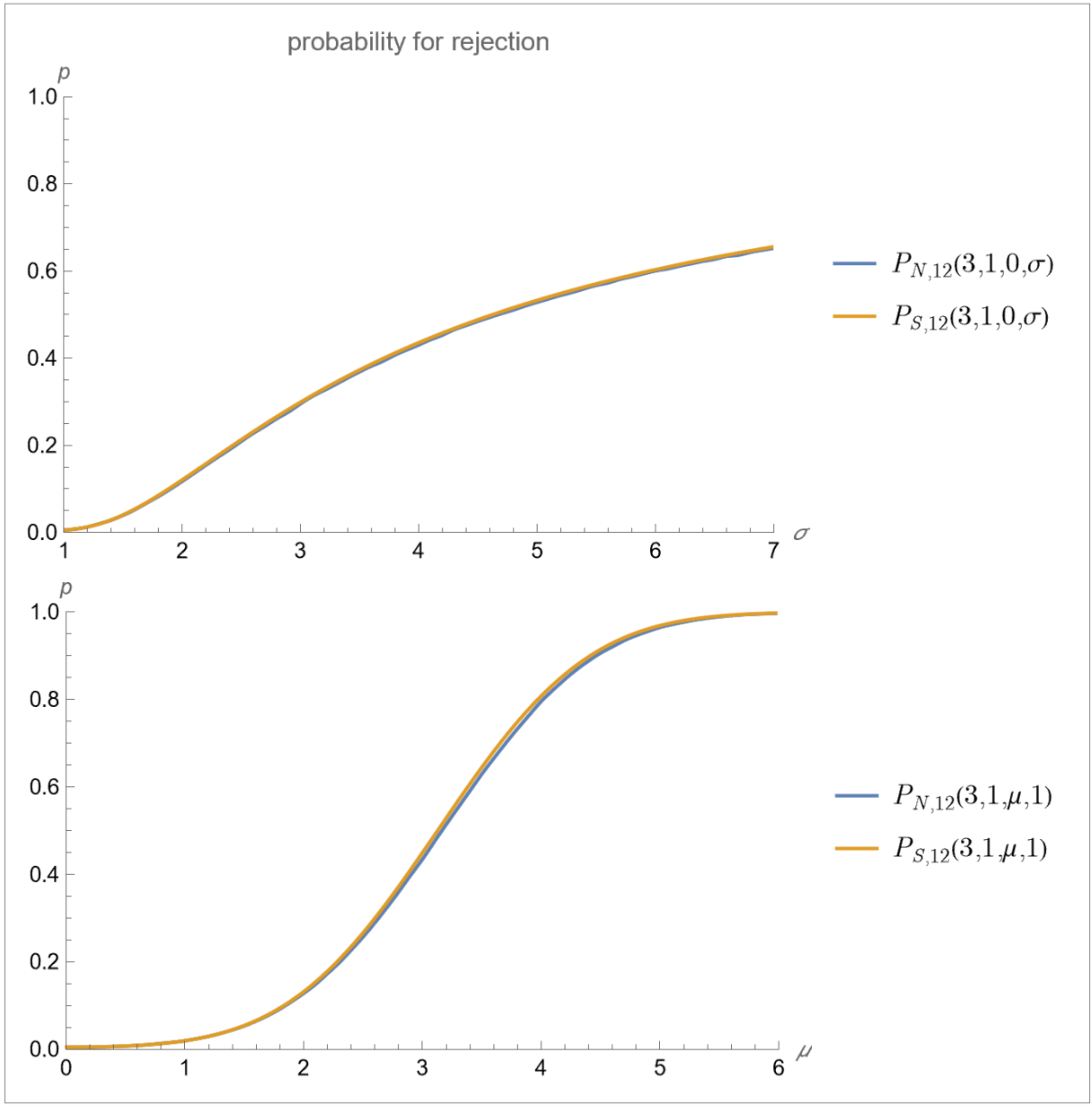

*Figure 64:* $P_{N,12}(3,1,\mu,\sigma)$ vs $P_{S,12}(3,1,\mu,\sigma)$

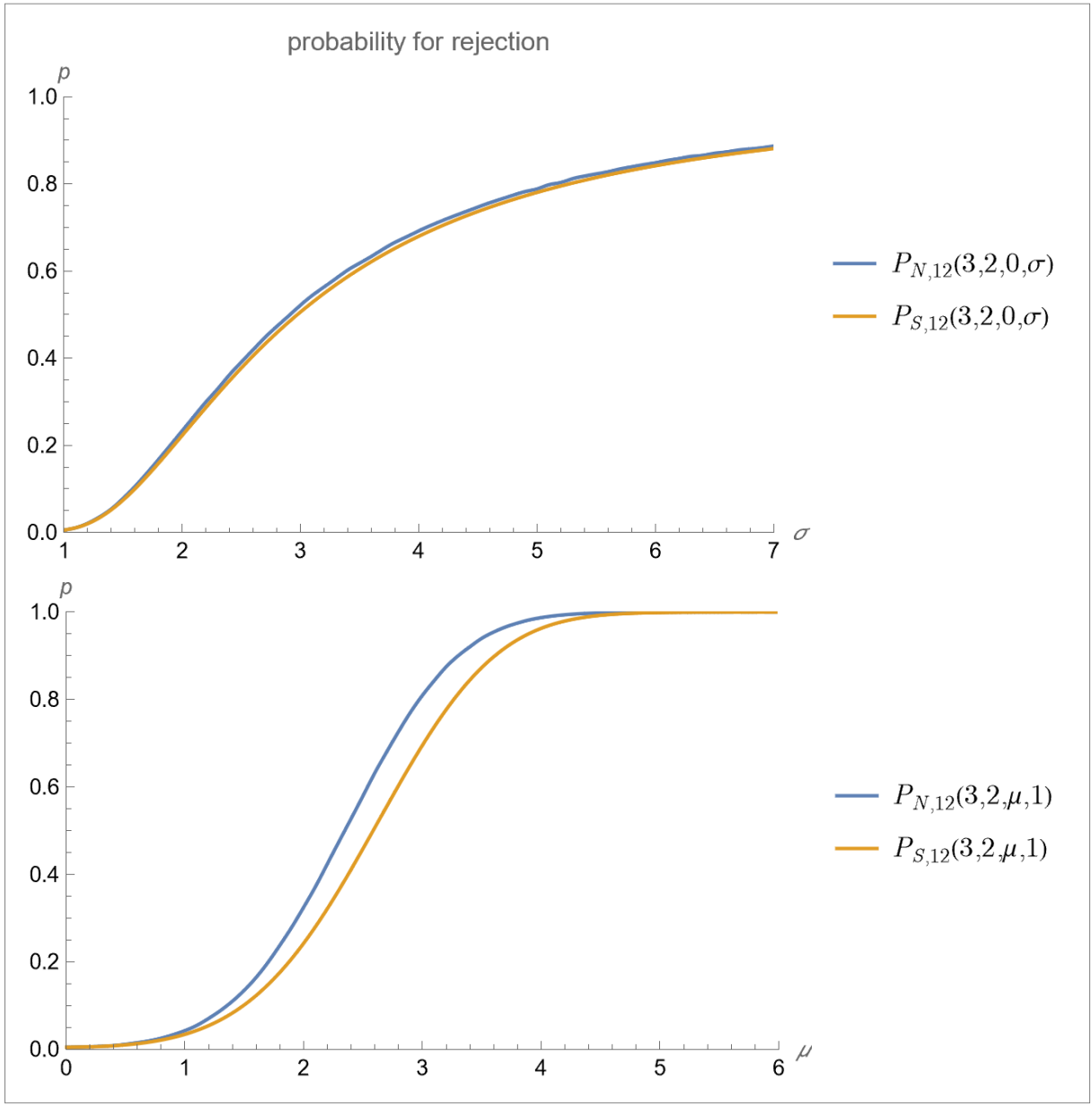

*Figure 65: $P_{N,12}(3,2,\mu,\sigma)$ vs $P_{S,12}(3,2,\mu,\sigma)$*

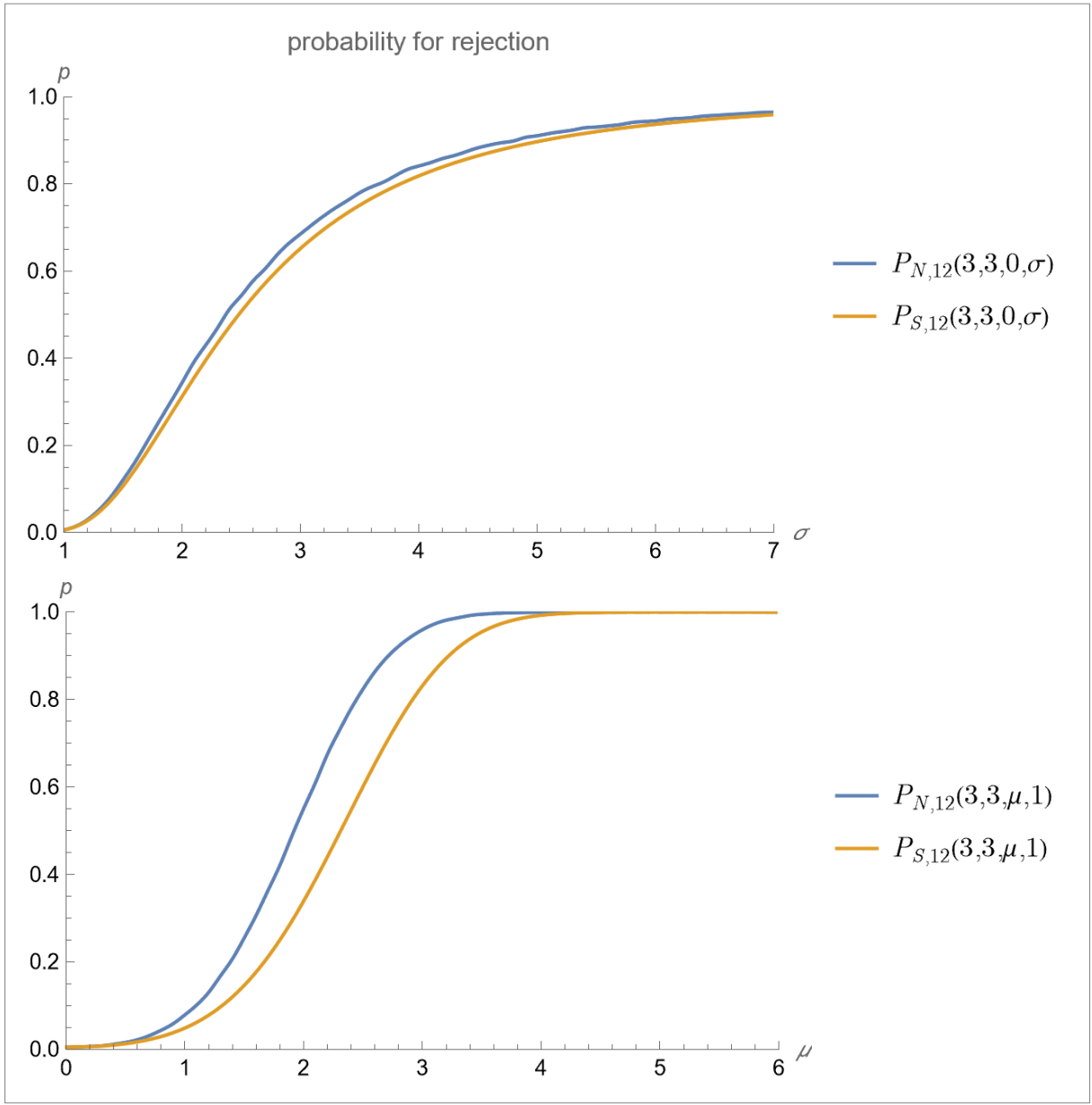

*Figure 66: $P_{N,12}(3,3,\mu,\sigma)$ vs $P_{S,12}(3,3,\mu,\sigma)$*

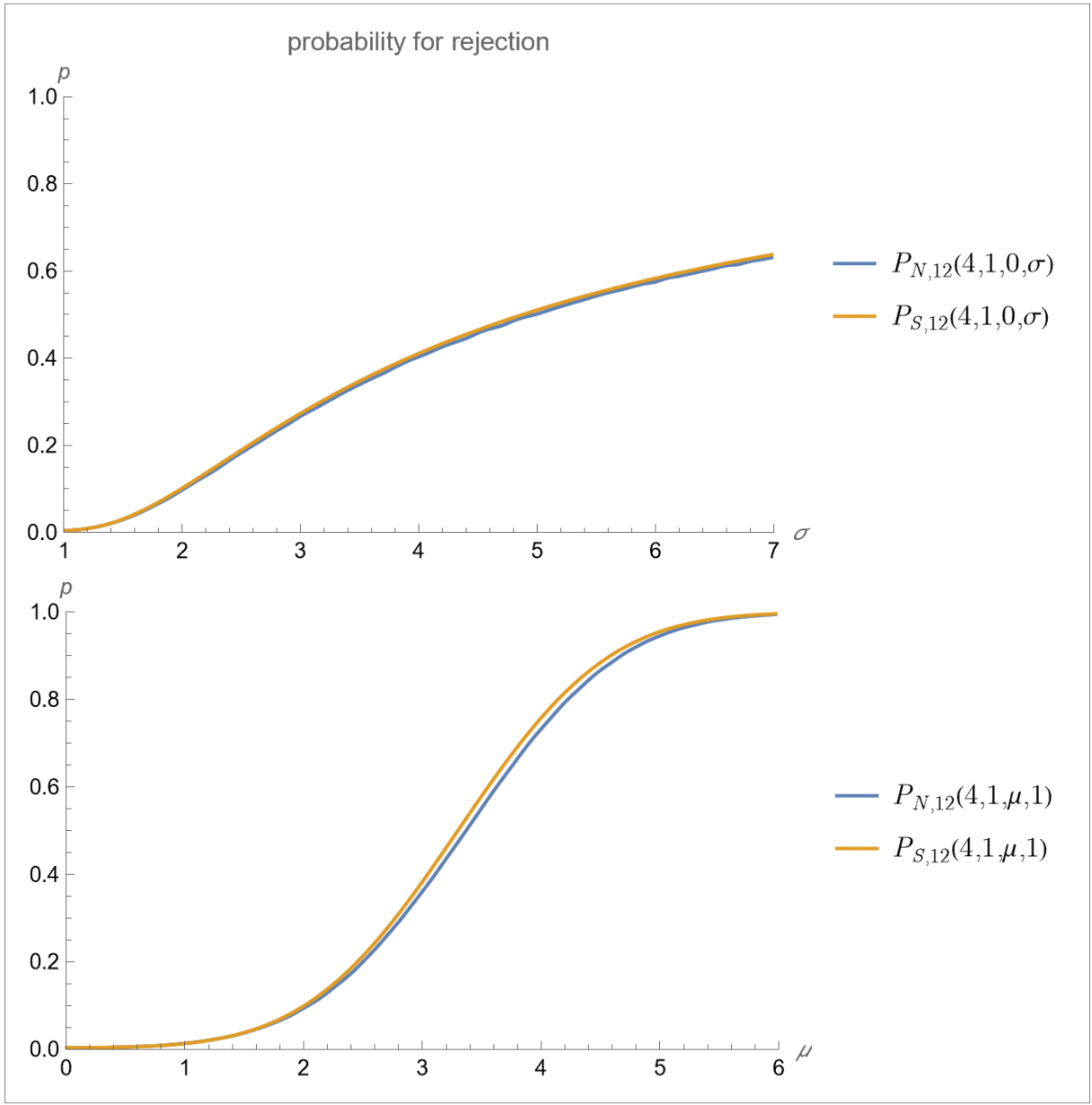

*Figure 67:* $P_{N,12}(4,1,\mu,\sigma)$ vs $P_{S,12}(4,1,\mu,\sigma)$

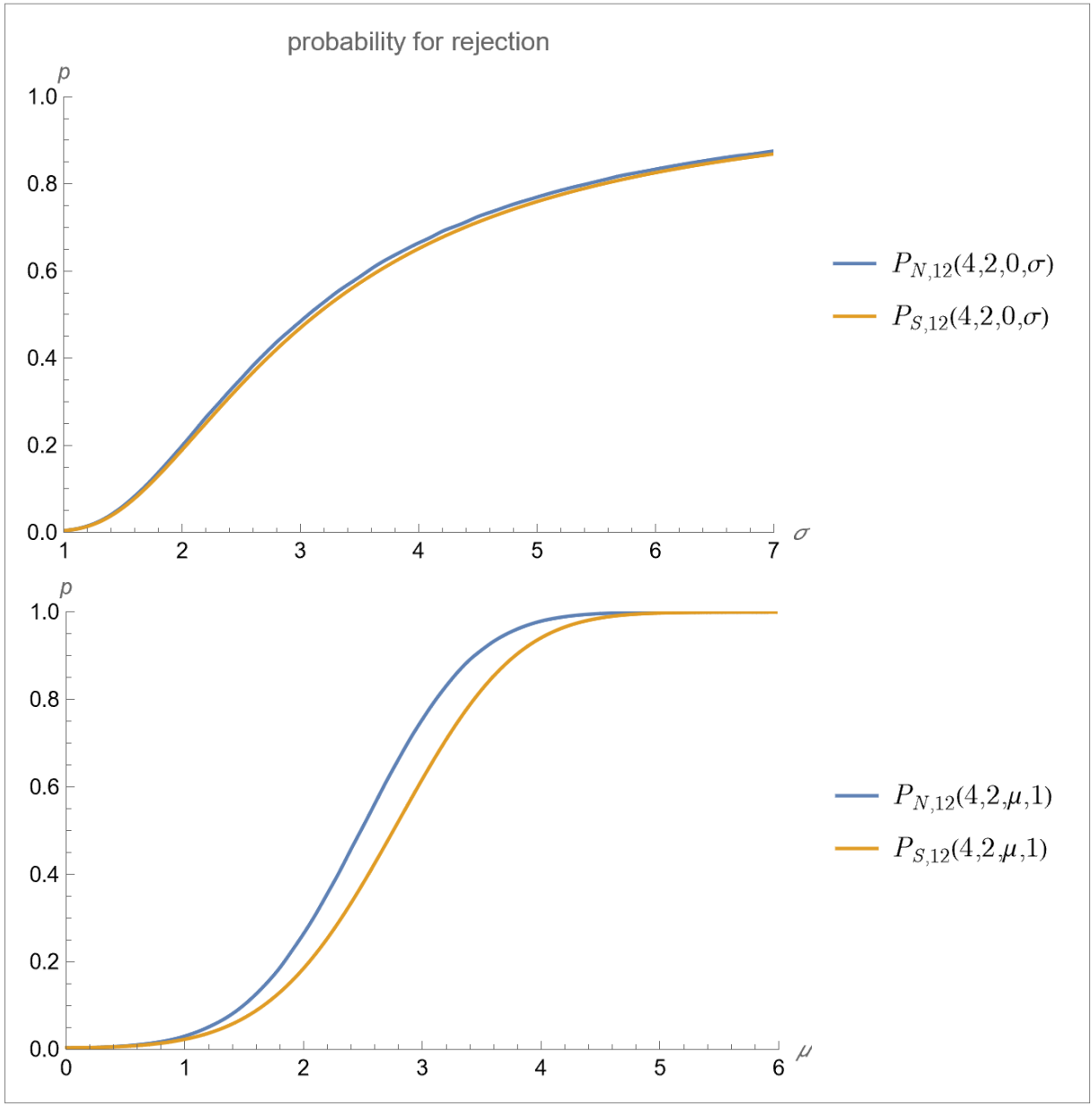

*Figure 68: $P_{N,12}(4,2,\mu,\sigma)$ vs $P_{S,12}(4,2,\mu,\sigma)$*

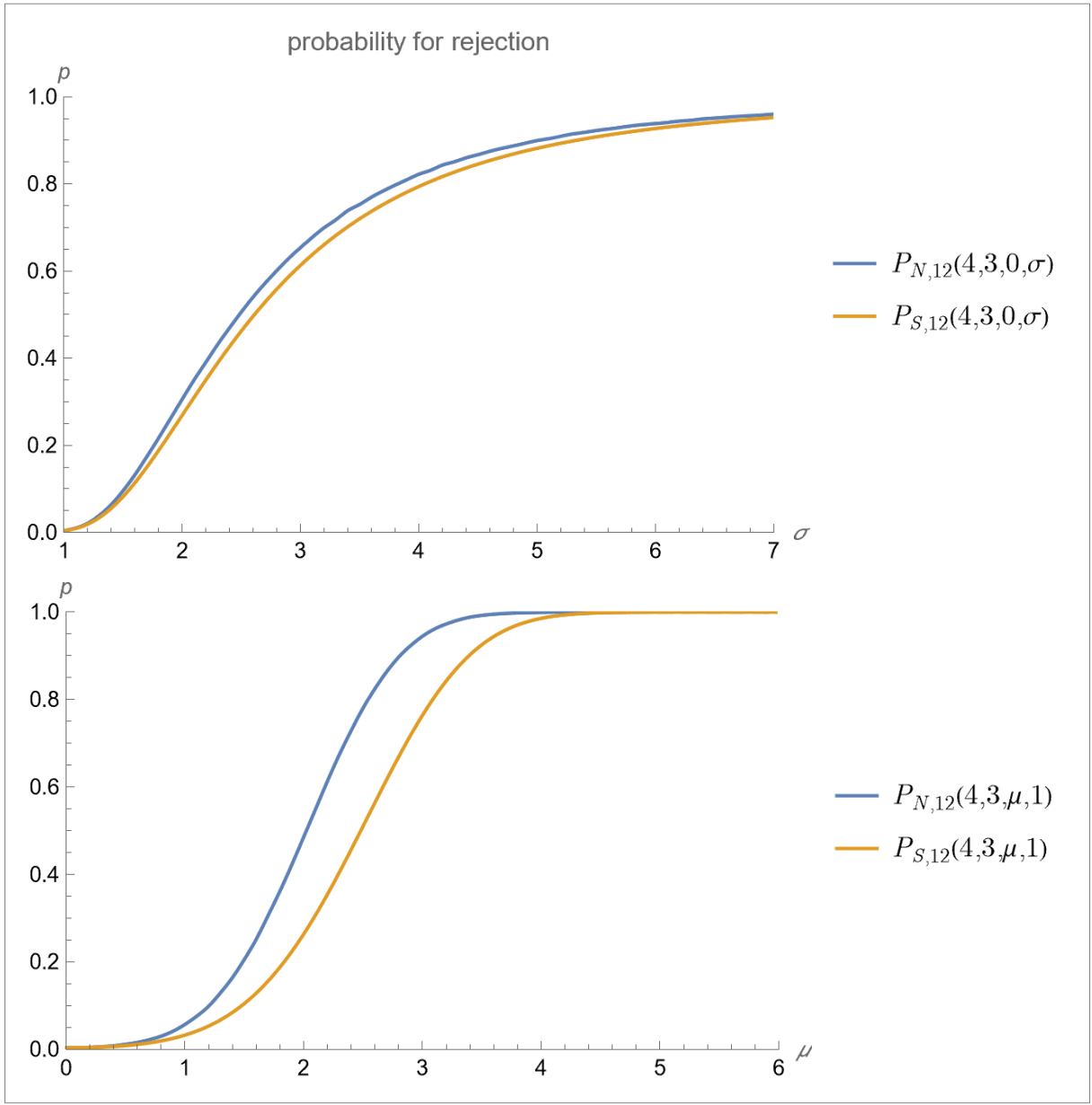

*Figure 69: $P_{N,12}(4,3,\mu,\sigma)$ vs $P_{S,12}(4,3,\mu,\sigma)$*

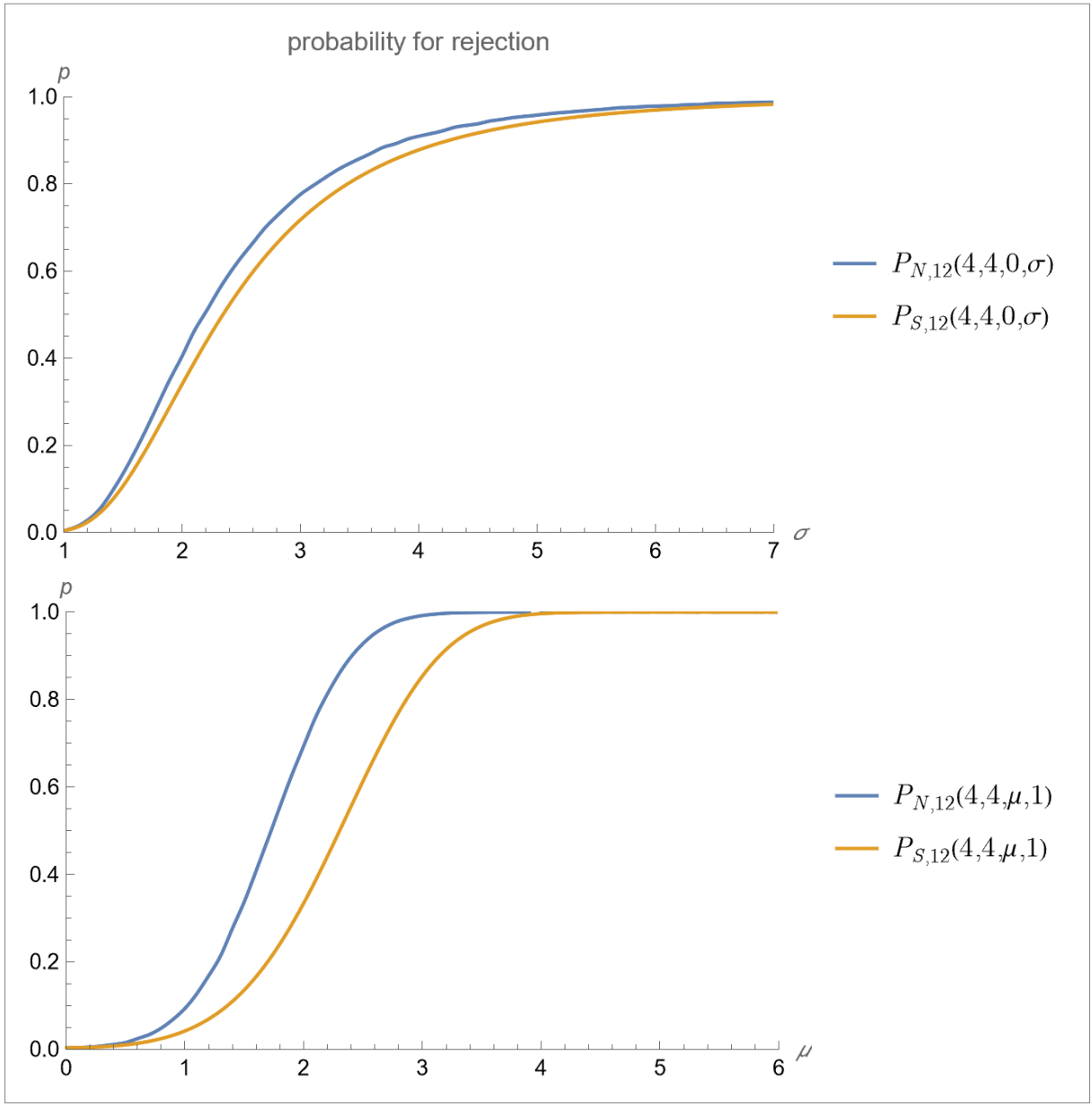

*Figure 70: $P_{N,12}(4,4,\mu,\sigma)$ vs $P_{S,12}(4,4,\mu,\sigma)$*

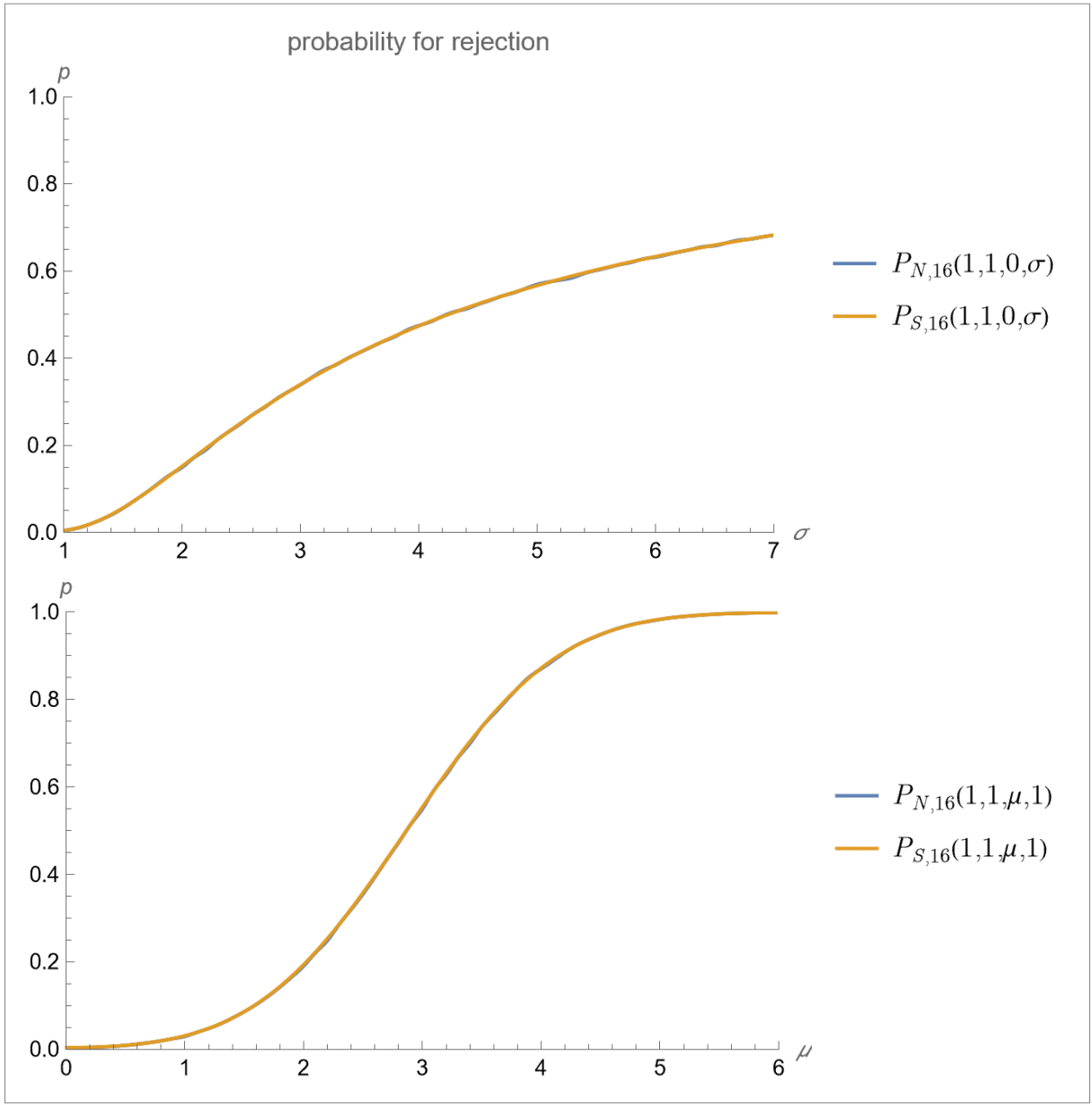

*Figure 71: $P_{N,16}(1,1,\mu,\sigma)$ vs $P_{S,16}(1,1,\mu,\sigma)$*

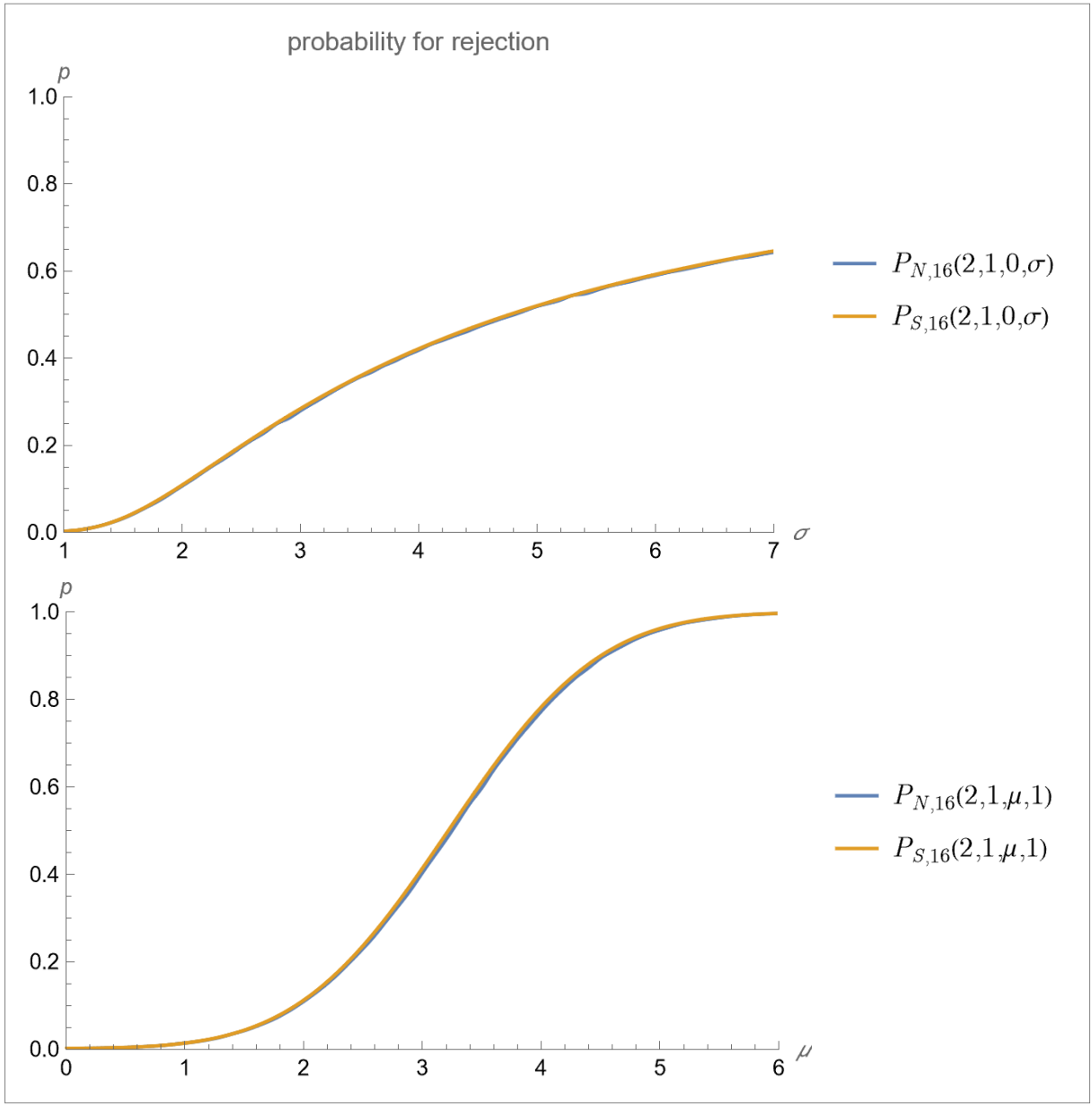

*Figure 72: $P_{N,16}(2,1,\mu,\sigma)$ vs $P_{S,16}(2,1,\mu,\sigma)$*

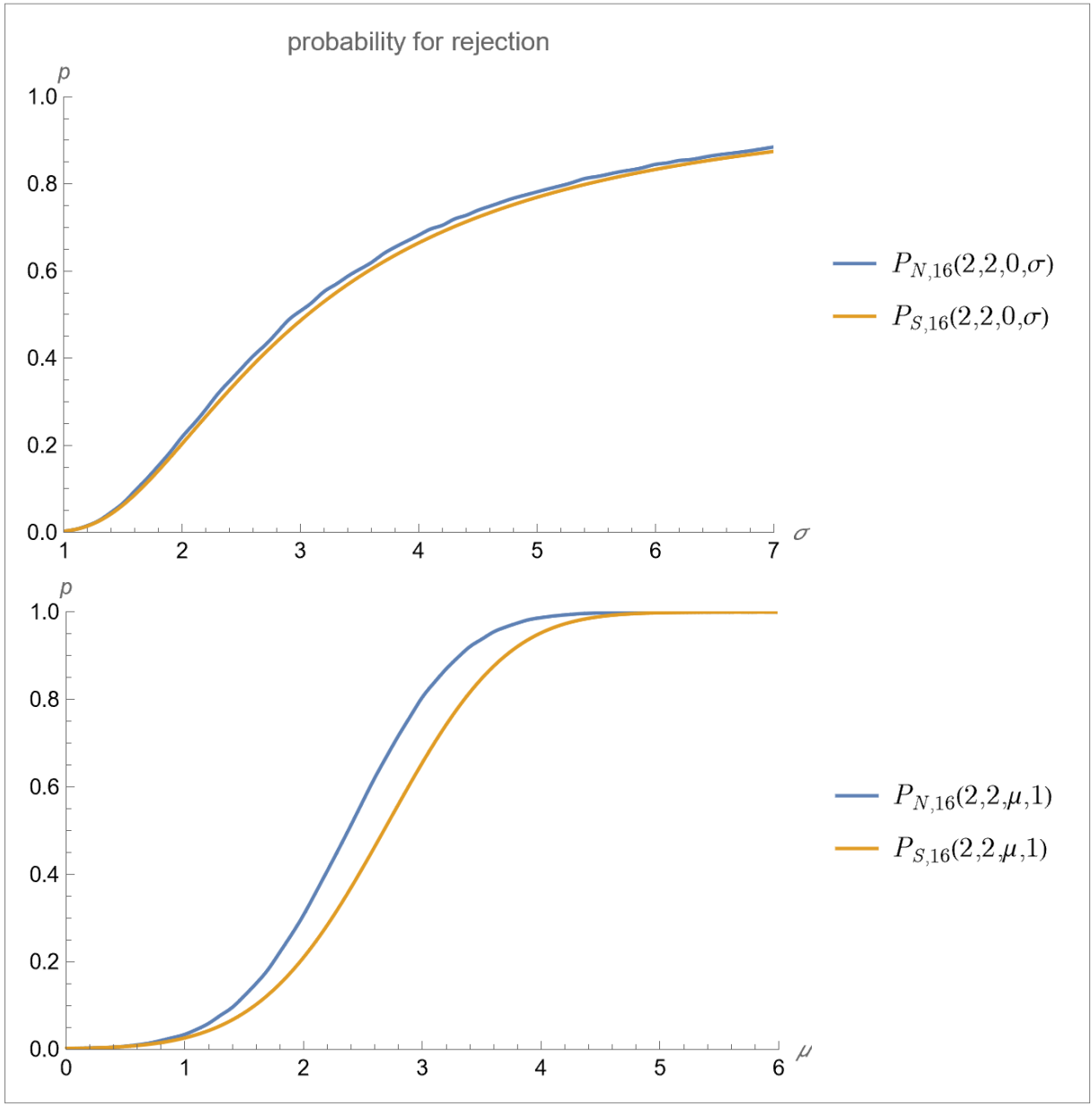

*Figure 73: $P_{N,16}(2,2,\mu,\sigma)$ vs $P_{S,16}(2,2,\mu,\sigma)$*

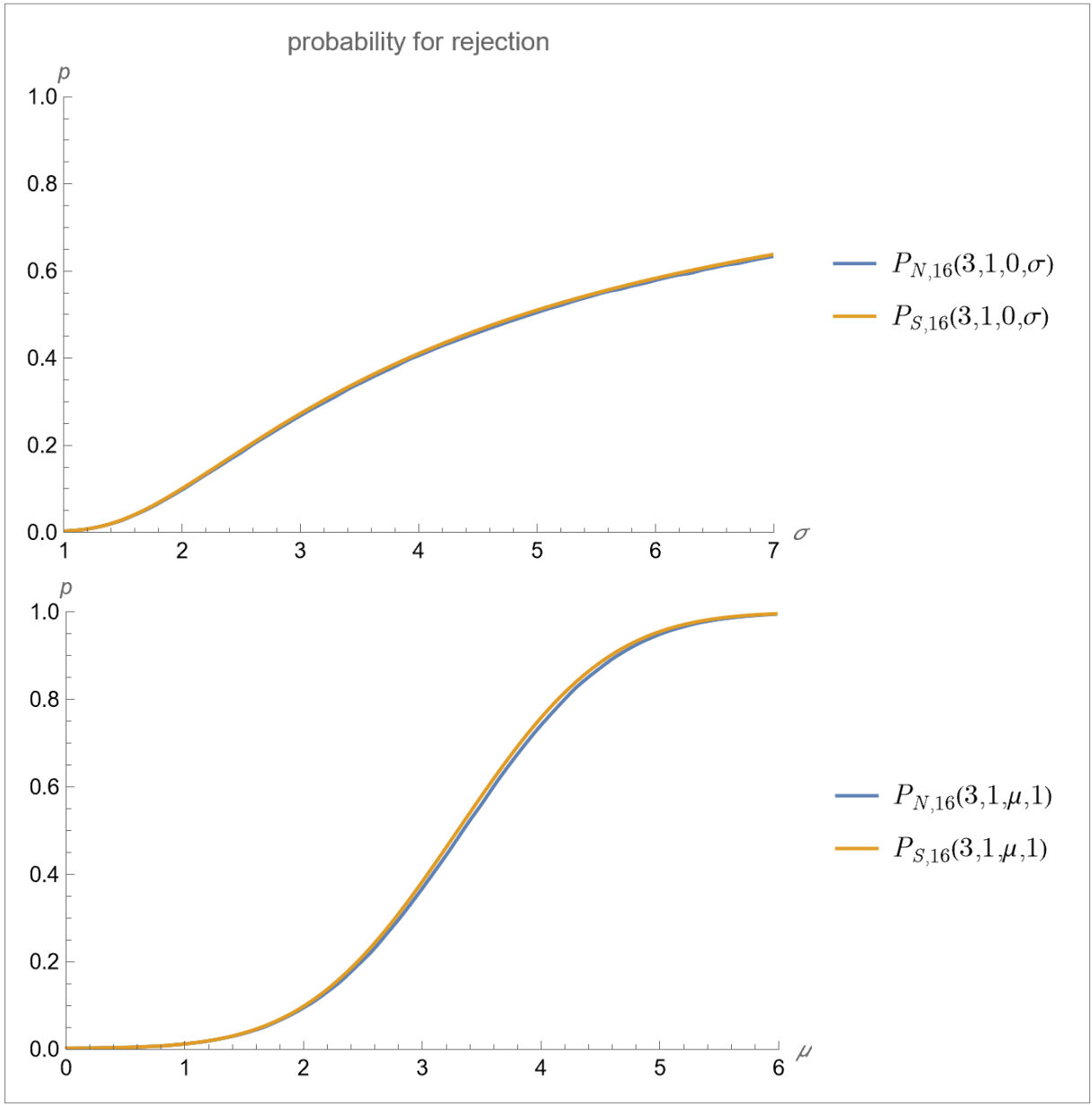

*Figure 74: $P_{N,16}(3,1,\mu,\sigma)$ vs $P_{S,16}(3,1,\mu,\sigma)$*

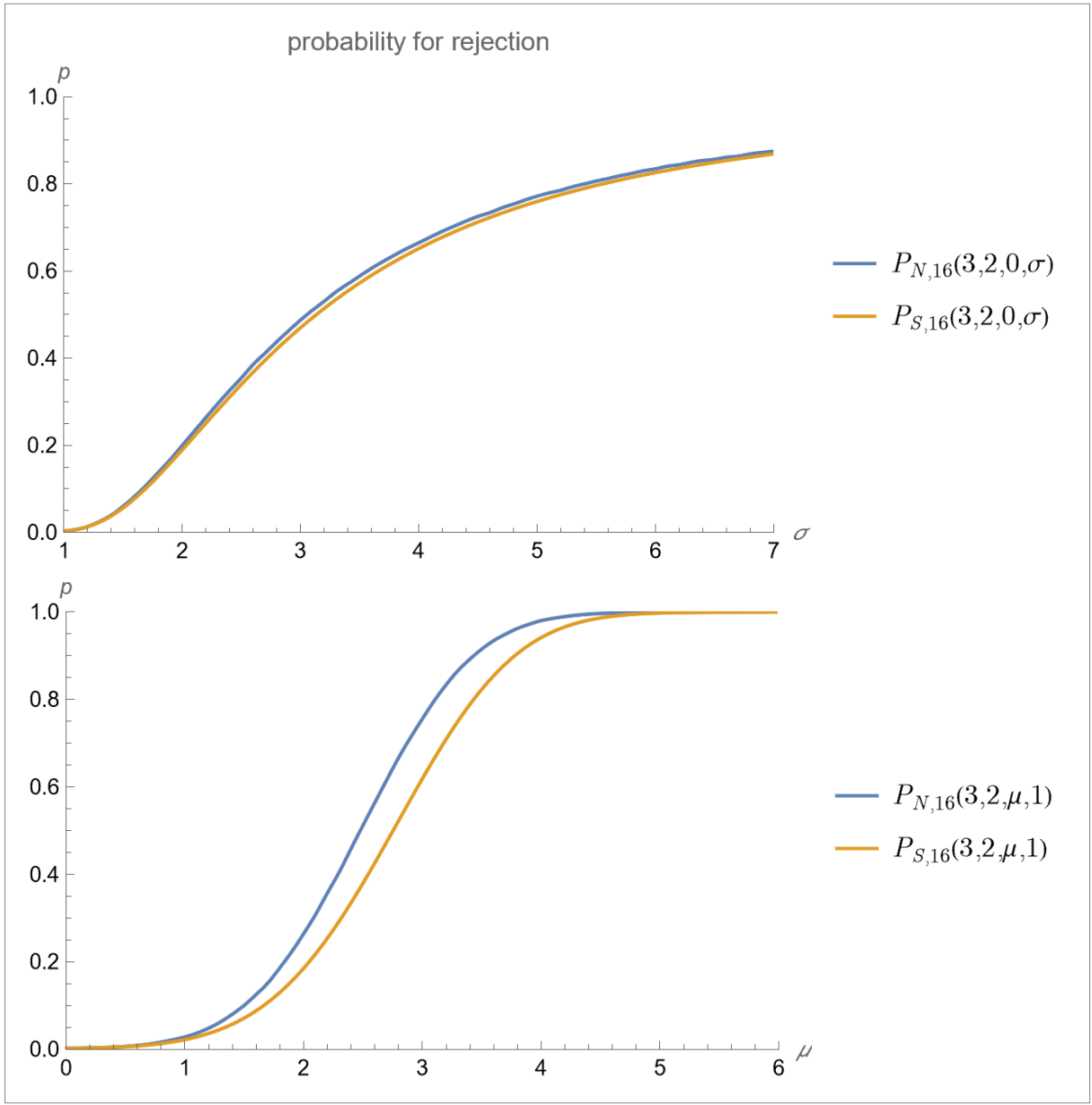

*Figure 75: $P_{N,16}(3,2,\mu,\sigma)$ vs $P_{S,16}(3,2,\mu,\sigma)$*

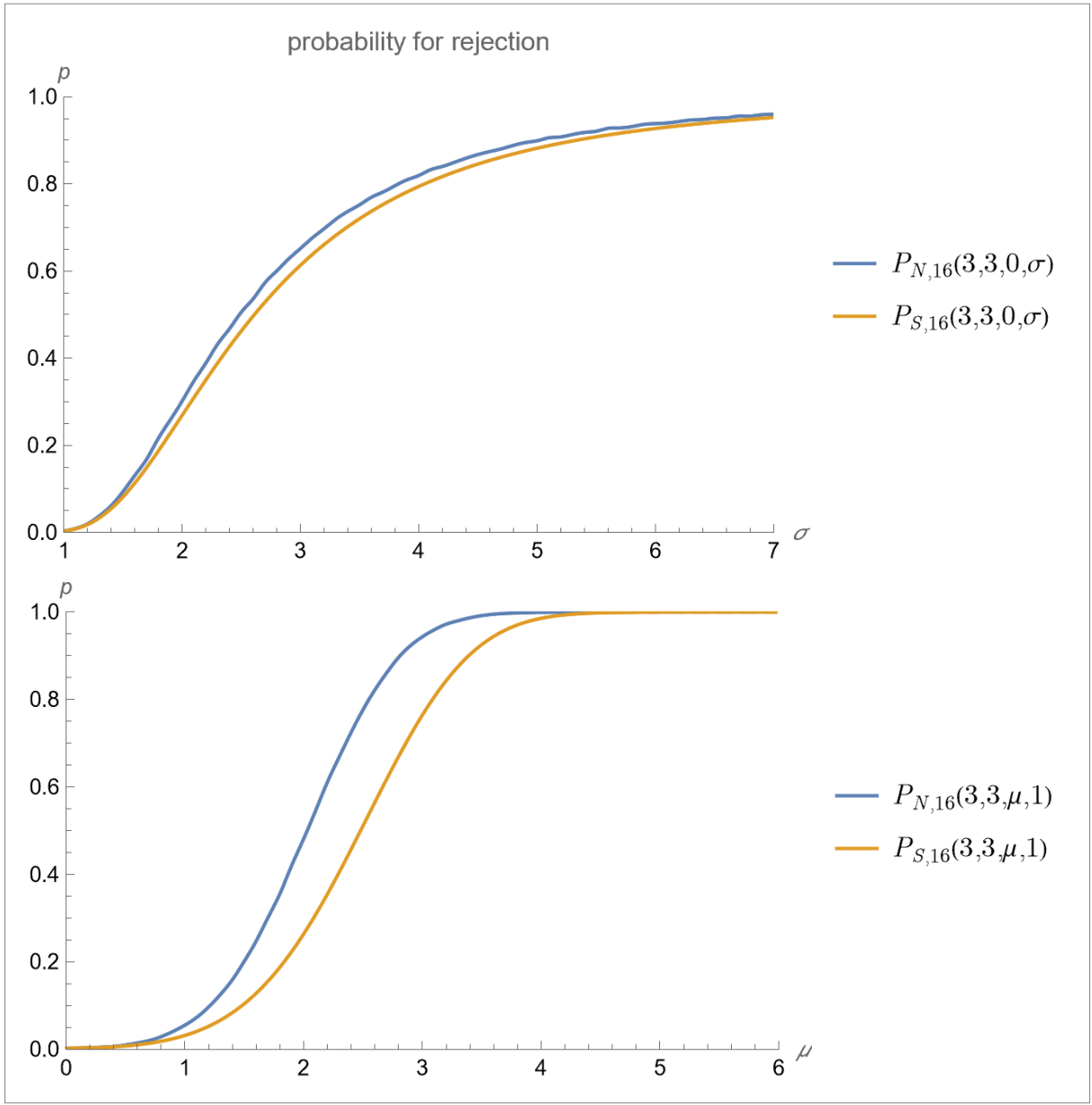

*Figure 76: $P_{N,16}(3,3,\mu,\sigma)$ vs $P_{S,16}(3,3,\mu,\sigma)$*

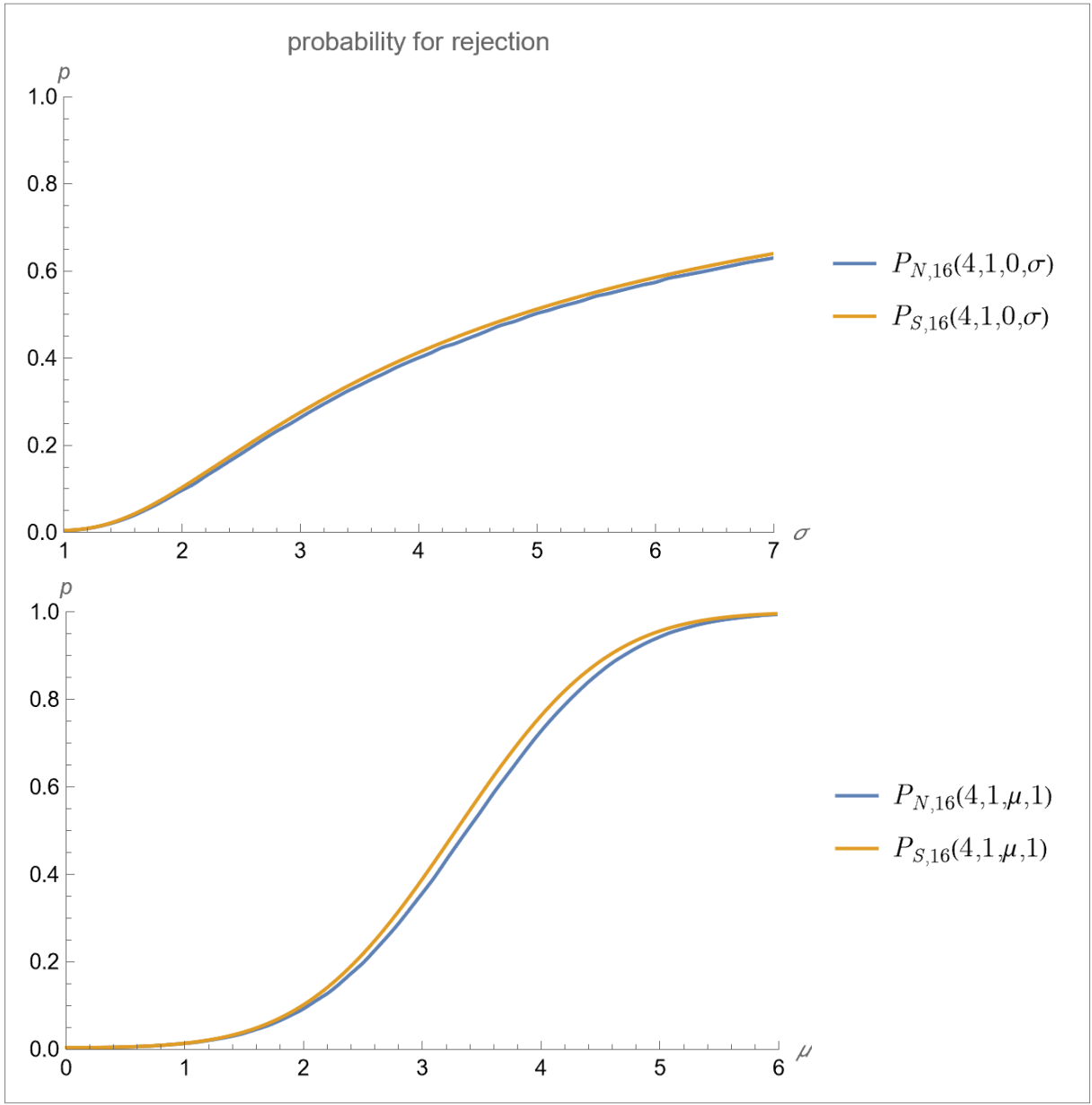

*Figure 77:* $P_{N,16}(4,1,\mu,\sigma)$ *vs* $P_{S,16}(4,1,\mu,\sigma)$

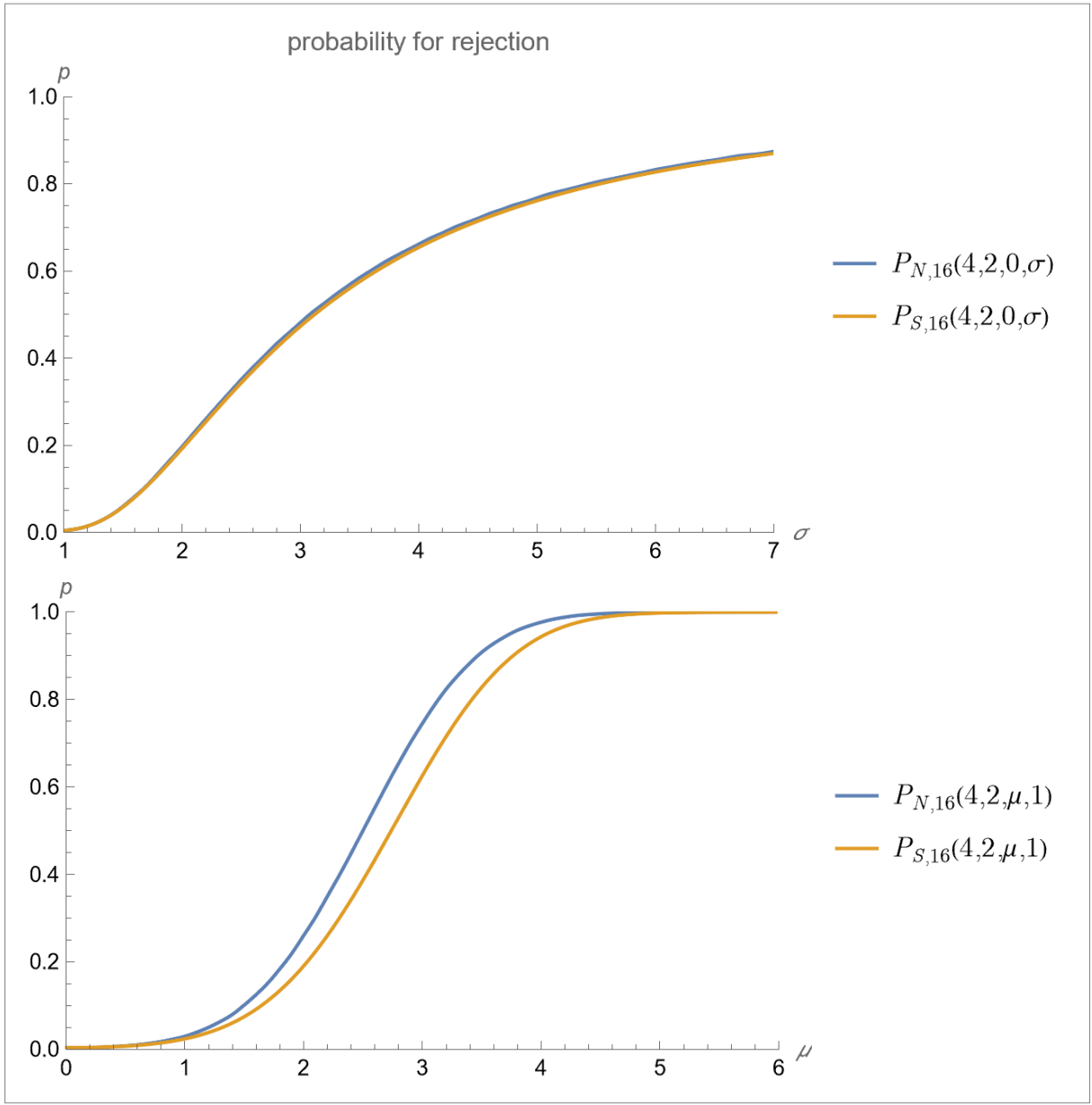

*Figure 78: $P_{N,16}(4,2,\mu,\sigma)$ vs $P_{S,16}(4,2,\mu,\sigma)$*

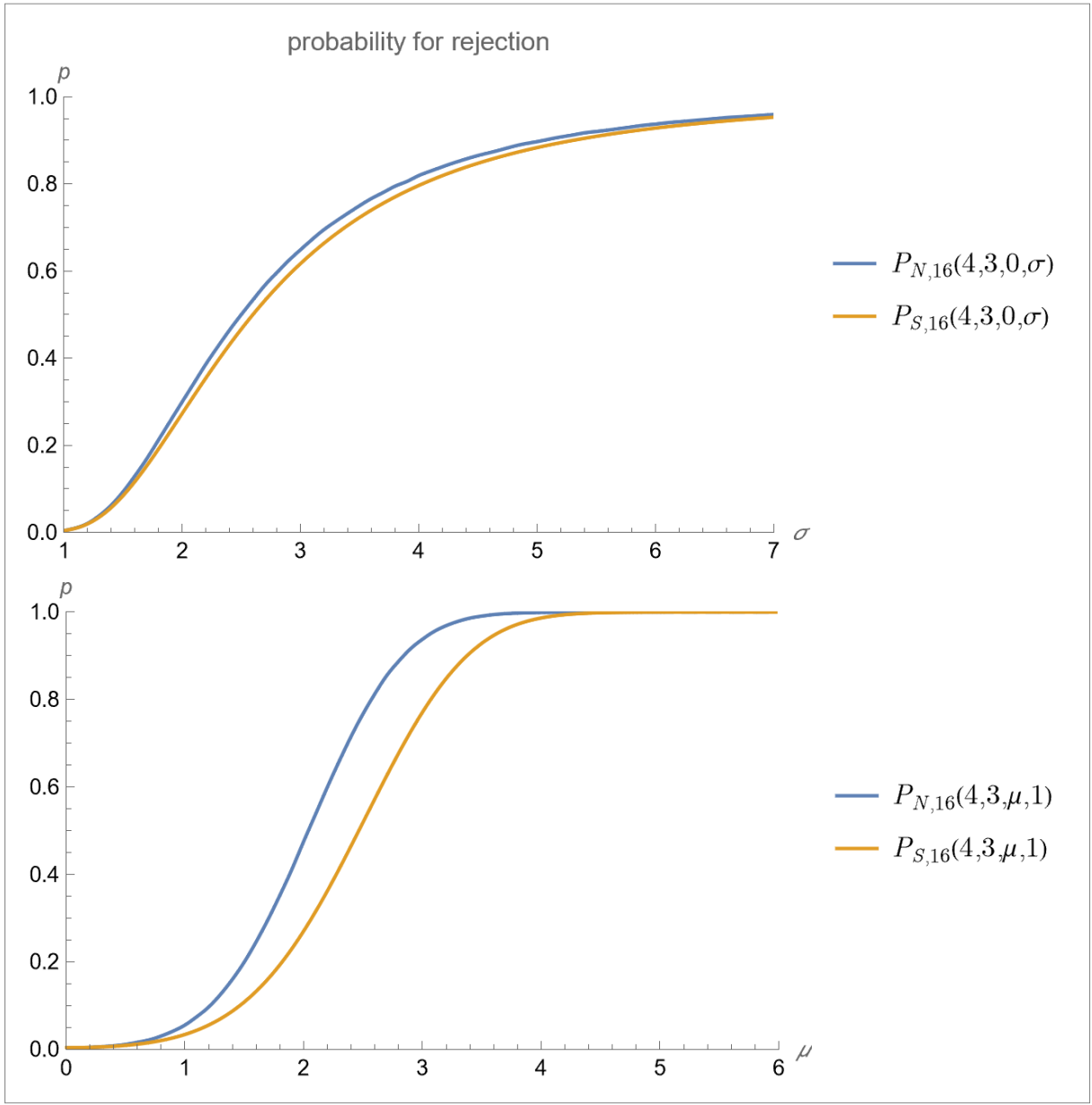

*Figure 79:* $P_{N,16}(4,3,\mu,\sigma)$ vs $P_{S,16}(4,3,\mu,\sigma)$

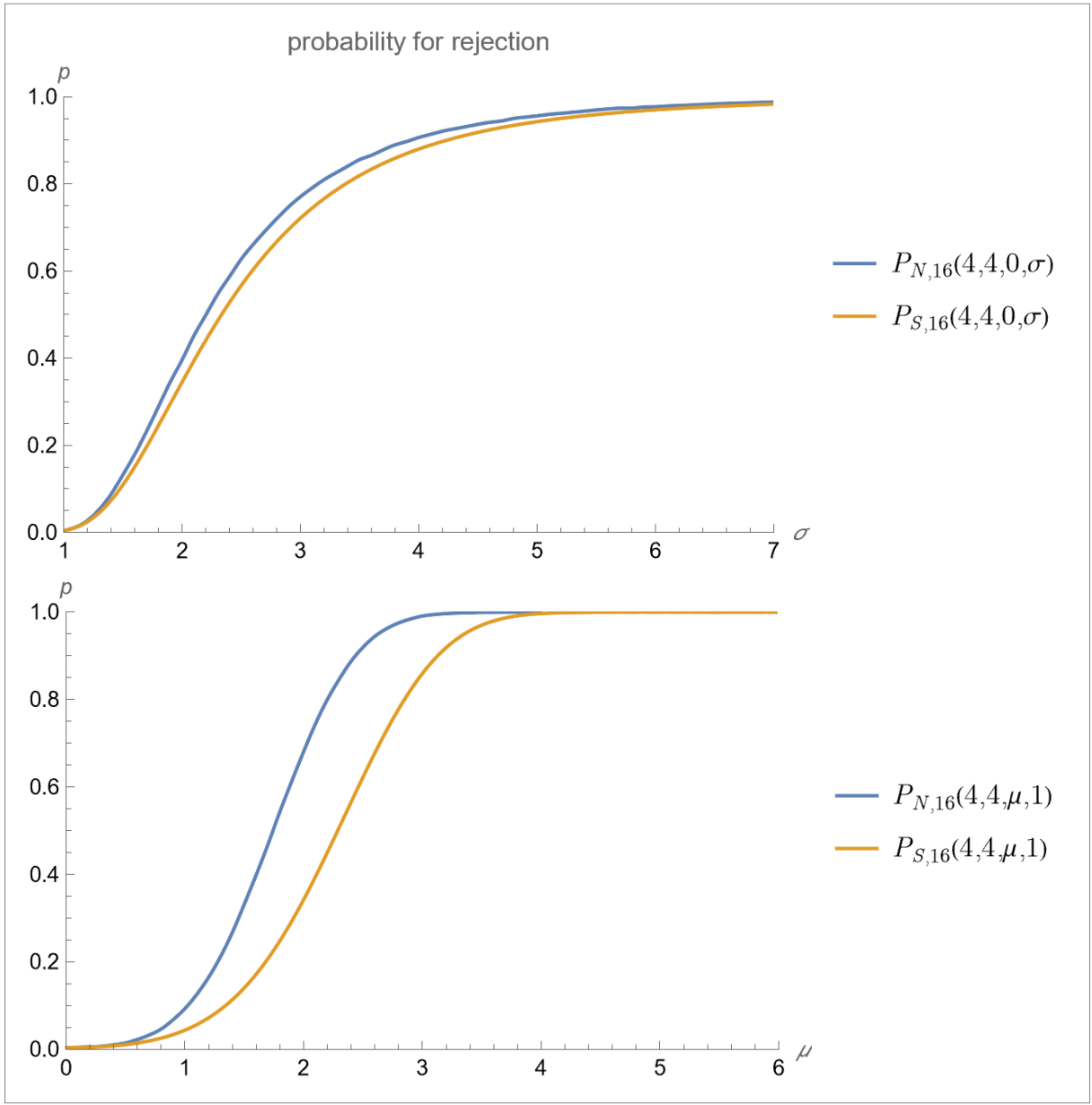

*Figure 80: $P_{N,16}(4,4,\mu,\sigma)$ vs $P_{S,16}(4,4,\mu,\sigma)$*

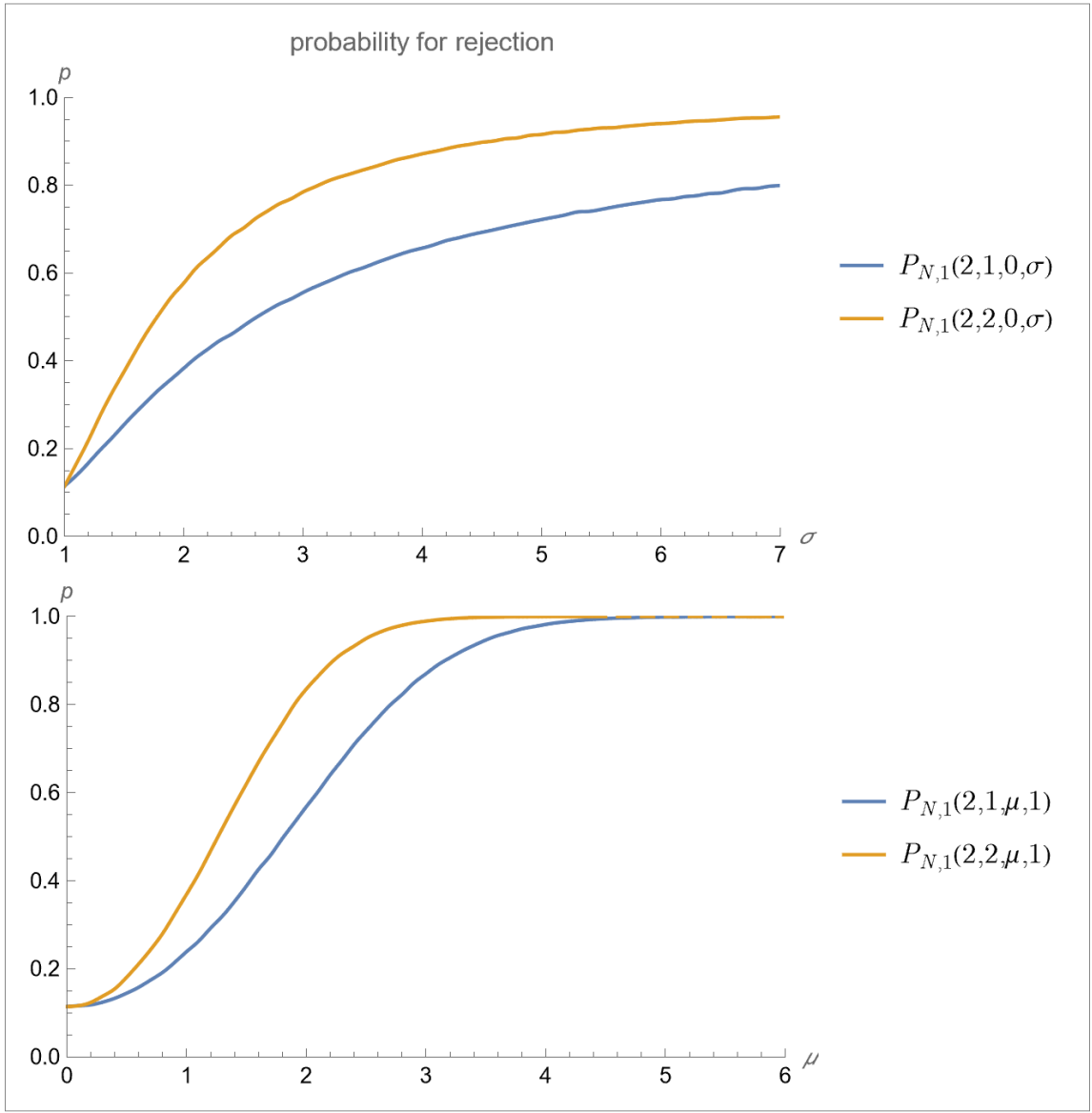

*Figure 81: $P_{N,1}(2, k, \mu, \sigma)$*

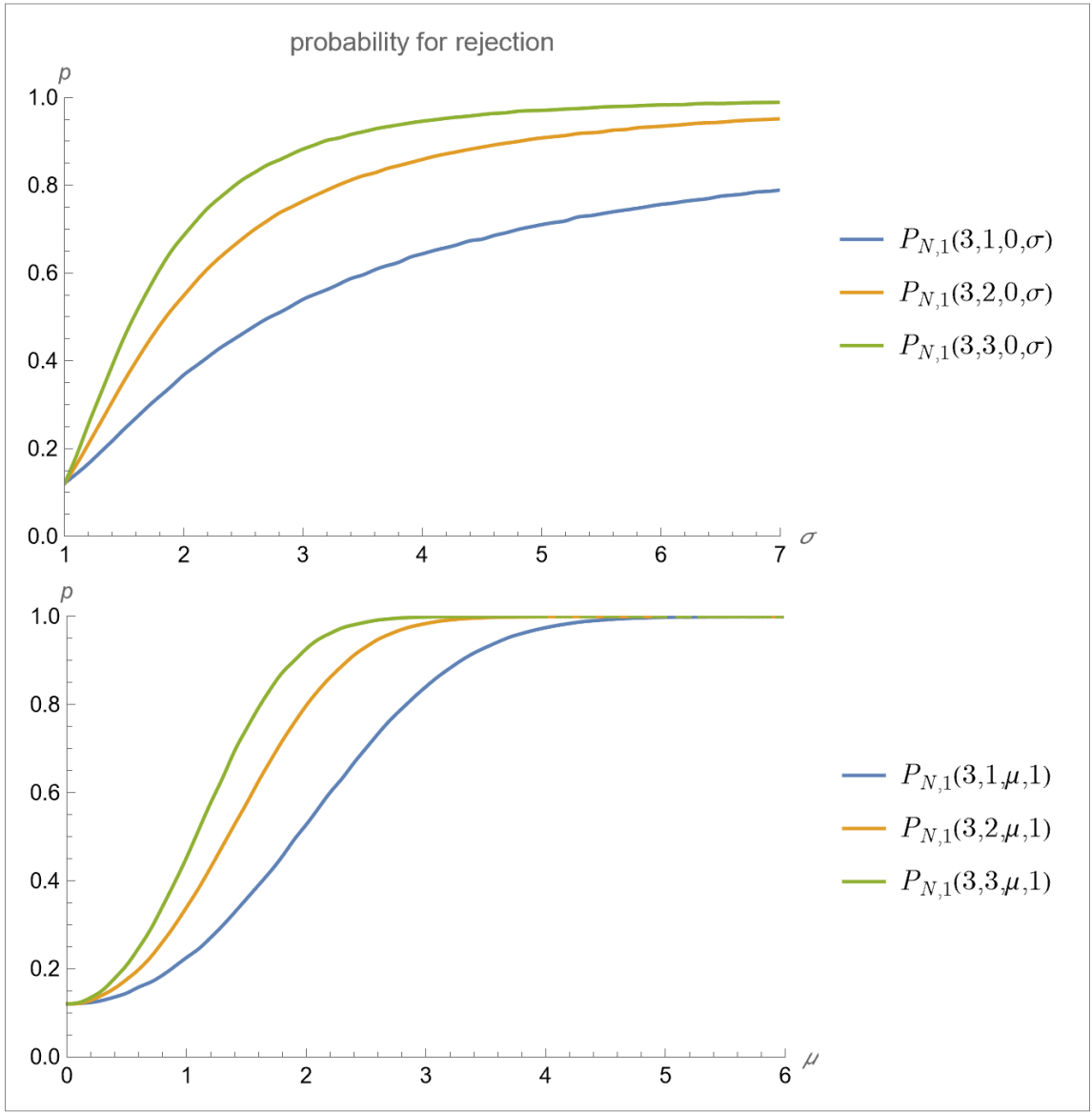

*Figure 82: $P_{N,1}(3, k, \mu, \sigma)$*

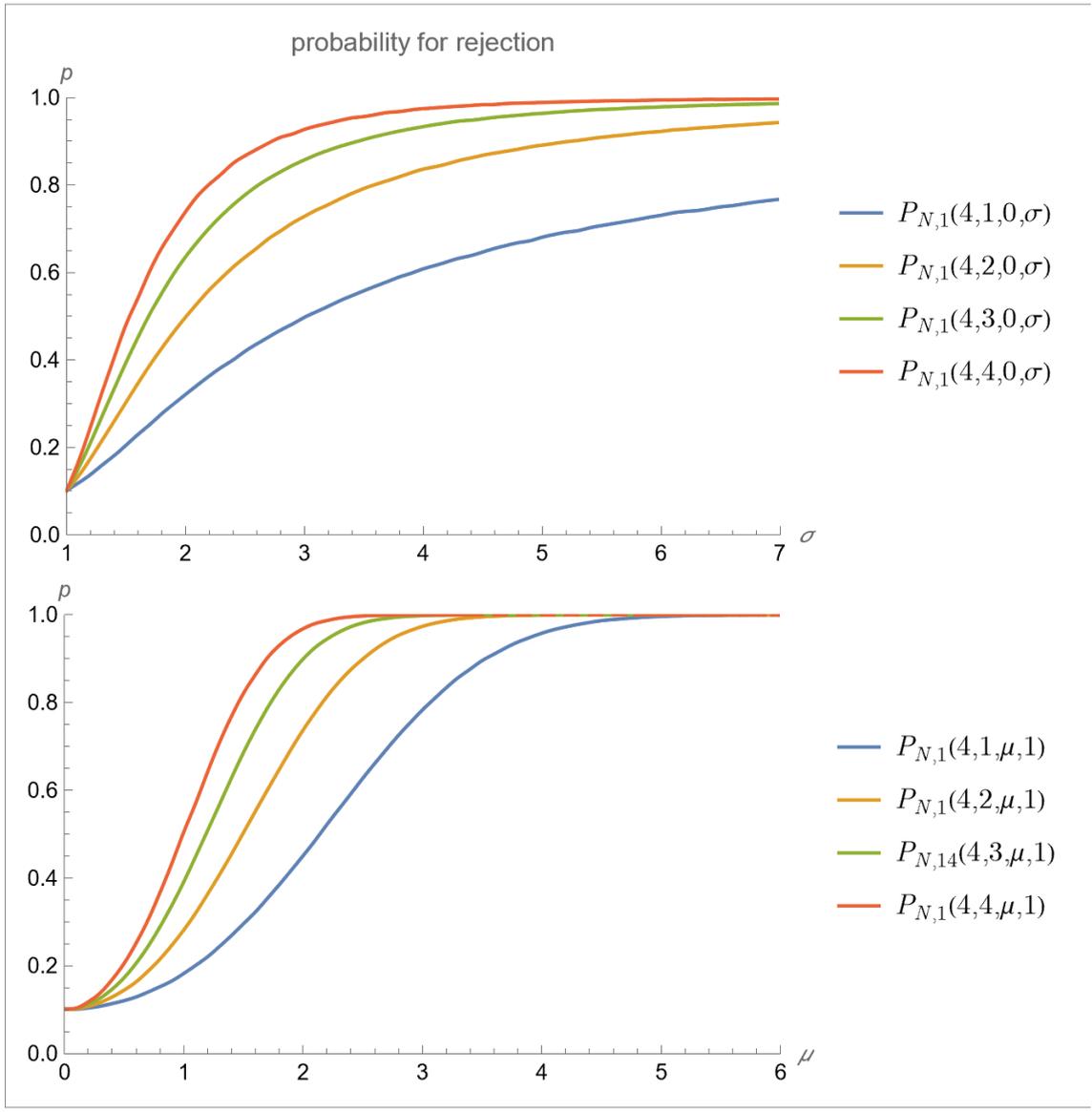

*Figure 83: $P_{N,1}(4, k, \mu, \sigma)$*

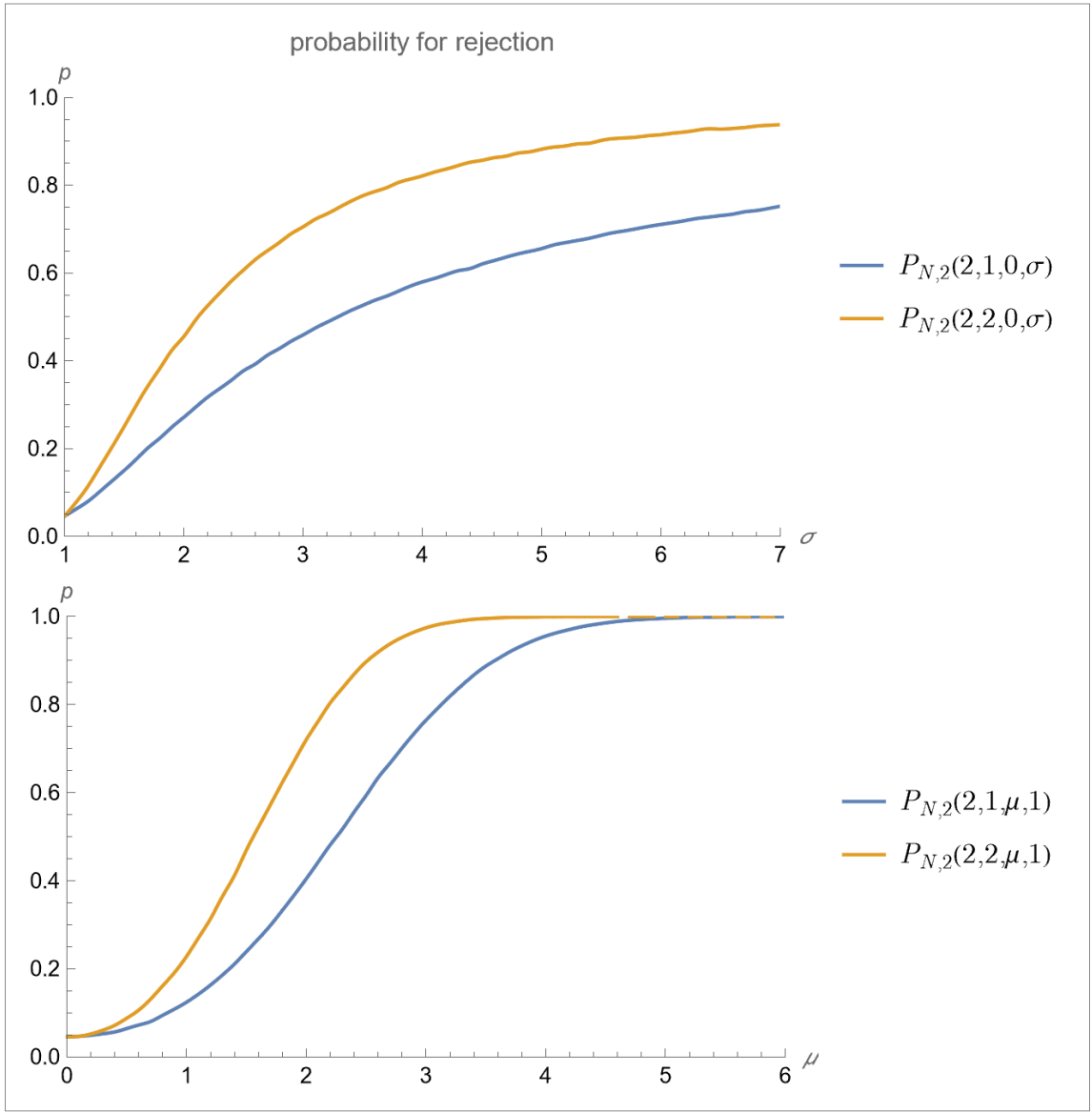

*Figure 84: $P_{N,2}(2, k, \mu, \sigma)$*

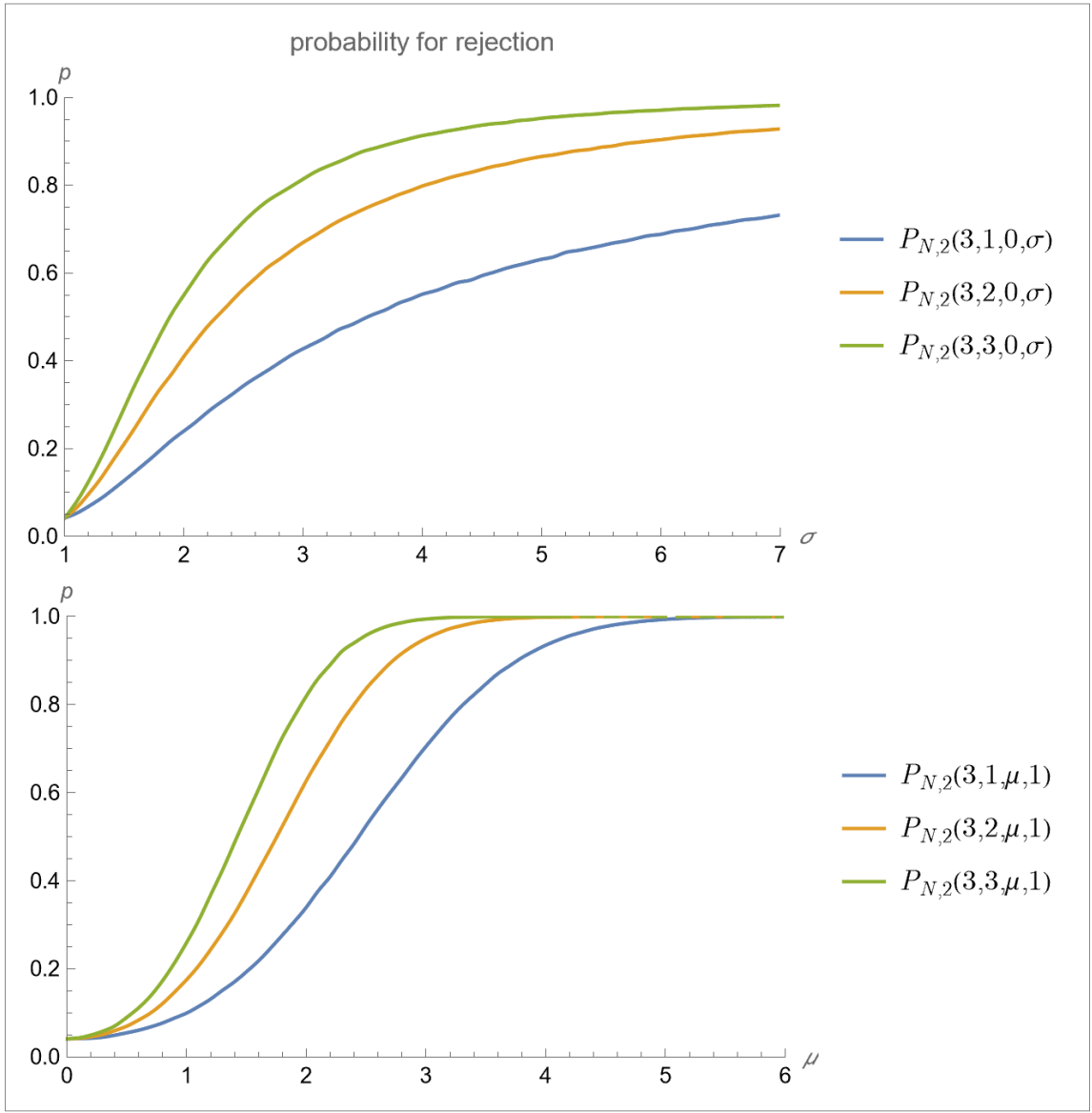

*Figure 85: $P_{N,2}(3,k,\mu,\sigma)$*

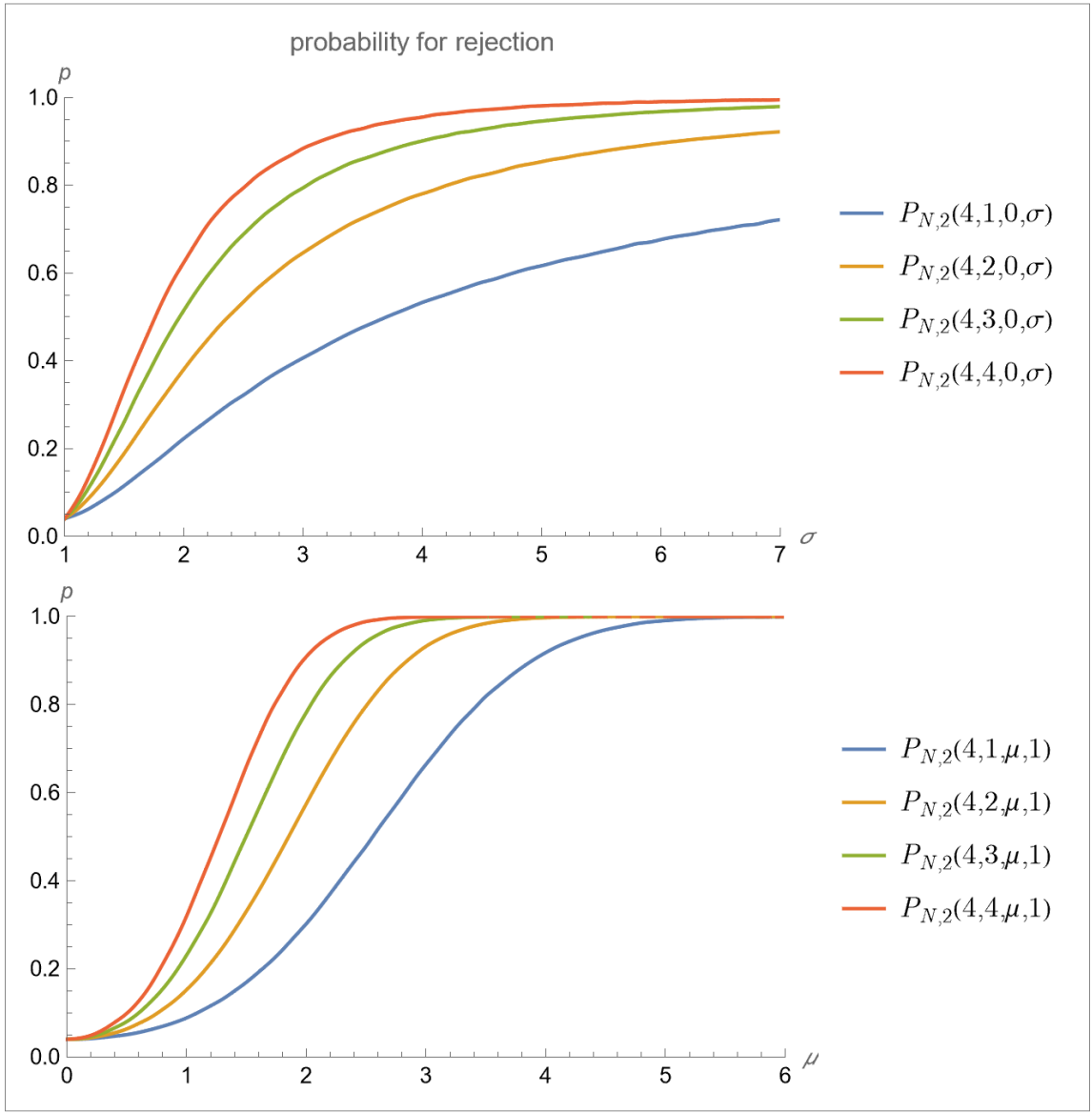

*Figure 86: $P_{N,2}(4, k, \mu, \sigma)$*

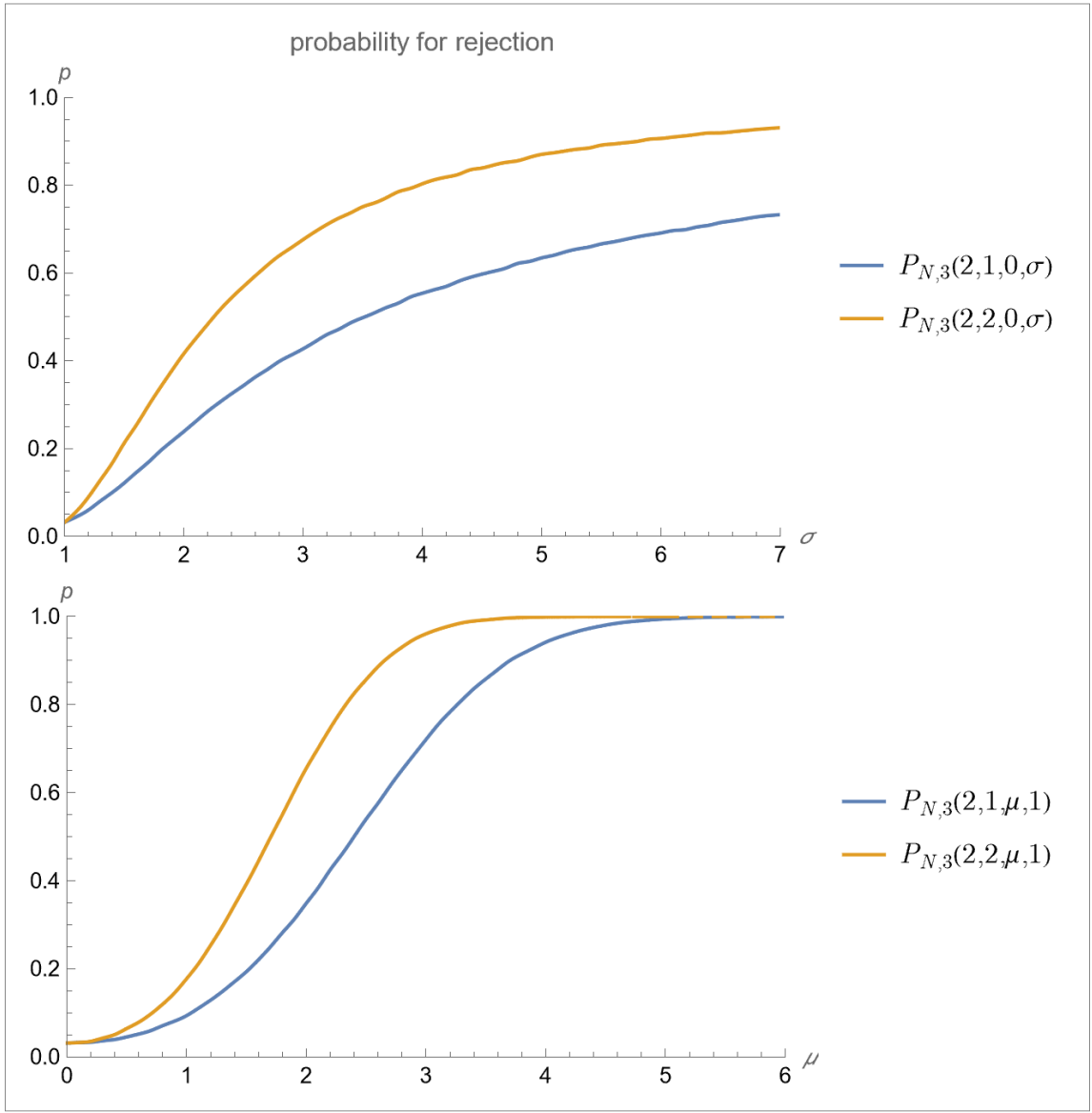

*Figure 87: $P_{N,3}(2,k,\mu,\sigma)$*

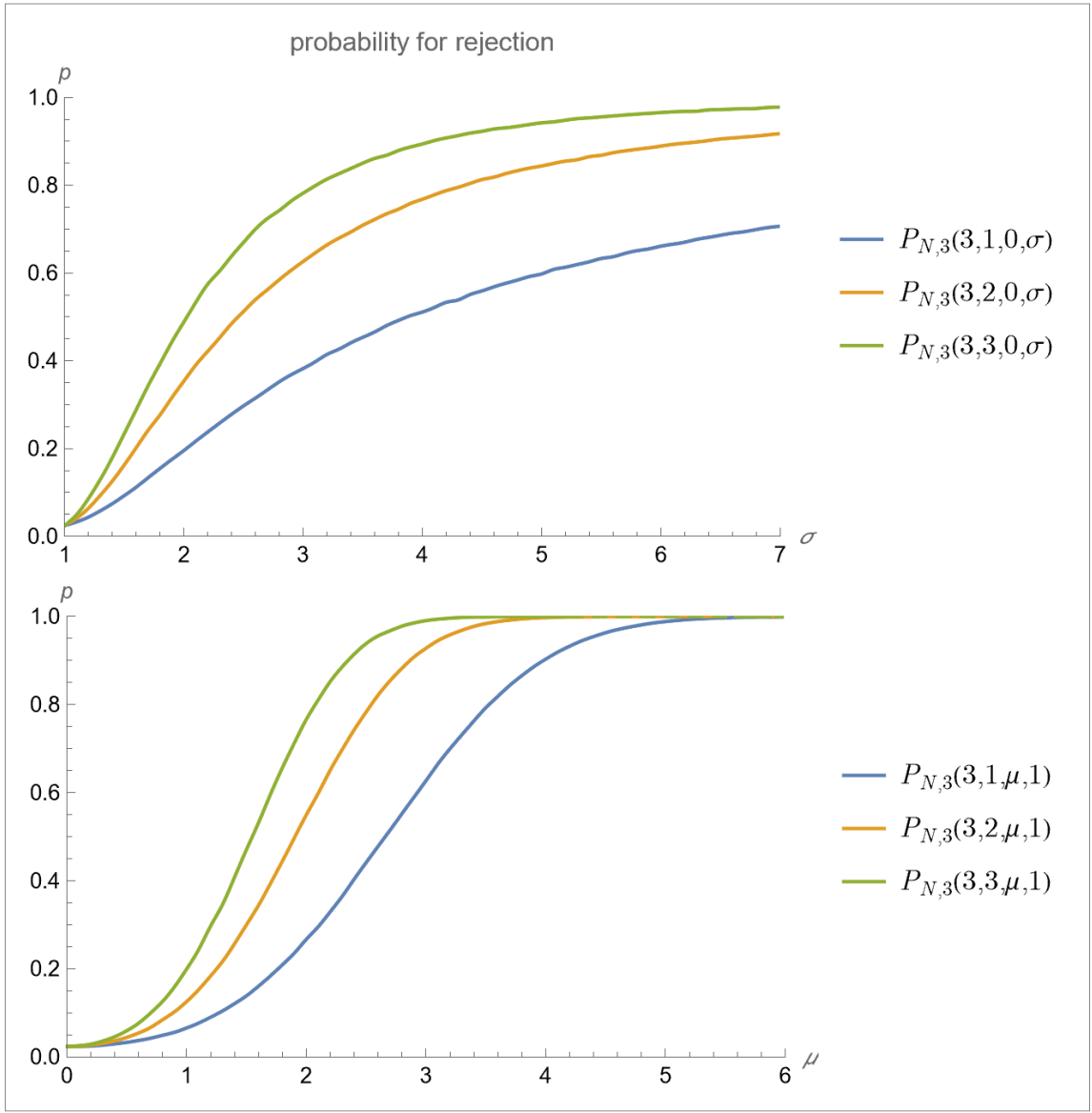

*Figure 88: $P_{N,3}(3,k,\mu,\sigma)$*

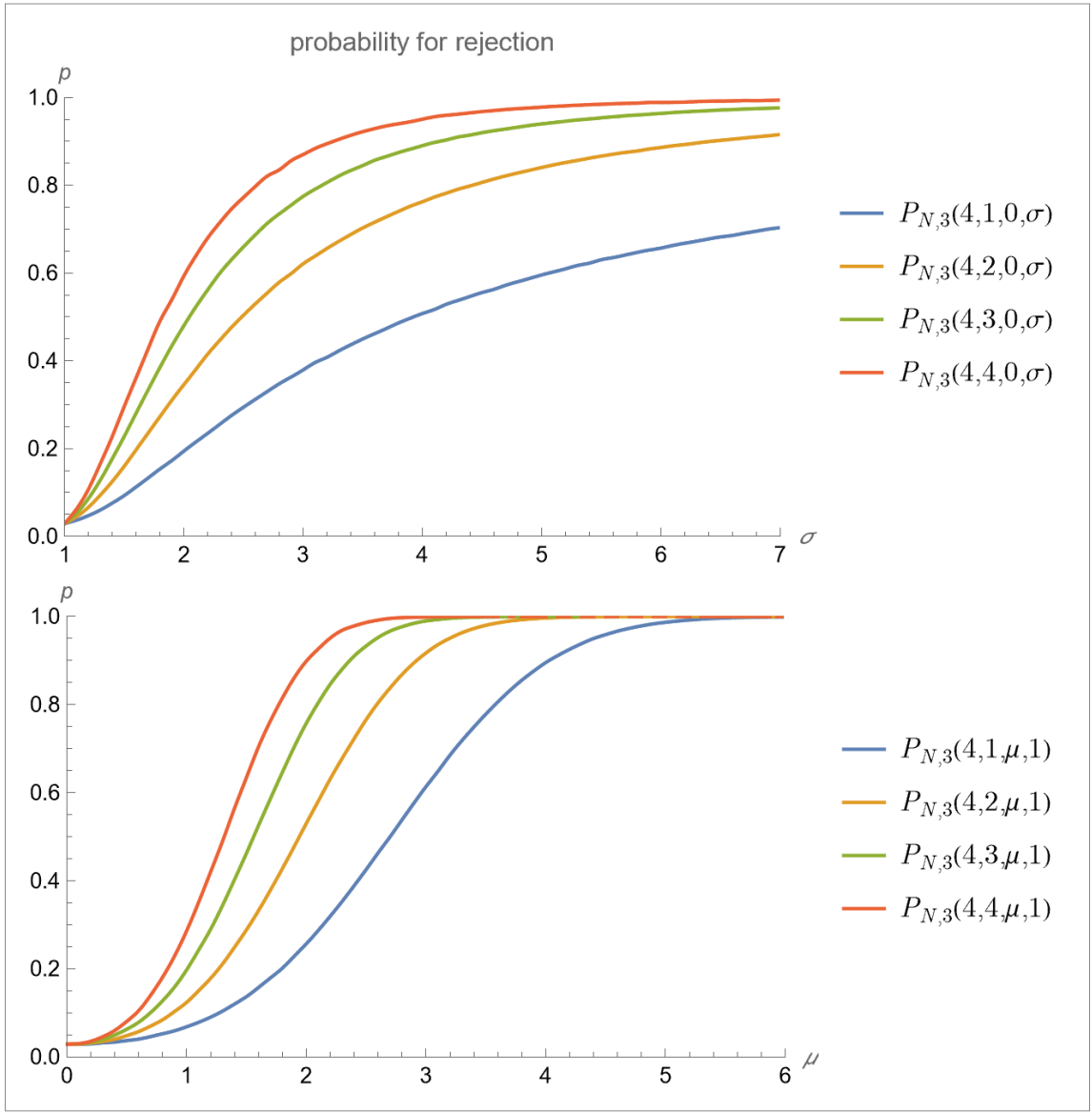

*Figure 89: $P_{N,3}(4, k, \mu, \sigma)$*

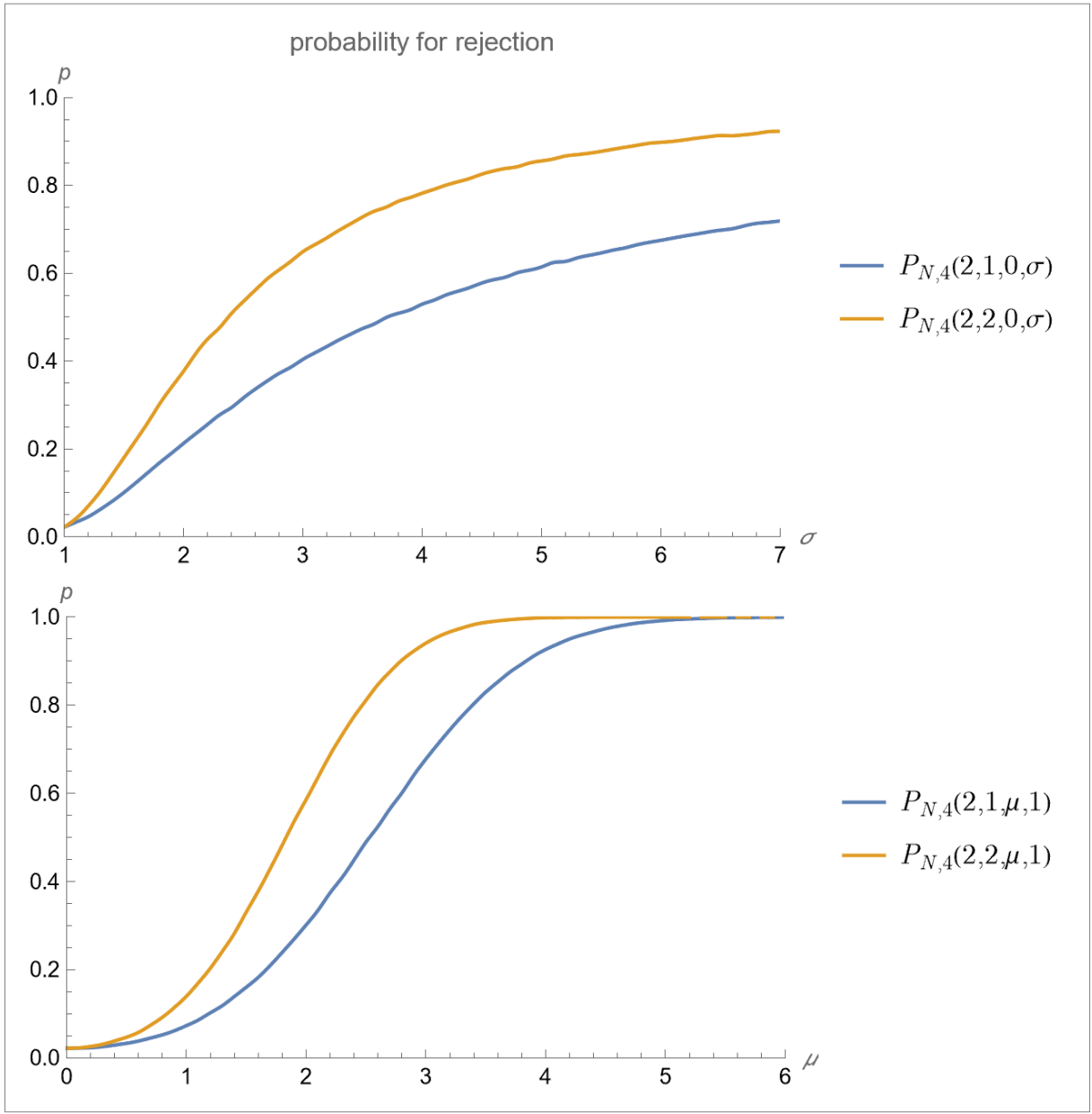

*Figure 90:* $P_{N,4}(2, k, \mu, \sigma)$

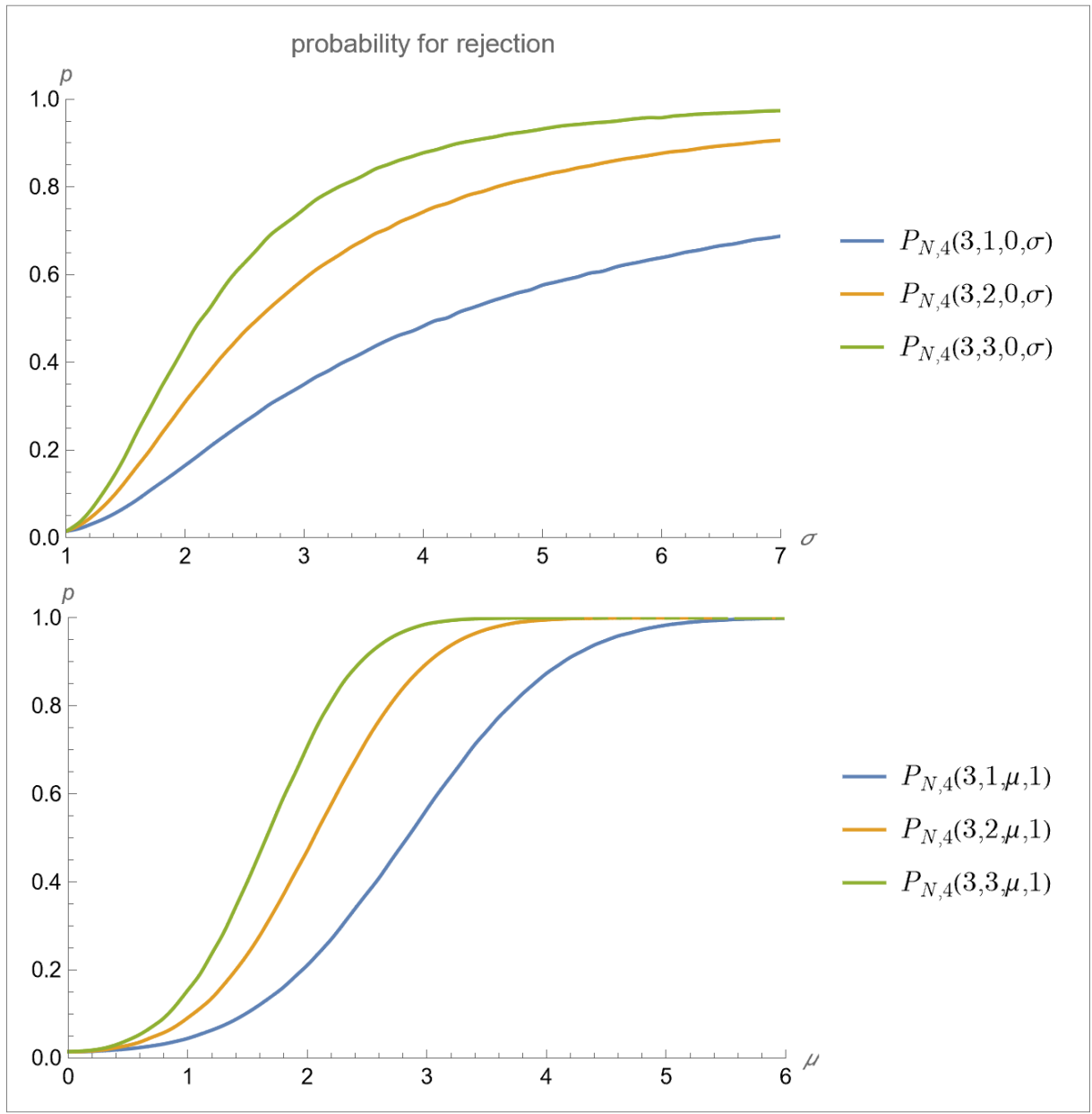

*Figure 91:* $P_{N,4}(3, k, \mu, \sigma)$

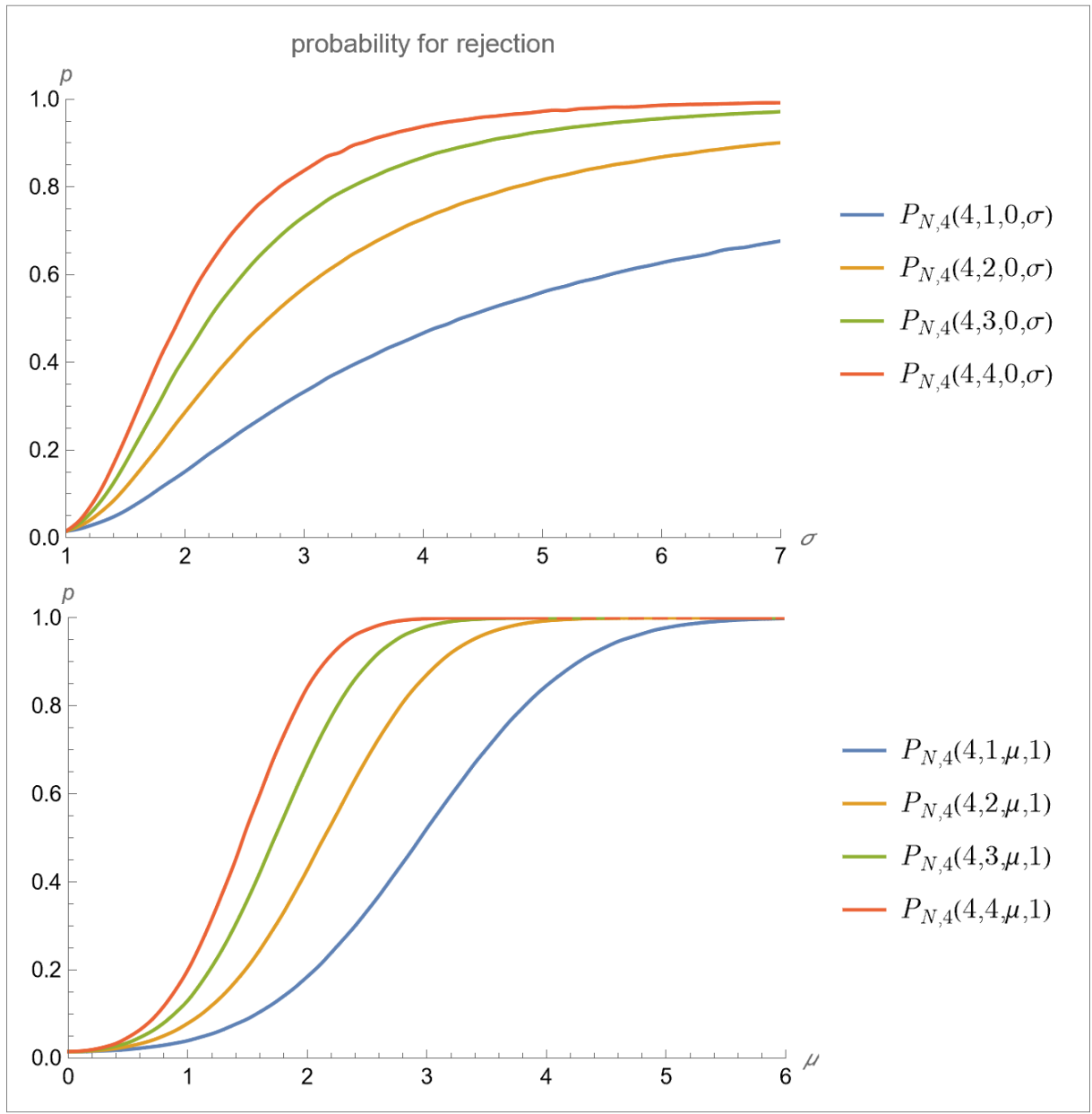

*Figure 92: $P_{N,4}(4, k, \mu, \sigma)$*

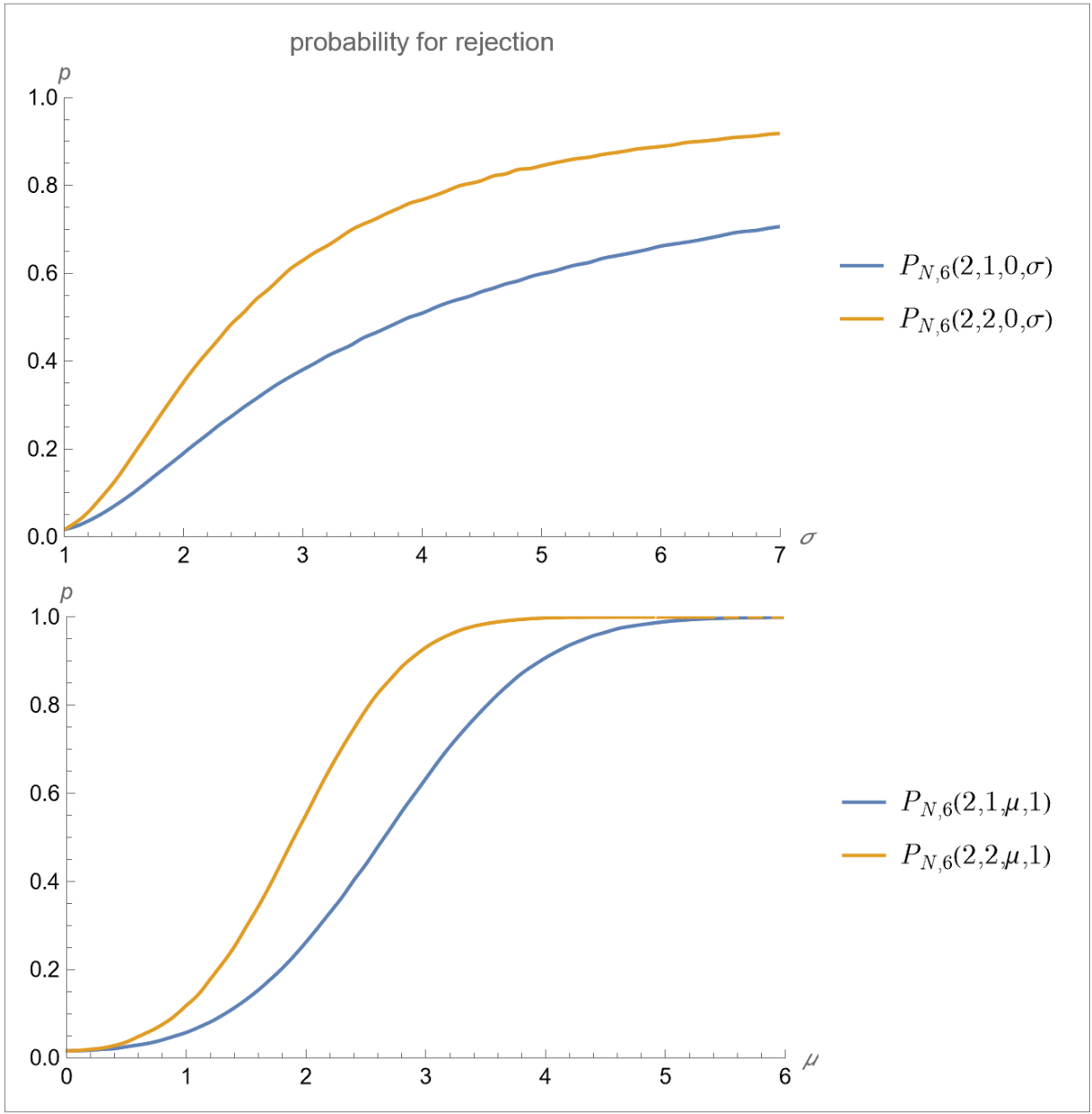

*Figure 93: $P_{N,6}(2,k,\mu,\sigma)$*

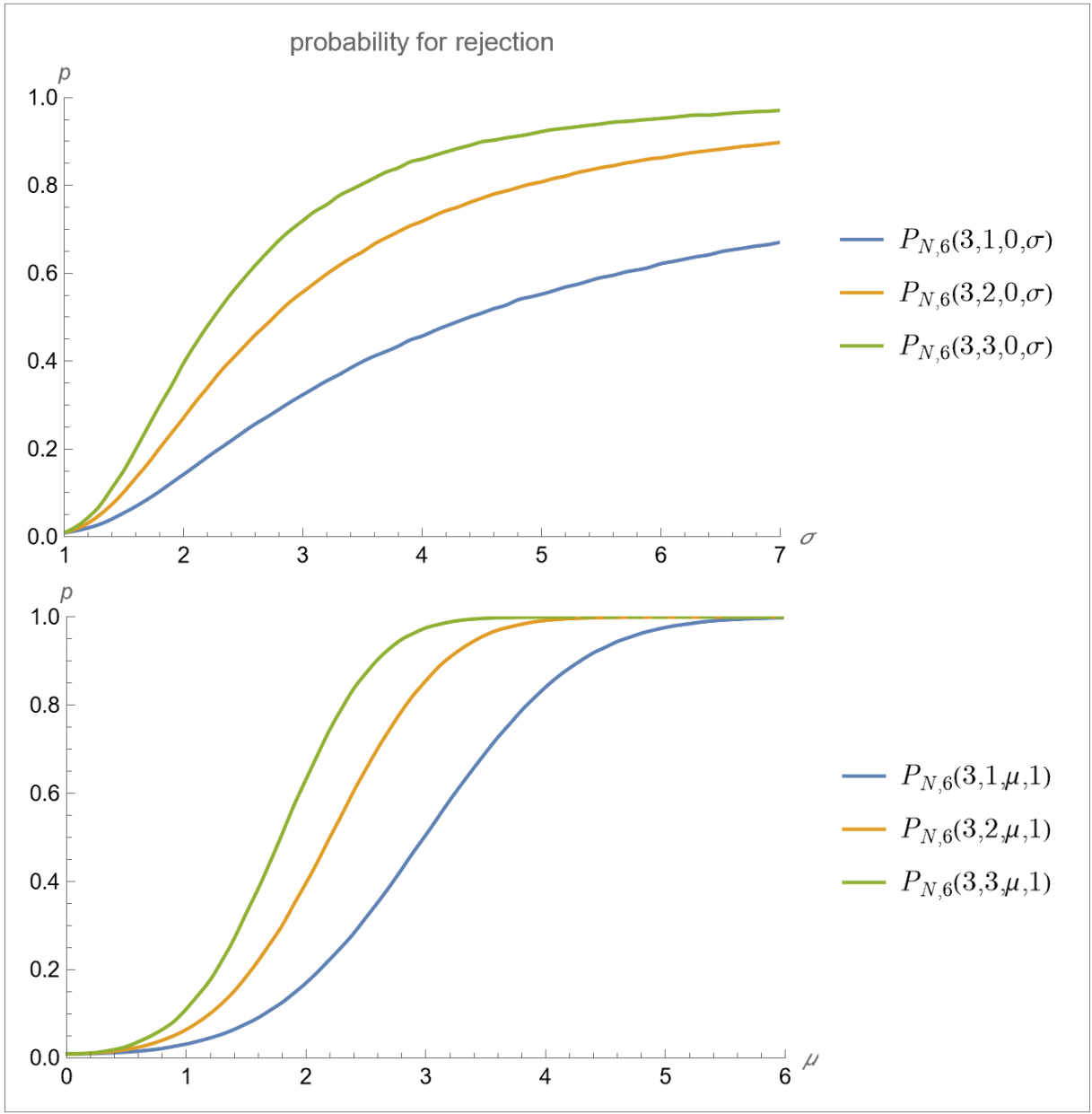

*Figure 94: $P_{N,6}(3,k,\mu,\sigma)$*

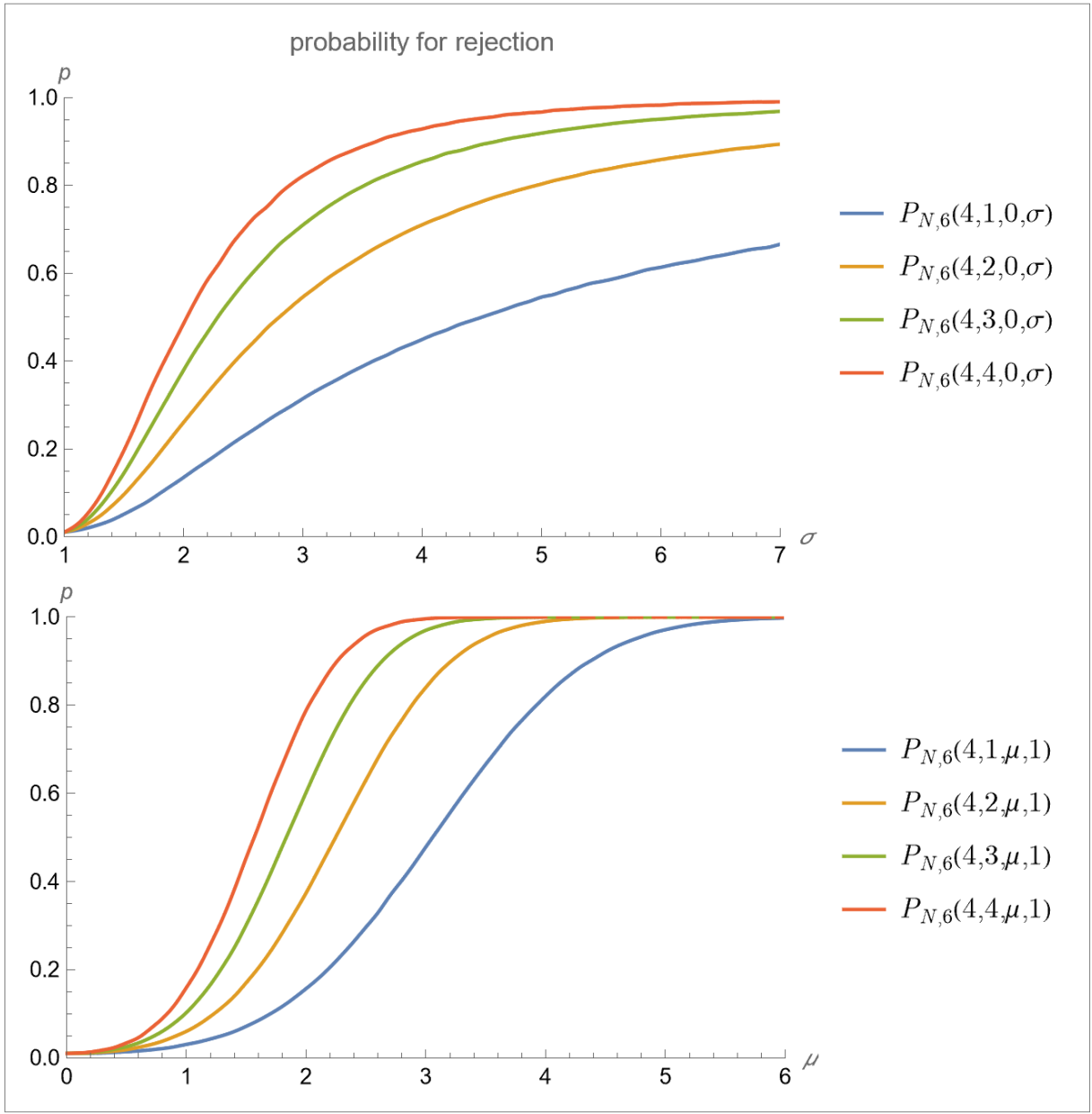

*Figure 95:* $P_{N,6}(4, k, \mu, \sigma)$

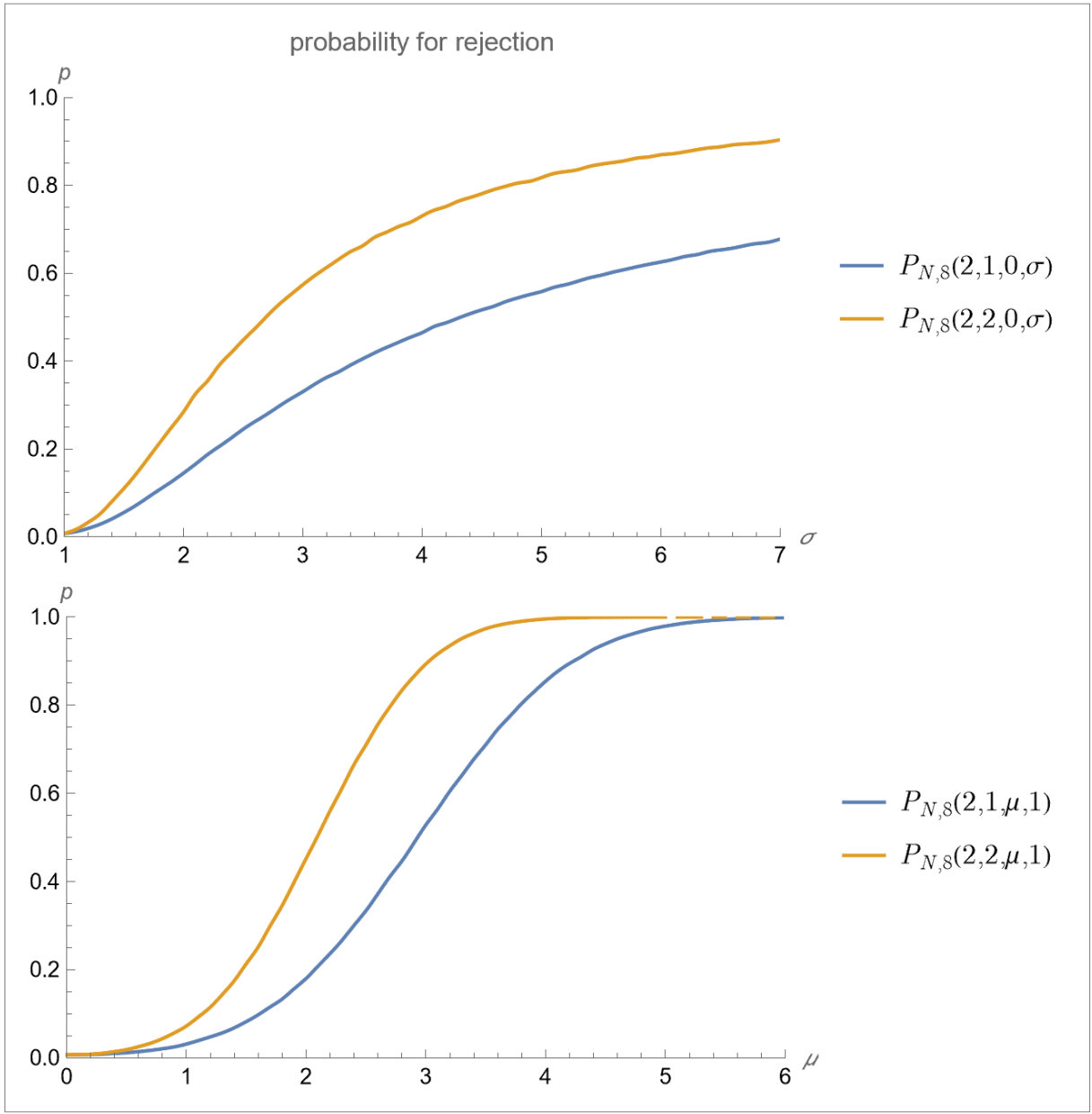

*Figure 96: $P_{N,8}(2,k,\mu,\sigma)$*

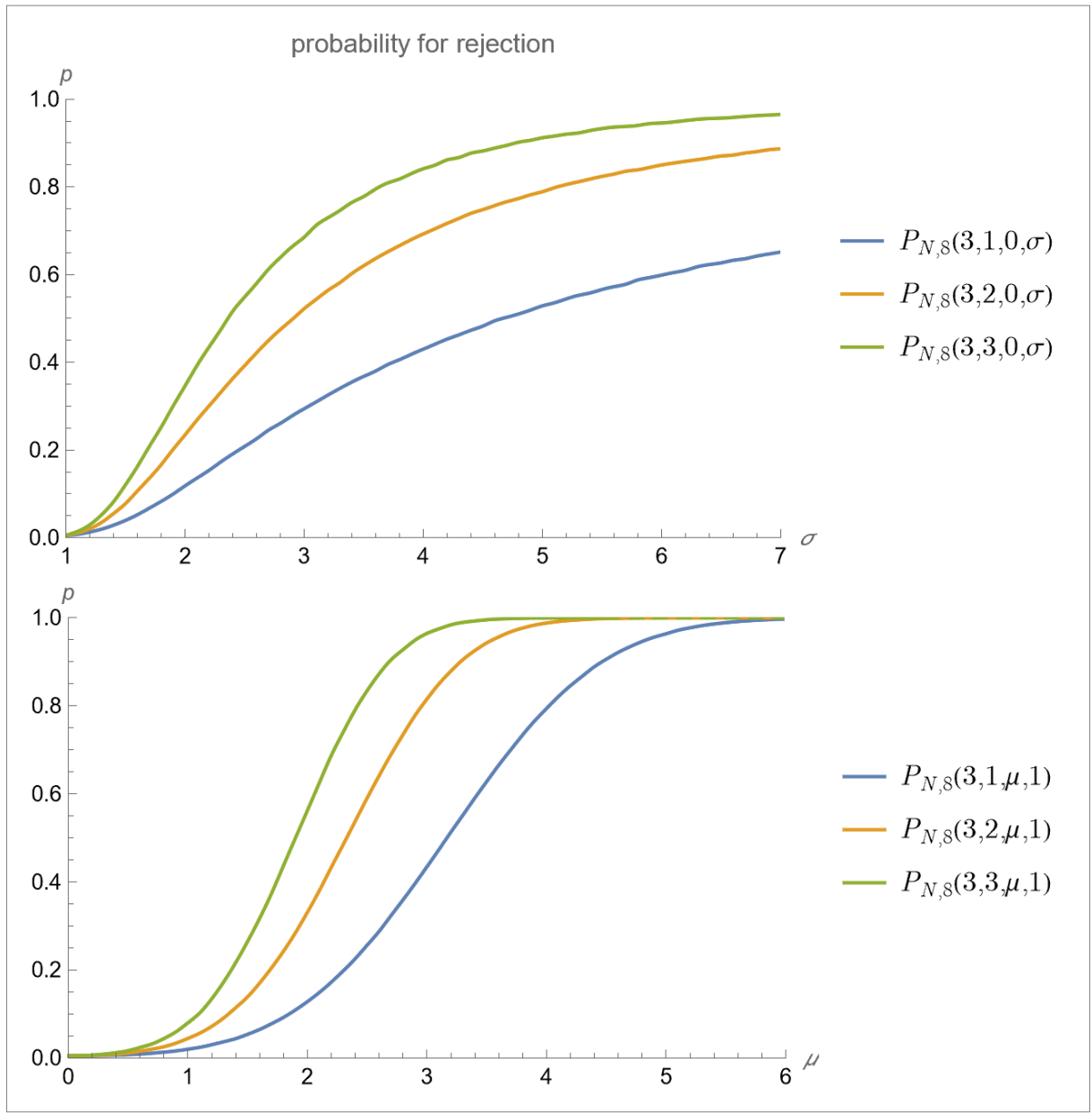

*Figure 97:* $P_{N,8}(3,k,\mu,\sigma)$

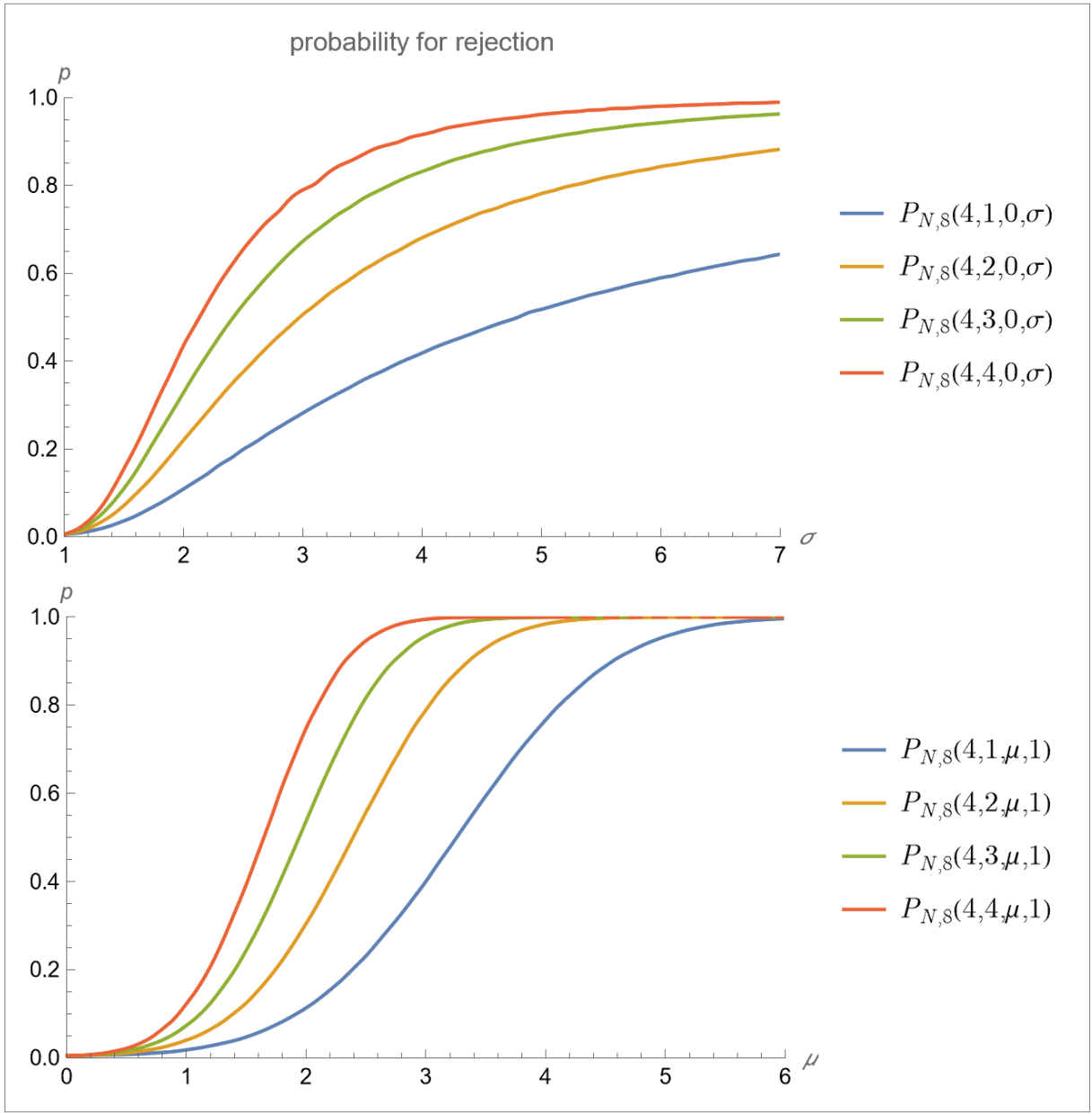

*Figure 98: $P_{N,8}(4, k, \mu, \sigma)$*

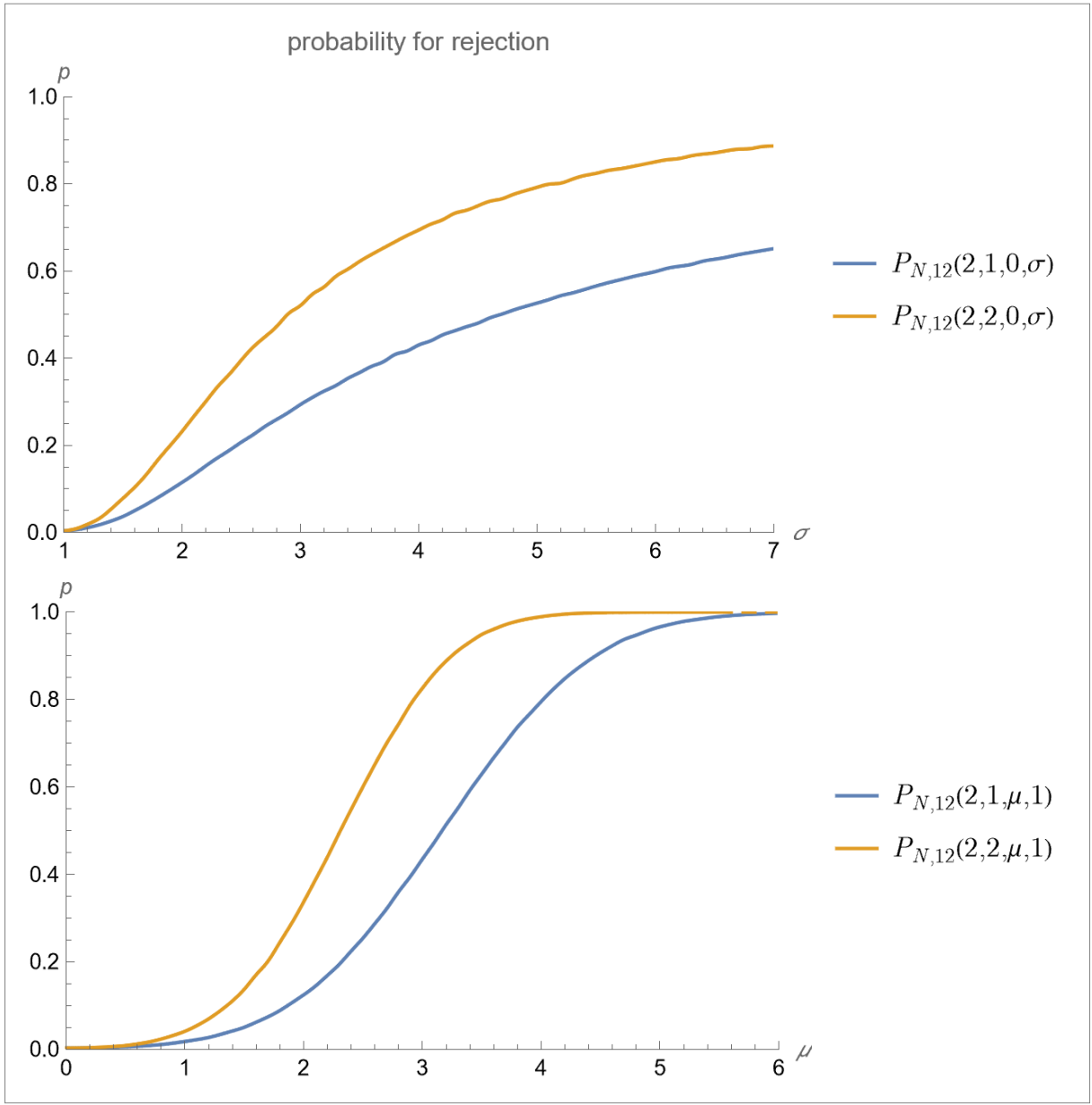

*Figure 99:* $P_{N,12}(2,k,\mu,\sigma)$

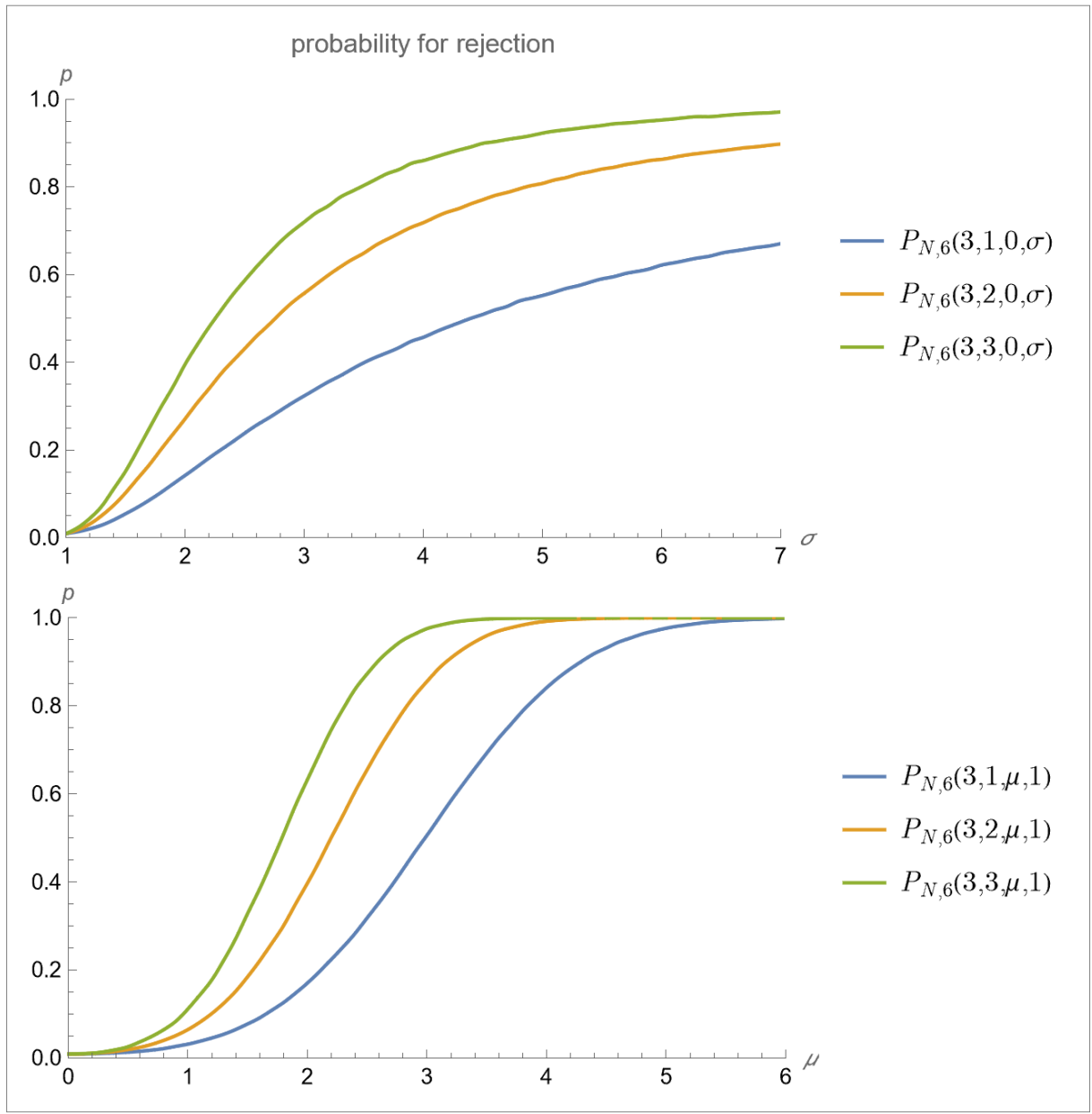

*Figure 100:* $P_{N,12}(3, k, \mu, \sigma)$

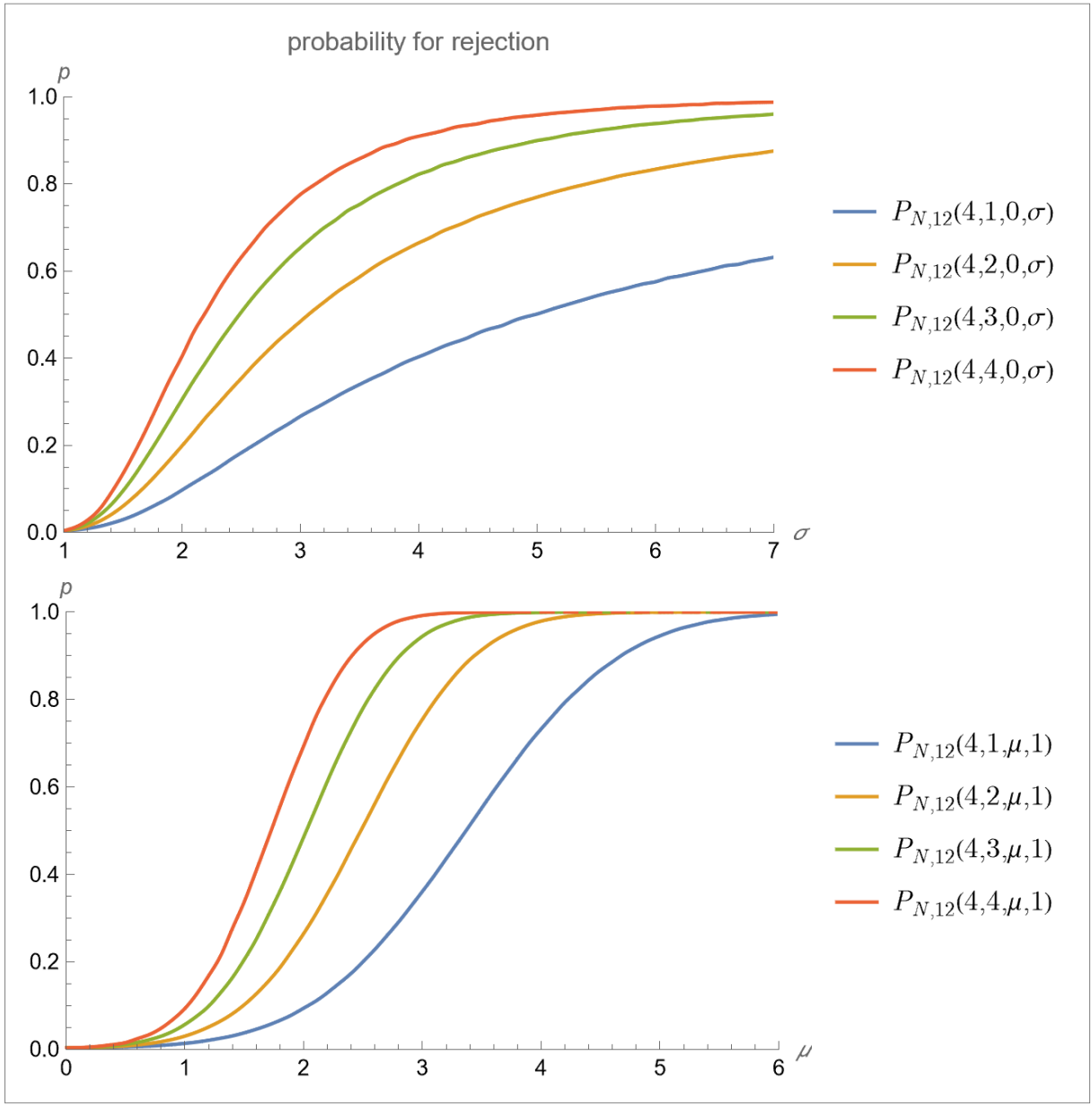

*Figure 101:* $P_{N,12}(4,k,\mu,\sigma)$

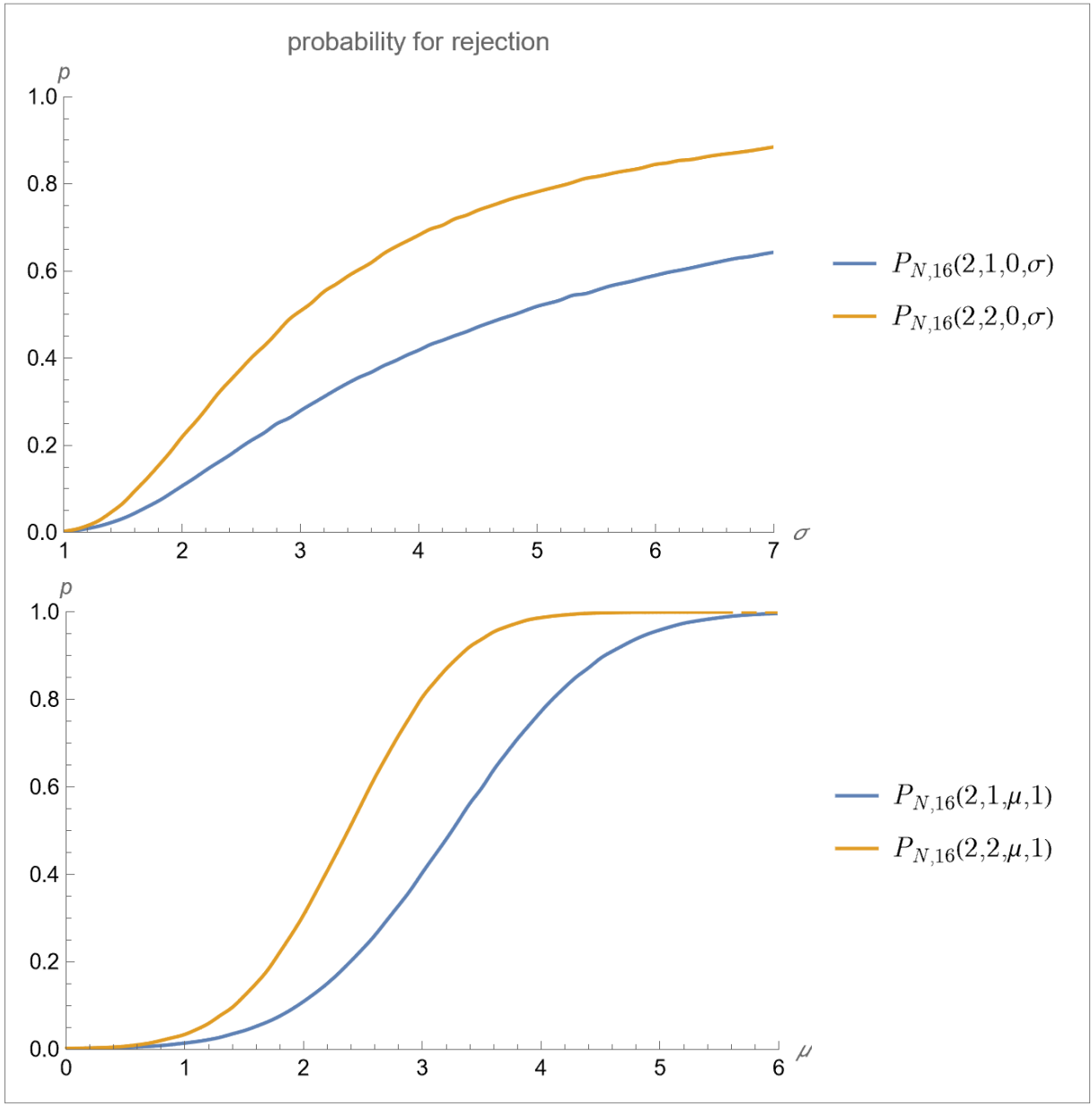

*Figure 102:* $P_{N,16}(2,k,\mu,\sigma)$

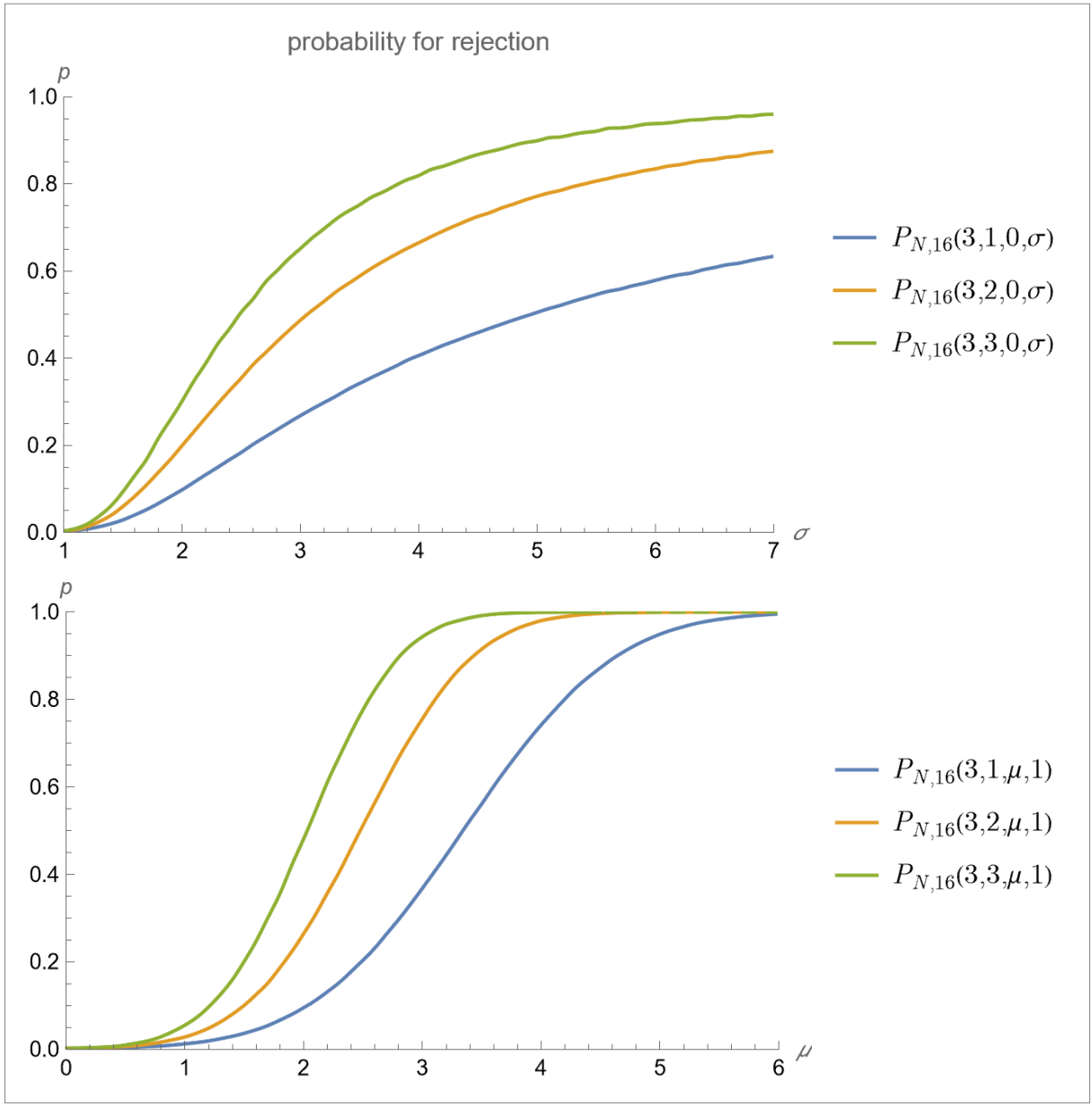

*Figure 103:* $P_{N,16}(3,k,\mu,\sigma)$

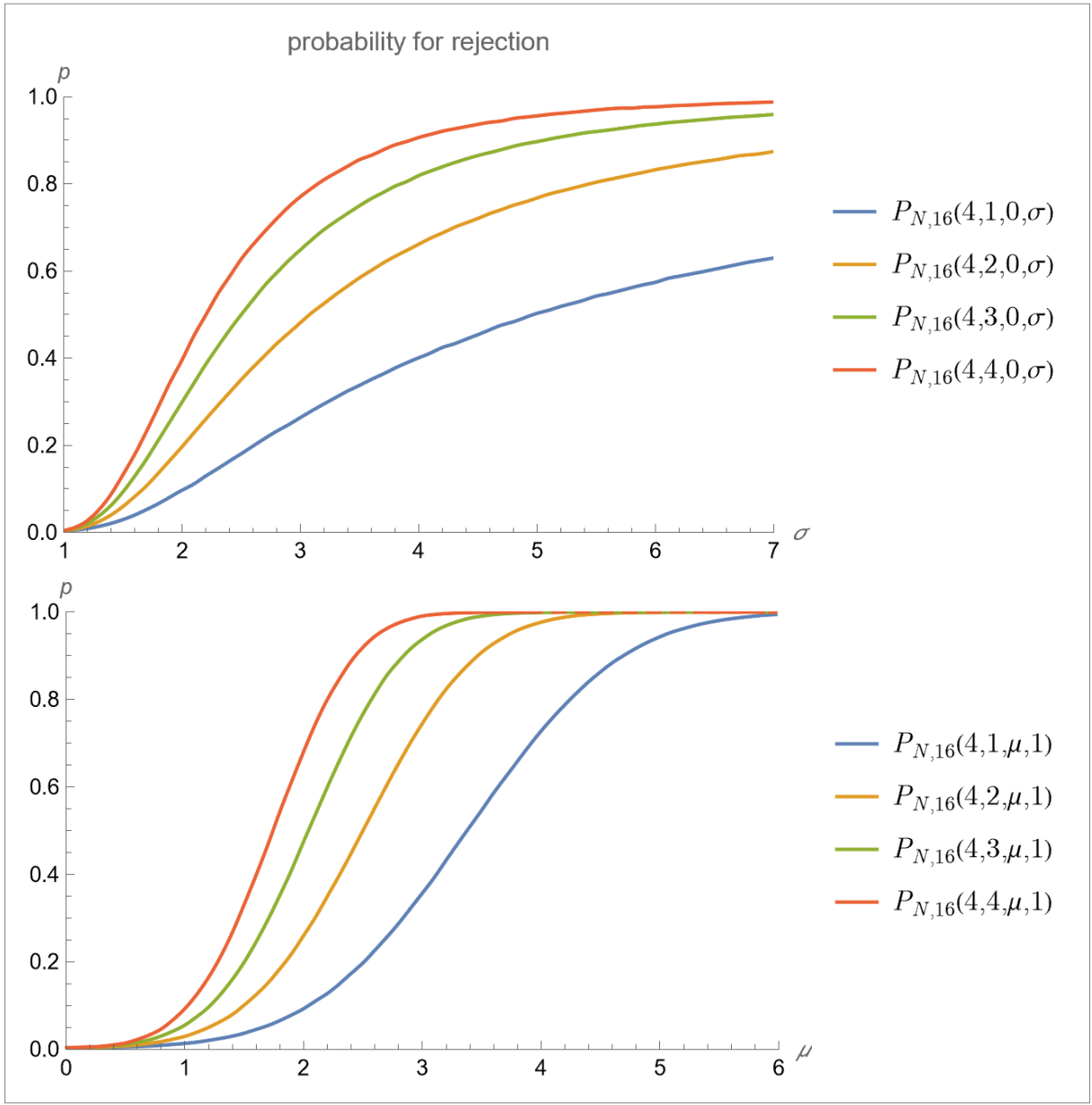

*Figure 104: $P_{N,16}(4,k,\mu,\sigma)$*

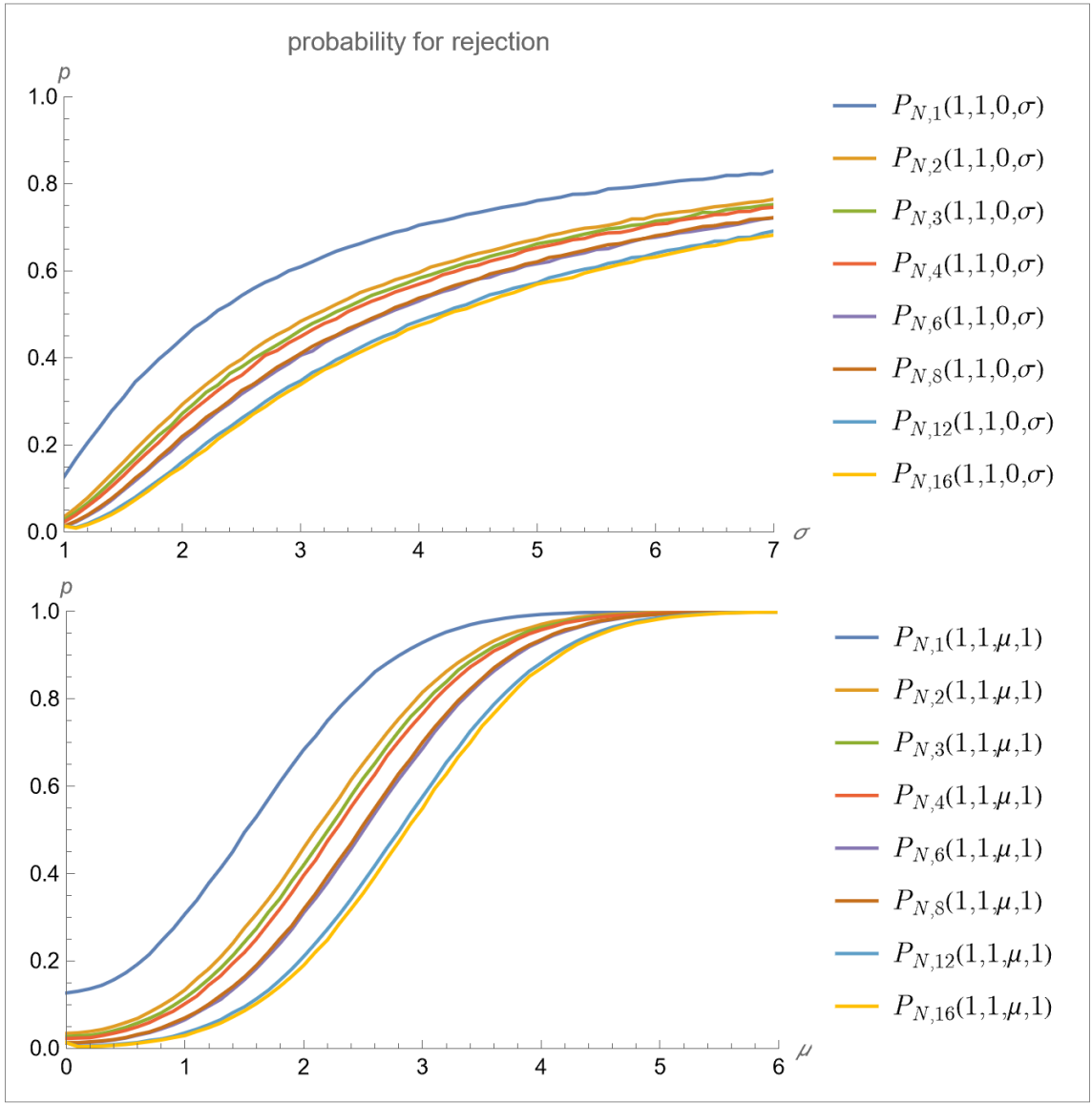

*Figure 105:* $P_{N,a}(1,1,\mu,\sigma)$

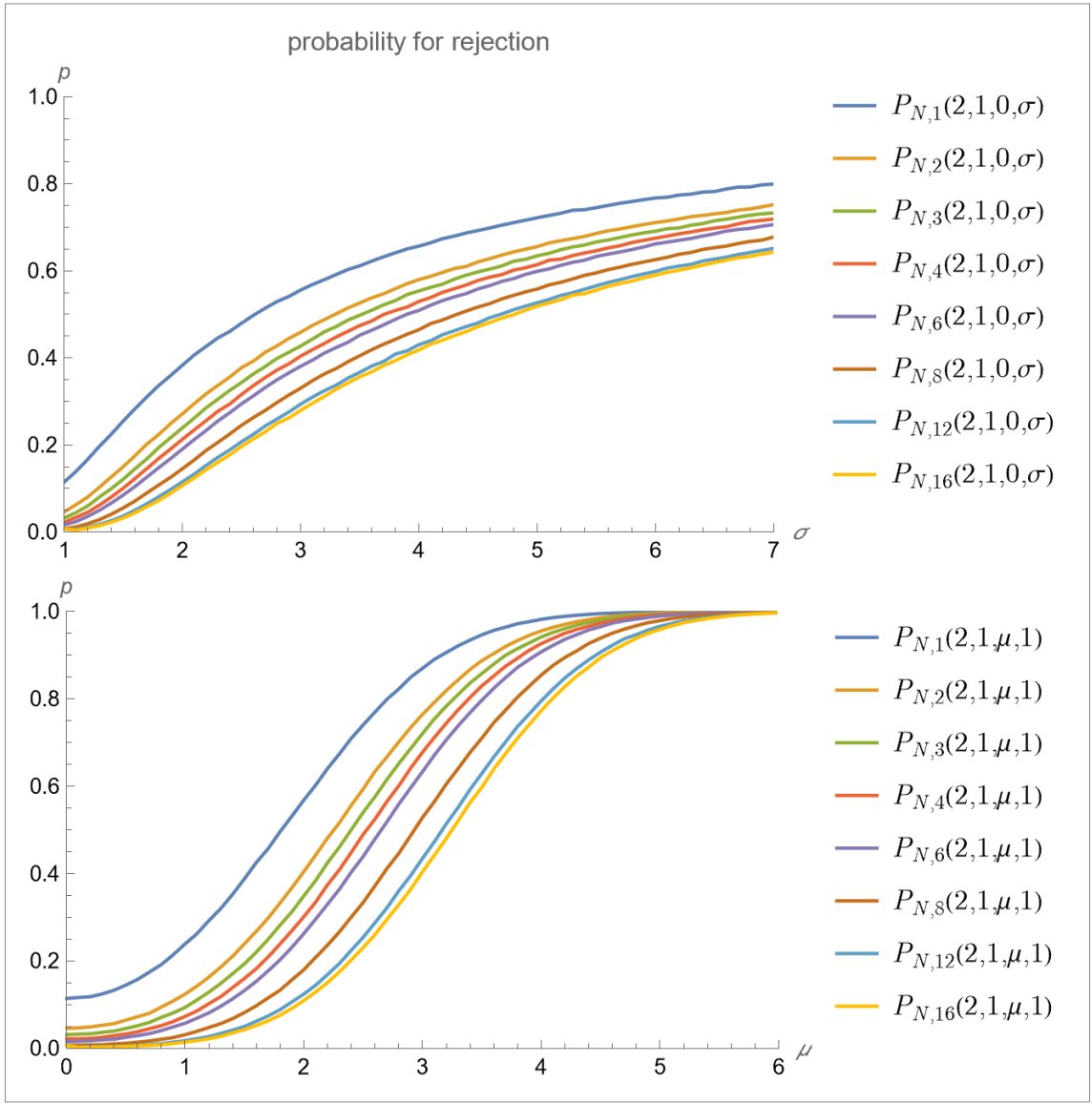

*Figure 106: $P_{N,a}(2,1,\mu,\sigma)$*

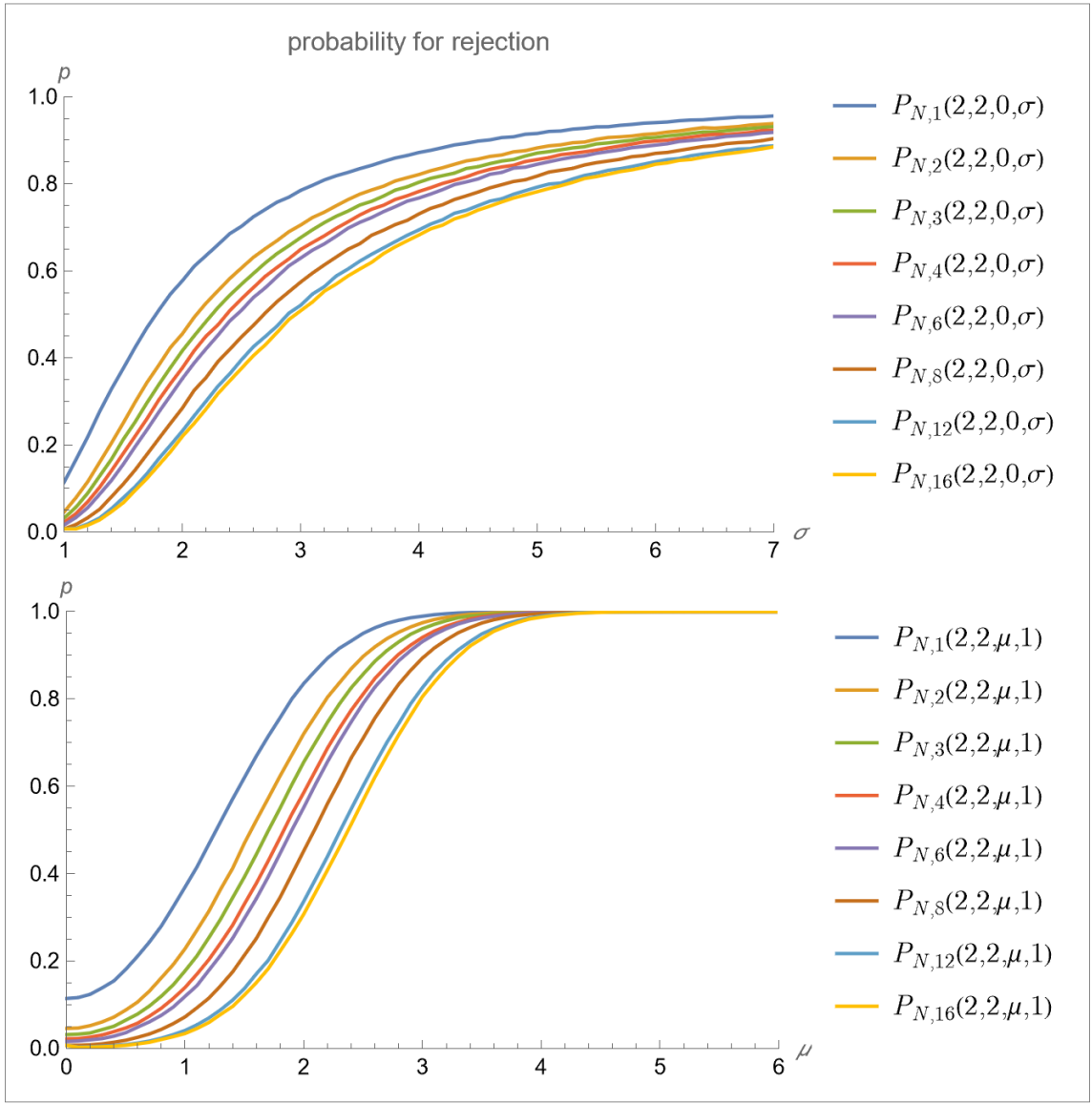

*Figure 107: $P_{N,a}(2,2,\mu,\sigma)$*

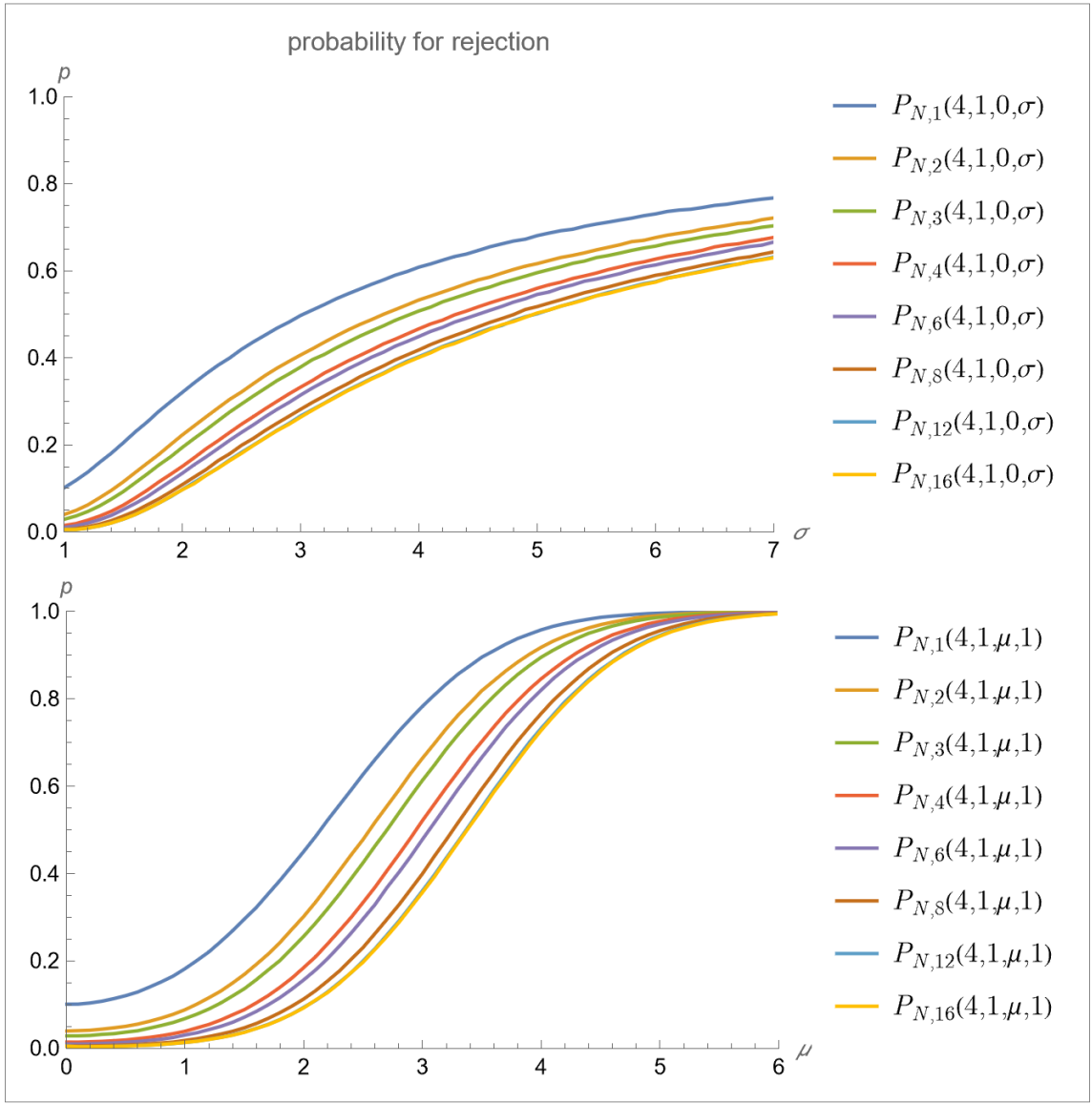

*Figure 108: $P_{N,a}(3,1,\mu,\sigma)$*

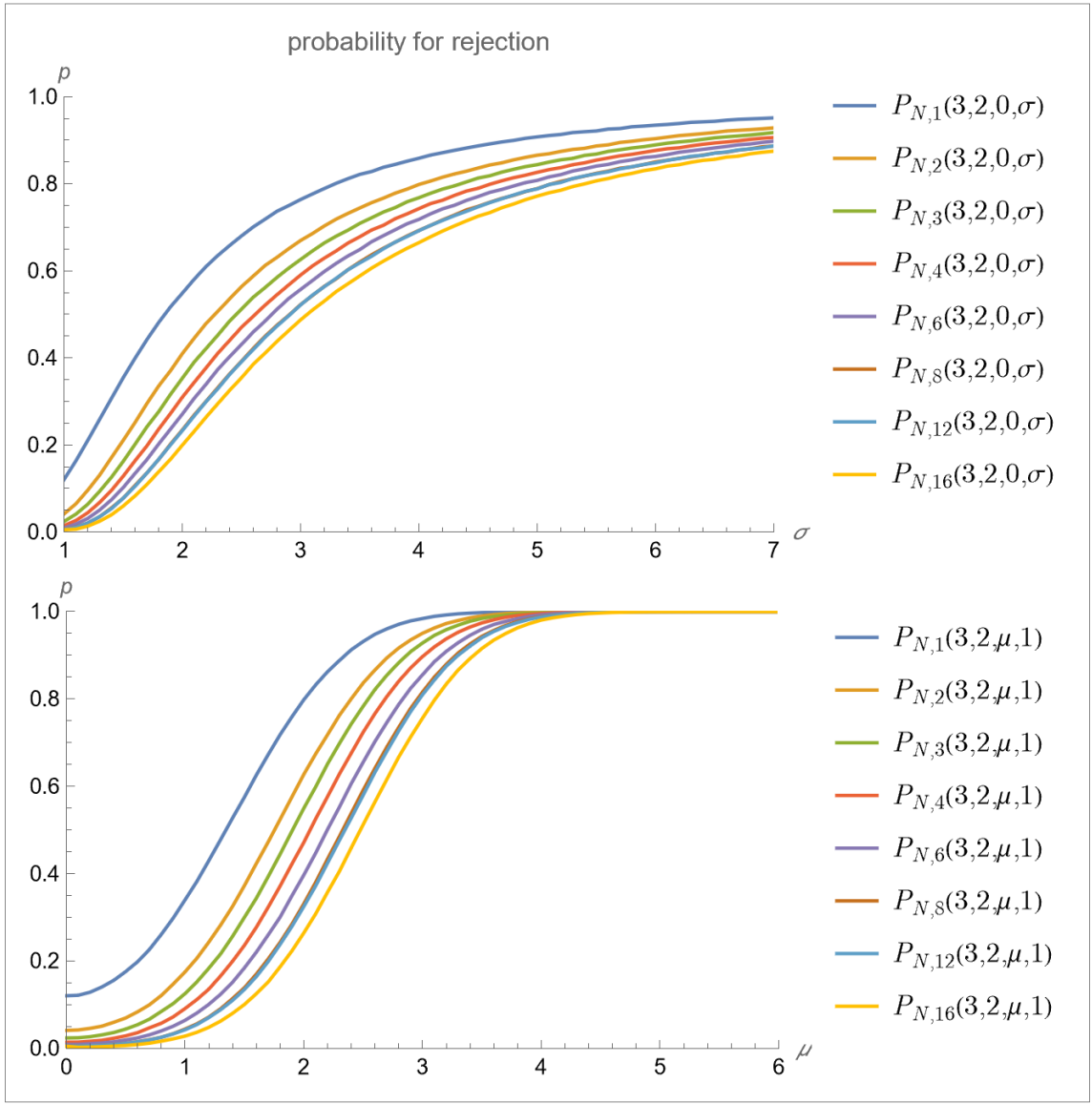

*Figure 109: $P_{N,a}(3,2,\mu,\sigma)$*

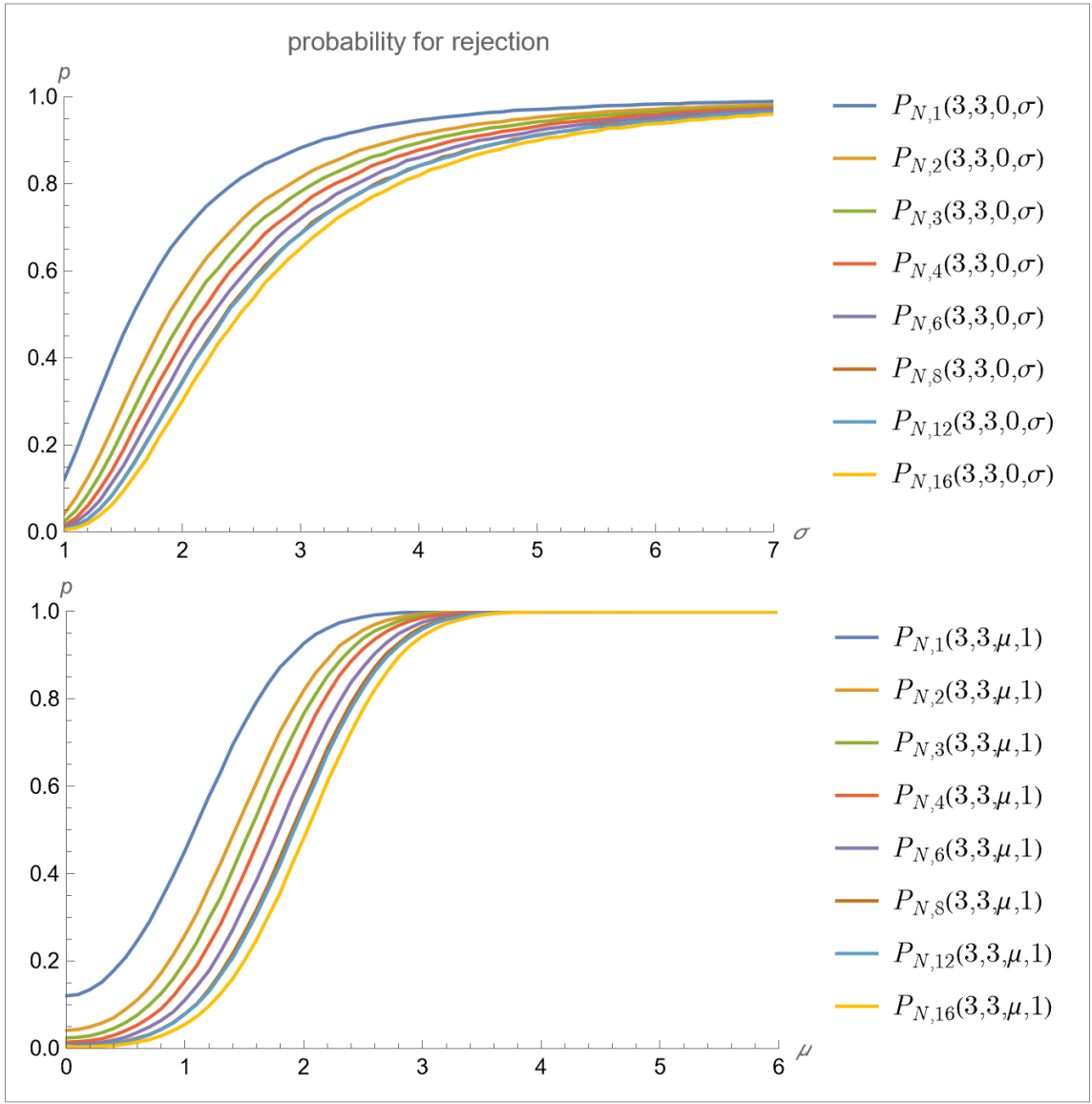

*Figure 110: $P_{N,a}(3,3,\mu,\sigma)$*

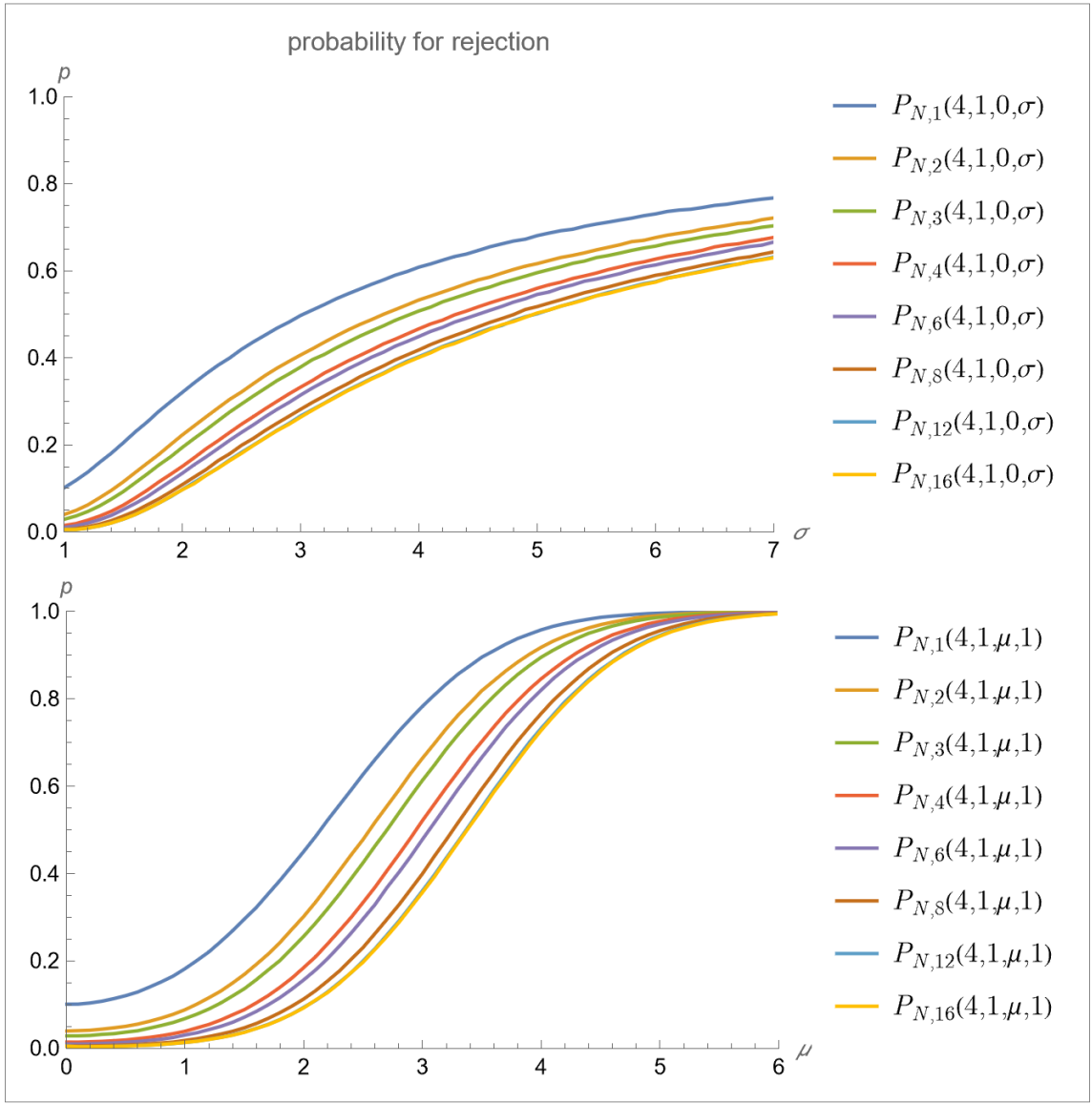

*Figure 111:* $P_{N,a}(4,1,\mu,\sigma)$

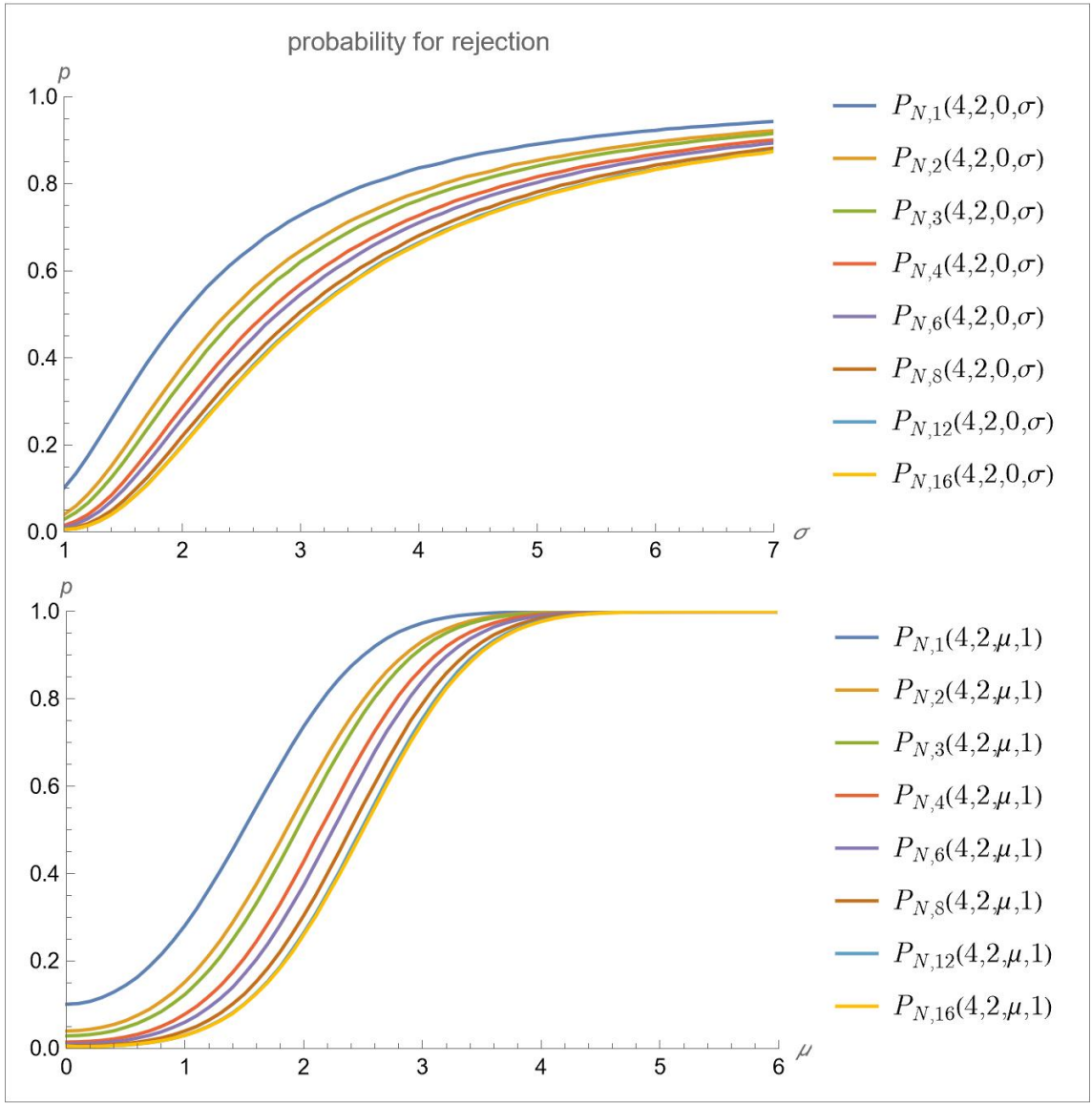

*Figure 112: $P_{N,a}(4,2,\mu,\sigma)$*

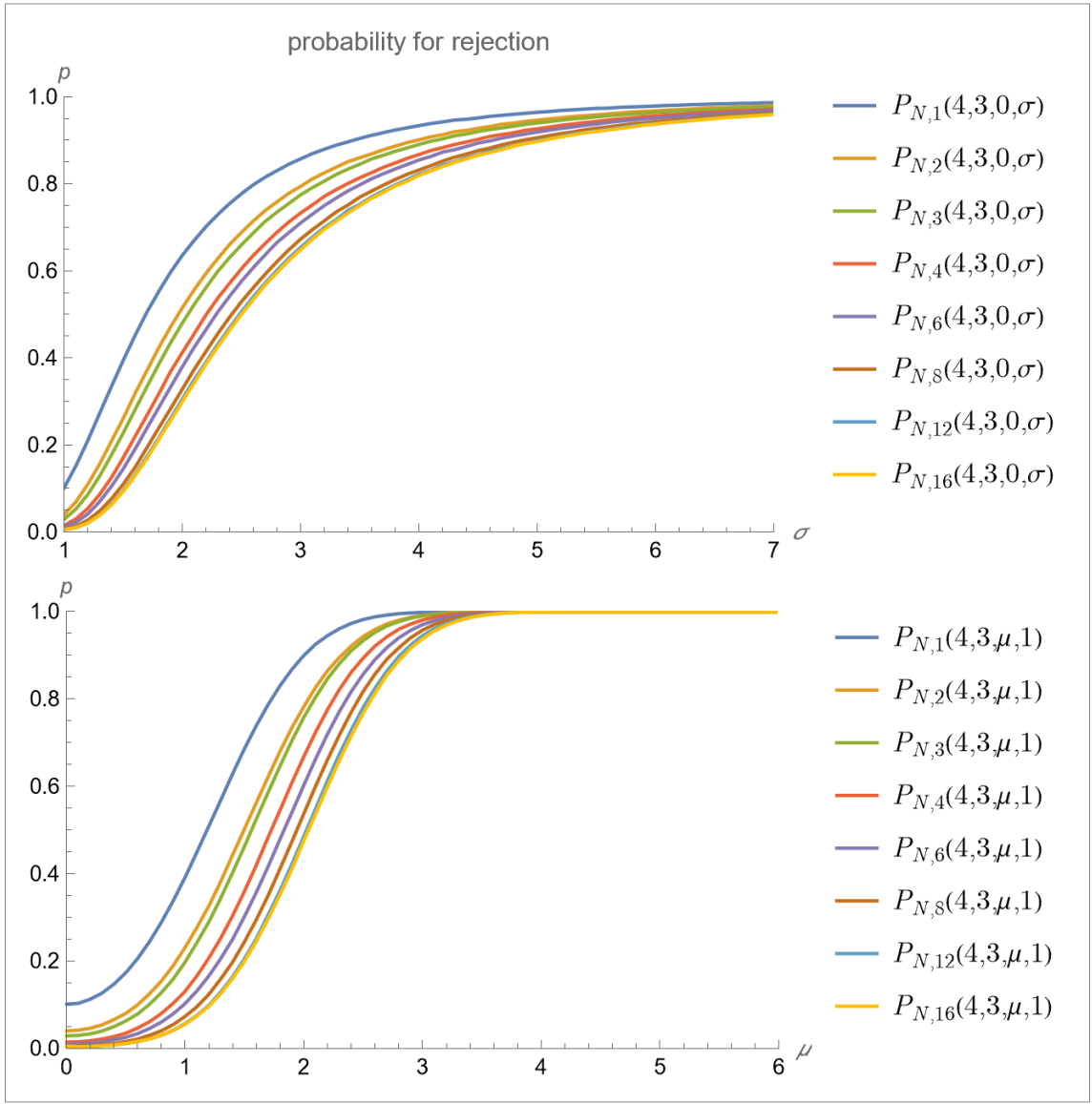

*Figure 113: $P_{N,a}(4,3,\mu,\sigma)$*

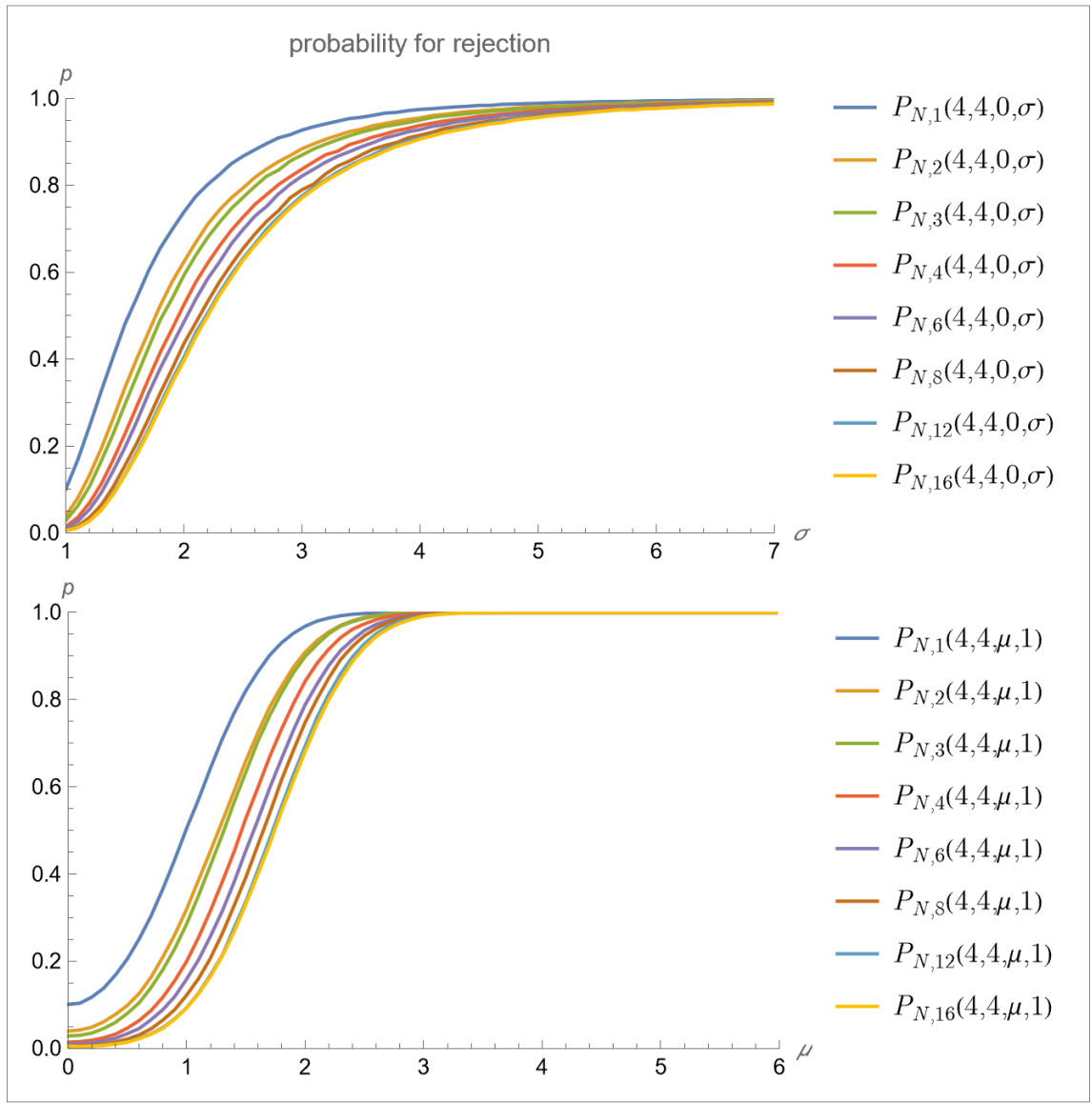

*Figure 114: $P_{N,a}(4,4,\mu,\sigma)$*

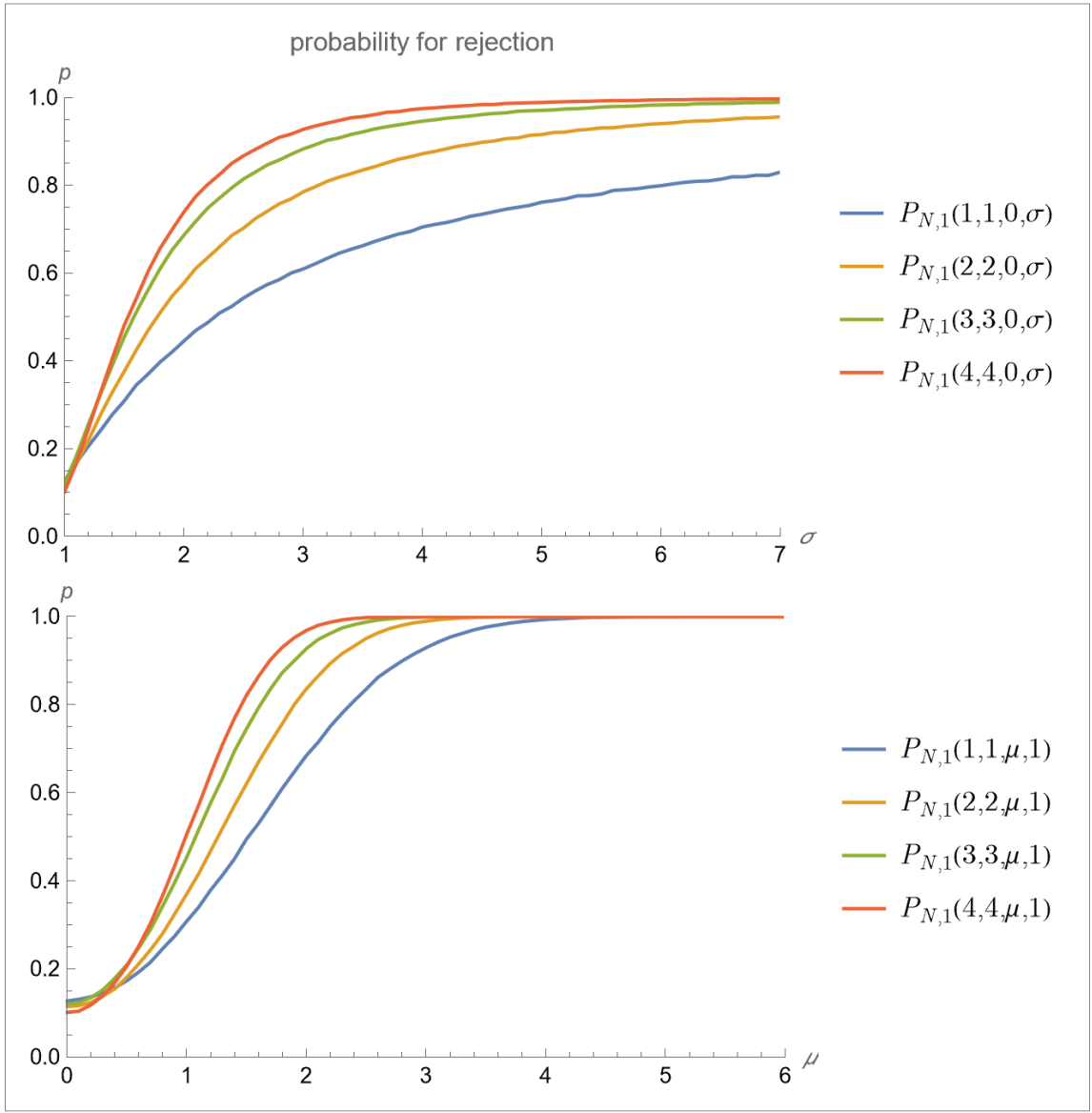

*Figure 115: $P_{N,1}(n,n,\mu,\sigma)$*

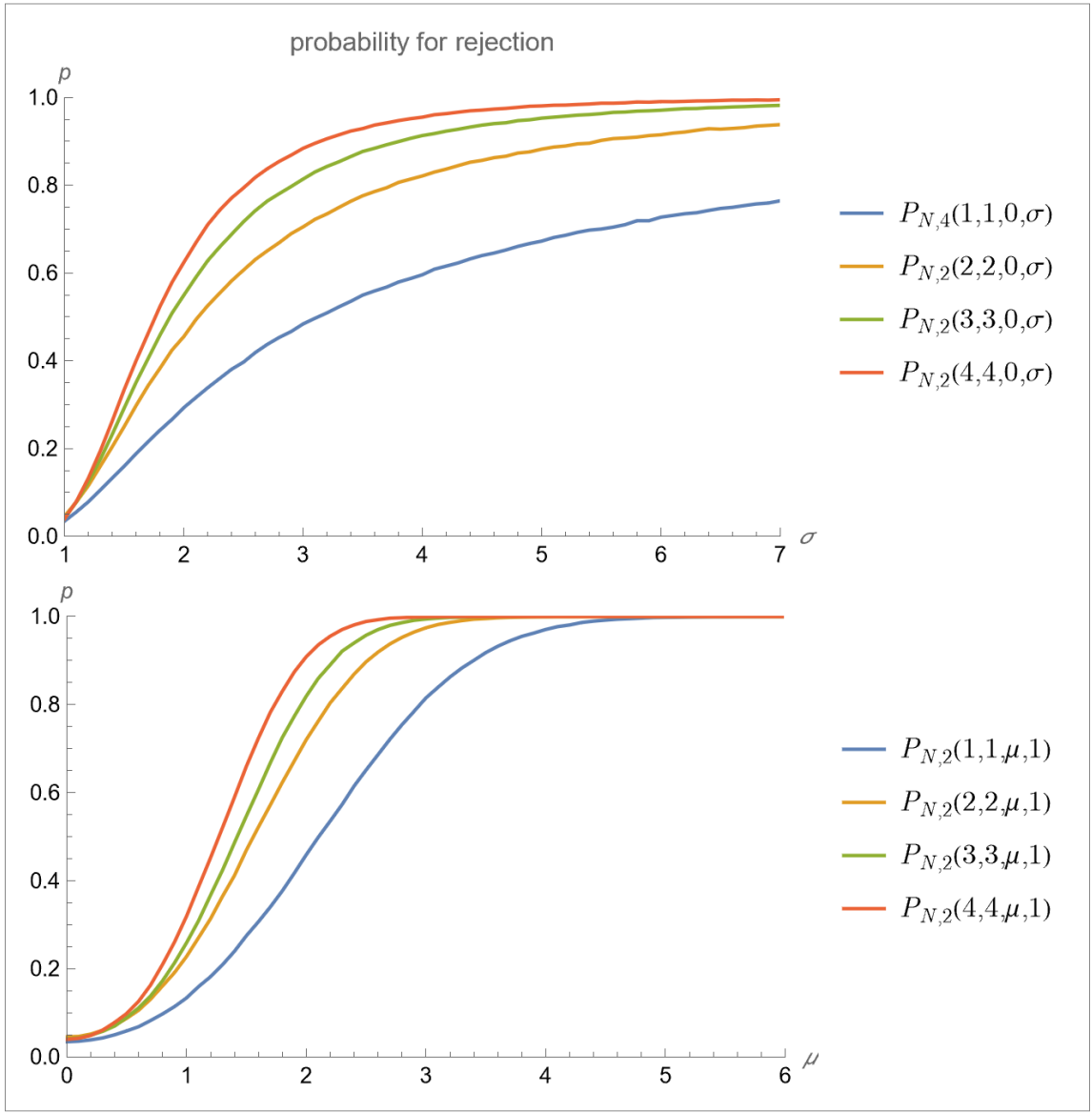

*Figure 116: $P_{N,2}(n,n,\mu,\sigma)$*

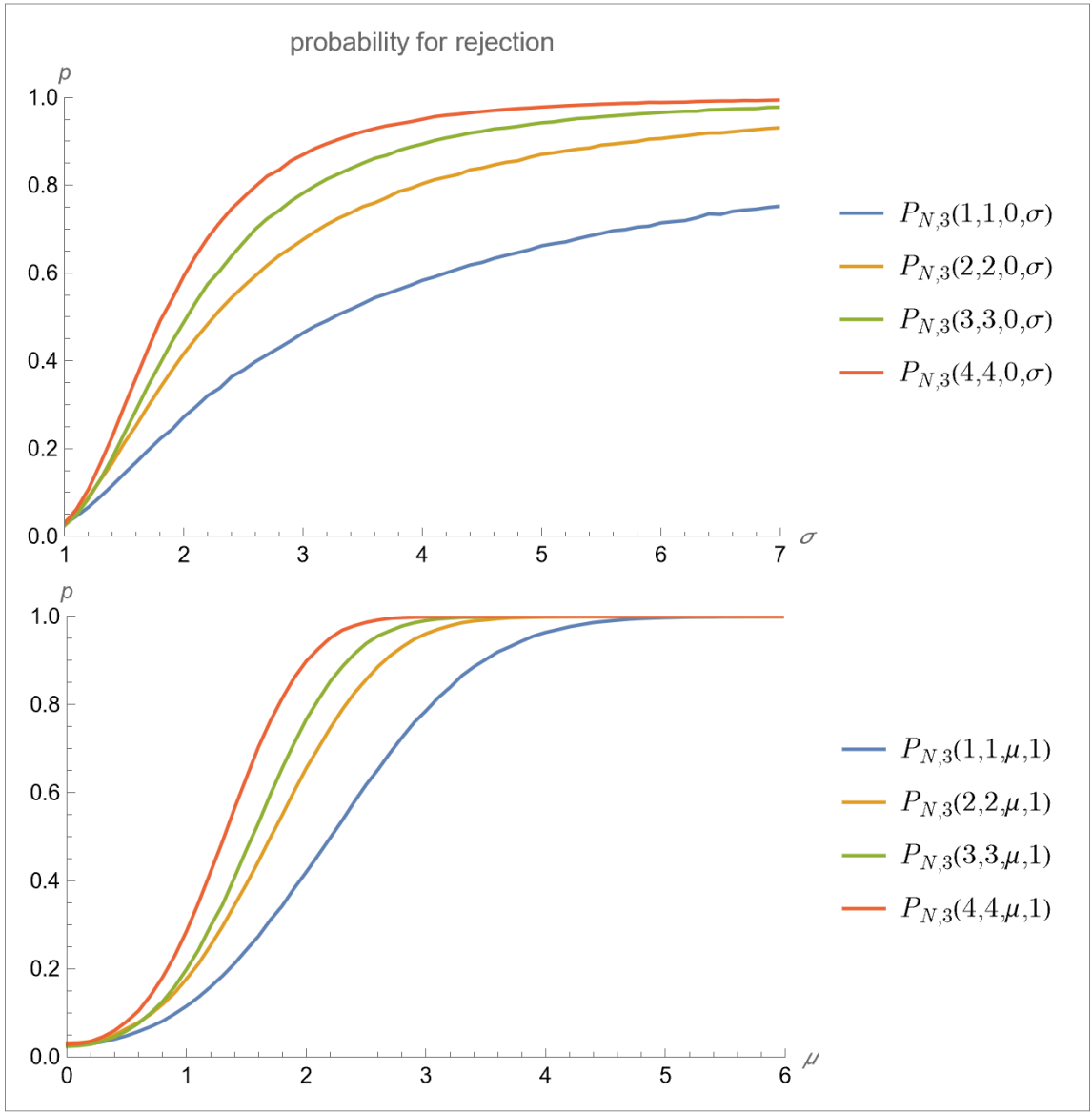

*Figure 117: $P_{N,3}(n,n,\mu,\sigma)$*

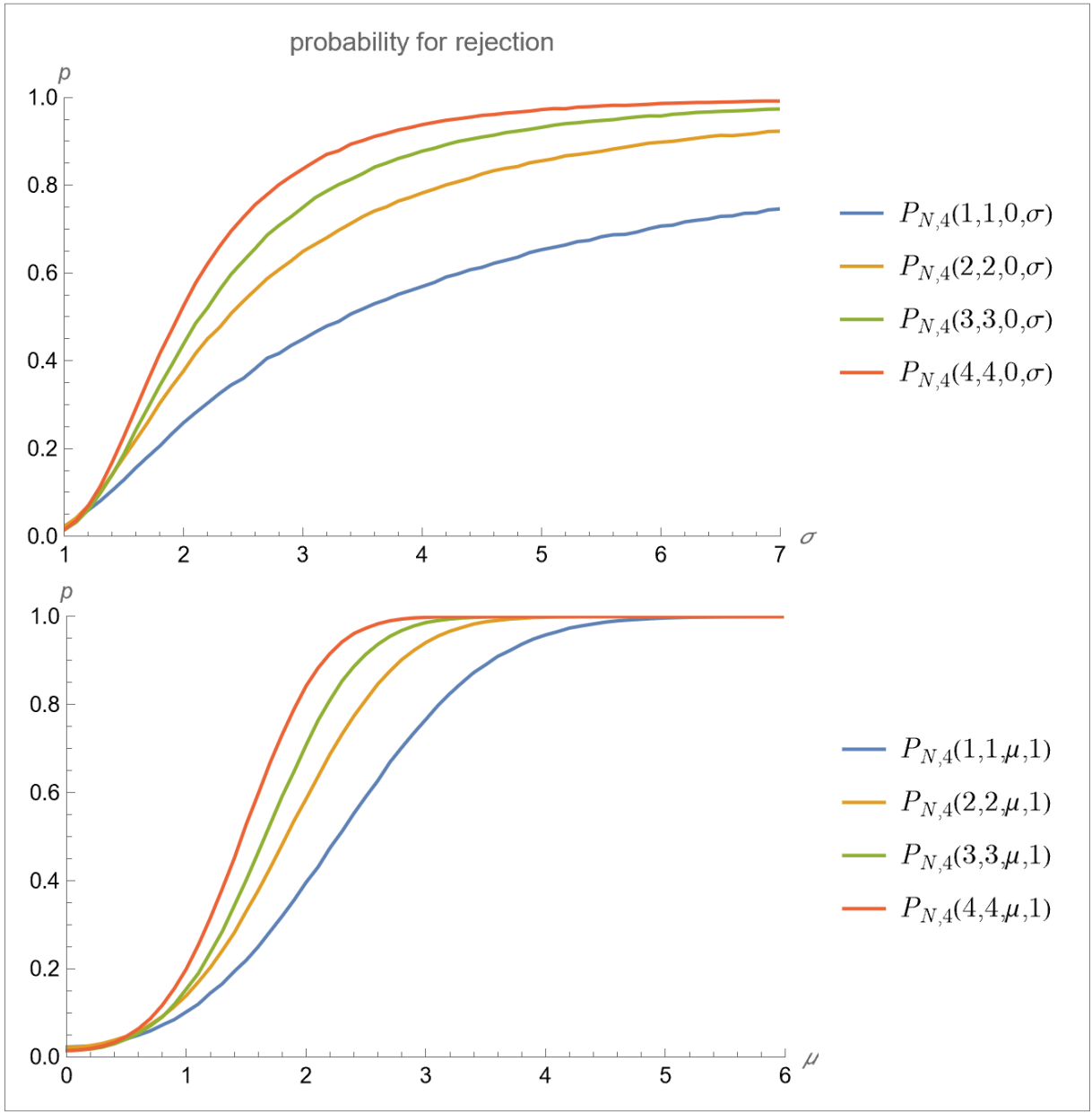

*Figure 118: $P_{N,4}(n,n,\mu,\sigma)$*

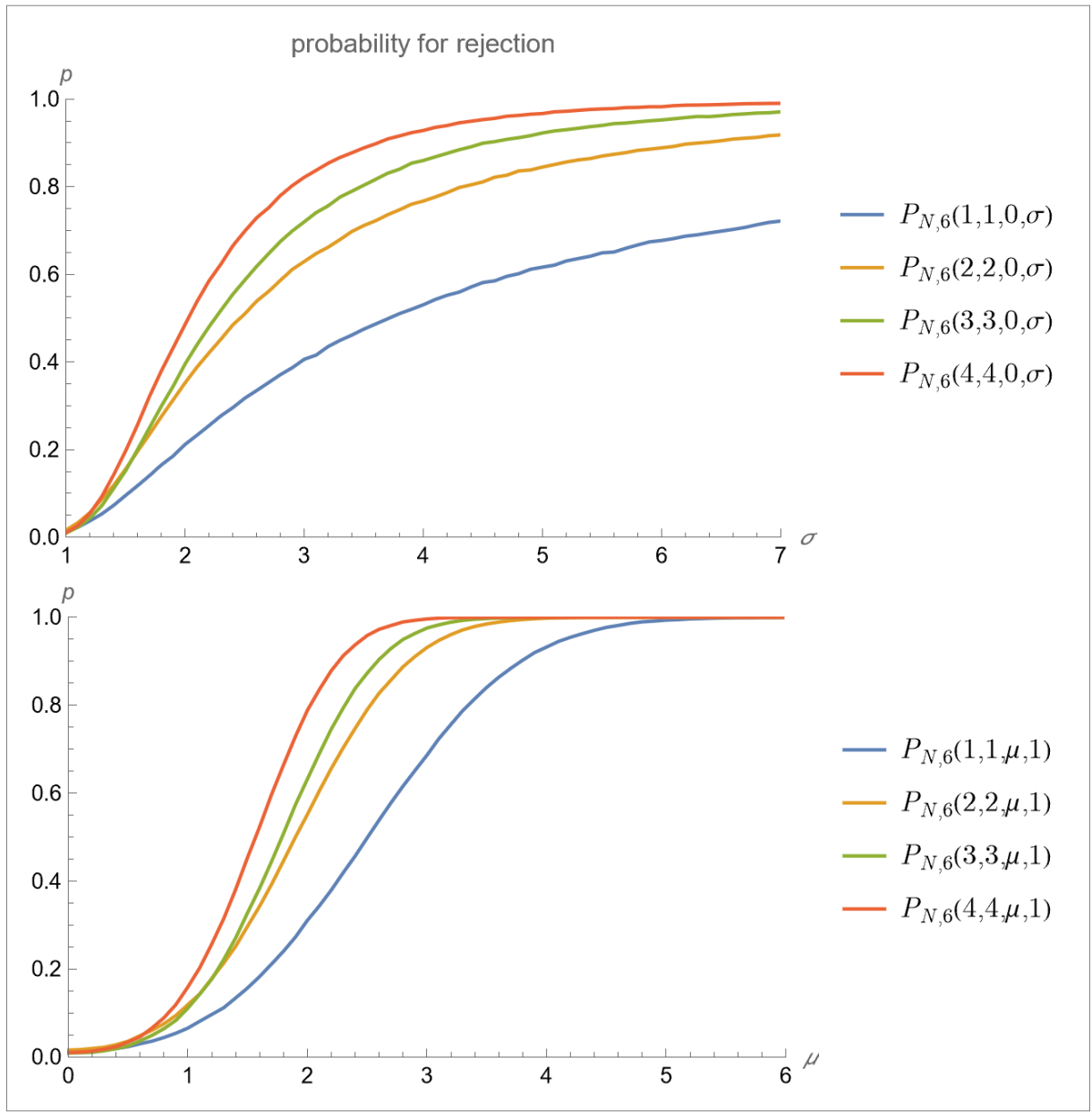

*Figure 119:* $P_{N,6}(n,n,\mu,\sigma)$

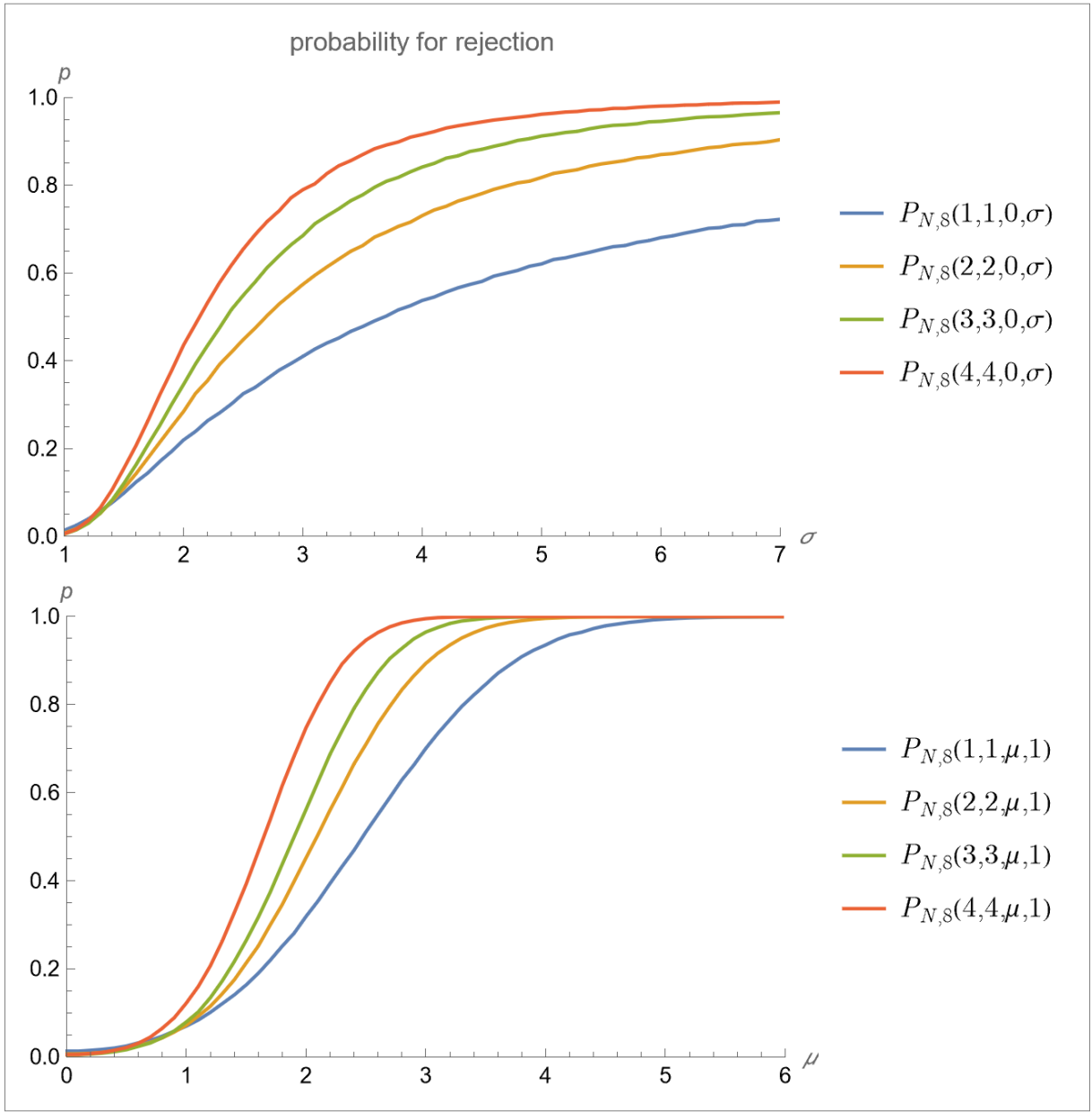

*Figure 120: $P_{N,8}(n,n,\mu,\sigma)$*

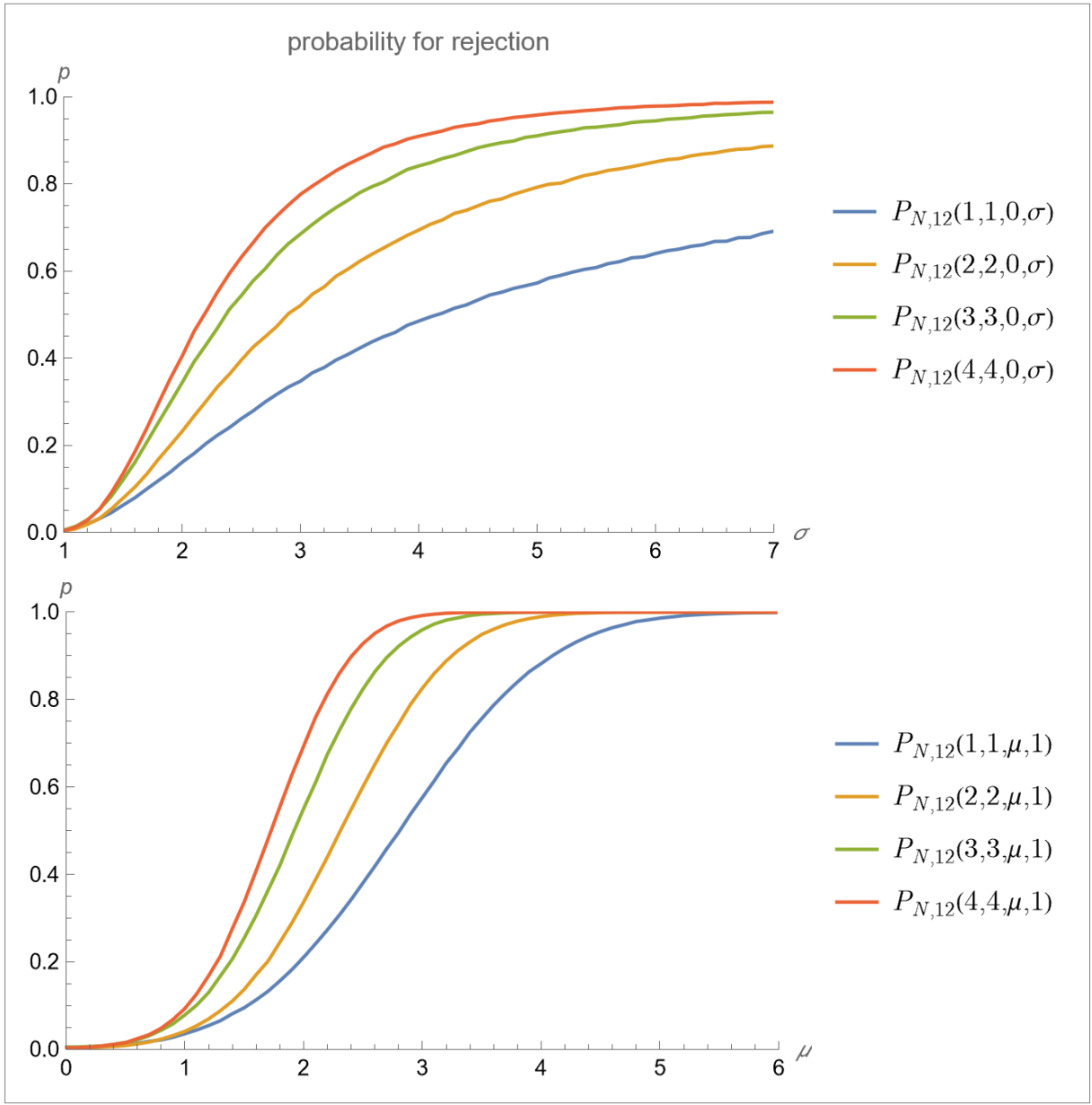

*Figure 121:* $P_{N,12}(n,n,\mu,\sigma)$

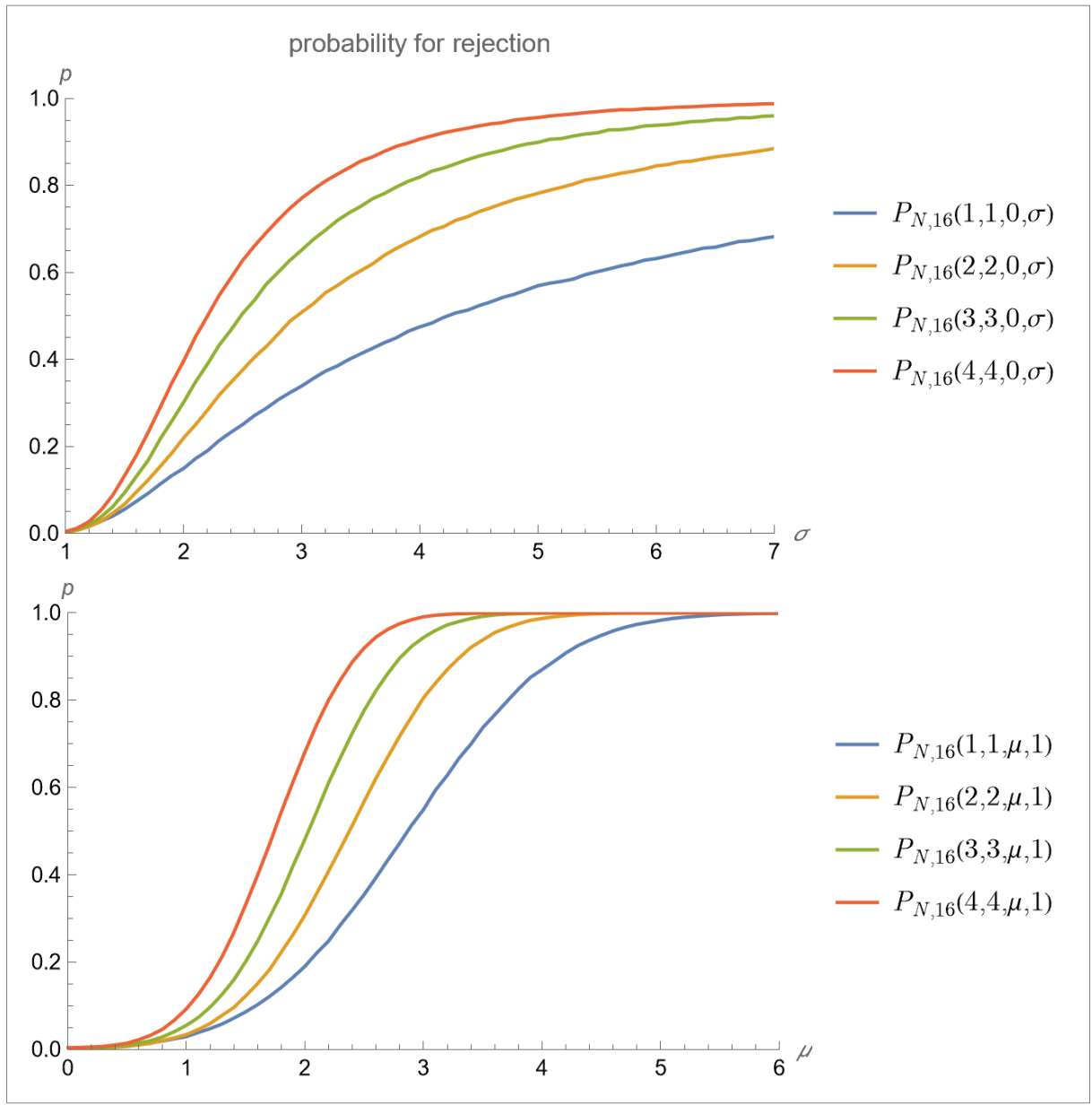

*Figure 122:* $P_{N,16}(n,n,\mu,\sigma)$

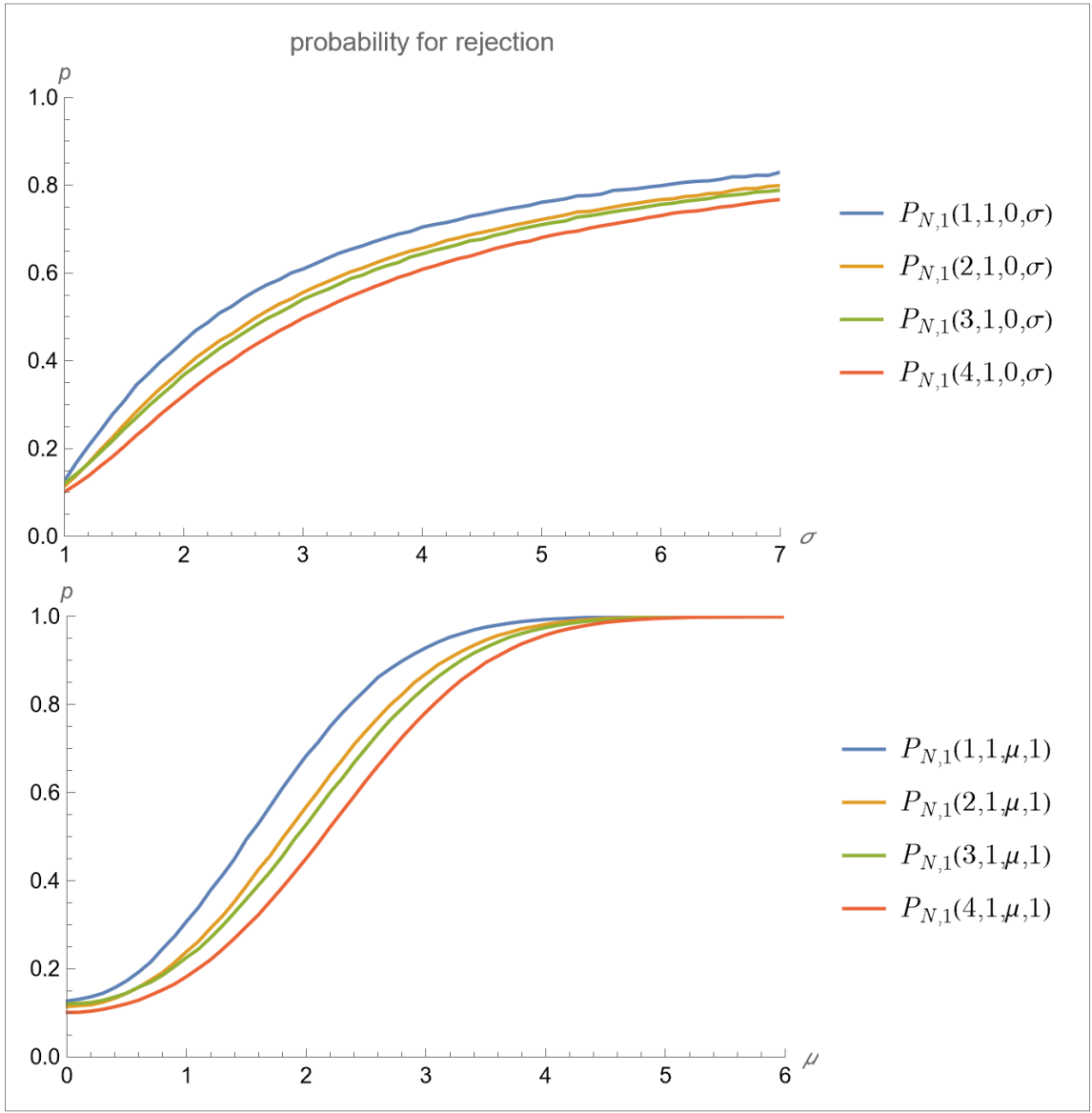

*Figure 123: $P_{N,1}(n, 1, \mu, \sigma)$*

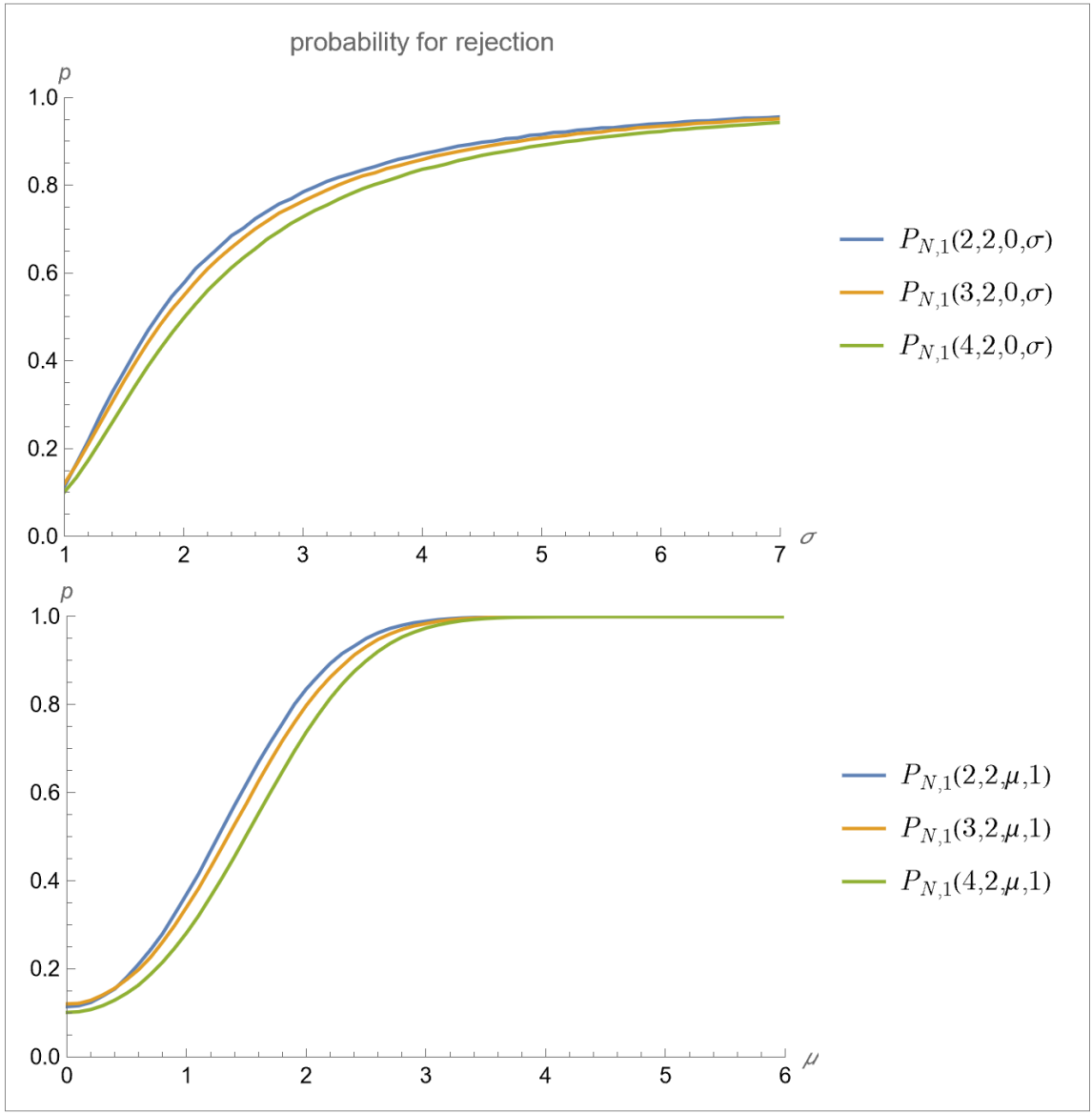

*Figure 124: $P_{N,1}(n, 2, \mu, \sigma)$*

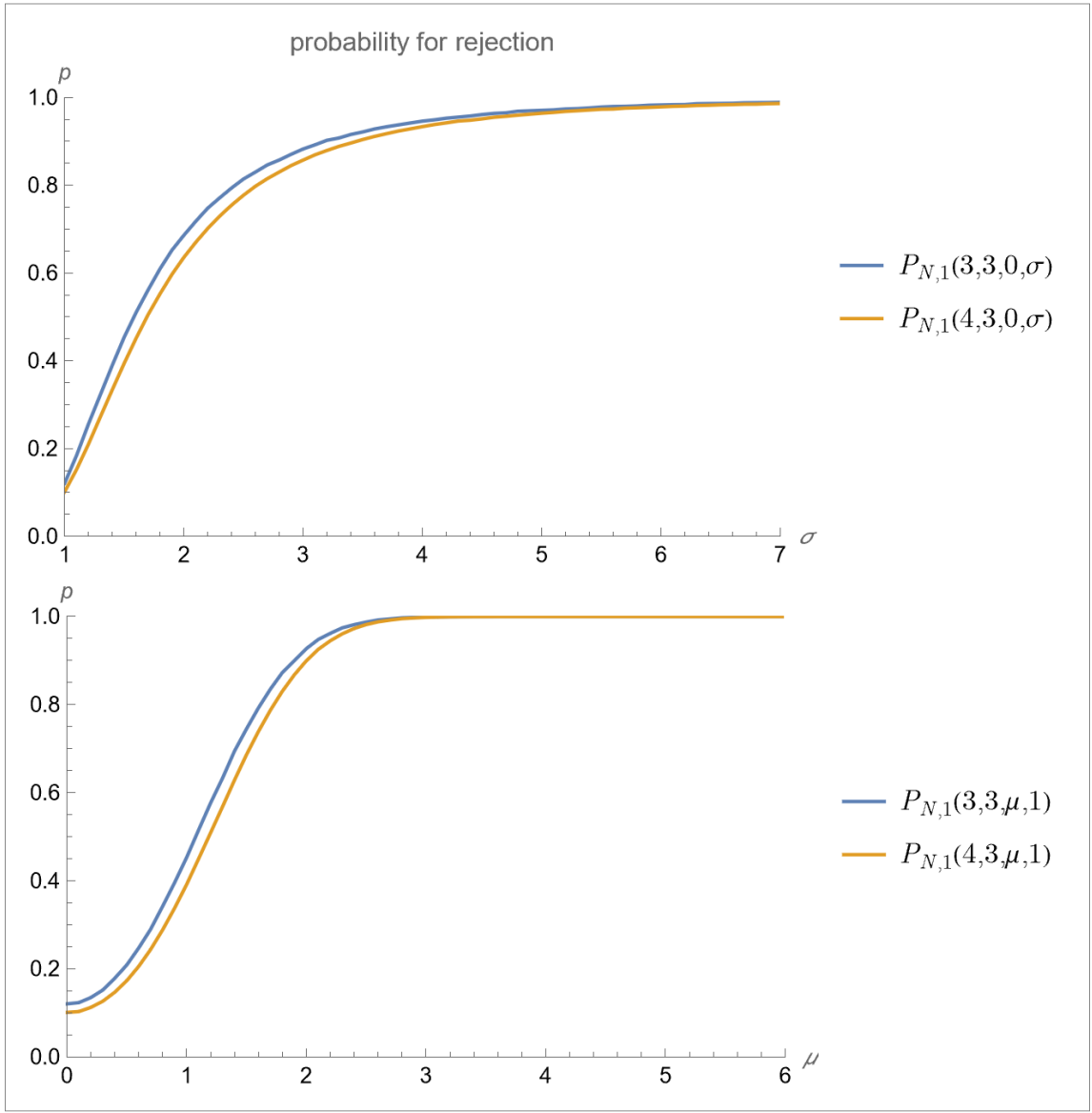

*Figure 125: $P_{N,1}(n, 3, \mu, \sigma)$*

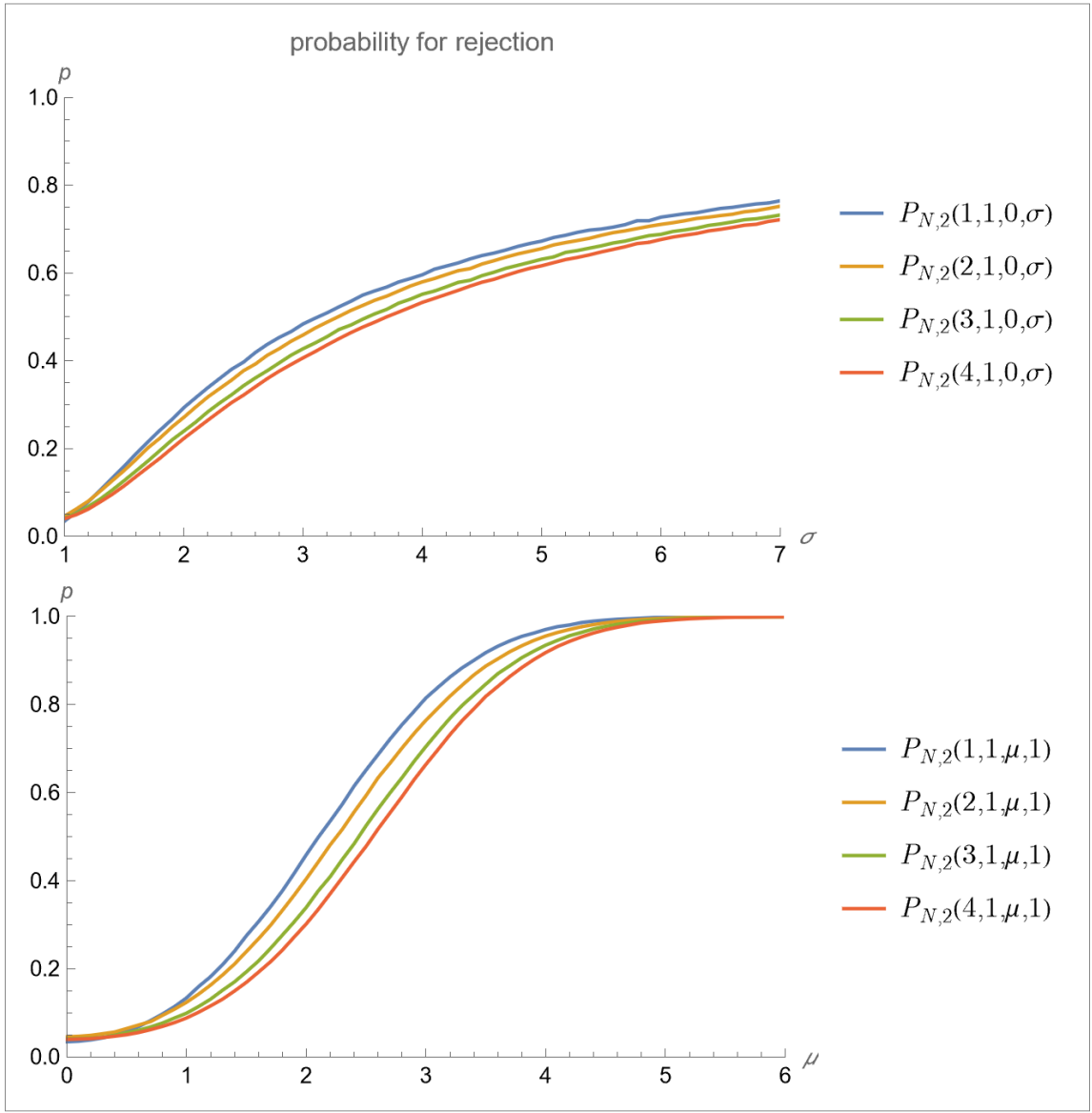

*Figure 126: $P_{N,2}(n, 1, \mu, \sigma)$*

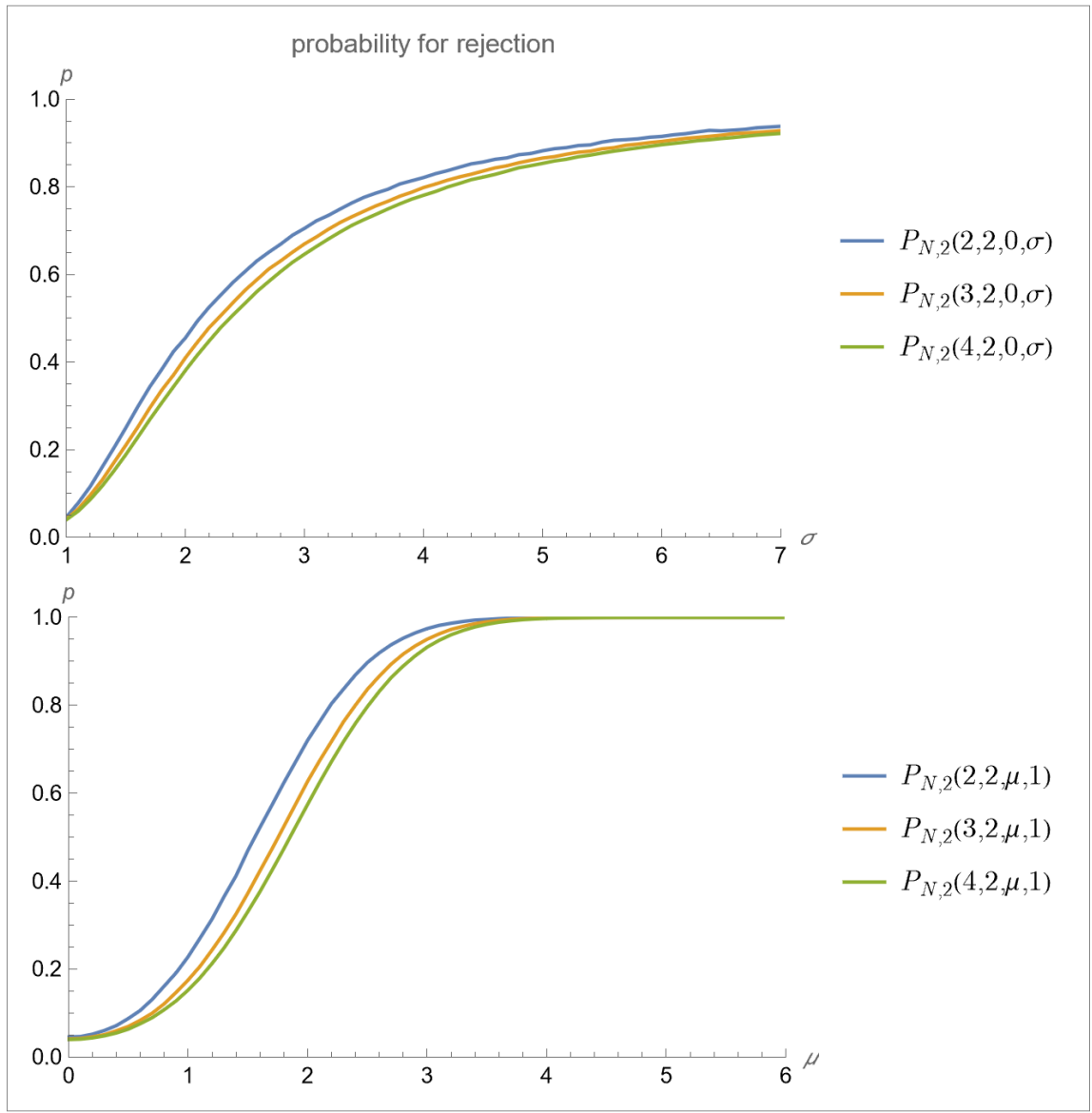

*Figure 127:* $P_{N,2}(n, 2, \mu, \sigma)$

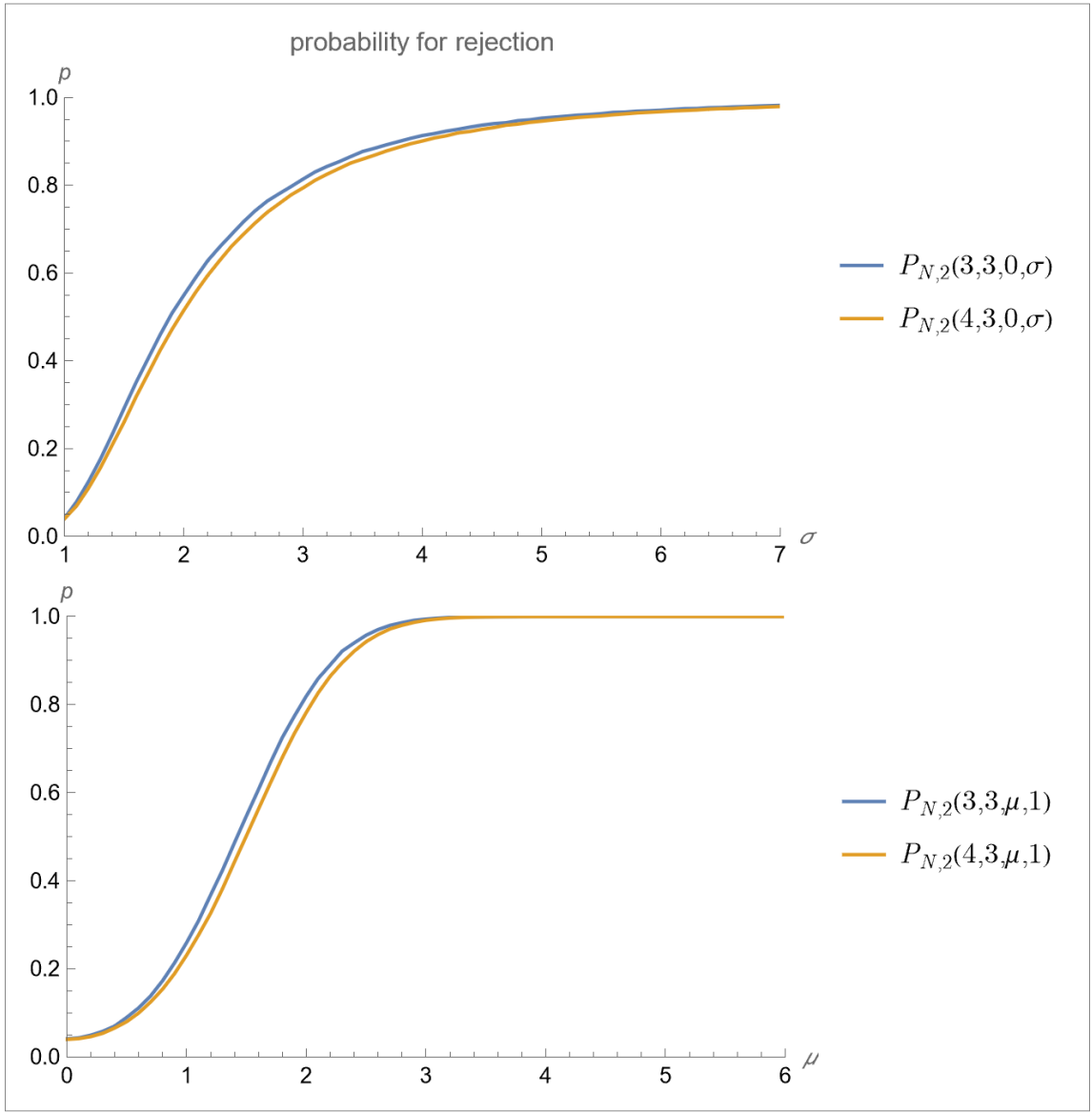

*Figure 128: $P_{N,2}(n, 3, \mu, \sigma)$*

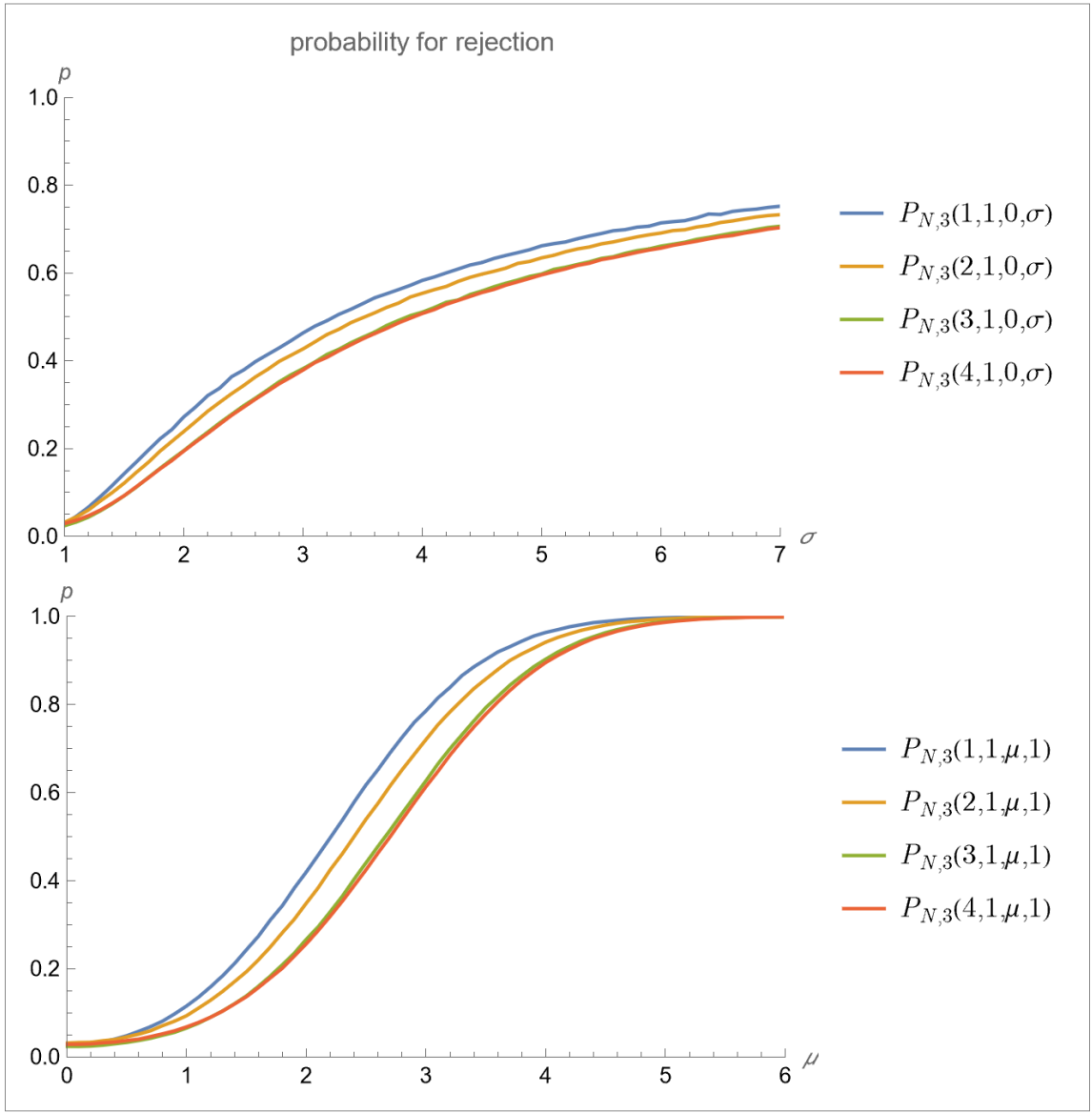

*Figure 129: $P_{N,3}(n, 1, \mu, \sigma)$*

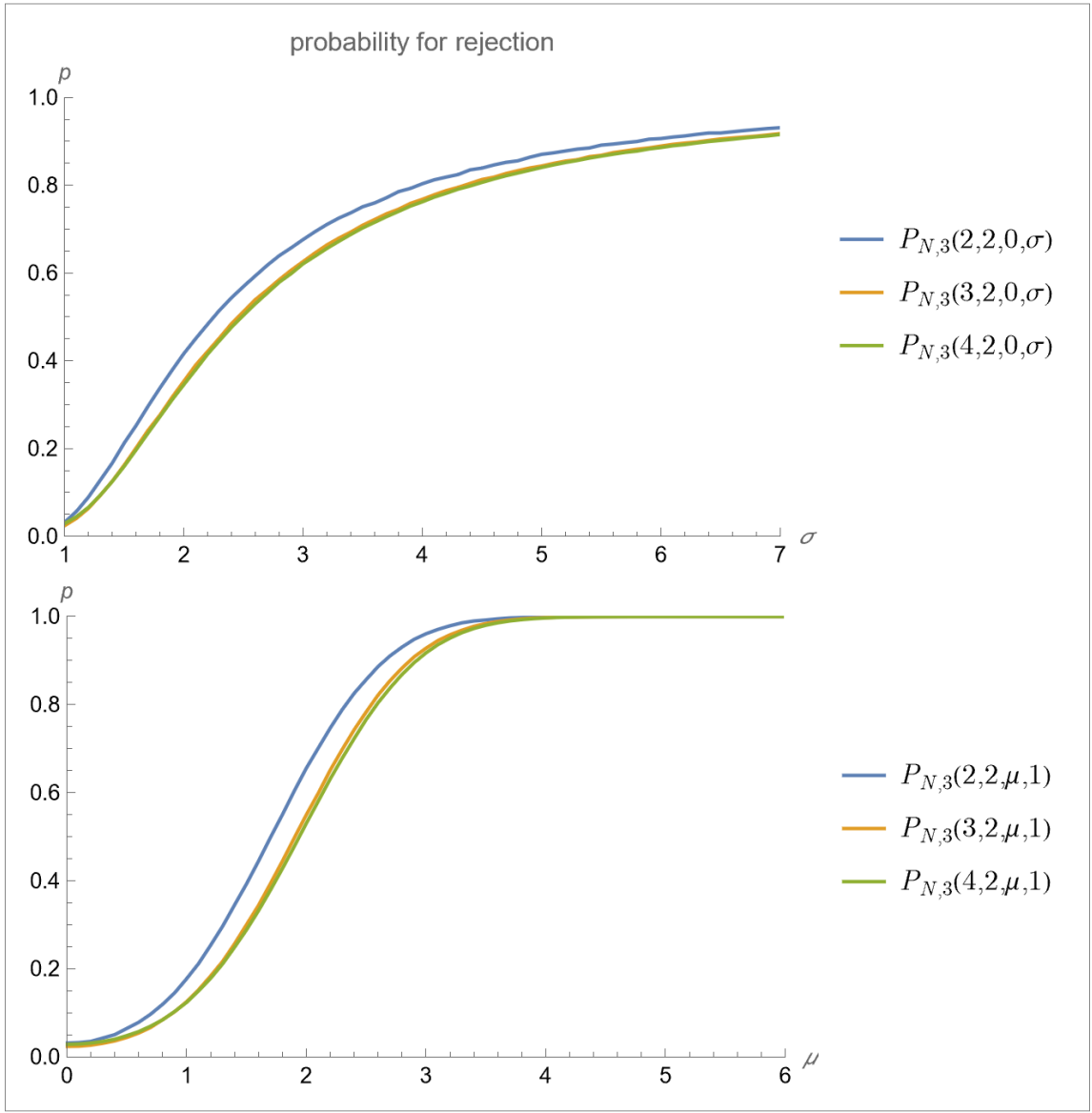

*Figure 130: $P_{N,3}(n, 2, \mu, \sigma)$*

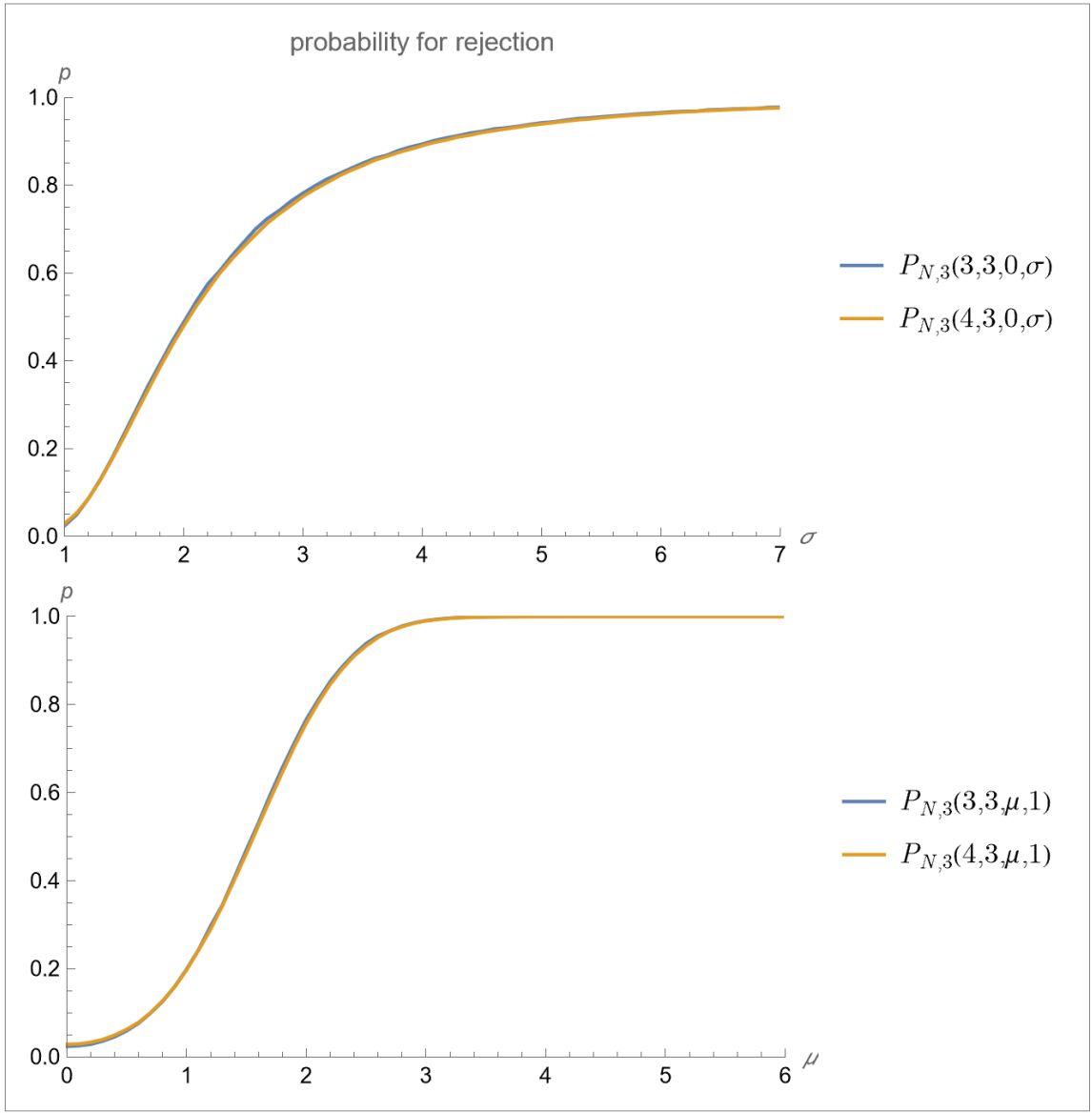

*Figure 131: $P_{N,3}(n, 3, \mu, \sigma)$*

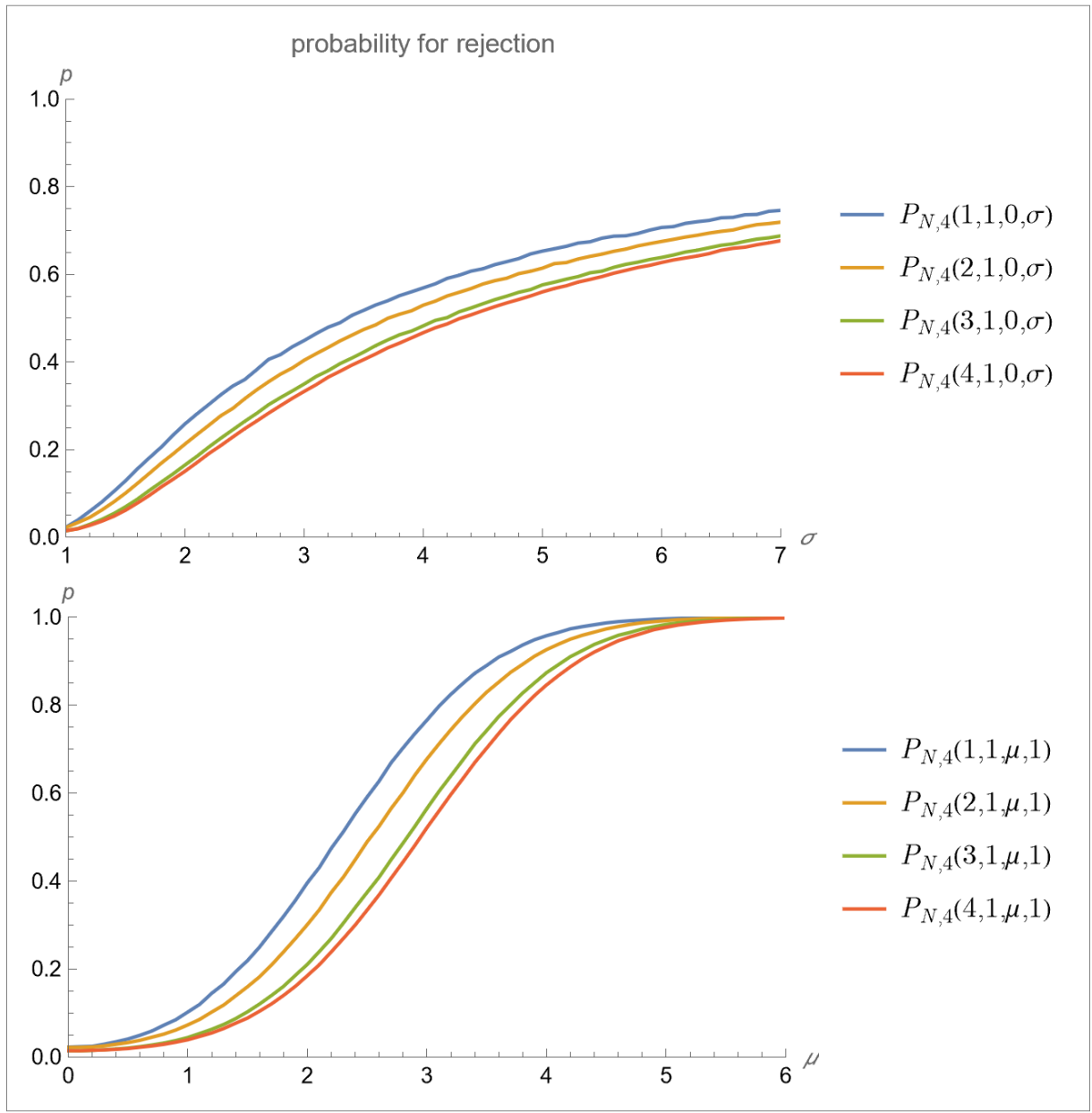

*Figure 132: $P_{N,4}(n,1,\mu,\sigma)$*

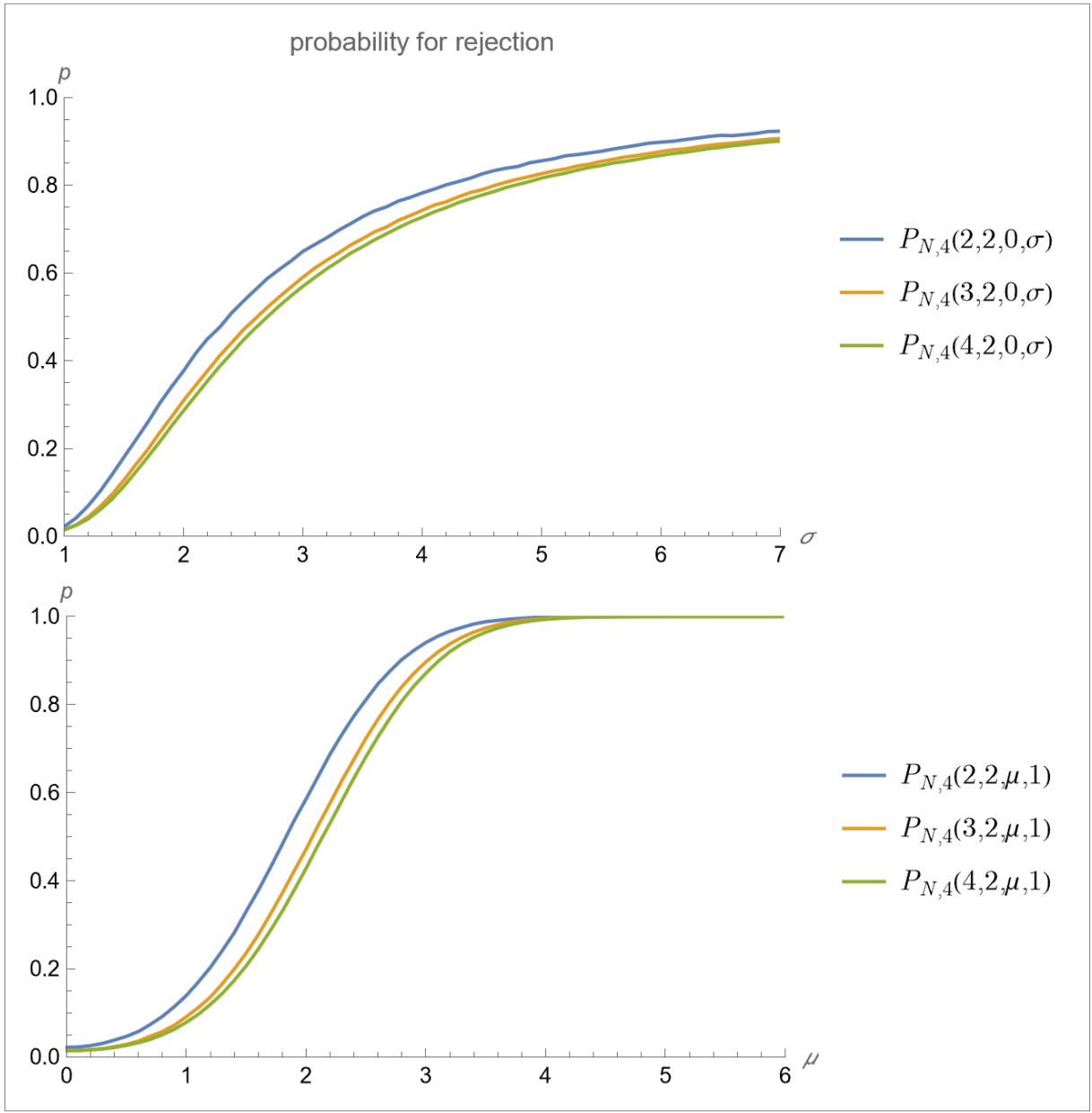

*Figure 133:* $P_{N,4}(n, 2, \mu, \sigma)$

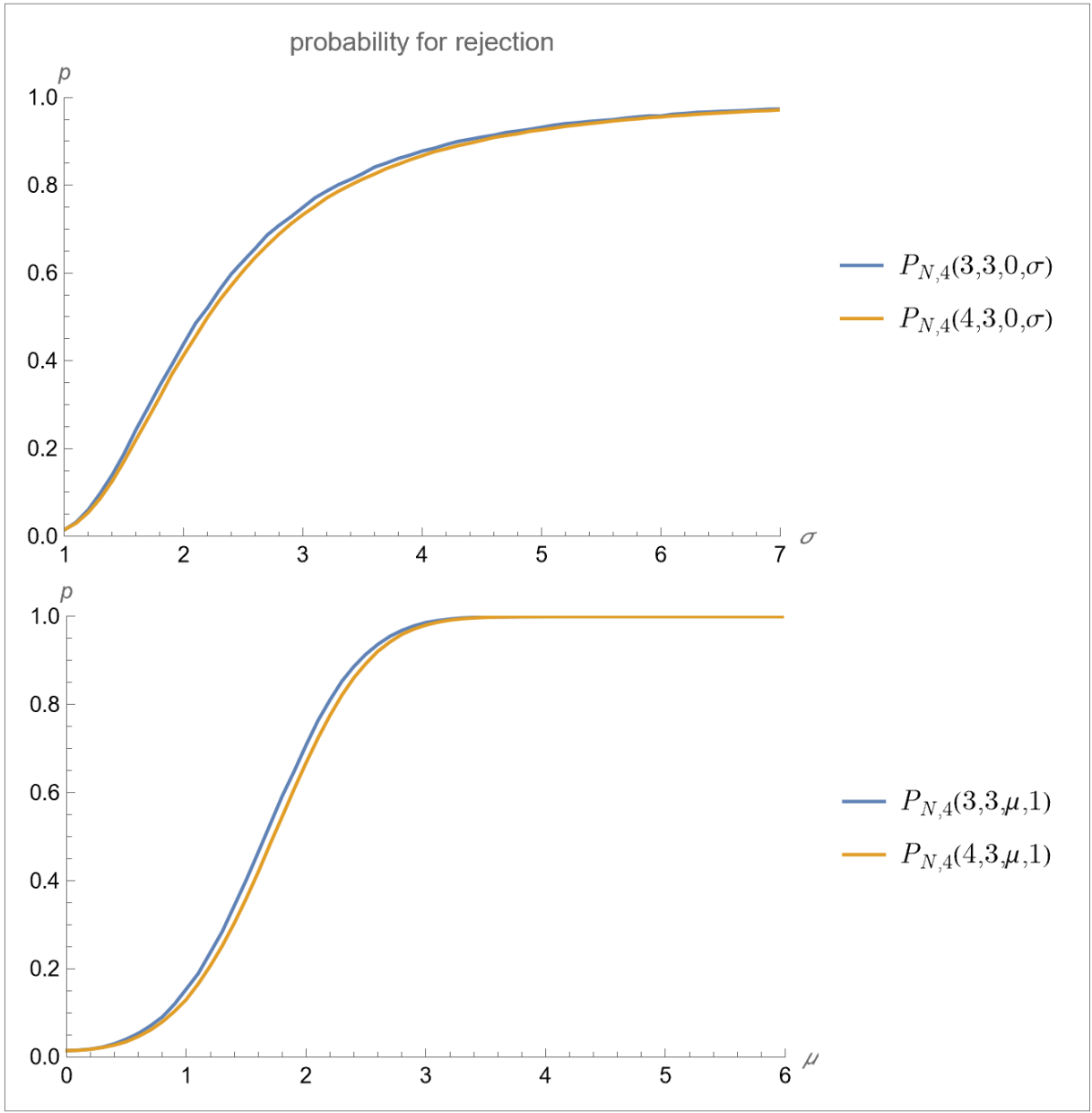

*Figure 134: $P_{N,4}(n,3,\mu,\sigma)$*

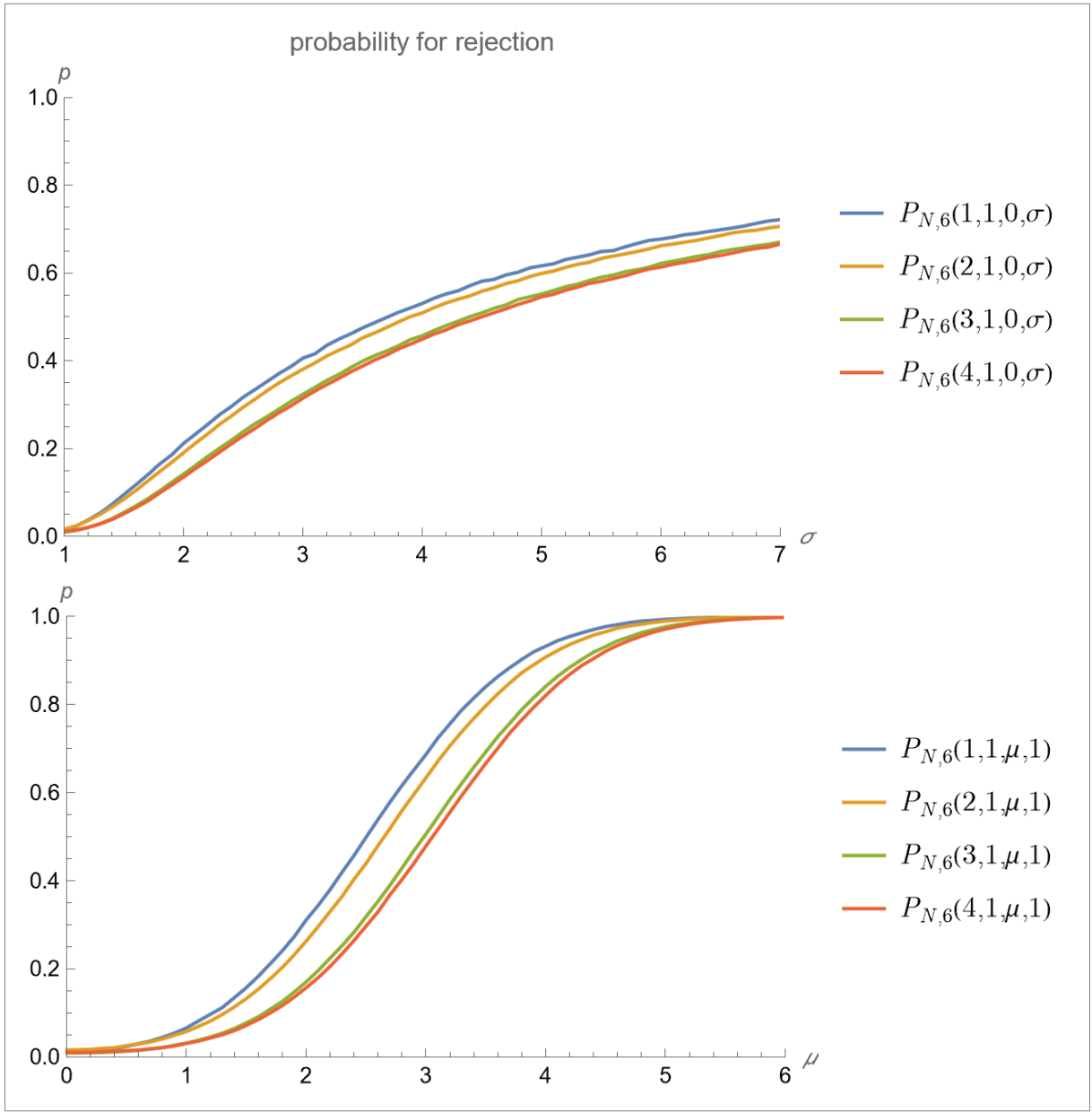

*Figure 135:* $P_{N,6}(n, 1, \mu, \sigma)$

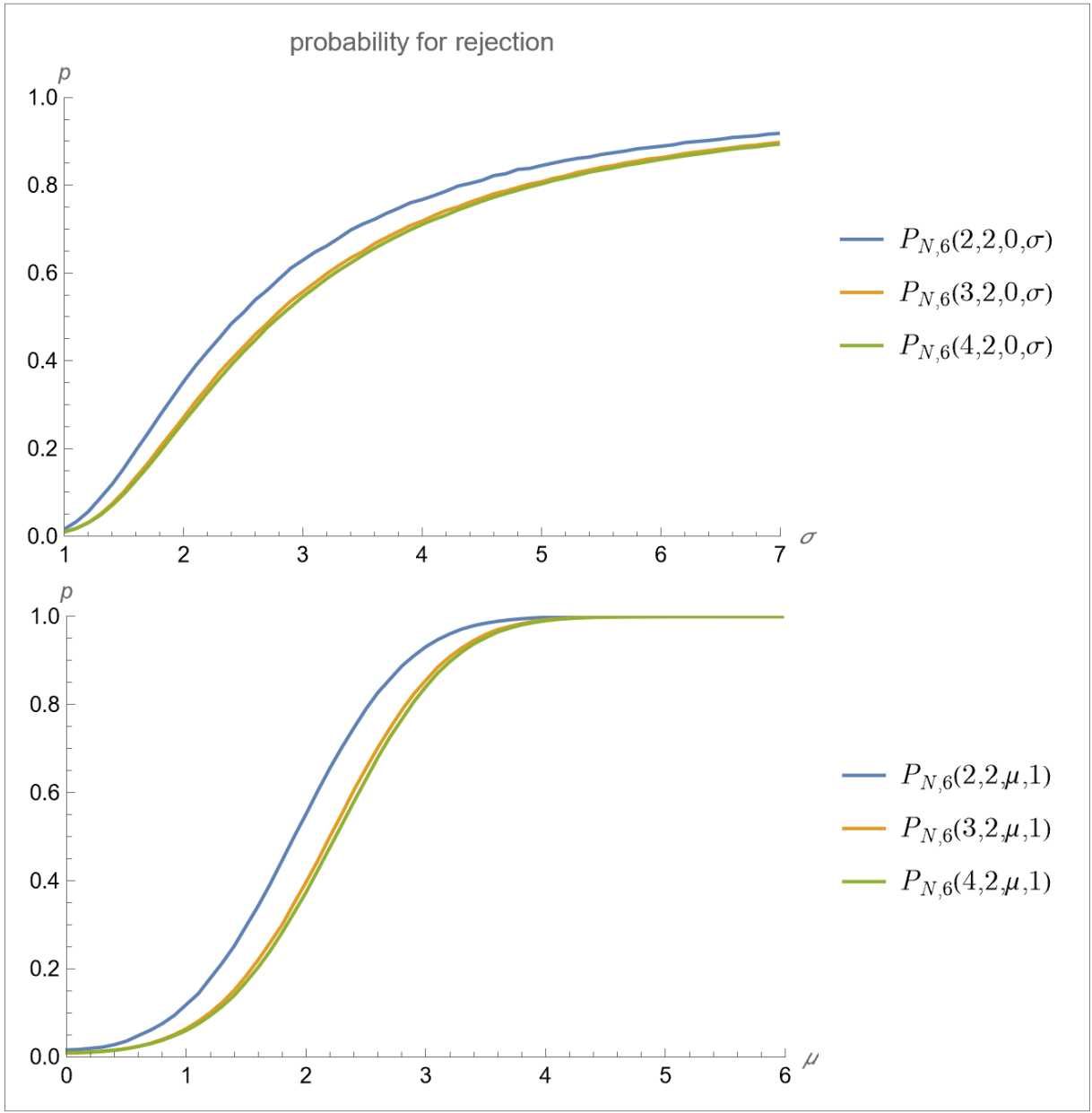

*Figure 136:* $P_{N,6}(n, 2, \mu, \sigma)$

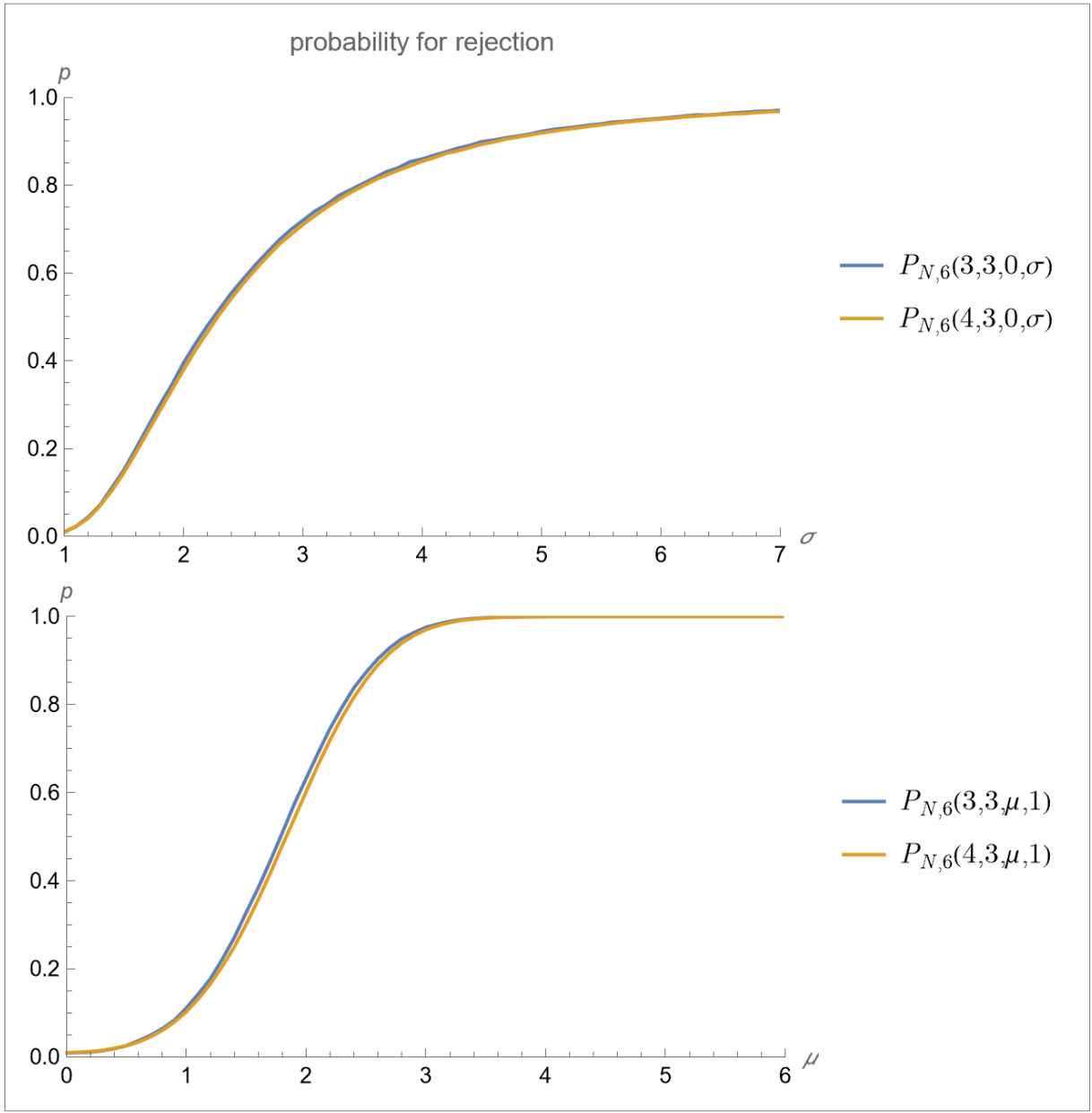

*Figure 137: $P_{N,6}(n, 3, \mu, \sigma)$*

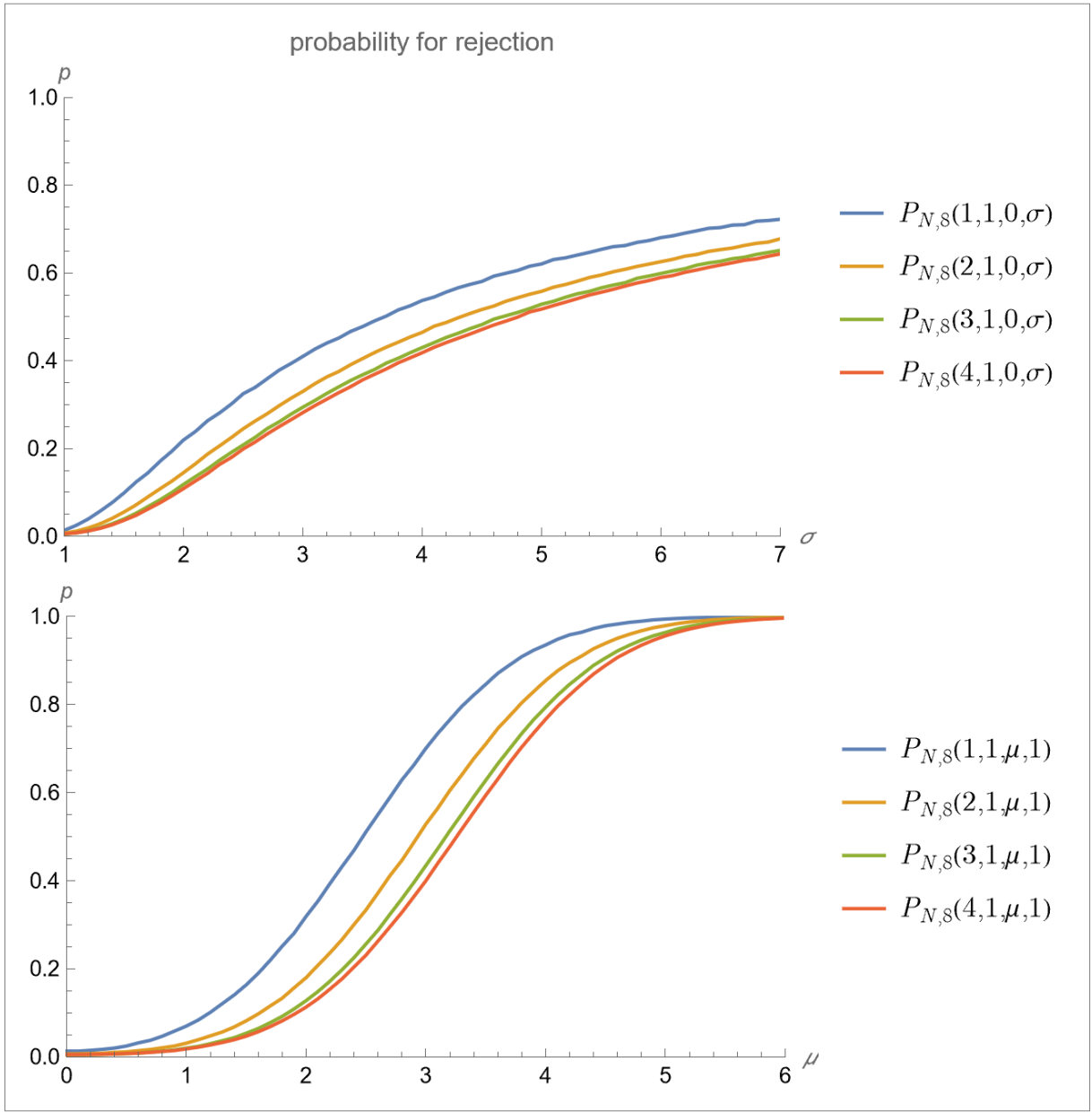

*Figure 138:* $P_{N,8}(n,1,\mu,\sigma)$

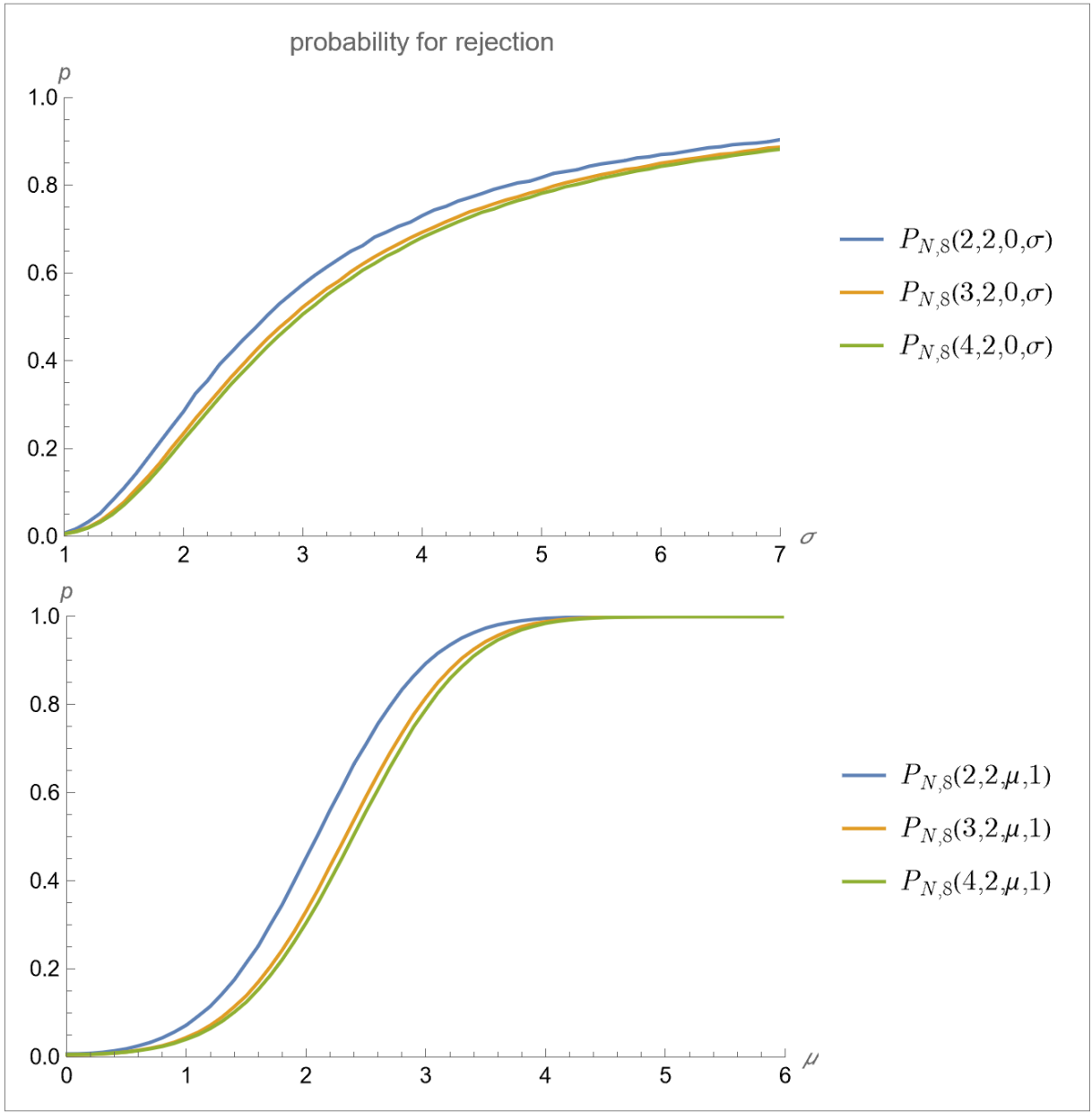

*Figure 139:* $P_{N,8}(n,2,\mu,\sigma)$

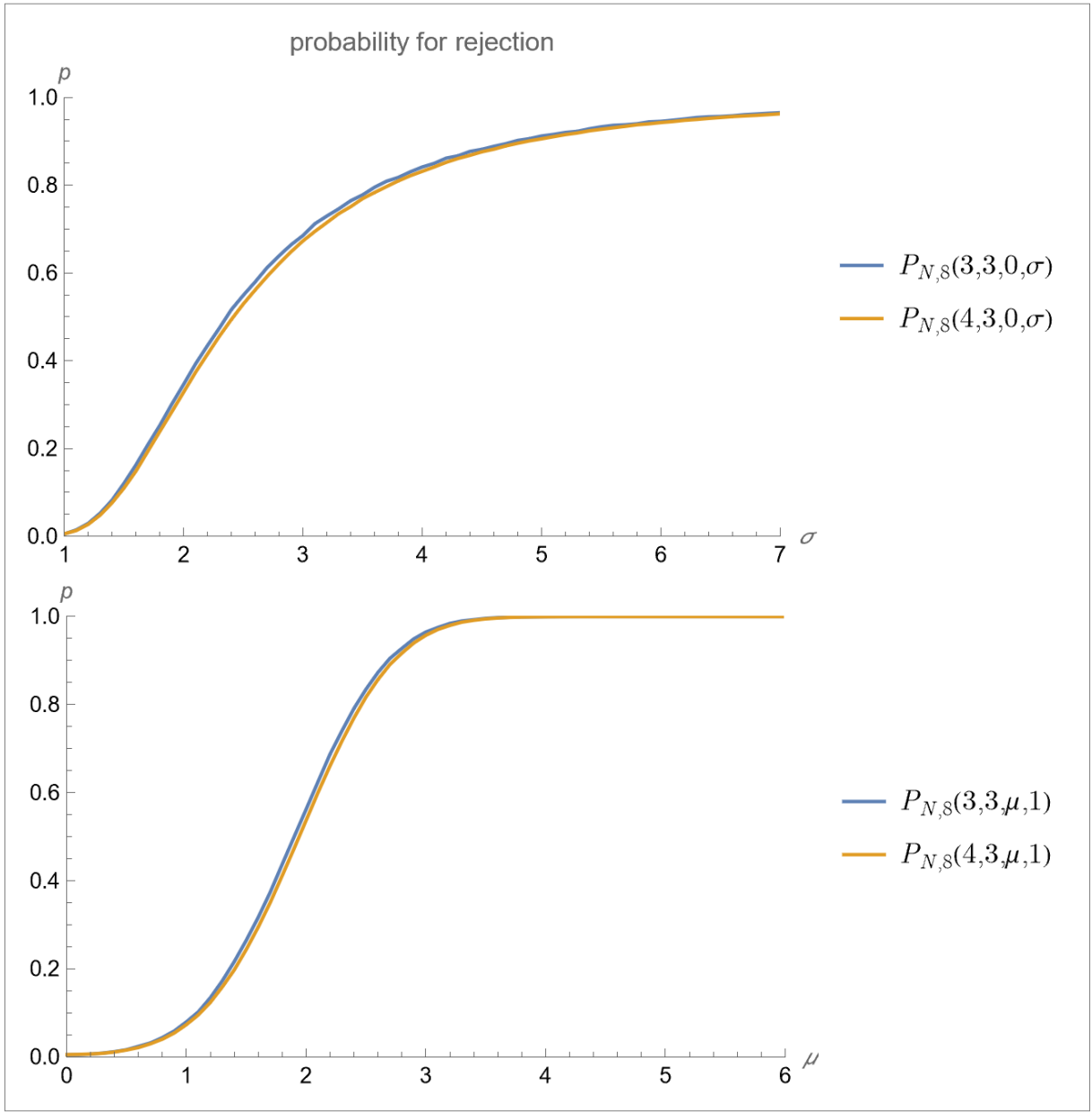

*Figure 140:* $P_{N,8}(n,3,\mu,\sigma)$

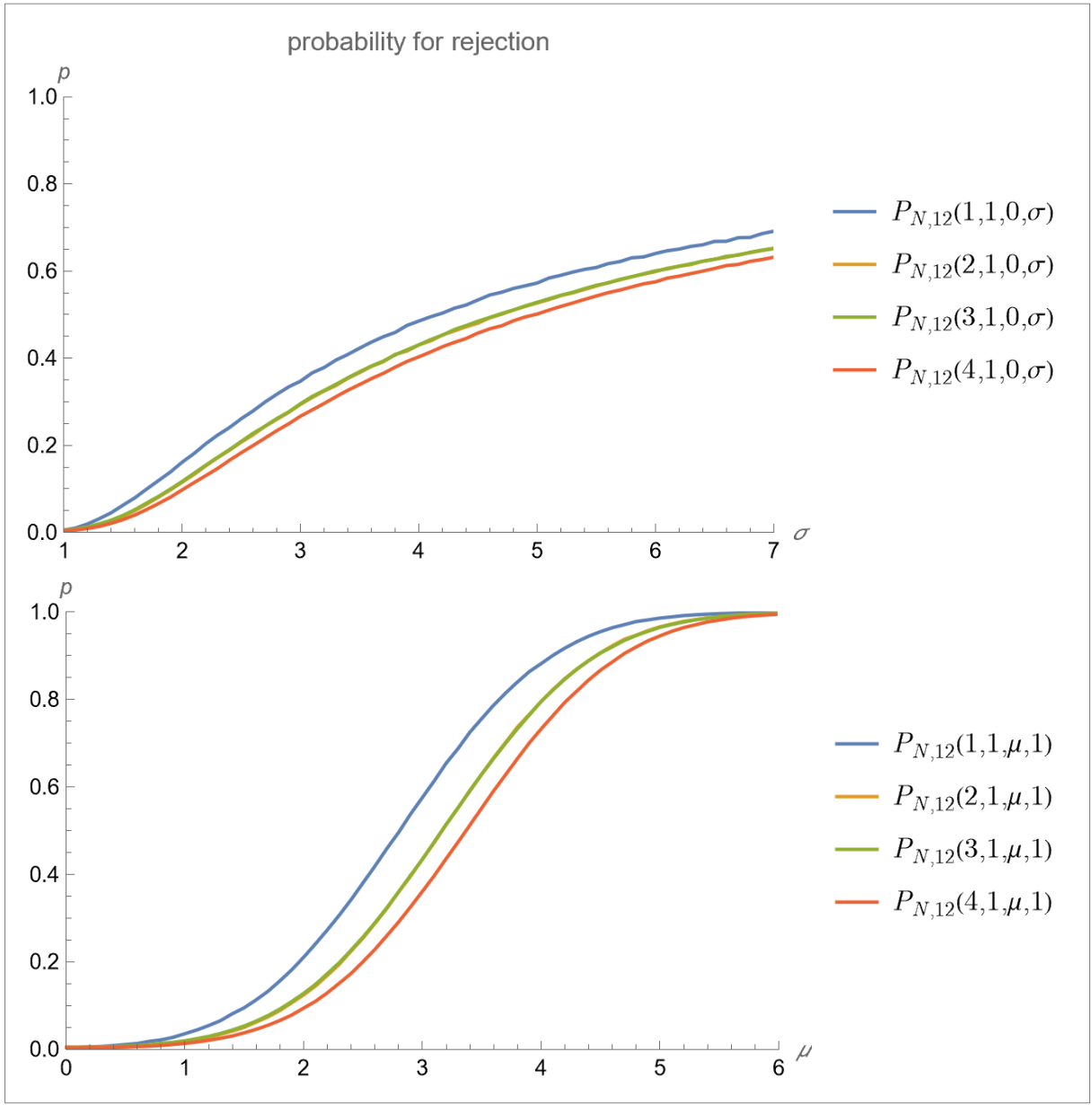

*Figure 141: $P_{N,12}(n,1,\mu,\sigma)$*

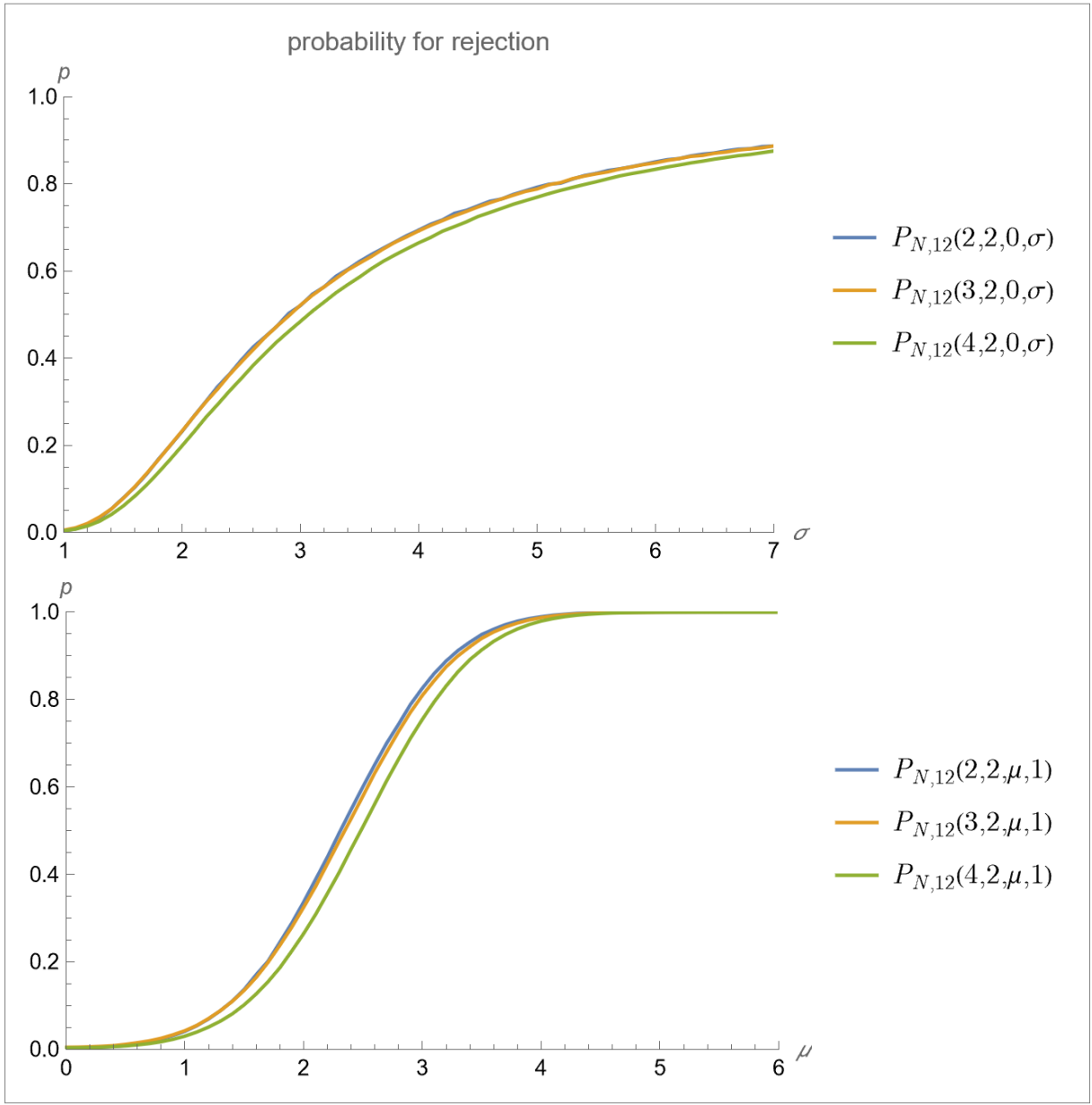

*Figure 142:* $P_{N,12}(n, 2, \mu, \sigma)$

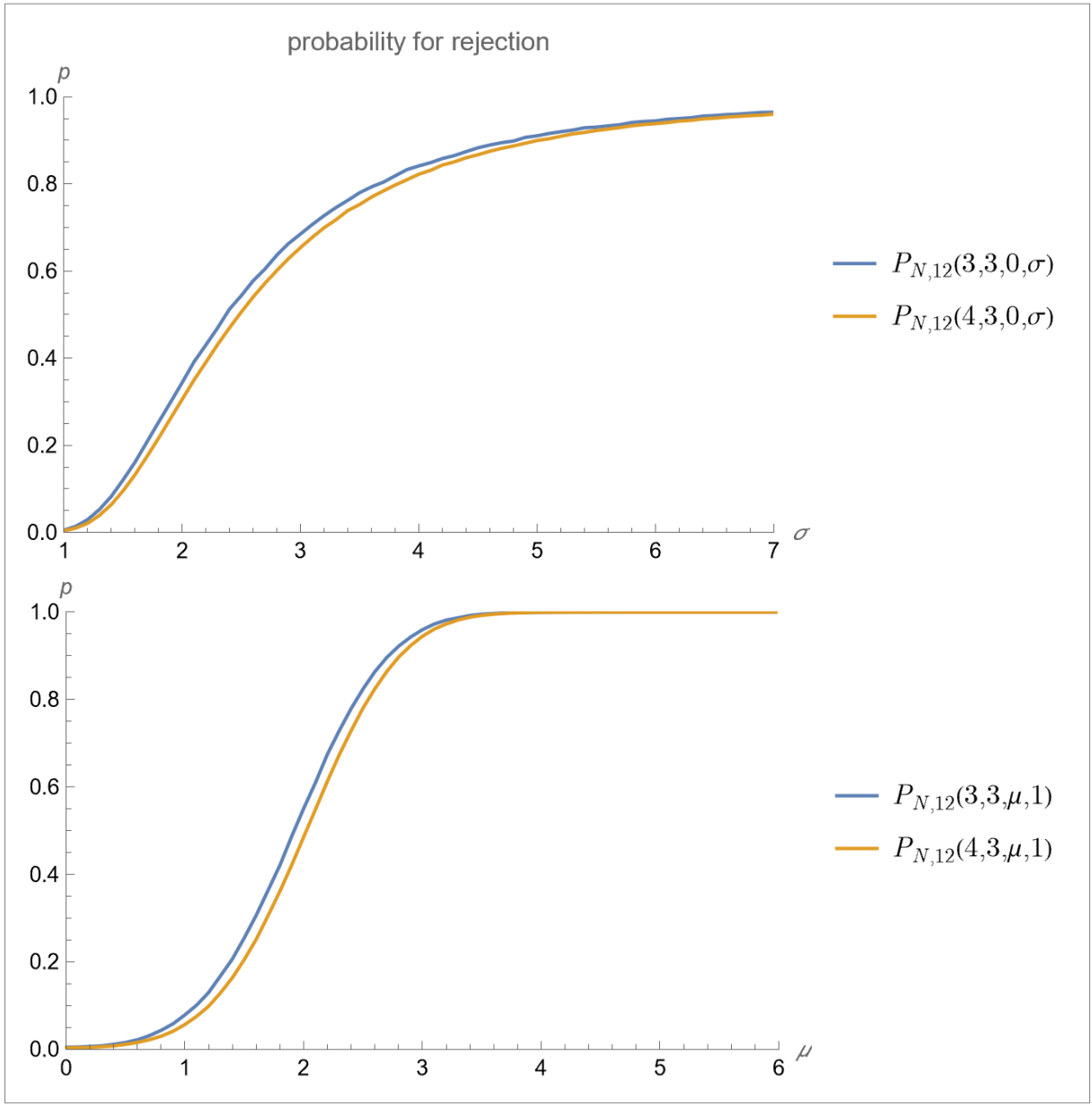

*Figure 143: $P_{N,12}(n, 3, \mu, \sigma)$*

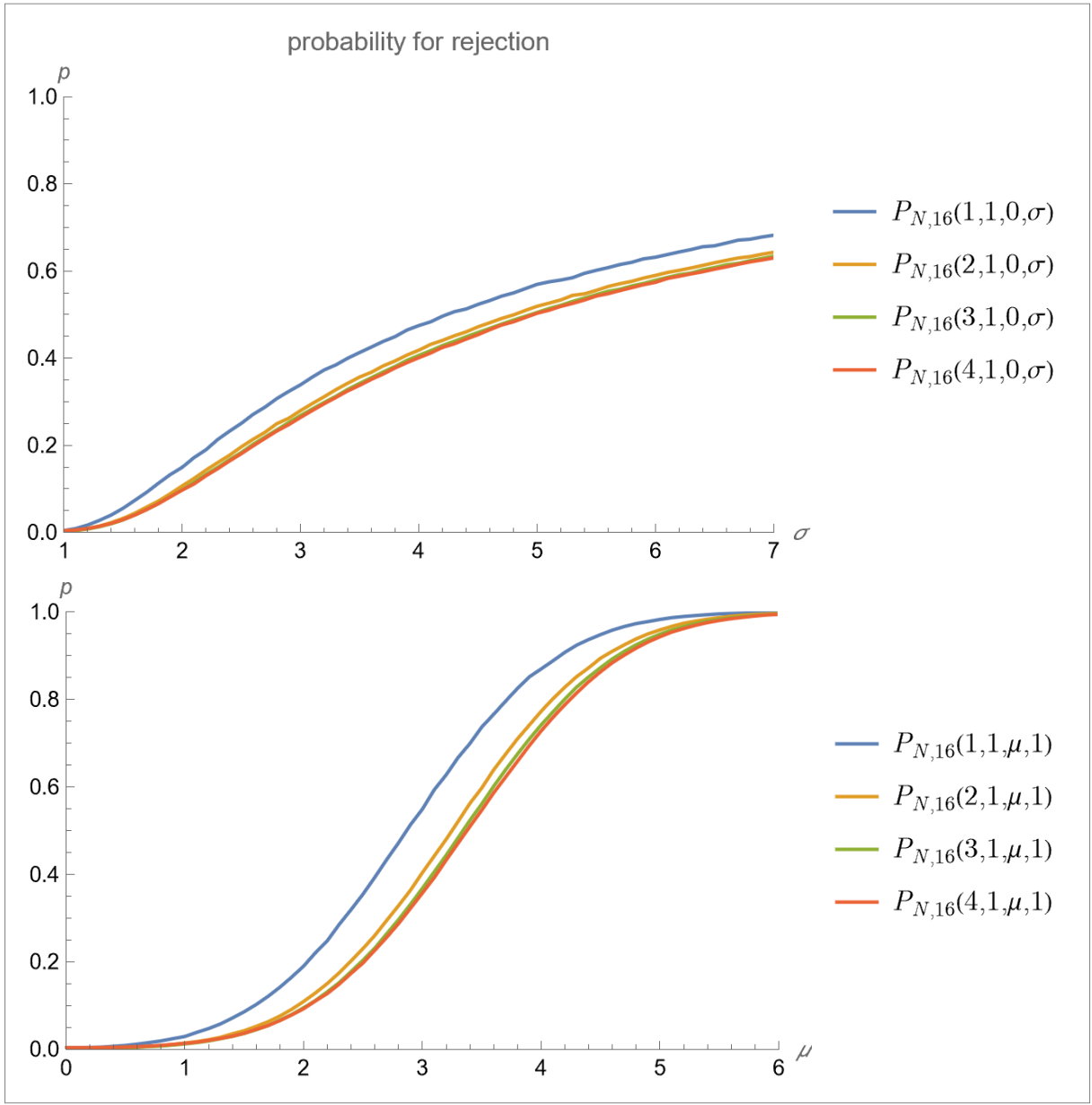

*Figure 144: $P_{N,16}(n, 1, \mu, \sigma)$*

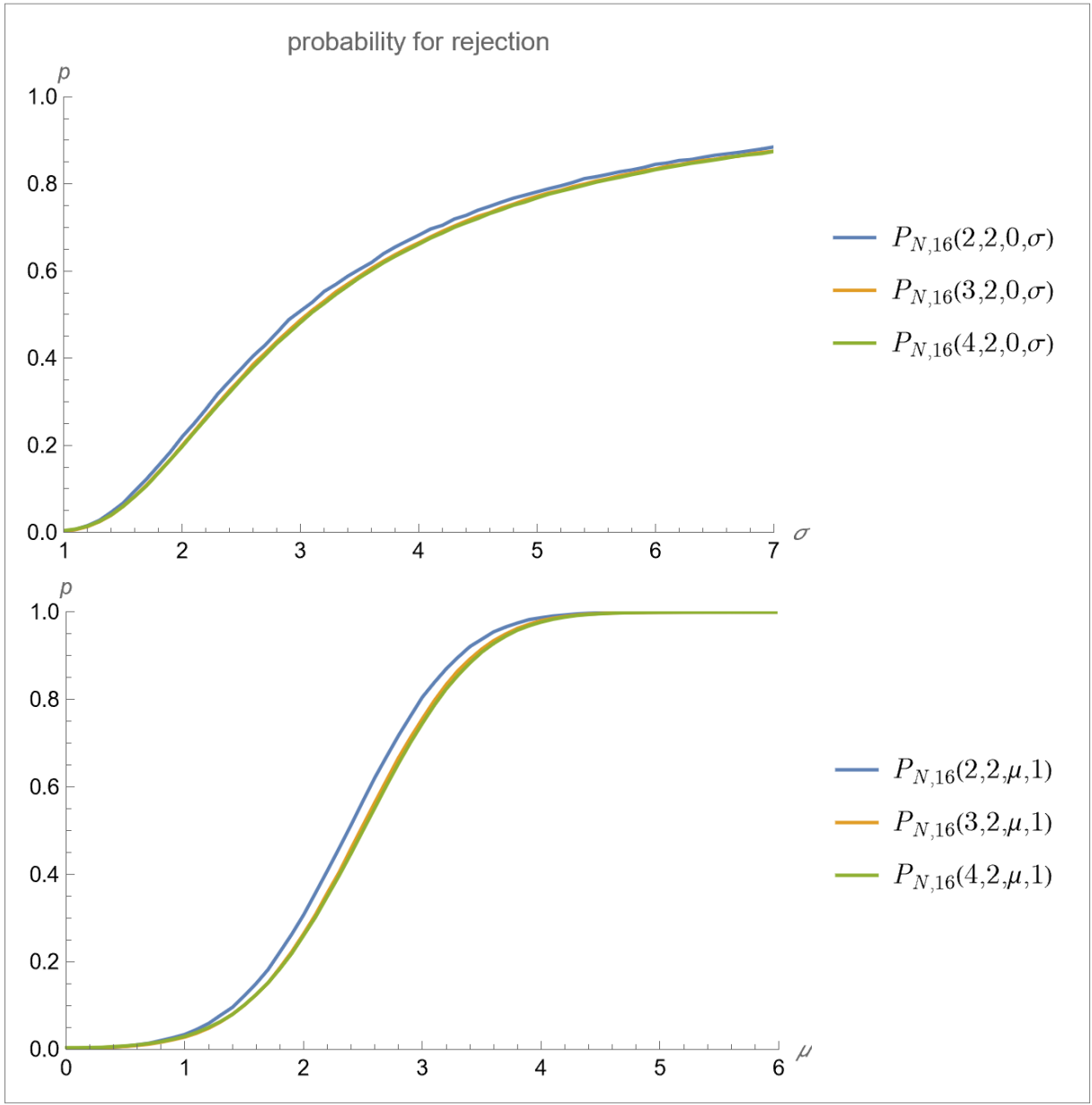

*Figure 145: $P_{N,16}(n, 2, \mu, \sigma)$*

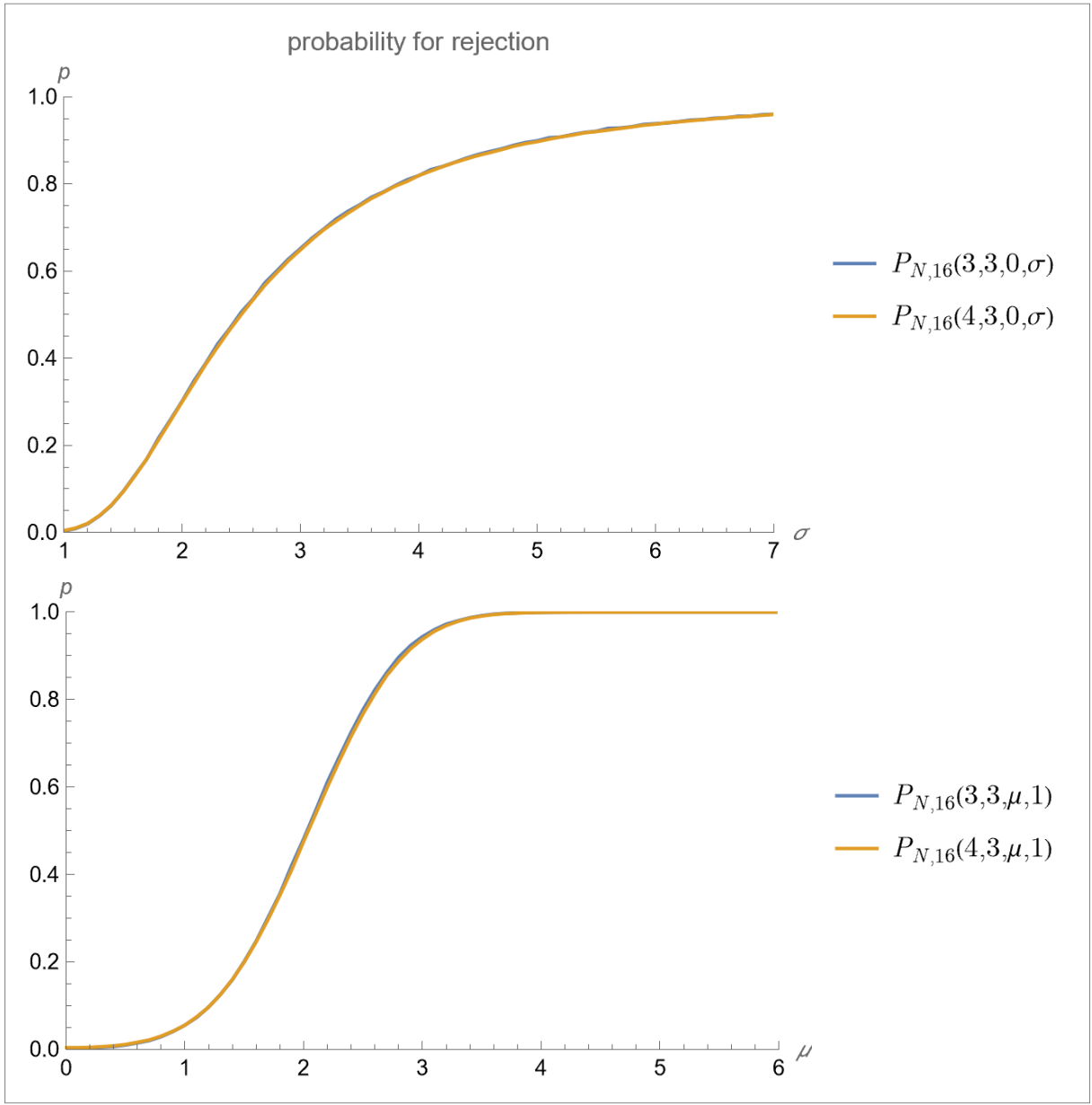

*Figure 146: $P_{N,16}(n, 3, \mu, \sigma)$*

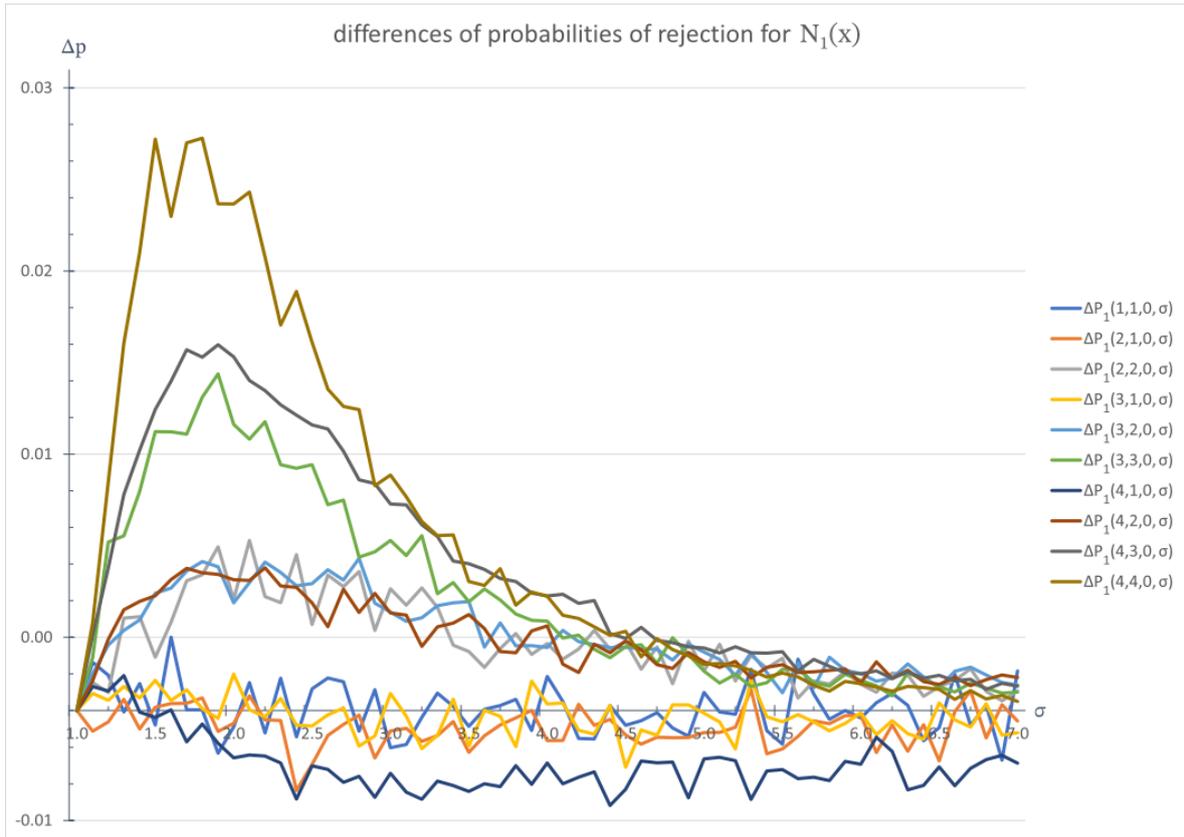

*Figure 147: $\Delta P_1(n, k, 0, \sigma)$*

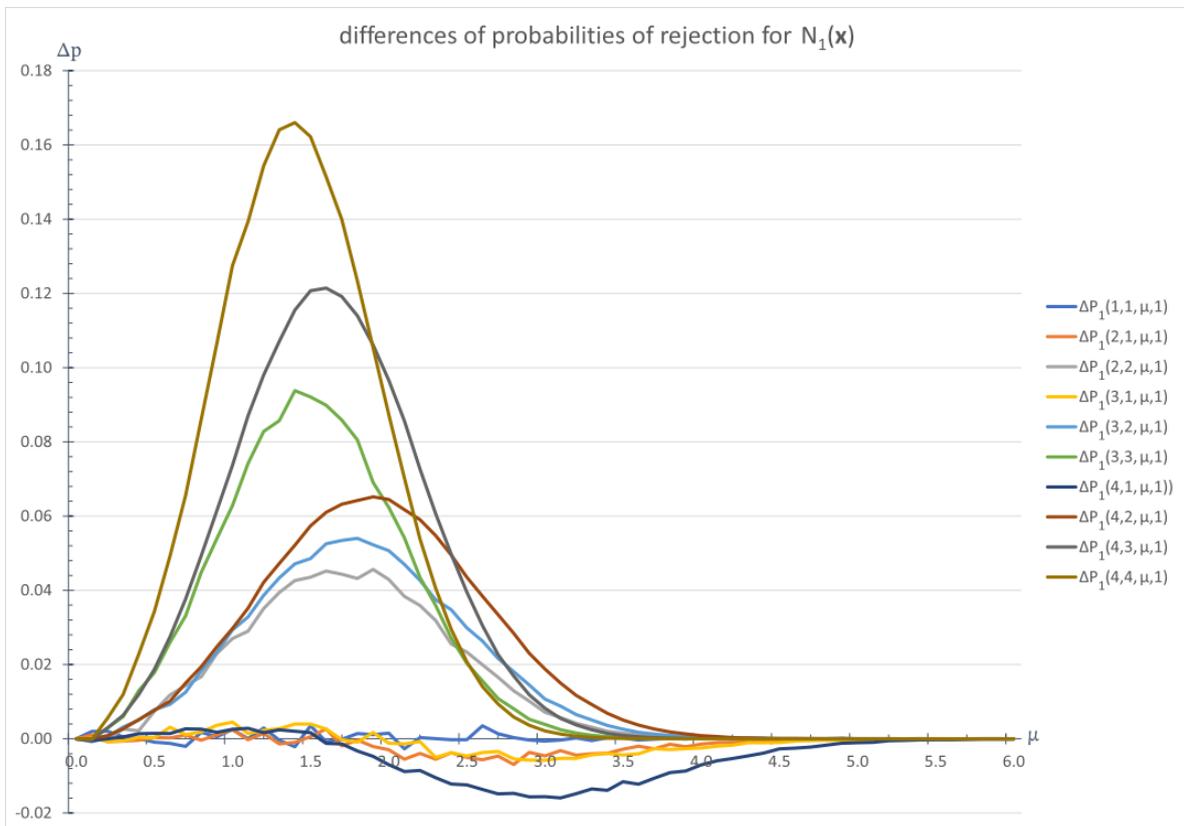

*Figure 148: $\Delta P_1(n, k, \mu, 1)$*

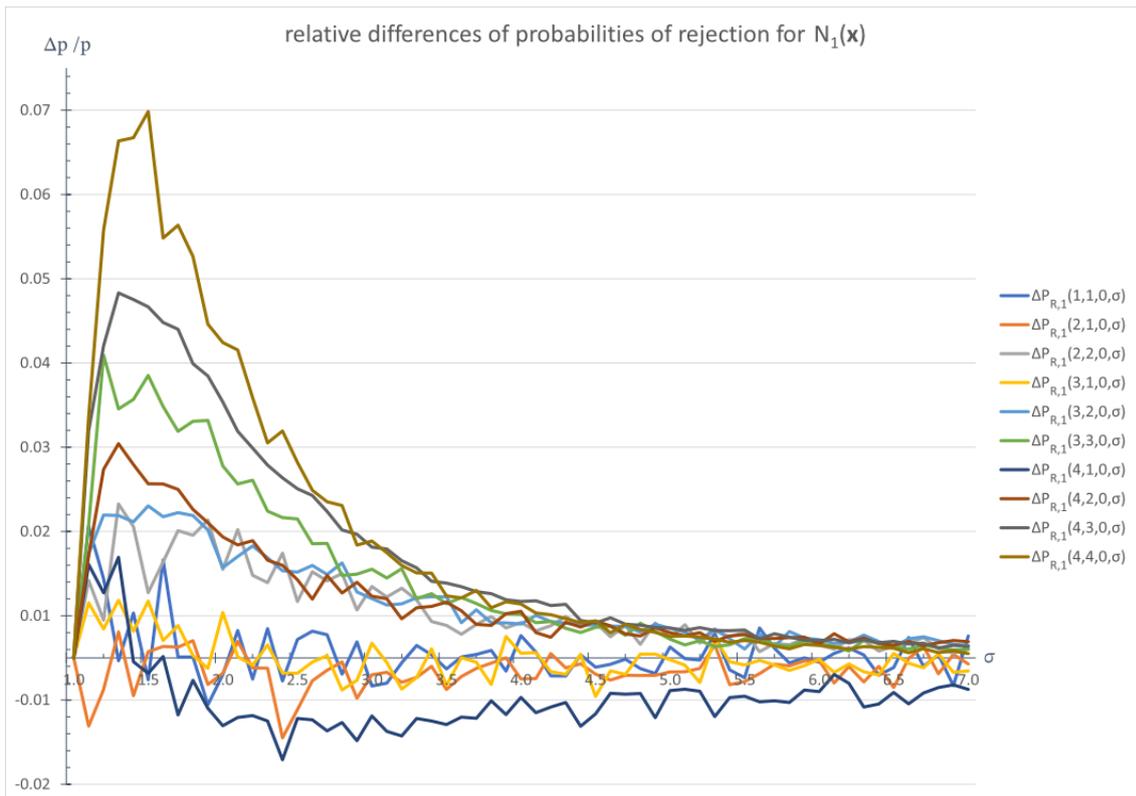

*Figure 149: $\Delta P_{R,1}(n,k,0,\sigma)$*

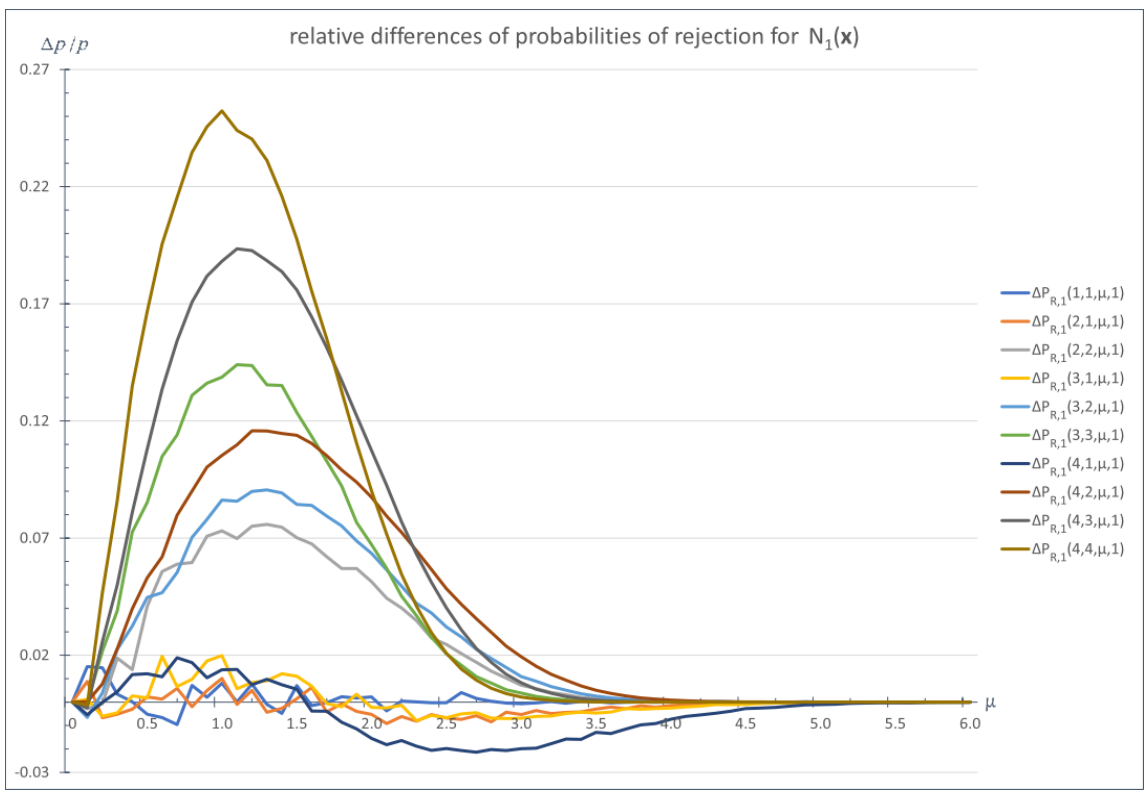

*Figure 150: $\Delta P_{R,1}(n,k,\mu,1)$*

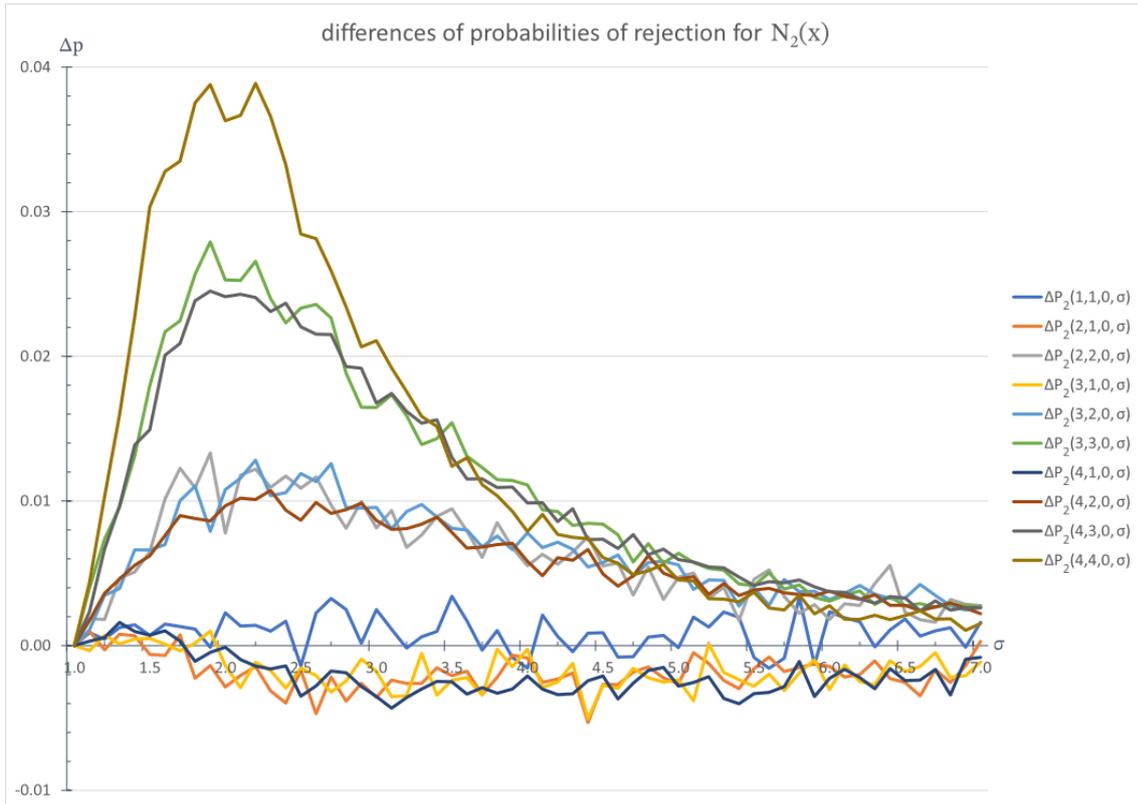

*Figure 151: $\Delta P_2(n,k,0,\sigma)$*

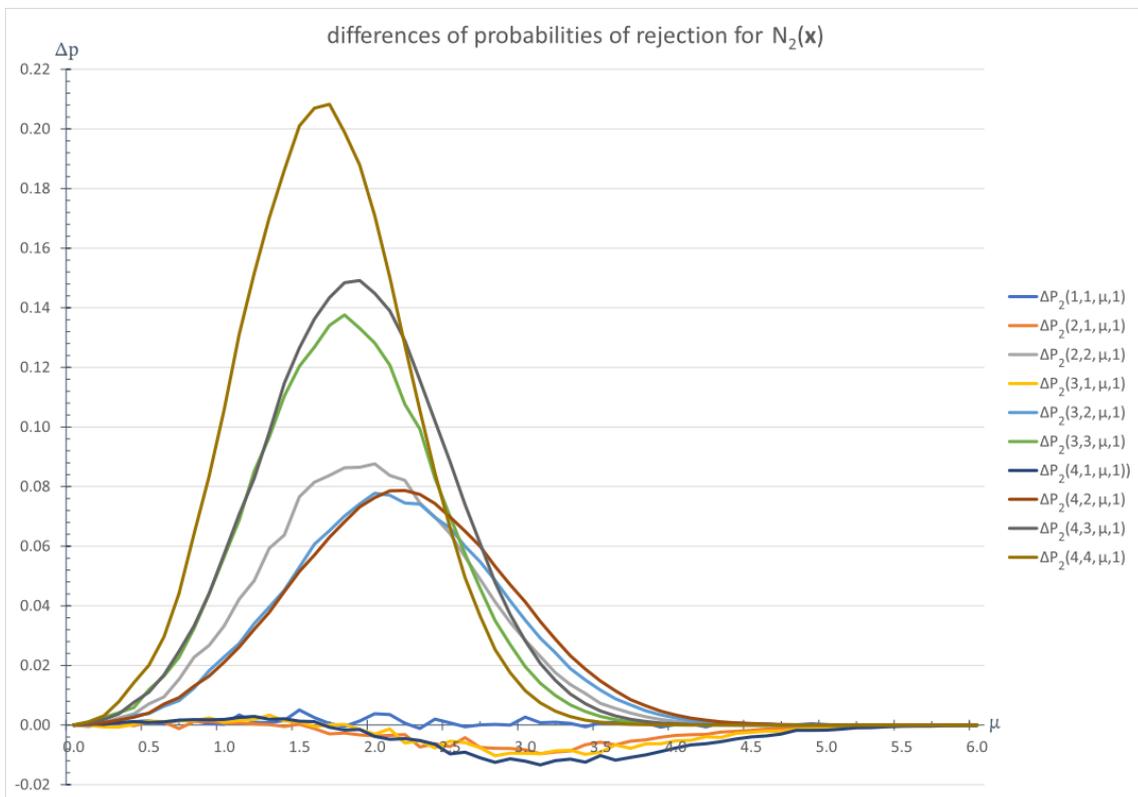

*Figure 152: $\Delta P_2(n,k,\mu,1)$*

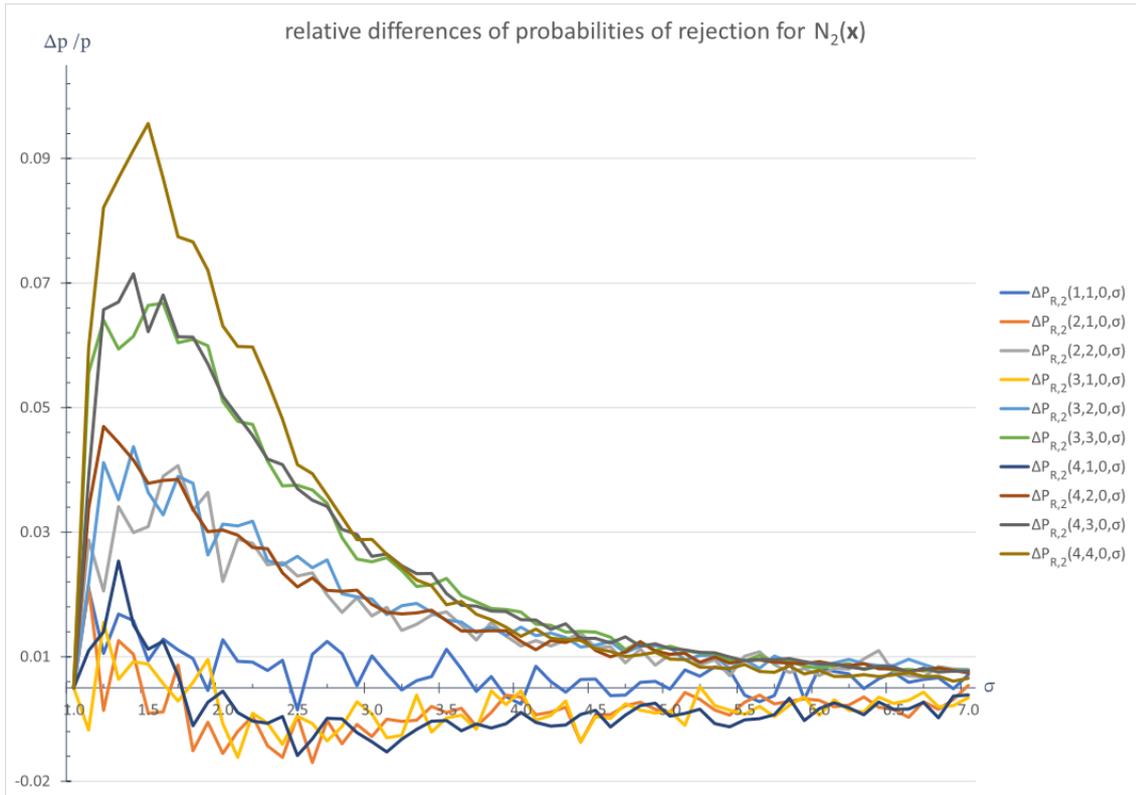

*Figure 153:* $\Delta P_{R,2}(n, k, 0, \sigma)$

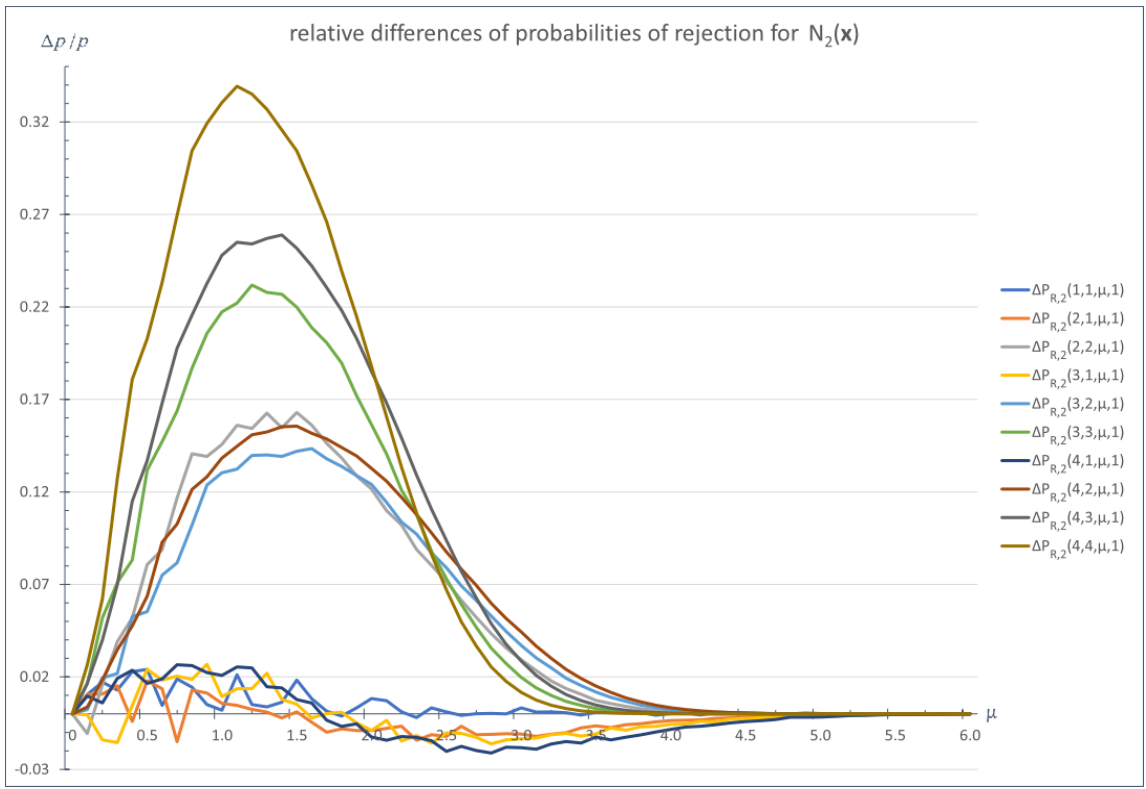

*Figure 154:* $\Delta P_{R,2}(n, k, \mu, 1)$

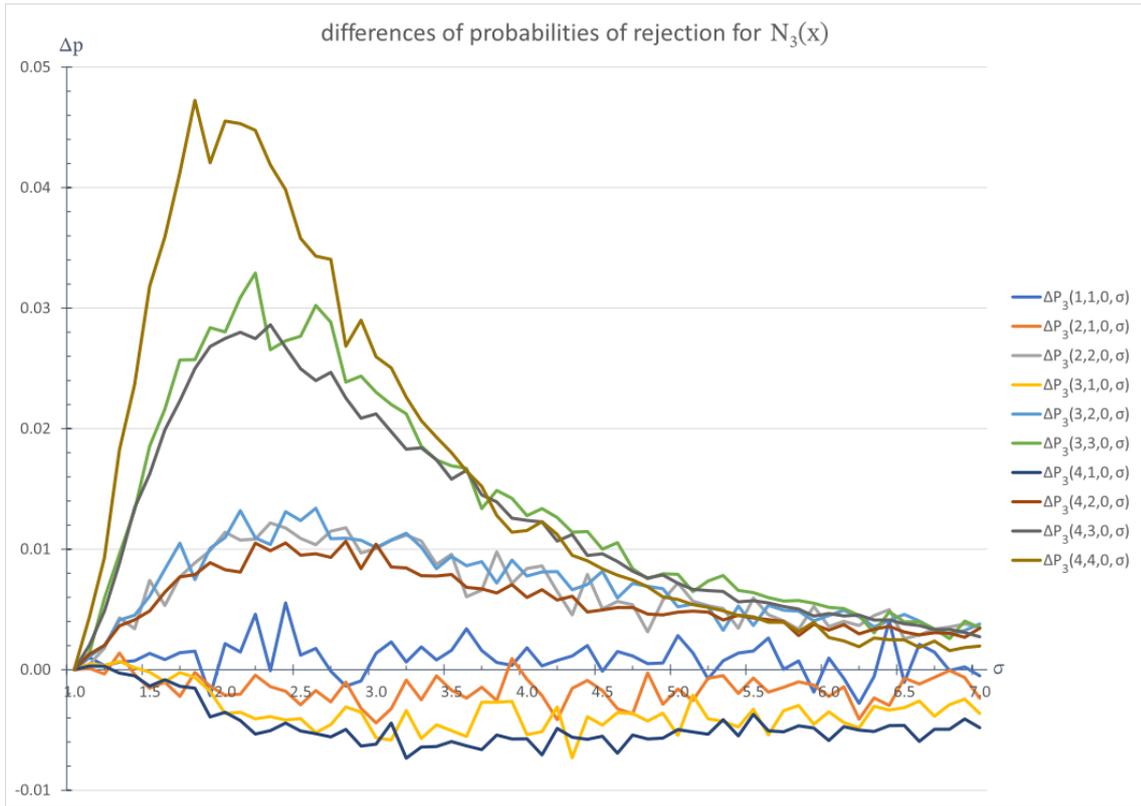

*Figure 155: $\Delta P_3(n, k, 0, \sigma)$*

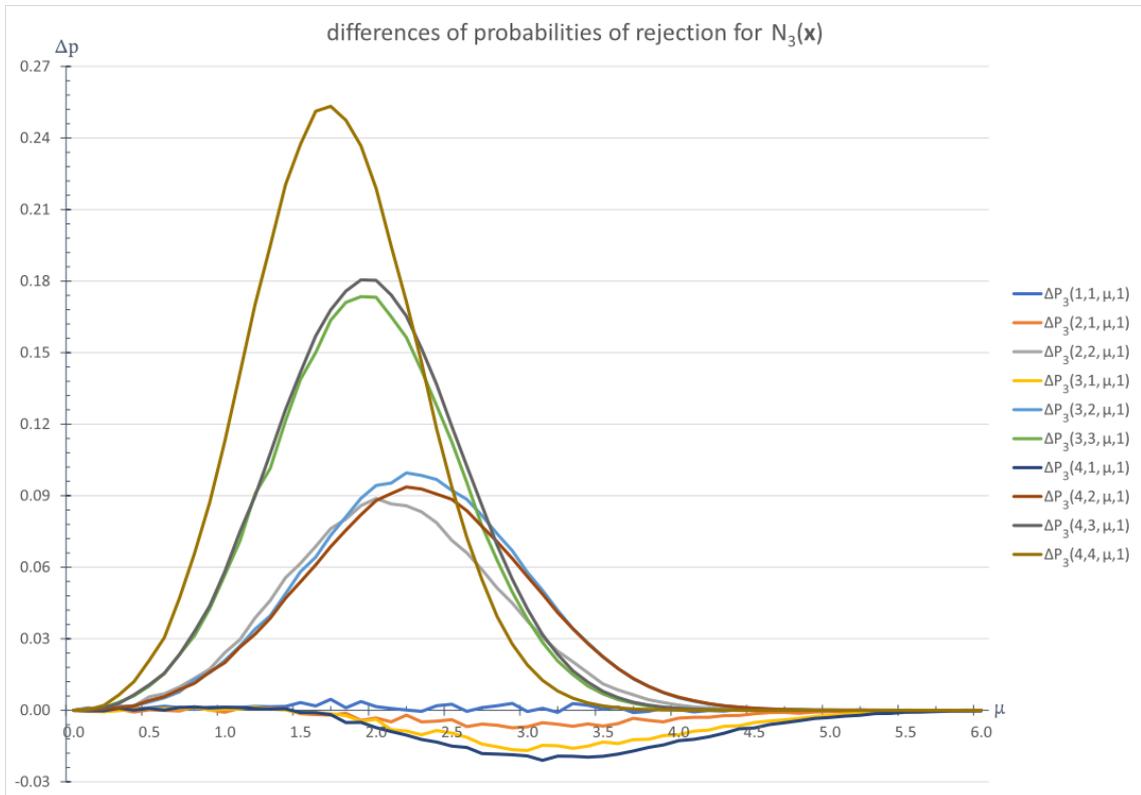

*Figure 156: $\Delta P_3(n, k, \mu, 1)$*

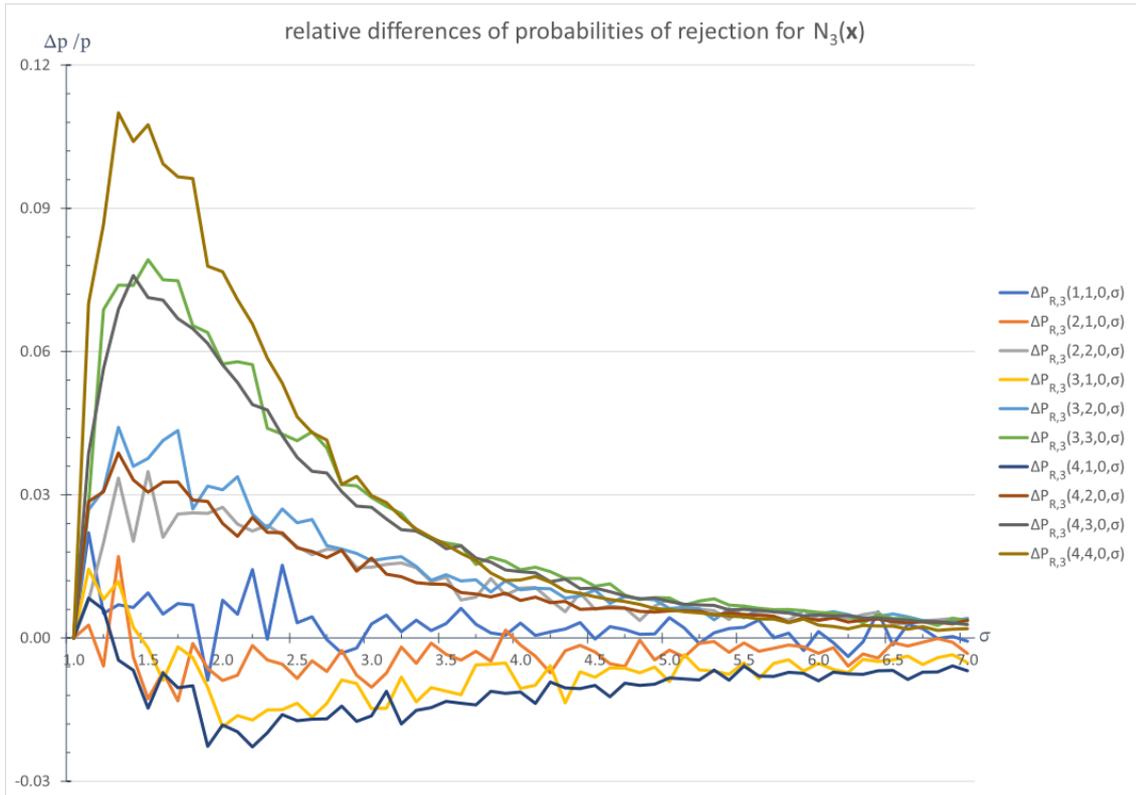

*Figure 157:* $\Delta P_{R,3}(n,k,0,\sigma)$

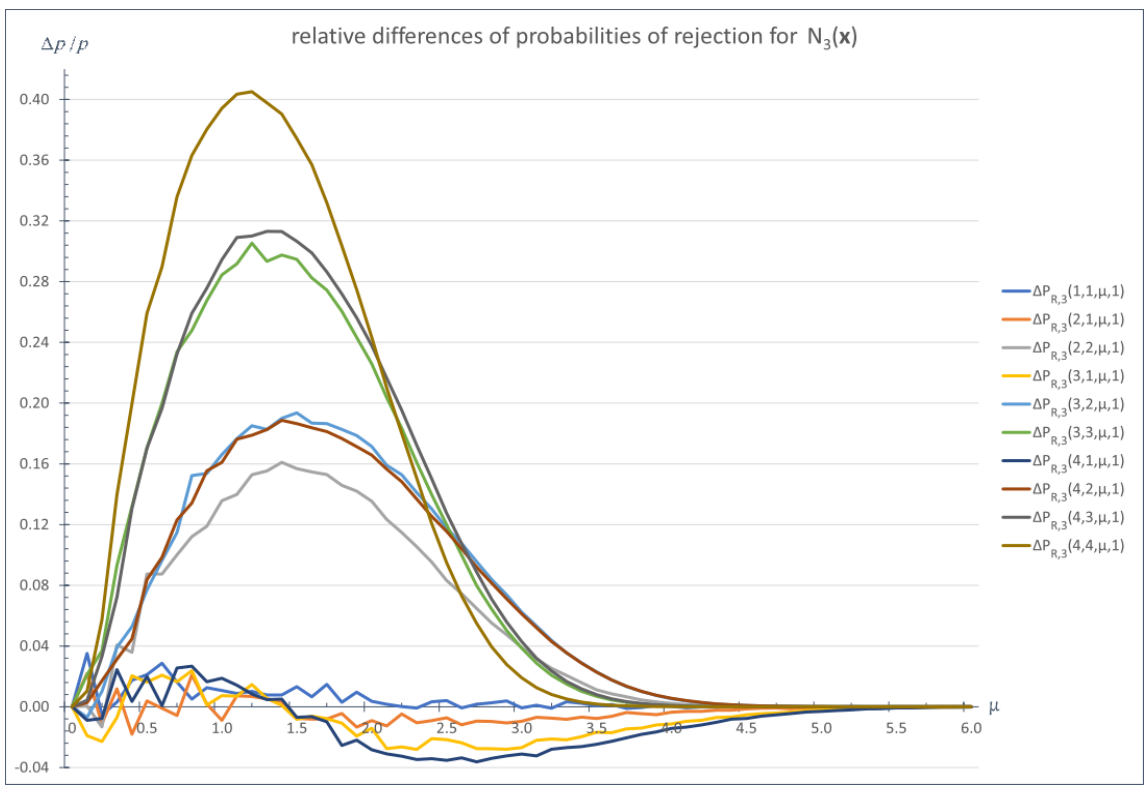

*Figure 158:* $\Delta P_{R,3}(n,k,\mu,1)$

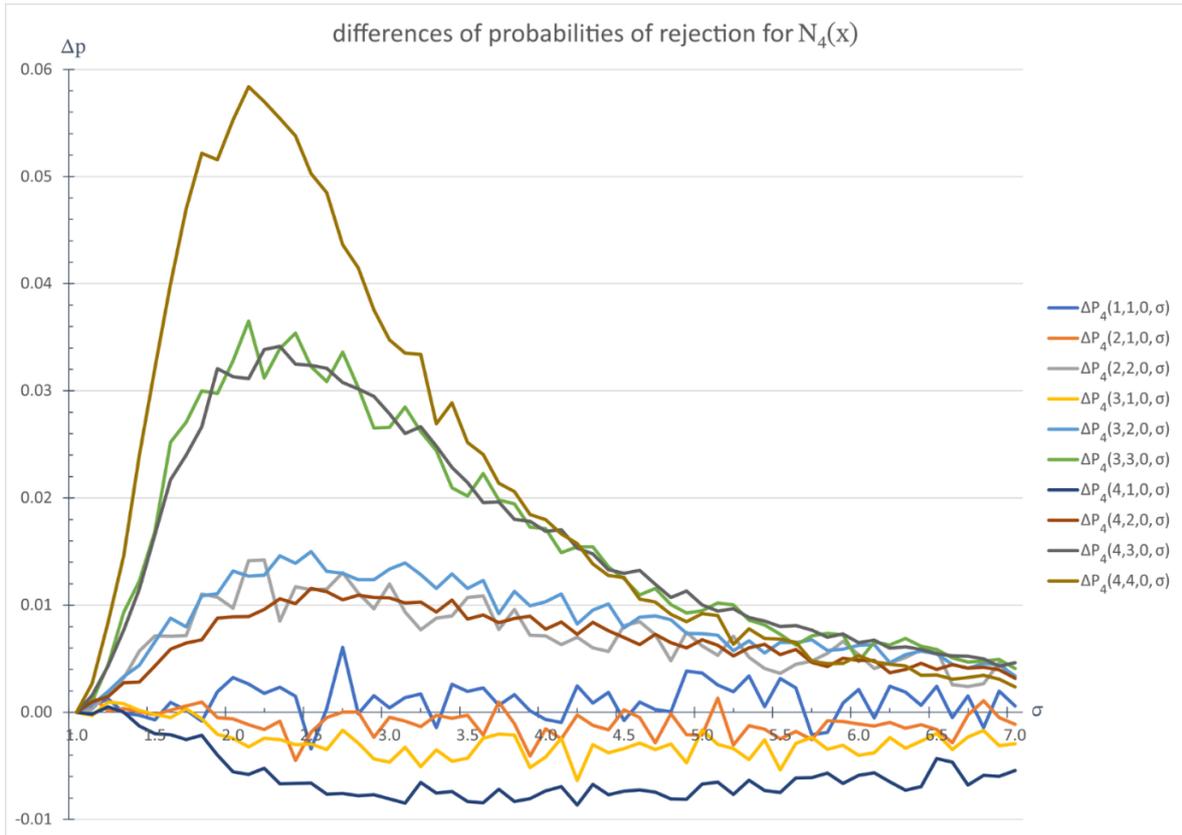

*Figure 159: $\Delta P_4(n,k,0,\sigma)$*

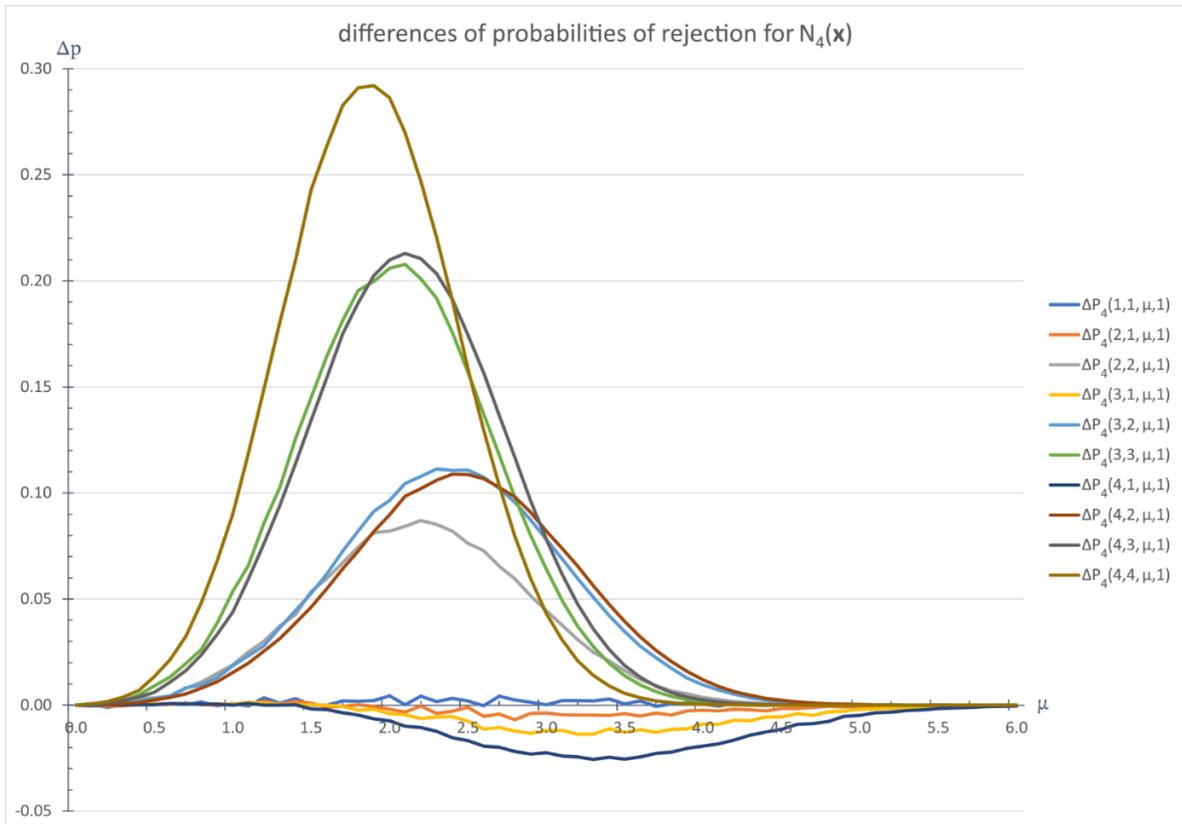

*Figure 160: $\Delta P_4(n,k,\mu,1)$*

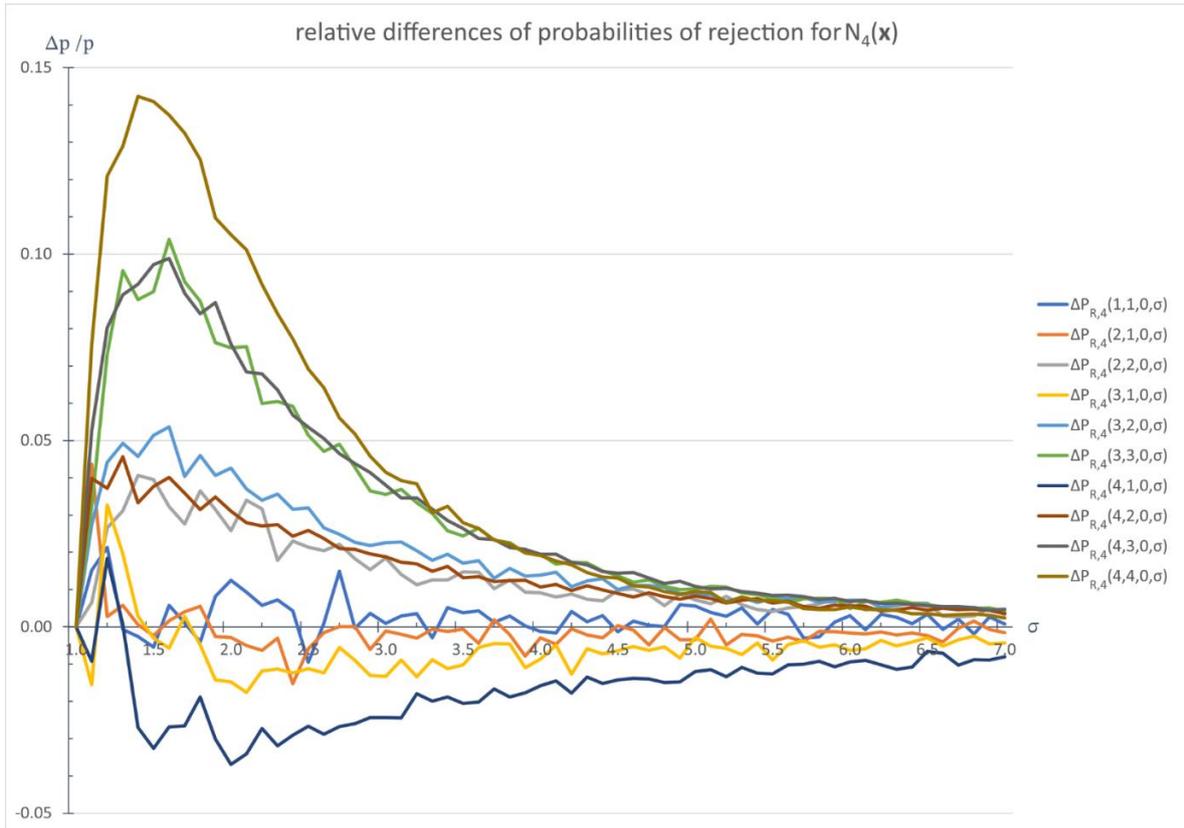

*Figure 161:* $\Delta P_{R,4}(n, k, 0, \sigma)$

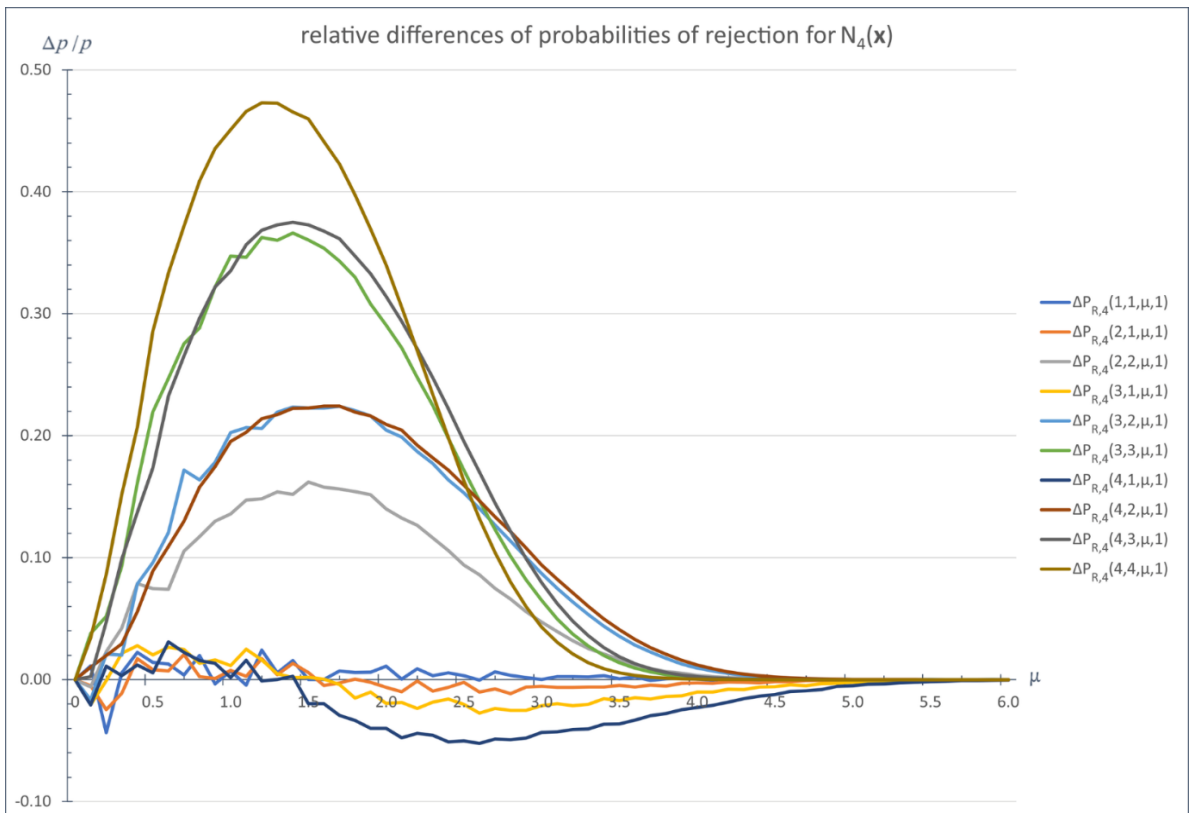

*Figure 162:* $\Delta P_{R,4}(n, k, \mu, 1)$

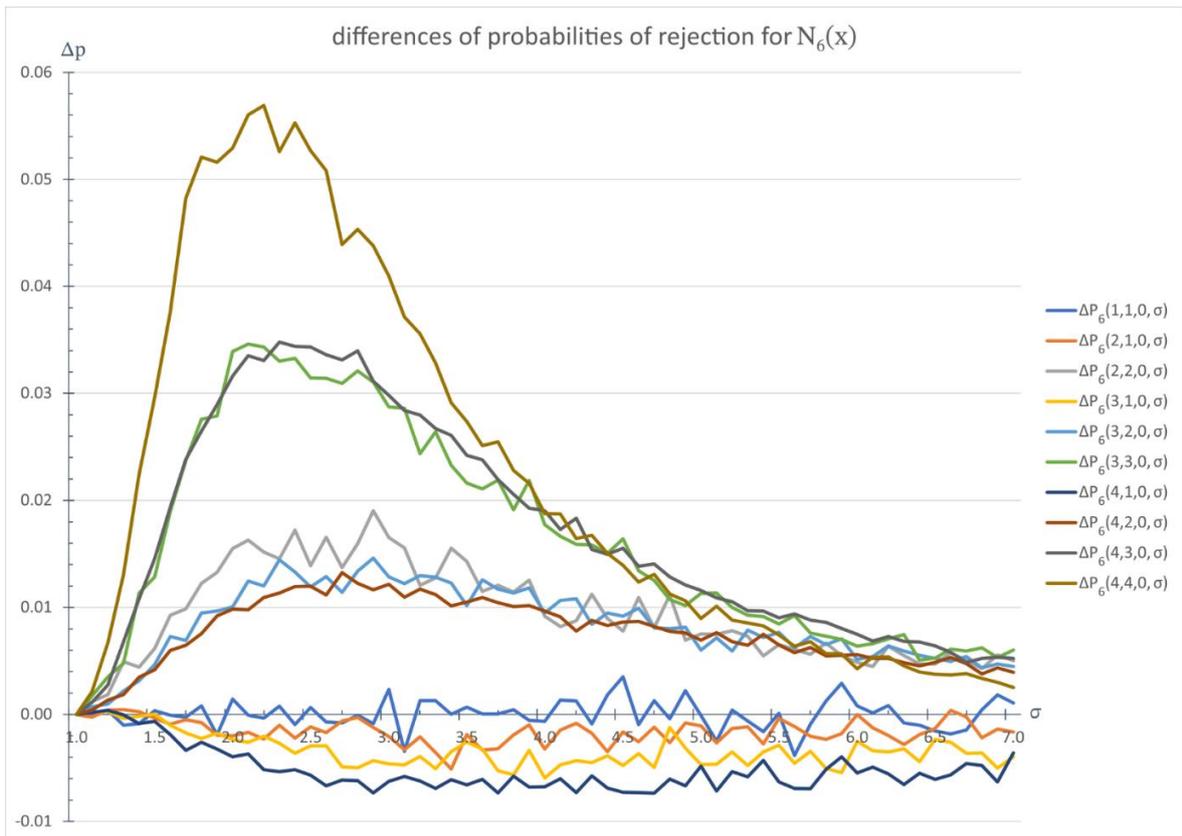

*Figure 163: $\Delta P_6(n,k,0,\sigma)$*

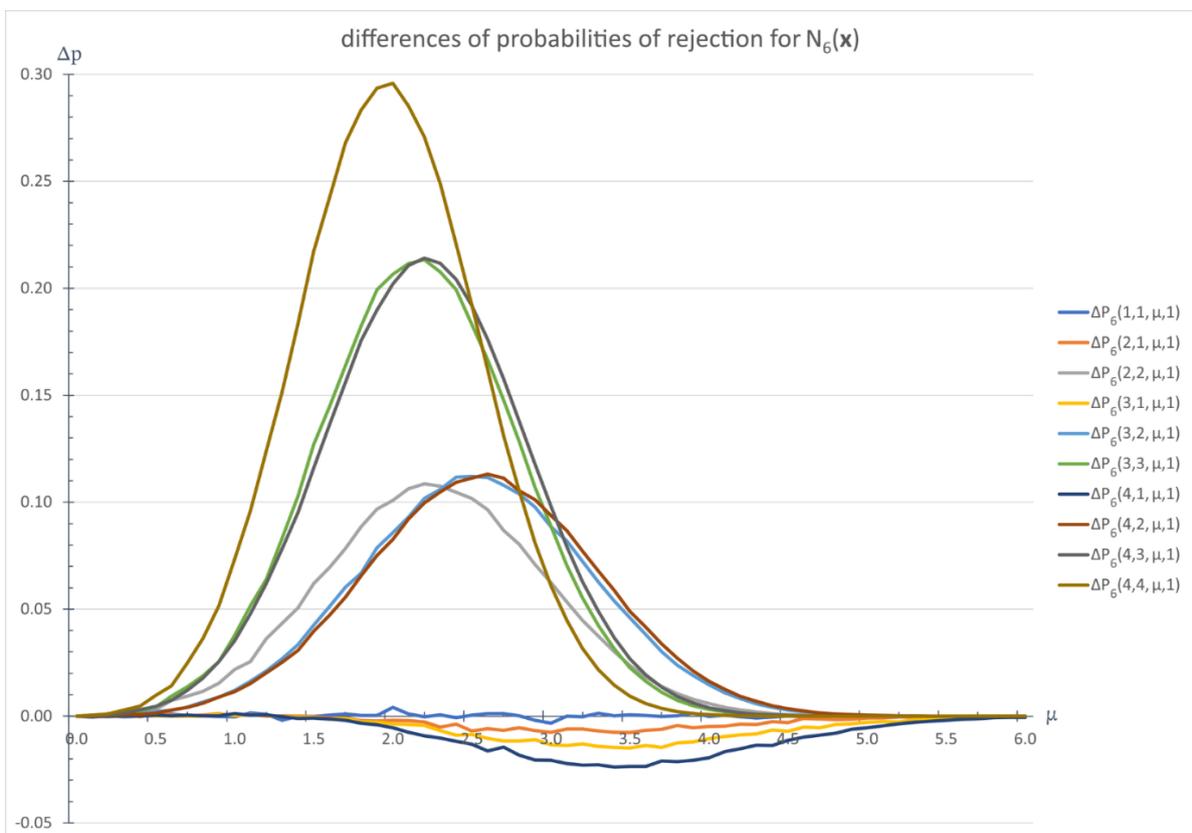

*Figure 164: $\Delta P_6(n,k,\mu,1)$*

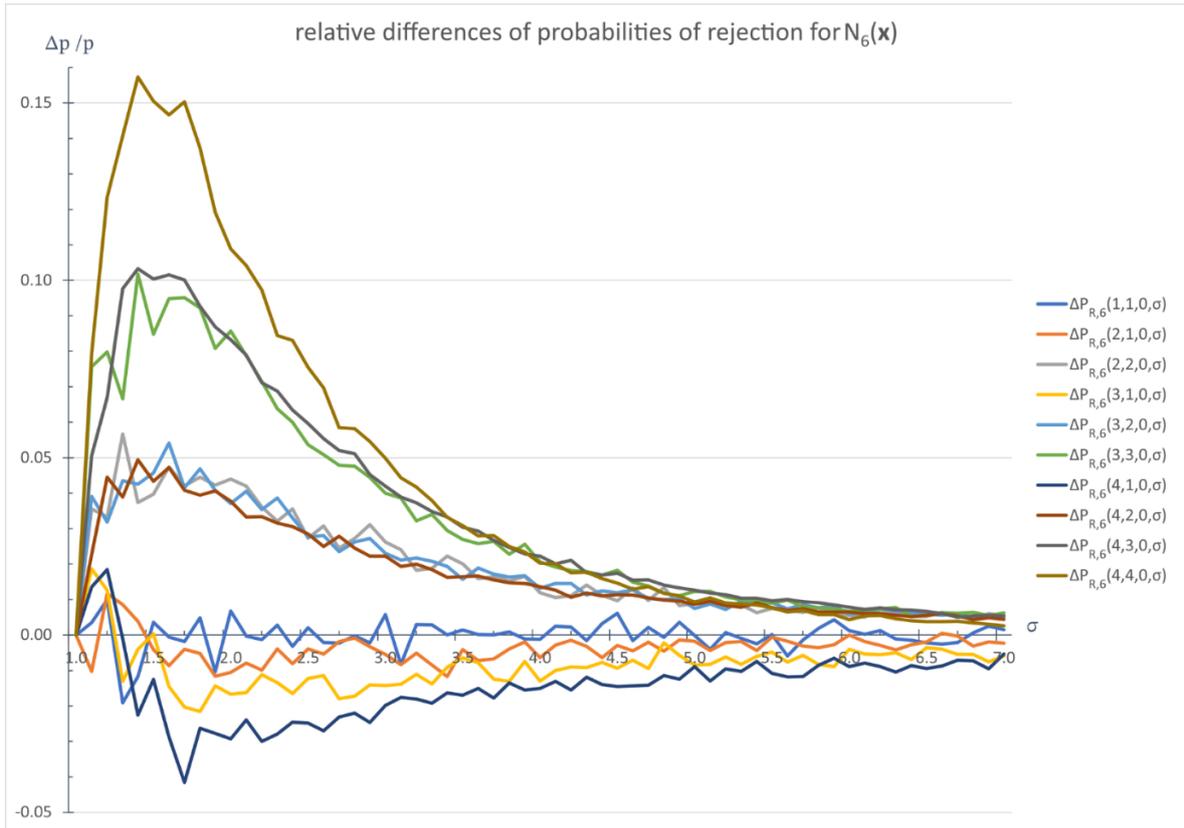

*Figure 165: $\Delta P_{R,6}(n, k, 0, \sigma)$*

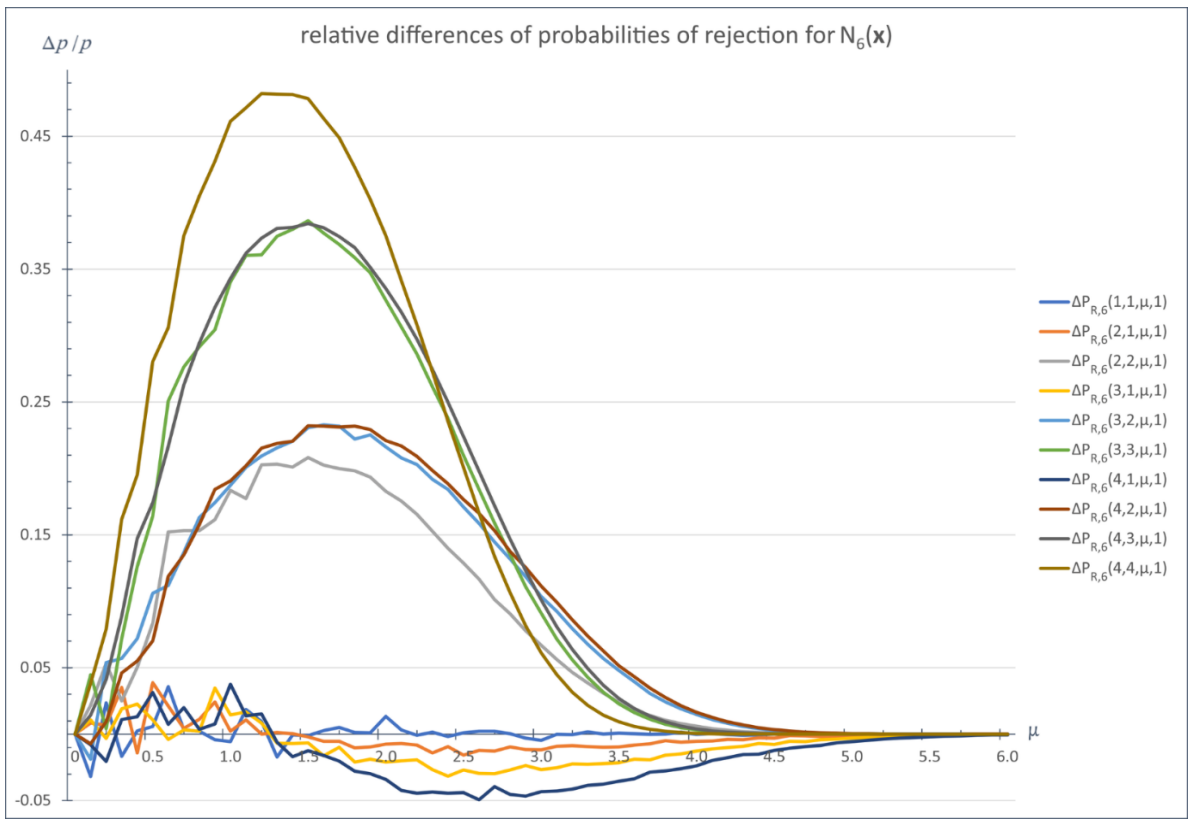

*Figure 166: $\Delta P_{R,6}(n, k, \mu, 1)$*

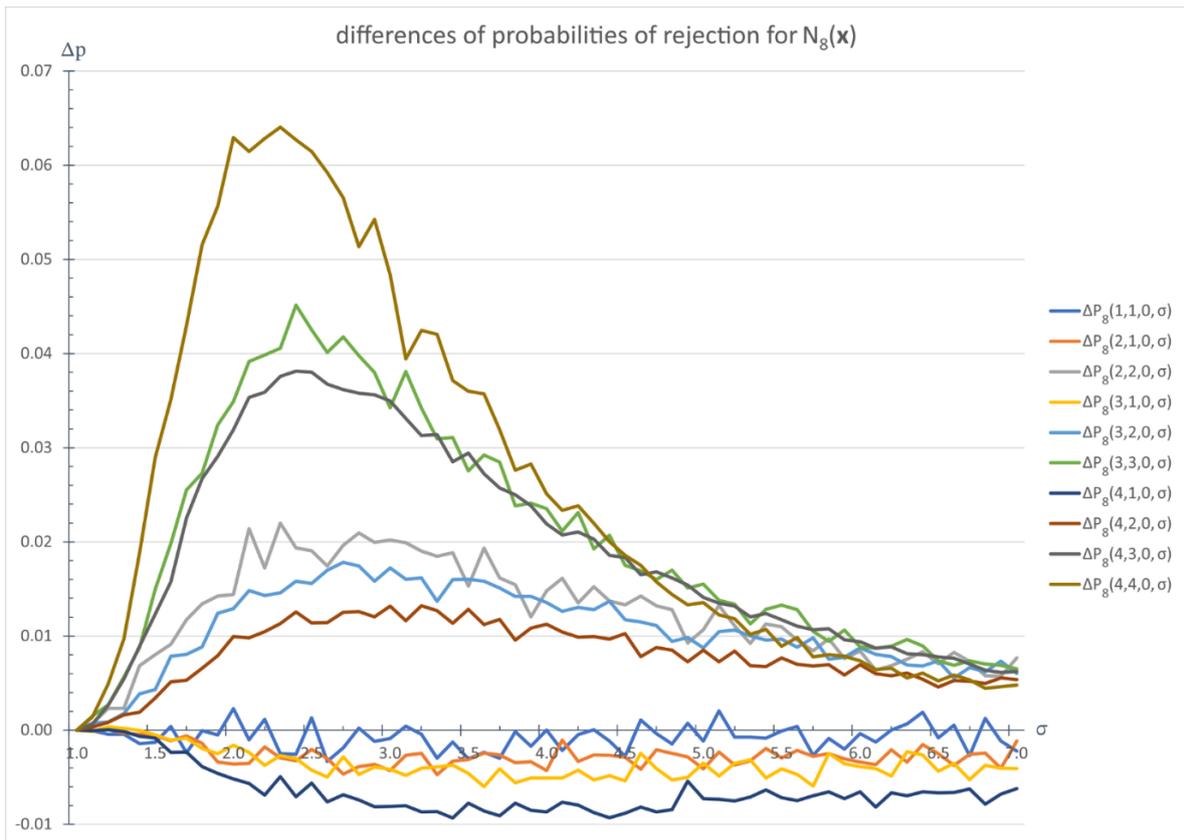

*Figure 167: $\Delta P_8(n, k, 0, \sigma)$*

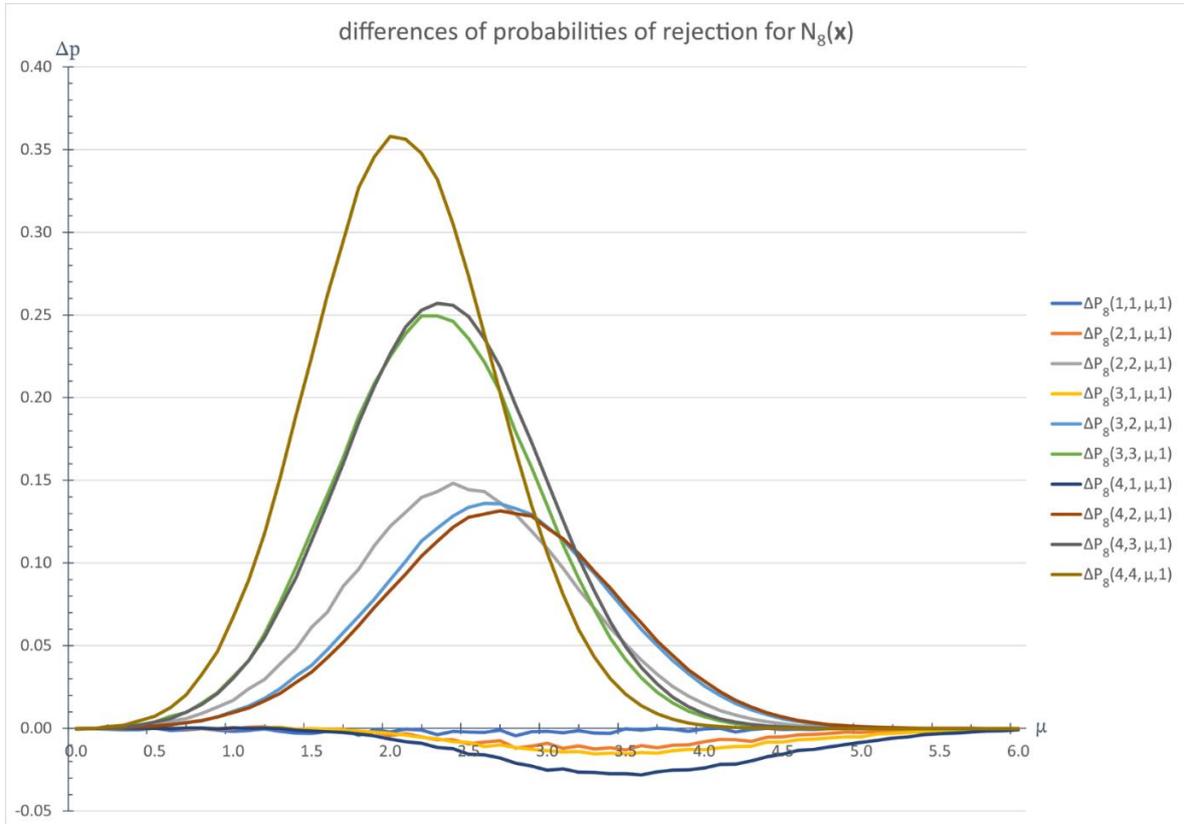

*Figure 168: $\Delta P_8(n, k, \mu, 1)$*

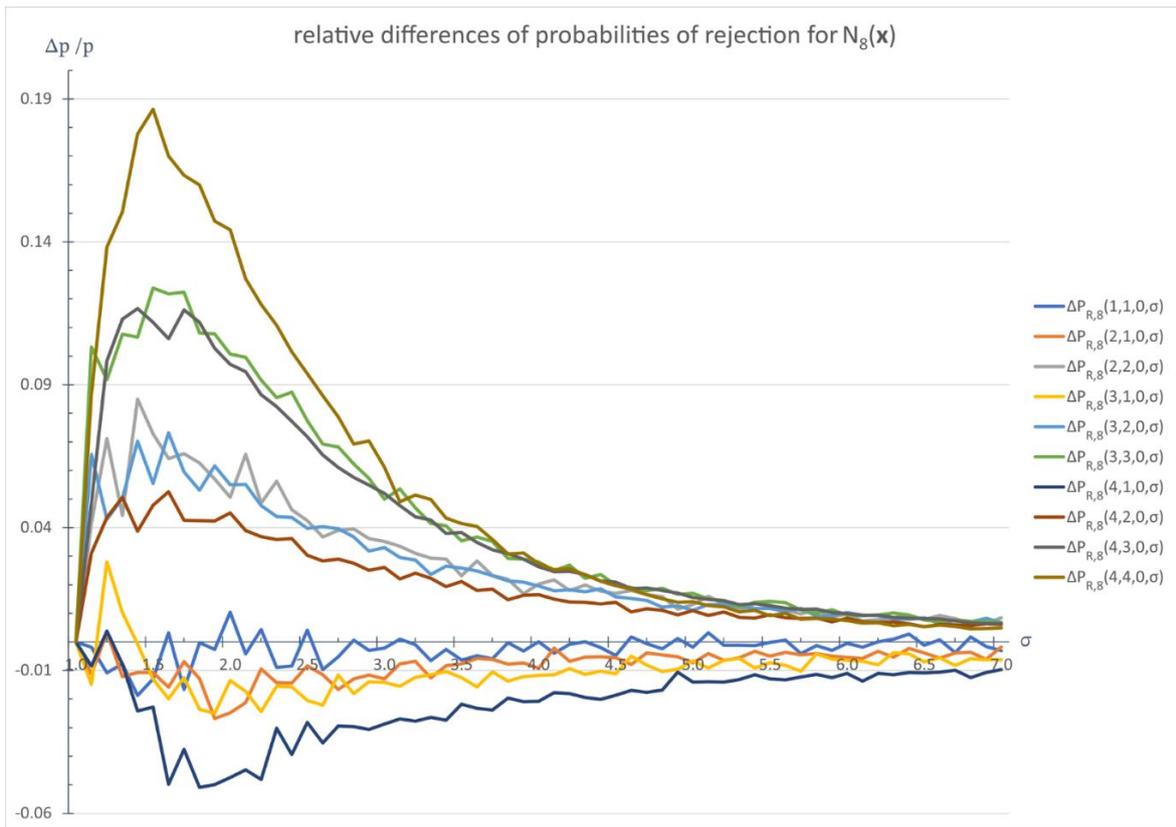

*Figure 169: $\Delta P_{R,8}(n,k,0,\sigma)$*

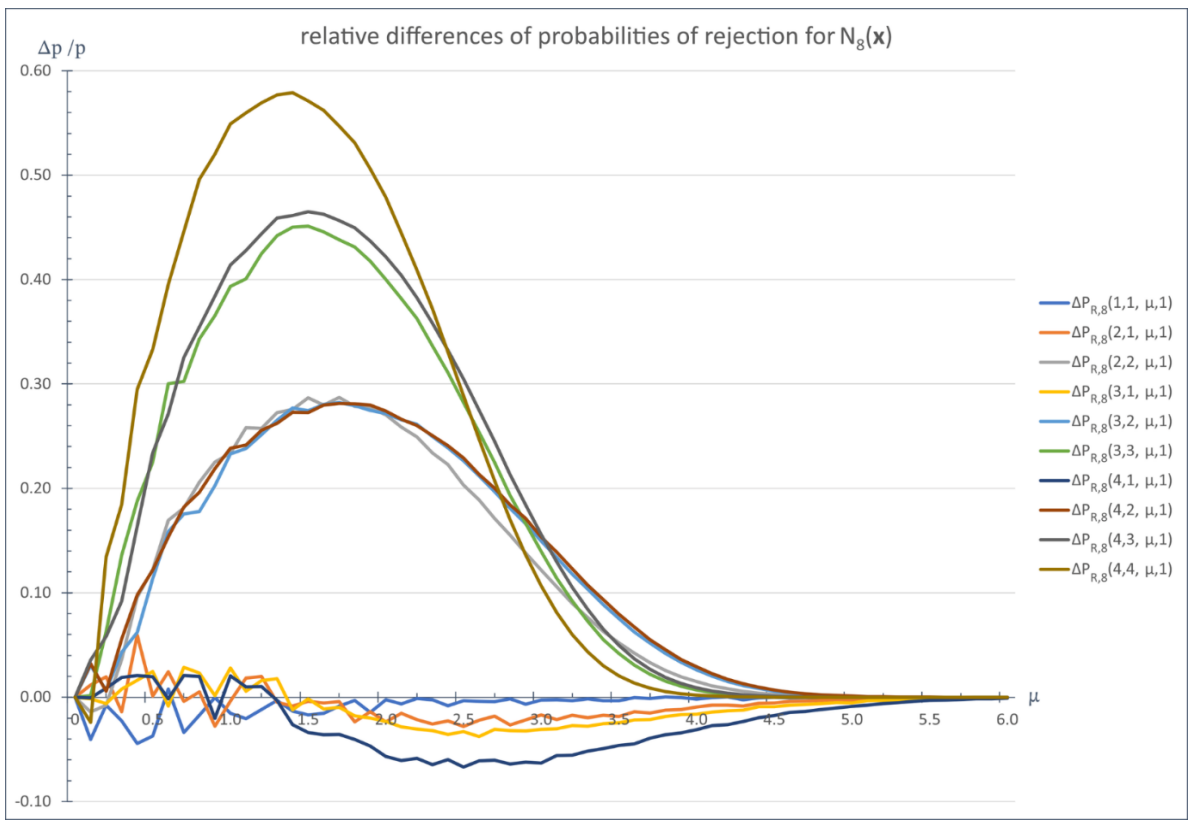

*Figure 170: $\Delta P_{R,8}(n,k,\mu,1)$*

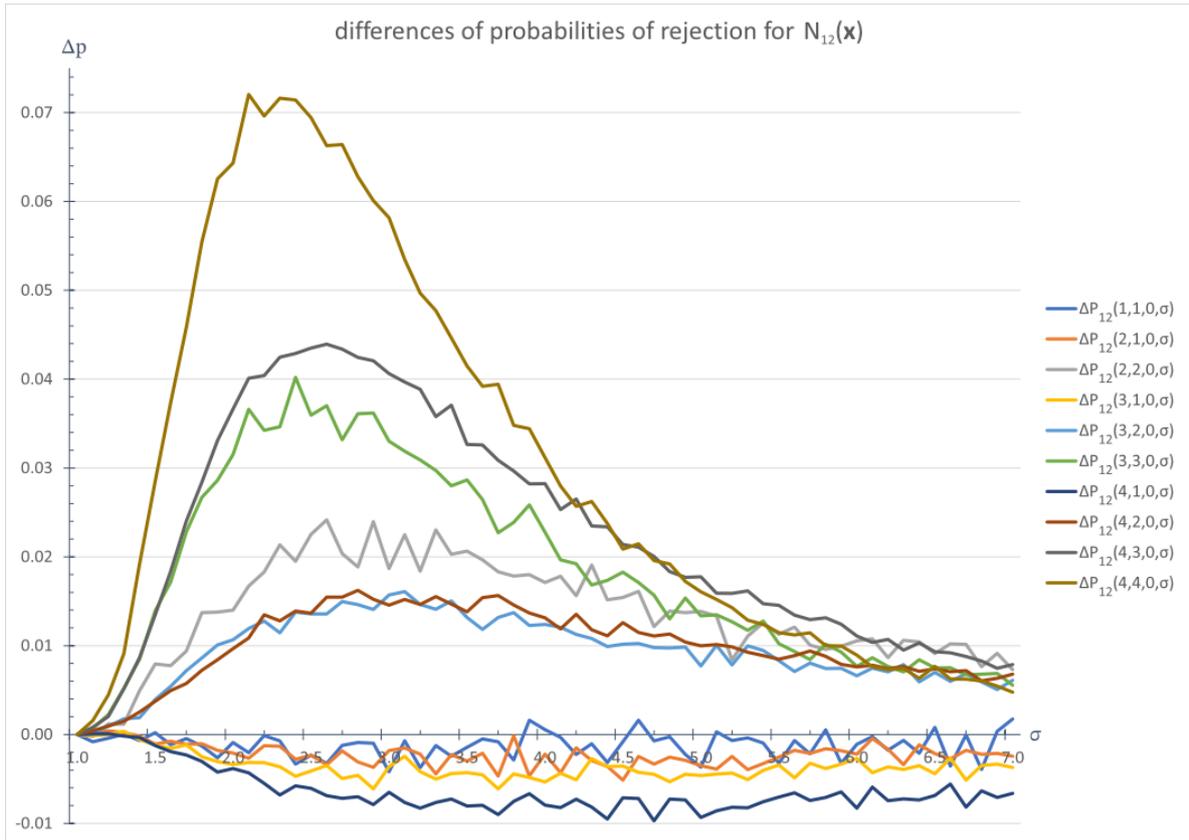

*Figure 171: $\Delta P_{12}(n, k, 0, \sigma)$*

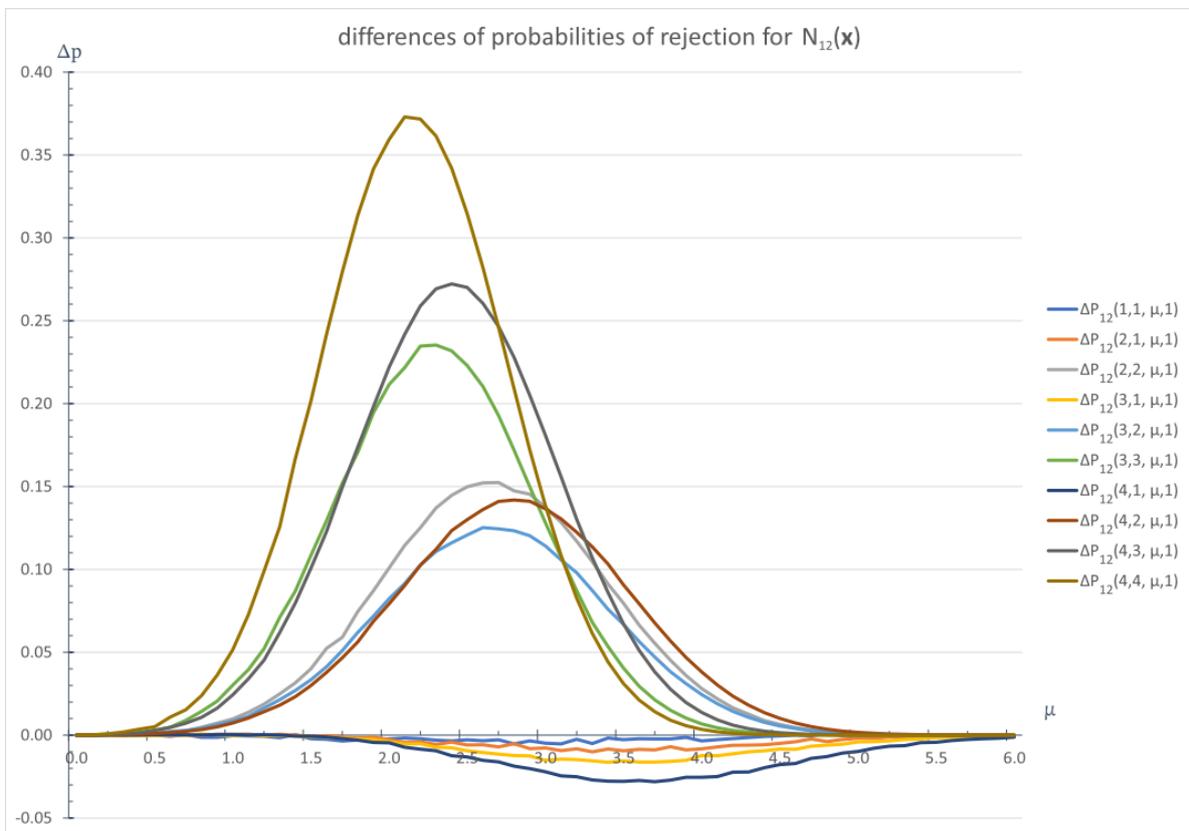

*Figure 172: $\Delta P_{12}(n, k, \mu, 1)$*

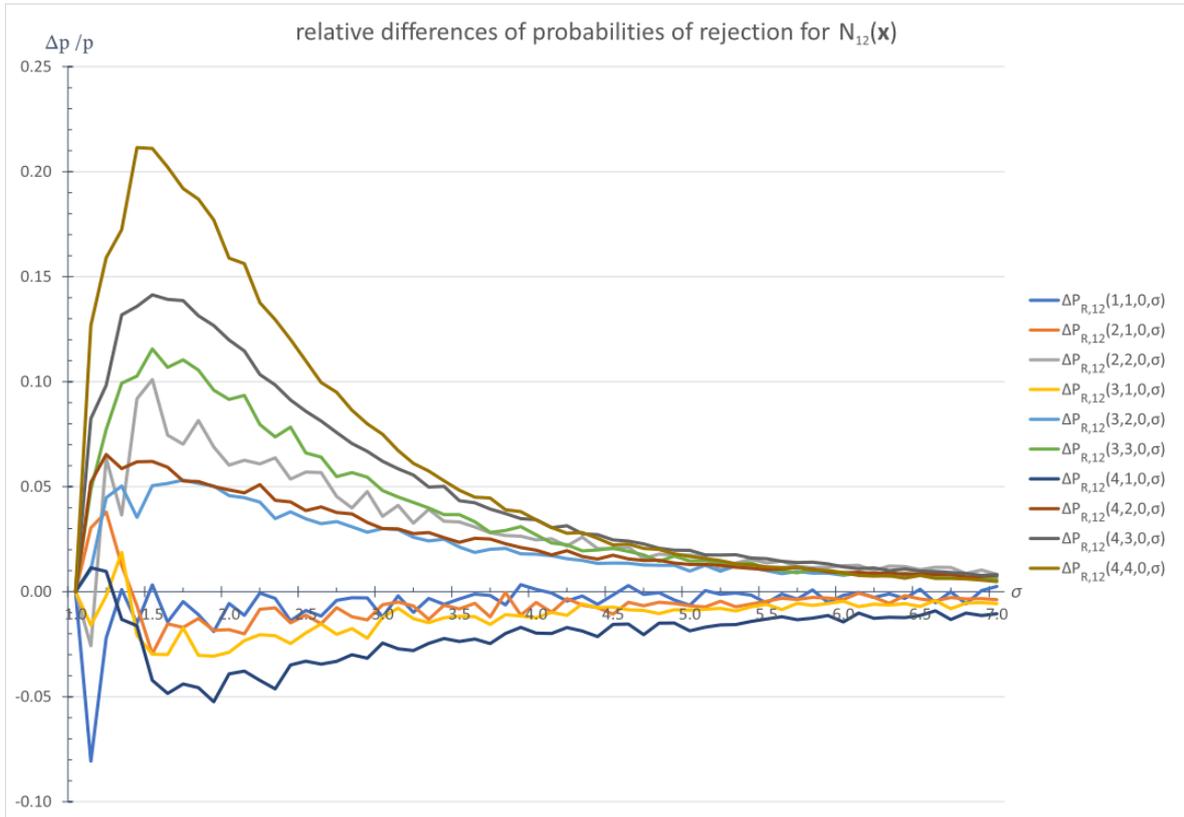

*Figure 173:* $\Delta P_{R,12}(n,k,0,\sigma)$

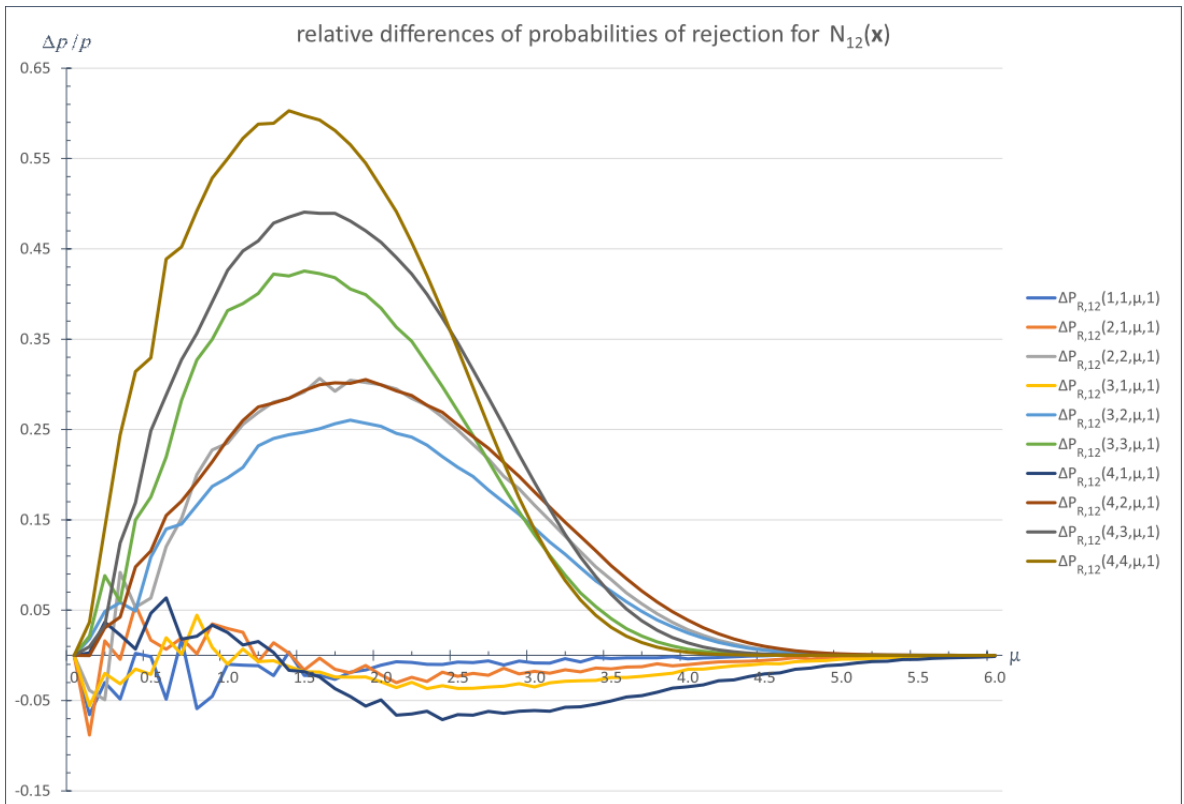

*Figure 174:* $\Delta P_{R,12}(n,k,\mu,1)$

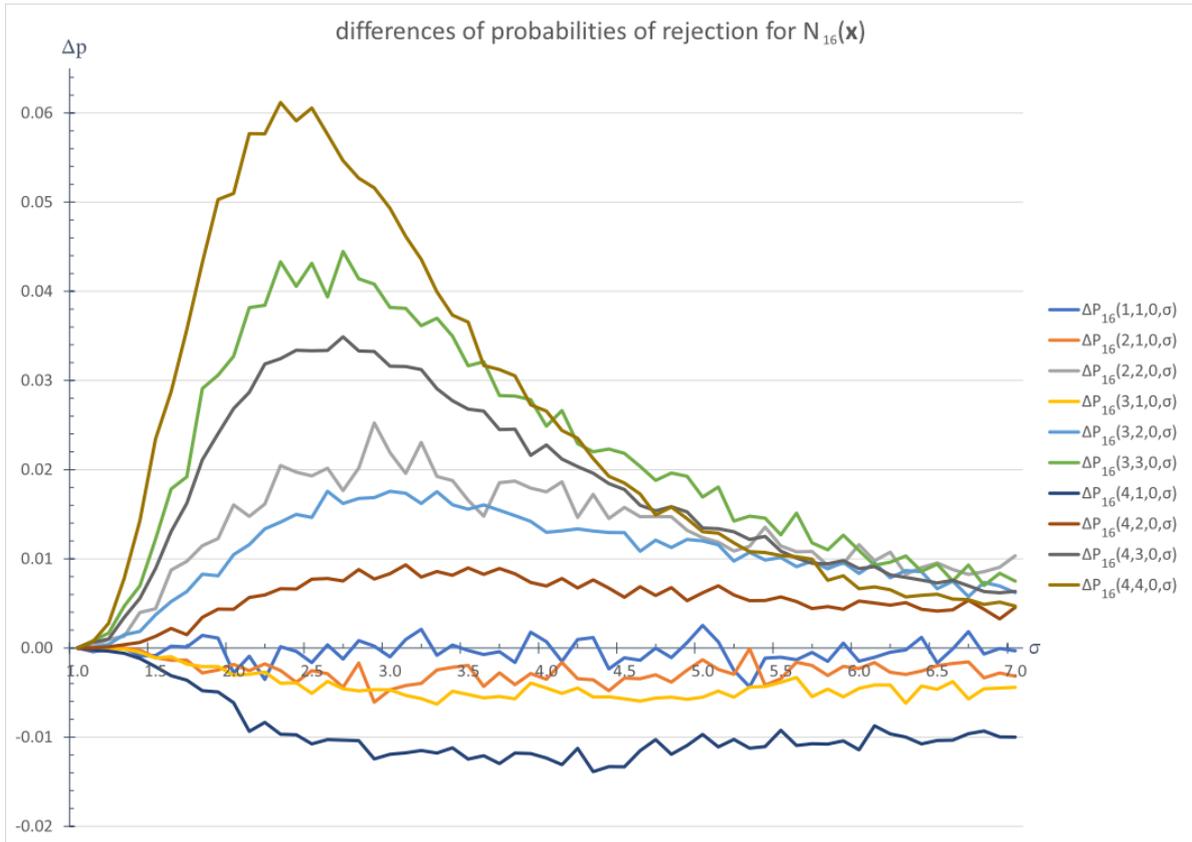

*Figure 175: $\Delta P_{16}(n, k, 0, \sigma)$*

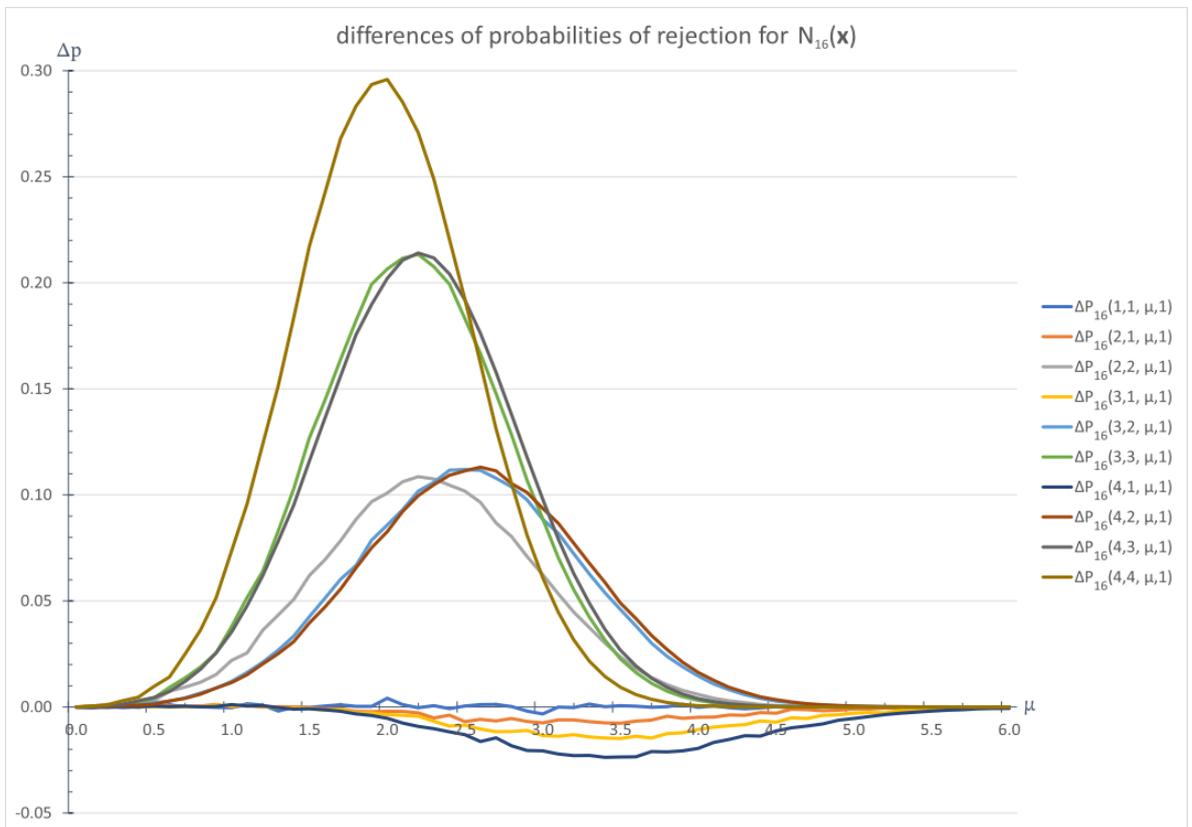

*Figure 176: $\Delta P_{16}(n, k, \mu, 1)$*

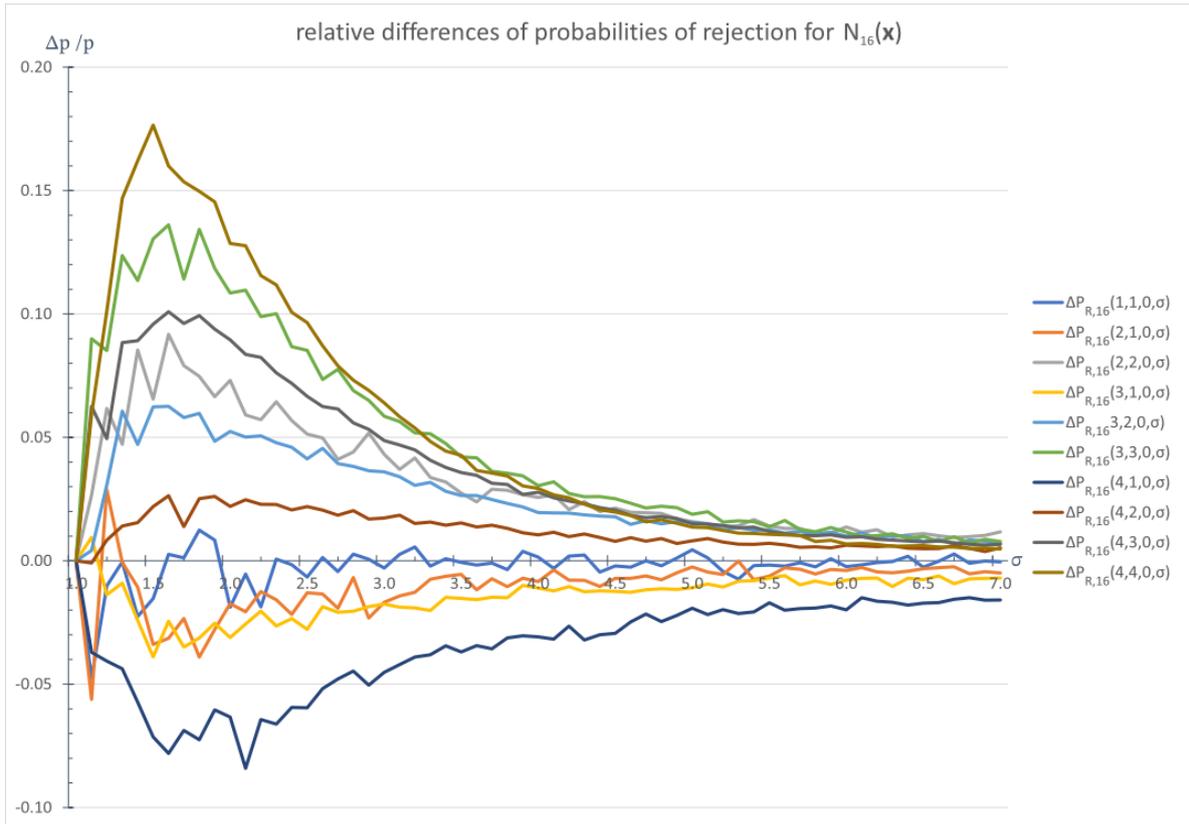

*Figure 177: $\Delta P_{R,16}(n,k,0,\sigma)$*

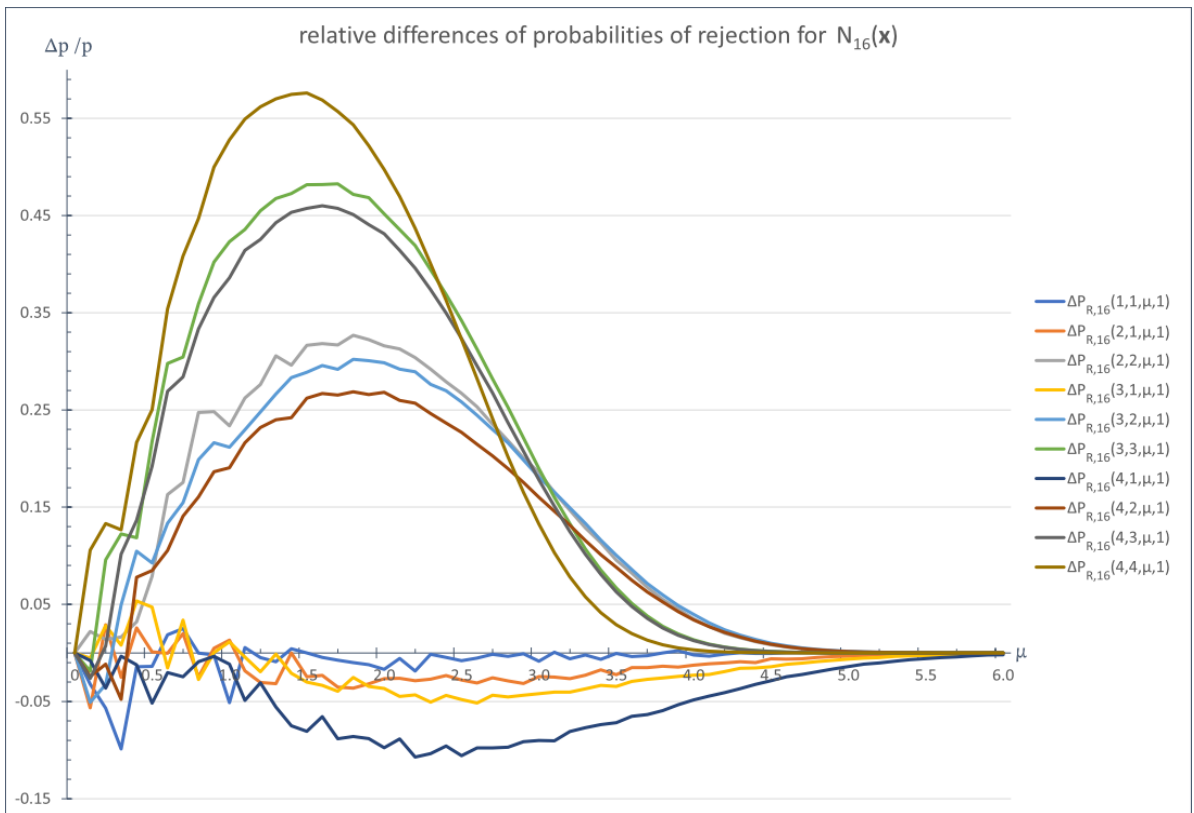

*Figure 178: $\Delta P_{R,16}(n,k,\mu,1)$*